\documentclass{book}
\usepackage{times}

\usepackage[utf8]{inputenc}
\usepackage[english]{babel}
\usepackage[letterpaper,top=2cm,bottom=2cm,left=3cm,right=3cm,marginparwidth=1.75cm]{geometry}

\usepackage[numbers]{natbib}
\usepackage{amsfonts}
\UseRawInputEncoding

\usepackage{pdflscape}
\usepackage{rotating}
\usepackage{longtable}
\usepackage{array}
\usepackage{float}
\usepackage{amsmath}
\usepackage{graphicx}
\usepackage{courier}
\usepackage[colorlinks=true, allcolors=blue]{hyperref}
\usepackage{enumitem}
\usepackage{tikz} 
\usepackage{listings}
\usepackage{xcolor}
\usepackage[T1]{fontenc}
\usepackage{epigraph}  

\definecolor{bgcolor}{rgb}{0.97,0.97,0.97}
\definecolor{codeblue}{rgb}{0.1,0.1,0.8}
\definecolor{codegreen}{rgb}{0,0.4,0}
\definecolor{codegray}{rgb}{0.4,0.4,0.4}
\definecolor{codepurple}{rgb}{0.5,0,0.5}
\definecolor{codered}{rgb}{0.6,0.2,0.2}
\definecolor{lightgray}{rgb}{0.9,0.9,0.9}
\definecolor{darkgray}{rgb}{0.6,0.6,0.6} 

\makeatletter
\renewcommand{\paragraph}{%
  \@startsection{paragraph}{4}{\z@}{1ex}{-1em}{\normalfont\normalsize\bfseries\color{gray}}}
\makeatother

\lstdefinestyle{python}{
    language=Python, 
    basicstyle=\ttfamily\small\color{black}, 
    keywordstyle=\bfseries\color{blue},  
    stringstyle=\color{orange!85!black},  
    commentstyle=\color{green!50!black},  
    showstringspaces=false,  
    numbers=left,  
    numberstyle=\tiny\color{gray},  
    stepnumber=1,  
    numbersep=8pt,  
    frame=single,  
    rulecolor=\color{black},  
    breaklines=true,  
    backgroundcolor=\color{gray!10},  
    tabsize=4,  
    captionpos=b,  
    morekeywords={self, None, True, False}, 
}

\lstdefinestyle{cmd}{
    language=bash,
    basicstyle=\ttfamily\small\color{black},  
    keywordstyle=\bfseries\color{blue},  
    stringstyle=\color{red!80!black},  
    commentstyle=\color{gray},  
    showstringspaces=false,  
    numbers=none,  
    frame=single,  
    rulecolor=\color{black},  
    breaklines=true,  
    backgroundcolor=\color{gray!10},  
    tabsize=4,  
    captionpos=b,  
}

\title{A Comprehensive Guide to Explainable AI: From Classical Models to LLMs}
\footnotesize
\author{
    Weiche Hsieh\textsuperscript{*$\dagger$}\thanks{National Tsing Hua University, \texttt{s112033645@m112.nthu.edu.tw}} \and
    Ziqian Bi\textsuperscript{*}\thanks{Indiana University, \texttt{bizi@iu.edu}} \and
    Chuanqi Jiang\textsuperscript{*}\thanks{National University of Singapore, \texttt{e0729764@u.nus.edu}} \and
    Junyu Liu\thanks{Kyoto University, \texttt{liu.junyu.82w@st.kyoto-u.ac.jp}} \and
    Benji Peng\thanks{AppCubic, \texttt{benji@appcubic.com}} \and
    Sen Zhang\thanks{Rutgers University, \texttt{sen.z@rutgers.edu}} \and
    Xuanhe Pan\thanks{University of Wisconsin-Madison, \texttt{xpan73@wisc.edu}} \and
    Jiawei Xu\thanks{Purdue University, \texttt{xu1644@purdue.edu}} \and
    Jinlang Wang\thanks{University of Wisconsin-Madison, \texttt{jinlang.wang@wisc.edu}} \and
    Keyu Chen\thanks{Georgia Institute of Technology, \texttt{kchen637@gatech.edu}} \and
    Pohsun Feng\thanks{National Taiwan Normal University, \texttt{41075018h@ntnu.edu.tw}} \and
    Yizhu Wen\thanks{University of Hawaii, \texttt{yizhuw@hawaii.edu}} \and
    Xinyuan Song\thanks{National University of Singapore, \texttt{songxinyuan@u.nus.edu}} \and
    Tianyang Wang\thanks{Xi'an Jiaotong-Liverpool University, \texttt{Tianyang.Wang21@student.xjtlu.edu.cn}} \and
    Ming Liu\thanks{Purdue University, \texttt{liu3183@purdue.edu}} \and
    Junjie Yang\thanks{Pingtan Research Institute of Xiamen University, \texttt{youngboy@xmu.edu.cn}} \and
    Ming Li\thanks{Georgia Institute of Technology, \texttt{mli694@gatech.edu}} \and
    Bowen Jing\thanks{University of Manchester, \texttt{bowen.jing@postgrad.manchester.ac.uk}} \and
    Jintao Ren\thanks{Aarhus University, \texttt{jintaoren@clin.au.dk}} \and
    Junhao Song\thanks{Imperial College London, \texttt{junhao.song23@imperial.ac.uk}} \and
    Hong-Ming Tseng\thanks{School of Visual Arts, \texttt{htseng@sva.edu}} \and
    Yichao Zhang\thanks{The University of Texas at Dallas, \texttt{yichao.zhang.us@gmail.com}} \and
    Lawrence K.Q. Yan\thanks{Hong Kong University of Science and Technology, \texttt{kqyan@connect.ust.hk}} \and
    Qian Niu\thanks{Kyoto University, \texttt{niu.qian.f44@kyoto-u.jp}} \and
    Silin Chen\thanks{Zhejiang University, \texttt{A1033439225@gmail.com}} \and
    Yunze Wang\thanks{University of Edinburgh, \texttt{Y.Wang-861@sms.ed.ac.uk}}
    Chia Xin Liang\thanks{JTB Technology Corp., \texttt{cxldun@gmail.com}}
}

\date{} 

\begin{document}

\maketitle

\begingroup
\renewcommand\thefootnote{}\footnote{
    \textsuperscript{*} Equal contribution \\
    \textsuperscript{$\dagger$} Corresponding author
}
\addtocounter{footnote}{0}
\endgroup

\epigraph{"The computer was born to solve problems that did not exist before."}{\textit{Bill Gates}}

\epigraph{"Design is where science and art break even."}{\textit{Robin Matthews}}

\epigraph{"Computers are good at following instructions, but not at reading your mind."}{\textit{Donald Knuth}}

\epigraph{"A good programmer is someone who always looks both ways before crossing a one-way street."}{\textit{Doug Linder}}

\epigraph{"The spread of computers and the Internet will put jobs in two categories. People who tell computers what to do, and people who are told by computers what to do."}{\textit{Marc Andreessen}}

\tableofcontents  

\chapter{Introduction}

\label{sec:Introduction}

\section{Background and Importance of Explainable AI (XAI)}

Artificial Intelligence (AI) has permeated numerous aspects of our daily lives, from predictive text on our smartphones to complex decision-making systems in healthcare and finance \cite{Jordan2015}. While AI has shown remarkable accuracy and efficiency, it is often criticized for being a 'black box,' particularly when it comes to complex models like deep learning and large language models (LLMs) \cite{Lipton2016}. This is where Explainable AI (XAI) comes into play \cite{Adadi2018}.

Explainable AI aims to make AI decisions transparent, understandable, and interpretable \cite{Doshi-Velez2017}. The lack of interpretability in AI systems has raised concerns about trust, accountability, and fairness \cite{Gilpin2018}. For instance, consider an AI system denying a bank loan application. Without explanation, the applicant is left in the dark, unable to understand why the decision was made or what could be improved for a future application.

Moreover, regulatory bodies like the European Union's General Data Protection Regulation (GDPR) emphasize the 'right to explanation,' increasing the demand for interpretable AI systems \cite{GDPR2016,Goodman2017}. Explainable AI not only builds trust with users but also facilitates debugging, compliance, and improved performance in AI systems \cite{Samek2017}. It addresses the fundamental question: How can we trust a system that we do not understand?

\section{Core Concepts and Definitions of XAI}

Before diving into the core concepts of Explainable AI, let's define some critical terms that will be used throughout this book:

\begin{itemize}
\item \textbf{Interpretability:} The degree to which a human can understand the cause of a decision. This often involves simplifying complex model predictions into human-comprehensible insights \cite{Lipton2016}.
\item \textbf{Transparency:} The openness and accessibility of the model's structure and data, which allows for external scrutiny. Transparent models like decision trees are considered intrinsically interpretable \cite{Rudin2019}.
\item \textbf{Fairness:} The assurance that AI systems do not produce biased results or discrimination based on sensitive attributes such as race, gender, or age \cite{Barocas2016}.
\item \textbf{Explainability:} The extent to which the internal mechanics of a machine learning model can be understood. Explainability goes a step further than interpretability by focusing on 'why' a decision was made \cite{Arrieta2020}.
\end{itemize}

These concepts are not mutually exclusive but rather interconnected aspects of XAI. For example, transparency aids interpretability, while interpretability facilitates explainability. It is essential to clarify these terms, as they form the foundation of our discussion on the techniques and applications of XAI.

\section{XAI, Transparency, Interpretability, and Fairness in AI}

The relationship between transparency, interpretability, and fairness is complex but crucial for the development of reliable AI systems \cite{Miller2019}. Let us illustrate these concepts with a few examples:

\begin{itemize}
\item \textbf{Transparency Example:} Imagine a simple linear regression model predicting house prices based on features like area, location, and age of the property. The model's coefficients can be easily inspected and interpreted, making it transparent \cite{Hastie2009}.
\item \textbf{Interpretability Example:} A decision tree used for medical diagnosis can provide clear, step-by-step reasoning for its predictions, making it interpretable even for non-experts \cite{Quinlan1986}.
\item \textbf{Fairness Example:} In a predictive policing model, if the training data includes biased crime reports, the model may disproportionately target specific demographics, raising fairness concerns \cite{Lum2016}.
\end{itemize}

\section{Structure of the Book and Reader's Guide}

This book is designed to guide readers through the fundamental concepts of Explainable AI (XAI), progressing to advanced techniques and exploring future research opportunities. Here is a brief overview of the chapters:

\begin{itemize}
\item \textbf{Chapter 2 - Theoretical Foundations of Explainable AI:} This chapter delves into the core reasons why interpretability is necessary in AI, discusses the inherent trade-offs between interpretability and model complexity, and outlines the challenges faced in achieving meaningful explanations.
\item \textbf{Chapter 3 - Interpretability of Traditional Machine Learning Models:} Focuses on classical models such as Decision Trees \cite{Quinlan1986}, Linear Regression \cite{Hastie2009}, Support Vector Machines \cite{Cortes1995}, and Bayesian Models, emphasizing their intrinsic interpretability and straightforward explanations.
\item \textbf{Chapter 4 - Interpretability of Deep Learning Models:} Explores the interpretability issues associated with deep learning models, including Convolutional Neural Networks (CNNs) and Recurrent Neural Networks (RNNs), and introduces techniques like feature visualization \cite{Zeiler2014} and attention mechanisms \cite{Bahdanau2015}.
\item \textbf{Chapter 5 - Interpretability of Large Language Models (LLMs):} Provides a comprehensive analysis of interpretability challenges specific to Large Language Models, including BERT \cite{Devlin2019}, GPT \cite{Radford2018}, and T5 \cite{Raffel2020}. The chapter covers techniques for probing, gradient-based analysis, and attention weight interpretation.
\item \textbf{Chapter 6 - Techniques for Explainable AI:} Introduces a variety of techniques for model interpretation, covering both intrinsic methods (like feature importance) and post-hoc methods (such as SHAP \cite{Lundberg2017}, LIME \cite{Ribeiro2016}, and Grad-CAM \cite{Selvaraju2017}). The chapter also includes advanced topics like counterfactual explanations and causal inference techniques.
\item \textbf{Chapter 7 - Applications of Explainable AI:} Discusses the practical use cases of XAI across different industries, including healthcare \cite{Samek2017,Tjoa2020}, finance, legal, and policy-making. The chapter provides examples of how XAI enhances decision support and addresses fairness issues.
\item \textbf{Chapter 8 - Evaluation and Challenges of Explainable AI:} Offers a detailed discussion on evaluating the quality of explanations using metrics like fidelity, stability, and comprehensibility. It also addresses key challenges such as the black-box nature of deep models and the trade-offs between accuracy and interpretability.
\item \textbf{Chapter 9 - Tools and Frameworks:} This chapter reviews the current landscape of XAI tools and frameworks, including model-agnostic tools like LIME and SHAP, deep learning-specific libraries like Captum \cite{Kokhlikyan2020}, and visualization frameworks for interactive explanations.
\item \textbf{Chapter 10 - Future Directions and Research Opportunities:} Concludes the book by examining emerging trends in XAI research, such as integrating XAI with legal compliance, exploring interpretability in federated learning, and addressing ethical concerns in AI explanations.
\end{itemize}

\textbf{Reader's  Guide:} This book is structured to be accessible to both newcomers and seasoned professionals in the field of AI. While each chapter builds on the concepts introduced in previous sections, the reader is encouraged to skip to specific chapters of interest if they are already familiar with certain topics. Throughout the book, practical examples are provided with Python code snippets to offer a hands-on understanding of the techniques discussed. These examples aim to bridge theory and practice, demonstrating the application of XAI methods in real-world scenarios. We hope this guide serves as a comprehensive resource on your journey to mastering Explainable AI.

For all the code examples, visit the GitHub repository: \url{https://github.com/Echoslayer/XAI_From_Classical_Models_to_LLMs.git}
\chapter{Theoretical Foundations of Explainable AI}
\label{sec:Foundations}

\section{Why is Interpretability Needed? The AI Black Box Problem}

The rise of artificial intelligence, particularly deep learning, has introduced remarkable advancements across numerous fields \citep{LeCun2015, Goodfellow2016}. However, with these advancements comes a critical issue: the 'Black Box' problem \citep{Lipton2016}. Many AI models, especially complex ones like neural networks and large language models (LLMs), are often regarded as black boxes due to their opaque decision-making processes \citep{Guidotti2018, Rudin2019}. The model might predict an outcome with high accuracy, but the reasoning behind the decision remains obscured. This lack of interpretability raises several significant concerns:

\begin{itemize}
\item \textbf{Trust and Accountability:} If an AI model makes a life-changing decision, such as diagnosing a medical condition \citep{Esteva2017} or approving a loan \citep{Khandani2010}, users need to understand the rationale behind it. Without this understanding, users cannot trust or verify the outcome \citep{Doshi-Velez2017}.
\item \textbf{Debugging and Improving Models:} Developers require insights into the decision-making process to diagnose errors or improve model performance. If we cannot interpret the model, finding the source of mistakes becomes guesswork \citep{Chandrasekaran2018}.
\item \textbf{Regulatory Compliance:} In domains like finance and healthcare, regulatory bodies demand that AI decisions are explainable. For instance, the European Union's General Data Protection Regulation (GDPR) includes the ''right to explanation,' which obligates organizations to provide clear reasoning for automated decisions \citep{GDPR2016, Wachter2017}.
\end{itemize}

The following diagram illustrates the AI black box problem:

\begin{center}
\begin{tikzpicture}

\node[draw, rectangle, minimum width=3cm, minimum height=1cm] (input) at (0, 0) {Input Data};
\node[draw, rectangle, minimum width=3cm, minimum height=1cm] (model) at (4, 0) {Complex Model};
\node[draw, rectangle, minimum width=3cm, minimum height=1cm] (output) at (8, 0) {Output Decision};

\draw[->] (input) -- (model);
\draw[->] (model) -- (output);

\draw[dashed, color=red] (model.north west) -- (model.south east);

\node[above] at (model.north) {\textbf{Black Box}};

\end{tikzpicture}
\end{center}

The red dashed line indicates the ''black box,' where the internal workings of the model are not directly observable. In response to this problem, Explainable AI (XAI) seeks to ''open the box' and provide interpretable explanations for model decisions \citep{Adadi2018}.

\section{Trade-off Between Interpretability and Model Complexity}

A common trade-off in AI is between \textbf{interpretability} and \textbf{model complexity} \citep{Rudin2019}. Models like decision trees and linear regression are interpretable by nature but often lack the flexibility to capture complex patterns in the data \citep{molnar2020interpretable}. On the other hand, deep learning models and LLMs have extraordinary predictive power but are notoriously difficult to interpret \citep{Lipton2016}.

Consider the following examples:

\begin{itemize}
\item \textbf{Linear Regression:} It is a simple, interpretable model where the coefficients directly indicate the relationship between features and the target variable \citep{Hastie2009}. However, it may not perform well on complex, non-linear datasets.
\item \textbf{Neural Networks:} These models can approximate complex functions and have shown state-of-the-art performance in tasks like image recognition and natural language processing \citep{Krizhevsky2012, Vaswani2017}. However, understanding the role of each neuron or layer in the decision-making process is challenging \citep{Montavon2018}.
\end{itemize}

This trade-off is illustrated in the diagram below:

\begin{center}
\begin{tikzpicture}
    \draw[->] (0,0) -- (6,0) node[right] {Model Complexity};
    \draw[->] (0,0) -- (0,4) node[above] {Interpretability};
    \node at (1,3) [circle, fill, inner sep=1.5pt, label=left:{\small Simple Models (High Interpretability)}] {};
    \node at (5,1) [circle, fill, inner sep=1.5pt, label=right:{\small Complex Models (Low Interpretability)}] {};
    \draw[dashed] (1,3) -- (5,1);
\end{tikzpicture}
\end{center}

The challenge is to find a balance, or use hybrid approaches that retain interpretability without sacrificing predictive power, a topic we will explore in depth in later chapters.

\section{Key Challenges in Achieving Interpretability}

Despite the growing interest in XAI, there are several key challenges in making models interpretable \citep{Adadi2018}:

\begin{itemize}
\item \textbf{Complexity of Modern AI Models:} Deep learning models have millions or even billions of parameters, making it nearly impossible to fully understand how each one contributes to the final decision \citep{LeCun2015}.
\item \textbf{Ambiguity in Interpretability:} There is no universal definition of interpretability. What is interpretable to one user (e.g., a data scientist) may not be interpretable to another (e.g., a medical professional) \citep{Lipton2016, Doshi-Velez2017}.
\item \textbf{Overfitting Risk:} Simplifying a model for interpretability can sometimes lead to oversimplification, reducing the model's predictive accuracy \citep{Ribeiro2016}.
\end{itemize}

\section{Different Levels and Types of Interpretability}

To understand interpretability, it is crucial to differentiate between various levels and types \citep{Montavon2018}:

\subsection{Interpretability vs. Visualization}

Interpretability should not be confused with visualization \citep{Choo2018}. Visualization involves techniques like plotting feature importance or activation maps, which can aid in understanding but are not explanations by themselves. We introduce the Iris dataset, a well-known dataset in the field of machine learning \citep{Fisher1936}, and build a simple machine learning model (further explained in Section \ref{sec:Traditional}). Using the SHAP tool \citep{Lundberg2017}, we attempt to interpret the model's predictions (further explained in Section \ref{sec:Techniques}). Here, the labels represent the species of iris flowers: \textit{setosa}, \textit{versicolor}, and \textit{virginica}. Below is the Python code used for this analysis:

\begin{lstlisting}[style=python, literate={\$}{{\$}}1]
import shap
import numpy as np
import pandas as pd
from sklearn.ensemble import RandomForestClassifier
from sklearn.datasets import load_iris
from sklearn.model_selection import train_test_split

Load the dataset

data = load_iris()
X = pd.DataFrame(data.data, columns=data.feature_names)
y = data.target

Split the dataset into training and testing sets

X_train, X_test, y_train, y_test = train_test_split(X, y, test_size=0.2, random_state=42)

Create and train a RandomForest model

model = RandomForestClassifier(random_state=42)
model.fit(X_train, y_train)

Initialize the SHAP Explainer

explainer = shap.Explainer(model.predict, X_test)

Compute SHAP values

shap_values = explainer(X_test)

Plot SHAP Summary Plot

shap.summary_plot(shap_values, X_test)
\end{lstlisting}

\begin{figure}[htbp]
\centering
\includegraphics[width=0.8\textwidth]{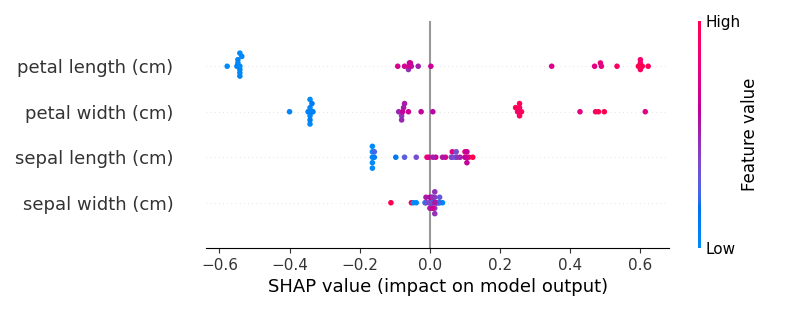}
\caption{SHAP Summary Plot for Feature Importance Visualization}
\label{fig:interpretability_vs_visualization}
\end{figure}

While the plot provides valuable insights into the relative importance of each feature, it does not fully explain why the model made a specific decision for an individual sample. In this example, the features like \texttt{petal length (cm)} and \texttt{petal width (cm)} appear to have the most significant impact on the predictions, pushing the decision boundary towards distinguishing the species \citep{Hastie2009}. However, without a deeper causal analysis, we cannot definitively claim why a particular sample was classified as \textit{versicolor} instead of \textit{virginica}. Thus, visualization here serves as a guide rather than a complete explanation. It's like using a magnifying glass: you can see the details better, but you still need the detective work to solve the mystery.

From this plot, we observe that the features with the largest SHAP values often correspond to those with the highest influence on the model's output, affirming our expectations about the importance of petal dimensions in distinguishing iris species \citep{Lundberg2017}. Hence, it can be concluded that while visualization aids interpretability, it is not a substitute for comprehensive explanations.

\subsection{Intrinsic Interpretability vs. Post-hoc Interpretability}

\begin{itemize}
\item \textbf{Intrinsic Interpretability:} Some models, like decision trees and linear regression, are interpretable by design \citep{Rudin2019}. Their simplicity allows for a straightforward understanding of how predictions are made.
\item \textbf{Post-hoc Interpretability:} More complex models often require additional methods for interpretation after training, such as LIME (Local Interpretable Model-agnostic Explanations) \citep{Ribeiro2016} or SHAP (Shapley Additive Explanations) \citep{Lundberg2017}.
\end{itemize}

The distinction is illustrated below:

\begin{center}
\begin{tikzpicture}

\node[draw, rectangle, minimum width=3cm, minimum height=1cm] (intrinsic) at (0, 0) {Intrinsic};
\node[draw, rectangle, minimum width=3cm, minimum height=1cm] (posthoc) at (5, 0) {Post-hoc};

\draw[->] (intrinsic) -- (posthoc);

\node[below] at (2.5, -0.5) {Complexity Increases $\Rightarrow$ Need for Post-hoc Techniques};

\end{tikzpicture}
\end{center}

\section*{Conclusion}

In this section, we covered the foundational concepts necessary to understand the field of Explainable AI. As we progress, we will dive deeper into the practical methods and tools for achieving interpretability, starting with traditional machine learning models in the next chapter.
\chapter{Interpretability of Traditional Machine Learning Models}
\label{sec:Traditional}

\section{Differences Between Interpretable and Non-interpretable Models}

When we discuss interpretability in machine learning, we refer to the ability to clearly understand and trace how a model reaches its predictions \cite{guidotti2018survey}. Interpretable models are those where a human observer can follow the decision-making process and directly link the input features to the output predictions \cite{lipton2016mythos}. In contrast, non-interpretable models, often referred to as "black box" models, have a more complex structure, making it challenging to understand the reasoning behind their predictions \cite{rudin2019stop}.

A common heuristic for differentiating between these models is as follows:

\begin{itemize}
    \item \textbf{Interpretable Models:} Models such as decision trees and linear models are considered interpretable. Their structure is designed in a way that each decision or coefficient can be explained and traced back to the input features \cite{quinlan1996learning, tibshirani1996regression}.
    \item \textbf{Non-interpretable Models:} Models like neural networks and ensemble methods (e.g., random forests and gradient boosting) are typically non-interpretable \cite{breiman2001random, chen2016xgboost}. Due to their complexity, consisting of numerous layers, nodes, and parameters, it is not straightforward to trace individual predictions \cite{lecun2015deep}.
\end{itemize}

To illustrate, a decision tree provides a transparent flow from the root node through various decision splits, ultimately leading to a leaf node that represents the final prediction. Each split can be interpreted based on the input features used at that decision point \cite{quinlan1986decision}. This kind of transparency makes decision trees highly interpretable and suitable for scenarios where explainability is critical.

On the other hand, consider a deep neural network with multiple hidden layers. The sheer number of neurons and weights creates a labyrinth of transformations that map the input to the output \cite{lecun2015deep}. Although this complexity often enhances the predictive power of the model, it significantly diminishes its interpretability, giving rise to what is known as the \textbf{Interpretability-Complexity Trade-off} \cite{rudin2019stop}. In essence, as the complexity of the model increases, its interpretability tends to decrease, and vice versa.

The trade-off between interpretability and model complexity is a fundamental issue in machine learning:

\begin{enumerate}
    \item \textbf{Increased Predictive Power:} Complex models like neural networks and ensemble methods generally achieve higher predictive accuracy due to their ability to capture intricate patterns in data \cite{lecun2015deep, breiman2001random}. However, the cost of this complexity is reduced interpretability.
    \item \textbf{Transparent Decision Process:} Simpler models, such as linear regression or shallow decision trees, offer a clear view of the decision-making process, but they may lack the flexibility to capture non-linear relationships in the data \cite{quinlan1986decision, tibshirani1996regression}.
\end{enumerate}
This trade-off is exemplified by two extremes:

\begin{itemize}
    \item \textbf{Example of an Interpretable Model:} In a linear regression model, the coefficients directly indicate the influence of each input feature on the output \cite{tibshirani1996regression}. If the coefficient of a feature is positive, it contributes positively to the prediction, and if it is negative, it contributes negatively. The magnitude of the coefficient reflects the strength of the relationship.
    \item \textbf{Example of a Non-interpretable Model:} In contrast, consider a deep convolutional neural network (CNN) trained to classify images \cite{lecun2015deep}. Each layer of the CNN applies multiple convolutions and transformations, extracting increasingly abstract features from the image. The final classification decision may depend on subtle patterns captured deep within the network, making it almost impossible for a human to trace back the reasoning process.
\end{itemize}

While the appeal of highly predictive, complex models is undeniable, there are significant domains, such as healthcare and finance, where interpretability is a key requirement \cite{caruana2015intelligible}. In these fields, decision-makers often need to justify their choices based on the model's predictions. Consequently, the tension between interpretability and predictive power remains an active area of research.

\paragraph{The Interpretability-Complexity Continuum}

To better understand the spectrum of interpretability, we can place common machine learning models along a continuum:

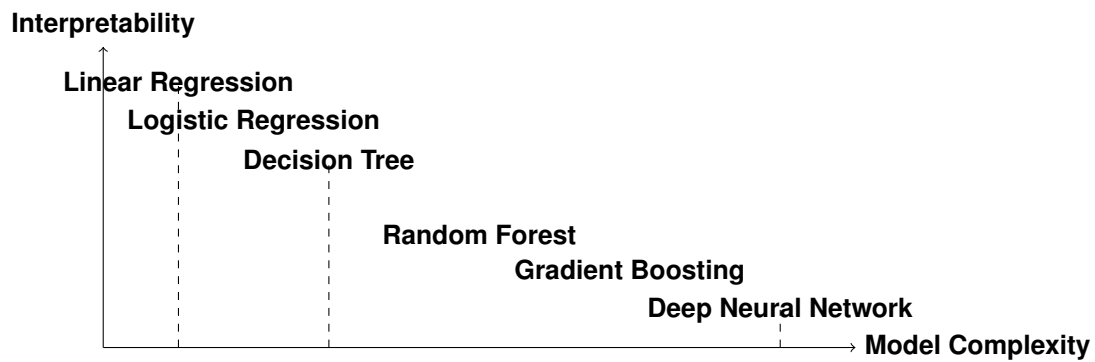
\begin{figure}[!ht]
    \centering
    \begin{tikzpicture}
        \draw[->] (0,0) -- (10,0) node[right]{\textbf{Model Complexity}};
        \draw[->] (0,0) -- (0,4) node[above]{\textbf{Interpretability}};

        \node at (1,3.5) {\textbf{Linear Regression}};
        \node at (2,3) {\textbf{Logistic Regression}};
        \node at (3,2.5) {\textbf{Decision Tree}};
        \node at (5,1.5) {\textbf{Random Forest}};
        \node at (7,1) {\textbf{Gradient Boosting}};
        \node at (9,0.5) {\textbf{Deep Neural Network}};

        \draw[dashed] (1,0) -- (1,3.5);
        \draw[dashed] (3,0) -- (3,2.5);
        \draw[dashed] (9,0) -- (9,0.5);
    \end{tikzpicture}
    \caption{The interpretability-complexity continuum of common machine learning models \cite{rudin2019stop}.}
    \label{fig:interpretability_complexity}
\end{figure}

In summary, while the choice of model often depends on the specific task and the desired balance between interpretability and predictive performance, it is crucial to consider the potential consequences of deploying a black-box model, especially in sensitive and regulated domains \cite{caruana2015intelligible}. The field of Explainable AI (XAI) aims to bridge this gap by developing techniques that make even the most complex models more understandable and trustworthy \cite{samek2017explainable}.

\section{Decision Trees}
Decision trees are widely regarded as one of the most interpretable models in machine learning \cite{Quinlan1986}. They possess a simple, intuitive flowchart-like structure where internal nodes represent decision rules based on feature values, branches denote the outcomes of these decisions, and leaf nodes hold the final predictions. The path from the root to a leaf node provides a clear and understandable decision-making process, which is crucial for explainable AI applications.

\paragraph{Structure and Interpretation of Decision Trees}
A decision tree splits data into subsets based on the values of input features, aiming to separate the data in a way that reduces uncertainty or "impurity." The most common criteria for splitting nodes include:

\begin{itemize}
    \item \textbf{Gini Impurity:} Measures the probability of incorrectly classifying a randomly chosen element if it were labeled according to the distribution of labels in the subset. It is calculated as:
    \[
    \text{Gini Impurity} = 1 - \sum_{i=1}^{n} p_i^2,
    \]
    where \( p_i \) is the proportion of instances of class \( i \) in the node.
    \item \textbf{Information Gain:} Based on entropy, this metric assesses the reduction in uncertainty after a split. It is calculated as:
    \[
    \text{Information Gain} = \text{Entropy(parent)} - \left( \sum \text{Weighted Entropy(children)} \right),
    \]
    where entropy quantifies the disorder or impurity in a subset.
\end{itemize}

To illustrate, consider a decision tree used to classify whether a customer will subscribe to a product:

\begin{center}
\begin{tikzpicture}
    \node (root) at (0,0) [rectangle, draw] {Age > 30?};
    \node (left) at (-3,-2) [rectangle, draw] {Income > \$40k?};
    \node (right) at (3,-2) [rectangle, draw] {Subscribed};
    \node (left_left) at (-5,-4) [rectangle, draw] {Not Subscribed};
    \node (left_right) at (-1,-4) [rectangle, draw] {Subscribed};

    \draw [->] (root) -- (left) node[midway, left] {No};
    \draw [->] (root) -- (right) node[midway, right] {Yes};
    \draw [->] (left) -- (left_left) node[midway, left] {No};
    \draw [->] (left) -- (left_right) node[midway, right] {Yes};
\end{tikzpicture}
\end{center}

In this simplified tree:
\begin{itemize}
    \item The root node evaluates if the customer's age is above 30.
    \item If \textbf{Yes}, the customer is predicted to subscribe.
    \item If \textbf{No}, the decision splits further based on income.
    \item Income above \$40k leads to a positive subscription prediction, while lower income predicts no subscription.
\end{itemize}

This structure highlights why decision trees are considered interpretable: every decision can be explained in terms of the input features, making it easy to justify the model's predictions.

\paragraph{Pruning Techniques and Interpretability}
Although decision trees are inherently interpretable, they can easily grow too deep and become overly complex, capturing noise in the data and leading to overfitting. To combat this, we employ \textbf{pruning}, which simplifies the tree by removing nodes that provide minimal additional predictive power.

The two main pruning strategies are:

\begin{itemize}
    \item \textbf{Pre-pruning (Early Stopping):} Limits the growth of the tree based on predefined criteria, such as maximum depth or minimum number of samples per leaf. This reduces the risk of overfitting and keeps the tree structure simpler.
    \item \textbf{Post-pruning:} First grows the tree to its full extent and then trims back nodes that do not significantly improve model performance. This method often results in a more balanced model with higher generalization capabilities.
\end{itemize}

\begin{lstlisting}[style=python, literate={\$}{{\$}}1]
from sklearn.tree import DecisionTreeClassifier

# Example of pre-pruning with maximum depth
clf = DecisionTreeClassifier(max_depth=4, min_samples_leaf=10)
clf.fit(X_train, y_train)
\end{lstlisting}

In this example, the tree's depth is limited to 4, and each leaf must contain at least 10 samples. This helps maintain interpretability without sacrificing much predictive power.

\paragraph{Advantages and Disadvantages of Logistic Regression}

\begin{itemize}
    \item \textbf{Advantages:}
    \begin{enumerate}
        \item \textbf{Easy to Interpret:} The coefficients provide a direct way to understand feature impacts on the probability of the outcome \cite{narkhede2021logistic}.
        \item \textbf{Probabilistic Output:} The model outputs a probability, making it useful for applications requiring risk estimation \cite{hosmer2013applied}.
    \end{enumerate}
    \item \textbf{Disadvantages:}
    \begin{enumerate}
        \item \textbf{Assumption of Linearity:} Logistic Regression assumes a linear relationship between the features and the log-odds of the outcome.
        \item \textbf{Limited to Binary Classification:} It is not naturally suited for multi-class problems without extensions like softmax regression \cite{bishop2013pattern}.
    \end{enumerate}
\end{itemize}

\paragraph{Python Code Example}
Feature importance in decision trees is determined by assessing the role each feature plays in reducing the impurity of a node during the splitting process. Impurity is a measure of the disorder or randomness within the node, typically evaluated using metrics like \textbf{Gini Impurity} or \textbf{Entropy}. When a feature significantly reduces impurity, it receives a higher importance score.

In essence, the more a feature contributes to reducing impurity across the tree splits, the more important it is considered. High feature importance implies that the model relies heavily on that feature for making predictions, making it a key factor in understanding the model's decision-making process.

For a practical demonstration, consider the following Python code, which showcases how to train a decision tree and extract feature importance.

\begin{lstlisting}[style=python, literate={\$}{{\$}}1]
import matplotlib.pyplot as plt
import numpy as np
from sklearn.tree import DecisionTreeClassifier
from sklearn.datasets import make_classification
from sklearn.model_selection import train_test_split

# 1. Generate synthetic data with 3 features for binary classification
X, y = make_classification(
    n_samples=100,          # Number of samples
    n_features=3,           # Number of features
    n_informative=3,        # Number of informative features
    n_redundant=0,          # No redundant features
    n_classes=2,            # Binary classification
    random_state=42         # Random seed for reproducibility
)

# 2. Split the data into training and testing sets (70% train, 30% test)
X_train, X_test, y_train, y_test = train_test_split(X, y, test_size=0.3, random_state=42)

# 3. Train a Decision Tree Classifier with a maximum depth of 4
clf = DecisionTreeClassifier(max_depth=4, random_state=42)
clf.fit(X_train, y_train)

# 4. Extract the feature importance scores from the trained classifier
feature_importance = clf.feature_importances_

# 5. Define feature names for the plot (e.g., Feature 1, Feature 2, Feature 3)
features = np.array(['Feature 1', 'Feature 2', 'Feature 3'])

# 6. Plot a horizontal bar chart to visualize feature importance
plt.barh(features, feature_importance)
plt.xlabel('Importance Score')                # Label for the x-axis
plt.title('Feature Importance in Decision Tree')  # Title of the plot
plt.show()                                   # Display the plot
\end{lstlisting}

\begin{figure}[!ht]
    \centering
    \includegraphics[width=0.7\textwidth]{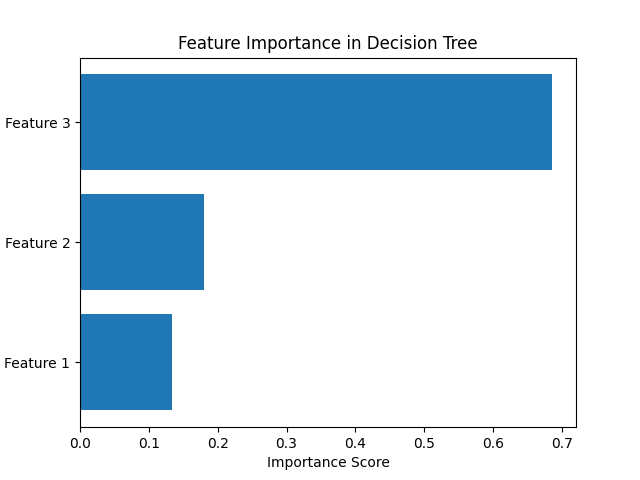}
    \caption{Visualization of Feature Importance in a Decision Tree}
    \label{fig:feature_importance}
\end{figure}

\paragraph{Explanation of Results}
The bar chart in Figure \ref{fig:feature_importance} illustrates the importance scores of each feature used in the decision tree. The x-axis represents the importance score, and the y-axis lists the features.

\begin{itemize}
    \item \textbf{Feature 3} has the highest importance score, indicating it is the most influential feature in the model's predictions. This suggests that the decision tree frequently relies on Feature 3 for splitting, as it contributes significantly to reducing impurity.
    \item \textbf{Feature 2} has a moderate importance score, playing an important but secondary role in the model. It provides useful information, but its influence is less pronounced compared to Feature 3.
    \item \textbf{Feature 1} shows the lowest importance score, implying that it has the least impact on the model's decisions. It likely provides minimal information for splitting, making it the least informative feature among the three.
\end{itemize}

This analysis highlights the relative contributions of the features, providing insight into the model's decision-making process. In this case, Feature 3 appears to be the primary driver of predictions, while Feature 1 has a minor role.

\paragraph{Key Takeaways}
\begin{enumerate}
    \item \textbf{Dominance of Feature 3:} The highest importance score for Feature 3 suggests it is the primary feature influencing the model's predictions.
    \item \textbf{Supportive Role of Feature 2:} Although not as influential as Feature 3, Feature 2 still contributes meaningfully to the decision process.
    \item \textbf{Minimal Impact of Feature 1:} The low importance score for Feature 1 indicates it may be less informative or redundant, possibly a candidate for feature elimination in further analysis.
\end{enumerate}

\paragraph{Limitations and Considerations}
While decision trees provide clear interpretability, they are not always the best choice for complex datasets with intricate patterns. In practice:

\begin{itemize}
    \item \textbf{Ensemble Methods:} Techniques like Random Forests and Gradient Boosting build upon decision trees to improve predictive performance, but they sacrifice some interpretability.
    \item \textbf{Regularization:} Setting constraints (e.g., maximum depth, minimum samples per split) helps mitigate overfitting but may still leave the model susceptible to small data changes.
    \item \textbf{Complexity-Interpretability Trade-off:} As decision trees become deeper and more complex, they can lose their interpretability, blending into the realm of black-box models.
\end{itemize}

\section{Linear Models}
Linear models, including Linear Regression and Logistic Regression, are some of the most interpretable machine learning models. They assume a linear relationship between the input features and the output, making it straightforward to understand the effect of each feature on the prediction. Despite their simplicity, linear models remain powerful, especially when the underlying data relationships are approximately linear. In cases where interpretability is crucial, such as in finance and healthcare, linear models often serve as a go-to choice.

\paragraph{Interpretation of Linear Regression and Coefficients}
Linear Regression is a model that predicts a continuous output \(y\) as a weighted sum of input features \(x_i\):

\[
y = \beta_0 + \beta_1 x_1 + \beta_2 x_2 + \cdots + \beta_n x_n
\]

Here:
\begin{itemize}
    \item \(\beta_0\) is the intercept, representing the expected value of \(y\) when all features \(x_i = 0\).
    \item \(\beta_i\) is the coefficient of feature \(x_i\), indicating the expected change in \(y\) for a one-unit increase in \(x_i\), assuming all other features are held constant.
\end{itemize}

The coefficients \(\beta_i\) are key to interpreting the model . A positive coefficient indicates that an increase in the feature leads to an increase in the predicted value, while a negative coefficient suggests the opposite.

\paragraph{Python Code Example}
To illustrate the interpretability of linear models, let's consider a simple example where we predict house prices based on two features: square footage and the number of bedrooms.

\begin{lstlisting}[style=python, literate={\$}{{\$}}1]
import numpy as np
from sklearn.linear_model import LinearRegression
import matplotlib.pyplot as plt

# Example data: [Square Footage, Number of Bedrooms]
X = np.array([[1500, 3], [2000, 4], [2500, 4], [3000, 5]])
y = np.array([300000, 400000, 500000, 600000])

# Initialize and train the Linear Regression model
model = LinearRegression()
model.fit(X, y)

# Output the intercept and coefficients
print("Intercept:", model.intercept_)
print("Coefficients:", model.coef_)

# Predict house prices using the trained model
y_pred = model.predict(X)
print("Predicted Prices:", y_pred)

# Plot the true vs predicted prices
plt.scatter(range(len(y)), y, color='blue', label='True Prices')
plt.scatter(range(len(y_pred)), y_pred, color='red', marker='x', label='Predicted Prices')
plt.xlabel('Sample Index')
plt.ylabel('House Price ($)')
plt.title('True vs Predicted House Prices')
plt.legend()
plt.show()
\end{lstlisting}

\begin{figure}[htbp]
    \centering
    \includegraphics[width=0.7\textwidth]{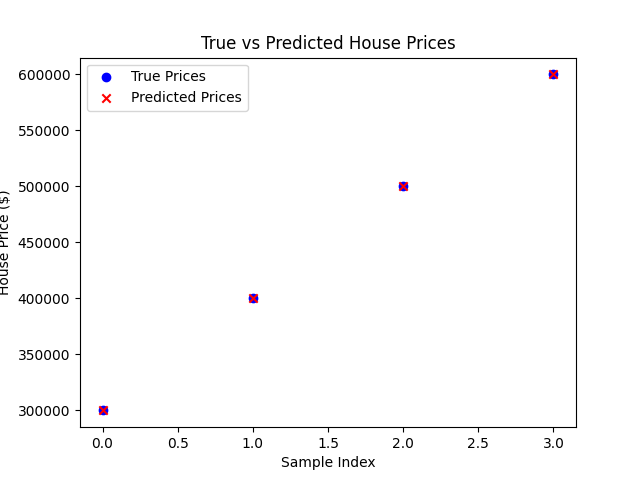}
    \caption{Comparison of True vs Predicted House Prices Using Linear Regression}
    \label{fig:linear_regression_results}
\end{figure}

\paragraph{Explanation of Results}
In Figure \ref{fig:linear_regression_results}, we observe a scatter plot comparing the true house prices (in blue) with the predicted prices from the linear regression model (in red). Let's break down the key findings:

\begin{itemize}
    \item \textbf{Intercept:} The intercept value indicates the base price of a house when both square footage and the number of bedrooms are zero. In this case, the intercept is a positive value, reflecting a baseline price component unrelated to the input features.
    \item \textbf{Coefficients:} The coefficients output by the model represent the change in house price for a one-unit increase in each feature:
    \begin{itemize}
        \item The coefficient for square footage is positive, indicating that as the square footage increases, the predicted house price also increases. Specifically, for each additional square foot, the price increases by the value of the coefficient (e.g., \$150 per square foot).
        \item The coefficient for the number of bedrooms is also positive, suggesting that more bedrooms are associated with higher house prices. For each additional bedroom, the predicted price increases by a fixed amount (e.g., \$25,000 per bedroom).
    \end{itemize}
    \item \textbf{Prediction Accuracy:} The plot shows that the predicted prices (red crosses) align well with the true prices (blue dots). This indicates a good fit, as the model captures the linear relationship between the input features and the target variable. In practical terms, this model can provide a quick and reasonably accurate estimate of house prices based on these features.
\end{itemize}

\paragraph{Key Takeaways}
\begin{enumerate}
    \item \textbf{Simplicity and Interpretability:} Linear Regression provides a straightforward interpretation of the relationship between features and the target variable. The coefficients directly indicate the magnitude and direction of influence of each feature.
    \item \textbf{Good Fit for Linear Relationships:} In this example, the model performs well because the relationship between the features (square footage and number of bedrooms) and house price is approximately linear.
    \item \textbf{Limitations:} While the model fits well here, linear regression assumes a linear relationship between the inputs and output. It may not capture more complex patterns or interactions between features, which we will address in later chapters with non-linear models.
\end{enumerate}

\paragraph{Limitations of Linear Regression}
While linear regression is simple and interpretable, it has several limitations:
\begin{enumerate}
    \item \textbf{Assumption of Linearity:} Linear regression assumes a linear relationship between features and the target. This may not hold true for complex datasets.
    \item \textbf{Sensitivity to Outliers:} Outliers can heavily influence the fitted line, leading to poor predictions.
    \item \textbf{Multicollinearity:} When features are highly correlated, it becomes difficult to determine the individual effect of each feature on the output.
\end{enumerate}

\paragraph{Python Code Example}
In this example, we demonstrate how Logistic Regression can be used to predict customer churn based on two features: monthly charges and tenure. The model aims to predict whether a customer will churn (i.e., leave the service) or not, using a binary outcome (0 = No churn, 1 = Churn).

\begin{lstlisting}[style=python, literate={\$}{{\$}}1]
import numpy as np
from sklearn.linear_model import LogisticRegression
import matplotlib.pyplot as plt

# Example data: [Monthly Charges, Tenure]
X = np.array([[30, 1], [40, 3], [50, 5], [60, 7]])
y = np.array([0, 0, 1, 1])  # 0 = No churn, 1 = Churn

# Initialize and train the Logistic Regression model
model = LogisticRegression()
model.fit(X, y)

# Output the intercept and coefficients
print("Intercept:", model.intercept_)
print("Coefficients:", model.coef_)

# Make predictions and predict probabilities
y_pred = model.predict(X)
y_prob = model.predict_proba(X)[:, 1]

print("Predicted Labels:", y_pred)
print("Predicted Probabilities (Churn):", y_prob)

# Visualization of the predicted probabilities
plt.scatter(range(len(y)), y, color='blue', label='True Labels (0=No churn, 1=Churn)')
plt.plot(range(len(y_prob)), y_prob, color='red', marker='x', linestyle='--', label='Predicted Probabilities')
plt.xlabel('Sample Index')
plt.ylabel('Probability of Churn')
plt.title('Logistic Regression: Churn Prediction')
plt.legend()
plt.show()
\end{lstlisting}

\begin{figure}[htbp]
    \centering
    \includegraphics[width=0.7\textwidth]{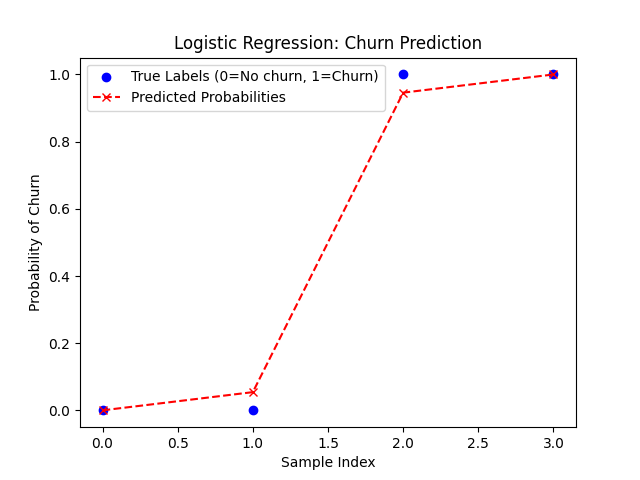}
    \caption{Logistic Regression Model Predictions for Customer Churn}
    \label{fig:logistic_regression_churn}
\end{figure}

\paragraph{Result Explanation}
In Figure \ref{fig:logistic_regression_churn}, we observe the predicted probabilities of customer churn (in red) compared with the true labels (in blue). Here's what the results indicate:

\begin{itemize}
    \item \textbf{Intercept and Coefficients:} The model's intercept and coefficients provide insight into the baseline churn probability and the influence of each feature:
    \begin{itemize}
        \item The intercept is negative, suggesting that without considering the features (monthly charges and tenure), the baseline probability of churn is low.
        \item The coefficient for monthly charges is positive, indicating that higher charges increase the likelihood of churn.
        \item The coefficient for tenure is also positive, suggesting that longer tenure is associated with a higher probability of churn. This might seem counterintuitive but could indicate customer dissatisfaction over time.
    \end{itemize}
    \item \textbf{Predicted Probabilities:} The red line represents the predicted probabilities of churn. As expected, the probabilities increase with higher monthly charges and longer tenure, aligning with the positive coefficients.
    \item \textbf{Model Predictions:} The predicted labels (0 or 1) are derived from the predicted probabilities using a default threshold of 0.5. The model correctly classifies all samples in this example, indicating a good fit to the data.
\end{itemize}

\paragraph{Key Takeaways}
\begin{enumerate}
    \item \textbf{Interpretable Coefficients:} Logistic Regression offers interpretable coefficients that help explain the relationship between features and the predicted outcome. A positive coefficient increases the log-odds of churn, while a negative coefficient decreases it.
    \item \textbf{Probabilistic Predictions:} Unlike linear regression, Logistic Regression predicts the probability of an event occurring. This probabilistic output is valuable in decision-making scenarios where risk assessment is crucial.
    \item \textbf{Limitations:} While the model performs well on this small dataset, it may struggle with more complex relationships or non-linear patterns. In such cases, more sophisticated models might be needed.
\end{enumerate}

\paragraph{Advantages and Disadvantages of Logistic Regression}
\begin{itemize}
    \item \textbf{Advantages:}
    \begin{enumerate}
        \item \textbf{Easy to Interpret:} The coefficients provide a direct way to understand feature impacts on the probability of the outcome.
        \item \textbf{Probabilistic Output:} The model outputs a probability, making it useful for applications requiring risk estimation.
    \end{enumerate}
    \item \textbf{Disadvantages:}
    \begin{enumerate}
        \item \textbf{Assumption of Linearity:} Logistic Regression assumes a linear relationship between the features and the log-odds of the outcome.
        \item \textbf{Limited to Binary Classification:} It is not naturally suited for multi-class problems without extensions like softmax regression.
    \end{enumerate}
\end{itemize}
\section{Interpretability of Support Vector Machines (SVM)}
Support Vector Machines (SVMs) are well-regarded for their robustness and ability to handle both linearly and non-linearly separable data \cite{James2013}. Although typically viewed as black-box models, SVMs with linear kernels can offer a degree of interpretability through their decision boundaries and support vectors \cite{molnar2020interpretable}. The support vectors are the critical data points that determine the position of the decision boundary, providing insights into how the model makes classifications.

\paragraph{Decision Boundaries and Support Vectors in SVM}
SVMs aim to find a hyperplane that best separates the data into different classes. The optimal hyperplane maximizes the margin, which is the distance between the hyperplane and the nearest data points from each class. These nearest points are known as \textbf{support vectors}, and they are fundamental to the SVM's decision-making process \cite{ShalevShwartz2014}.

The general equation of the decision boundary (hyperplane) is:

\[
\mathbf{w} \cdot \mathbf{x} + b = 0,
\]

where:
\begin{itemize}
    \item \(\mathbf{w}\) is the weight vector, determining the orientation of the hyperplane.
    \item \(b\) is the bias term, shifting the hyperplane.
    \item \(\mathbf{x}\) represents the feature vector.
\end{itemize}

The support vectors satisfy the condition:

\[
\mathbf{w} \cdot \mathbf{x_i} + b = \pm 1,
\]

where \(\mathbf{x_i}\) are the support vectors. These points lie exactly on the boundary of the margin.

\paragraph{Python Code Example}
Let's consider a simple example using a linear SVM classifier to separate two classes of data \cite{Pedregosa2011}.

\begin{lstlisting}[style=python, literate={\$}{{\$}}1]
import numpy as np
from sklearn import datasets
from sklearn.svm import SVC
import matplotlib.pyplot as plt

# Load a sample dataset with two features
X, y = datasets.make_classification(n_samples=100, n_features=2,
                                    n_informative=2, n_redundant=0, n_repeated=0,
                                    n_classes=2, n_clusters_per_class=1,
                                    random_state=42)

# Initialize and train a linear SVM classifier
clf = SVC(kernel='linear')
clf.fit(X, y)

# Extract the weight vector and bias term
w = clf.coef_[0]
b = clf.intercept_[0]

# Define the decision boundary
x_points = np.linspace(min(X[:, 0]), max(X[:, 0]), 100)
y_points = -(w[0] / w[1]) * x_points - b / w[1]

# Plot the data points and decision boundary
plt.scatter(X[:, 0], X[:, 1], c=y, cmap='coolwarm', edgecolors='k', label='Data Points')
plt.plot(x_points, y_points, color='red', label='Decision Boundary')

# Highlight the support vectors
plt.scatter(clf.support_vectors_[:, 0], clf.support_vectors_[:, 1],
            s=100, facecolors='none', edgecolors='k', linewidths=1.5,
            label='Support Vectors')

plt.xlabel('Feature 1')
plt.ylabel('Feature 2')
plt.title('SVM Decision Boundary with Support Vectors')
plt.legend()
plt.show()
\end{lstlisting}

\begin{figure}[htbp]
    \centering
    \includegraphics[width=0.7\textwidth]{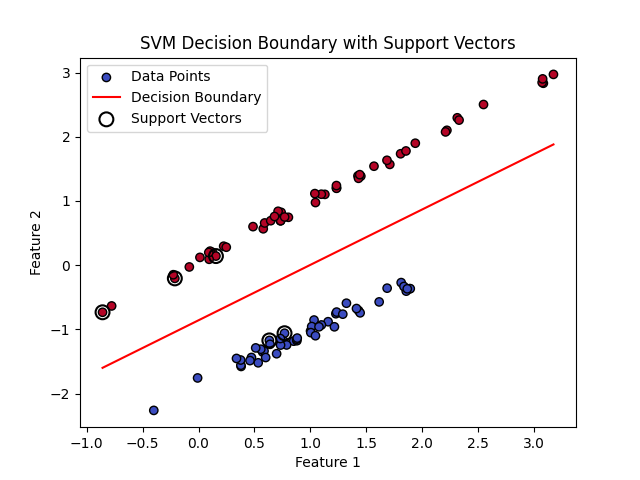}
    \caption{Visualization of SVM Decision Boundary and Support Vectors}
    \label{fig:svm_decision_boundary}
\end{figure}

\paragraph{Result Explanation}
In Figure \ref{fig:svm_decision_boundary}, we can observe the following key elements:

\begin{itemize}
    \item \textbf{Decision Boundary:} The red line represents the hyperplane separating the two classes. This line is determined by the weight vector \(\mathbf{w}\) and the bias term \(b\). The decision boundary divides the feature space into two regions, each corresponding to a different class label (blue and red points).
    \item \textbf{Support Vectors:} The support vectors are highlighted with larger, unfilled circles. These are the data points closest to the decision boundary and lie on the margin boundaries. They are critical in defining the margin and the orientation of the hyperplane. As shown in the figure, a few support vectors from both classes lie exactly on the margin.
    \item \textbf{Margin:} The margin is the region between the two parallel lines that pass through the support vectors. The SVM algorithm aims to maximize this margin, which improves the model's generalization ability. A larger margin indicates a more robust classifier, less sensitive to small variations in the data.
    \item \textbf{Class Separation:} The plot shows a clear separation between the two classes, represented by blue and red points. The linear decision boundary successfully divides the two classes, indicating that the data is linearly separable. This is expected given the use of a linear kernel and a well-defined dataset with two distinct clusters.
    \item \textbf{Edge Cases and Misclassifications:} In this particular plot, there are no visible misclassified points, as all data points are correctly separated by the decision boundary. However, in real-world scenarios, the presence of noise or overlapping data could result in some misclassifications, affecting the placement of the support vectors.
\end{itemize}

\paragraph{Key Observations}
\begin{enumerate}
    \item The support vectors play a crucial role in defining the decision boundary. Even if we remove other data points, the position of the boundary would remain unchanged as long as the support vectors are preserved.
    \item The decision boundary is linear, as expected from using a linear kernel. This simplicity makes the model interpretable and easy to understand.
    \item The model's performance would likely decrease if the data were not linearly separable. In such cases, using a non-linear kernel (e.g., RBF or polynomial) would help capture more complex patterns, though at the cost of reduced interpretability \cite{Murdoch2019}.
\end{enumerate}

\paragraph{Advantages and Disadvantages of SVM Interpretability}
\begin{itemize}
    \item \textbf{Advantages:}
    \begin{enumerate}
        \item \textbf{Clear Decision Boundary:} In the case of a linear kernel, the decision boundary is straightforward and interpretable, especially in low-dimensional spaces.
        \item \textbf{Influential Data Points:} By focusing on support vectors, we can identify the critical data points that the model relies on for classification.
    \end{enumerate}
    \item \textbf{Disadvantages:}
    \begin{enumerate}
        \item \textbf{Limited to Linear Kernels:} Interpretability is significantly reduced when using non-linear kernels (e.g., RBF), as the decision boundary becomes complex and difficult to visualize.
        \item \textbf{Sensitivity to Outliers:} The presence of outliers can drastically affect the support vectors, altering the decision boundary and potentially reducing the model's robustness.
    \end{enumerate}
\end{itemize}

\paragraph{Limitations and Considerations}
While SVMs provide some level of interpretability through their decision boundaries and support vectors, this is mainly applicable when using a linear kernel. For more complex datasets requiring non-linear decision boundaries, the interpretability diminishes as the kernel function introduces non-linear transformations. In such cases, post-hoc interpretability techniques, such as LIME or SHAP (discussed in later chapters), may be necessary to understand the model's predictions \cite{Ribeiro2016, Lundberg2017}.

\section{Rule-based Systems}

Rule-based systems are among the earliest forms of artificial intelligence, originating in the realm of expert systems \cite{Russell2010}. They consist of a set of human-defined rules that dictate the model's behavior. These rules are typically expressed in the form of logical IF-THEN statements, where the IF clause defines a condition and the THEN clause defines the action or outcome. Due to their explicit nature, rule-based systems are inherently interpretable, making them ideal for applications where transparency and traceability are essential, such as medical diagnosis, legal decision-making, and financial regulation \cite{Letham2015}.

\paragraph{Principle and Formulation}
The core of a rule-based system can be expressed mathematically as a series of logical rules:

\[
\text{IF } (x_1 \text{ meets condition}) \text{ AND } (x_2 \text{ meets condition}) \text{ THEN } y = \text{Outcome}
\]

Each rule can be seen as a Boolean function \( f(x) \) that outputs 1 (true) if the condition is satisfied and 0 (false) otherwise. Formally, a rule-based model \( R(x) \) can be expressed as:

\[
R(x) = \sum_{i=1}^{n} w_i \cdot f_i(x)
\]

where \( w_i \) is the weight assigned to rule \( i \), and \( f_i(x) \) represents the Boolean condition for rule \( i \). The outcome \( y \) is determined by evaluating the relevant rules for a given input \( x \) \cite{Piltaver2016}.

\paragraph{Interpretable Nature}
One of the key advantages of rule-based systems is their inherent interpretability. The decision-making process can be easily traced by examining which rules were triggered for a particular input. Unlike complex black-box models like deep neural networks, rule-based systems allow for clear, step-by-step reasoning, making them suitable for high-stakes domains where understanding the "why" behind a decision is critical \cite{Lipton2016}.

\paragraph{Python Code Example}

To demonstrate a simple rule-based system, let's consider a classic example of a medical diagnosis rule set. In this example, the system decides whether a patient has a common cold based on symptoms such as fever and cough.

\begin{lstlisting}[style=python, literate={\$}{{\$}}1]
def diagnose(symptoms):
    """
    Diagnoses a condition based on the provided symptoms using a simple rule-based system.

    Args:
        symptoms (dict): A dictionary where keys are symptom names (e.g., 'fever', 'cough') 
                         and values are booleans indicating whether the symptom is present.

    Returns:
        str: The diagnosis based on the provided symptoms.
    """
    if symptoms.get('fever') and symptoms.get('cough'):
        return "Common Cold"
    elif symptoms.get('fever'):
        return "Fever of unknown origin"
    elif symptoms.get('cough'):
        return "Possible respiratory infection"
    else:
        return "No specific diagnosis"

if __name__ == "__main__":
    # Collect symptoms from the user
    fever = input("Do you have a fever? (yes/no): ").strip().lower() == 'yes'
    cough = input("Do you have a cough? (yes/no): ").strip().lower() == 'yes'

    # Create a dictionary of symptoms
    symptoms = {'fever': fever, 'cough': cough}

    # Get the diagnosis
    diagnosis = diagnose(symptoms)

    # Display the diagnosis
    print(f"Diagnosis: {diagnosis}")
\end{lstlisting}

\paragraph{Result Explanation}
In this Python example, the function \texttt{diagnose()} implements a basic rule-based system using straightforward IF-ELSE logic to determine the diagnosis based on the patient's symptoms. The decision rules are simple and interpretable, making this approach highly transparent.

Let's break down the diagnosis rules:
\begin{itemize}
    \item \textbf{Rule 1:} If both \texttt{fever} and \texttt{cough} symptoms are present, the system concludes that the patient likely has a "Common Cold".
    \item \textbf{Rule 2:} If only \texttt{fever} is reported, it may indicate a "Fever of unknown origin", suggesting that further investigation might be necessary.
    \item \textbf{Rule 3:} If only \texttt{cough} is present, it implies a "Possible respiratory infection", which could range from mild to severe, depending on other symptoms not considered in this simple rule set.
    \item \textbf{Rule 4:} If neither symptom is present, the function returns "No specific diagnosis", indicating no immediate concerns based on the current rules.
\end{itemize}

This rule-based approach is limited to binary symptom inputs (i.e., presence or absence of symptoms), which keeps it simple but may overlook nuances in symptom severity or other important medical factors.

\begin{lstlisting}[style=cmd]
$ python diagnose.py
Do you have a fever? (yes/no): yes
Do you have a cough? (yes/no): yes
Diagnosis: Common Cold

$ python diagnose.py
Do you have a fever? (yes/no): yes
Do you have a cough? (yes/no): no
Diagnosis: Fever of unknown origin

$ python diagnose.py
Do you have a fever? (yes/no): no
Do you have a cough? (yes/no): yes
Diagnosis: Possible respiratory infection

$ python diagnose.py
Do you have a fever? (yes/no): no
Do you have a cough? (yes/no): no
Diagnosis: No specific diagnosis
\end{lstlisting}

This minimal example demonstrates the core principle of rule-based systems: explicit, interpretable rules provide direct, understandable decisions \cite{Kurgan2011}. However, the simplicity of this system also highlights its limitations:

\begin{enumerate}
    \item \textbf{Lack of Scalability:} As the number of symptoms and medical conditions increases, the rule set may become unwieldy and difficult to maintain \cite{Mitchell2018}.
    \item \textbf{Binary Symptom Representation:} The current implementation only accounts for the presence or absence of symptoms, ignoring severity or other medical nuances \cite{Shortliffe2014}.
    \item \textbf{Potential for Rule Conflicts:} In larger rule-based systems, conflicting rules could arise, requiring conflict resolution strategies such as rule prioritization or a certainty factor \cite{Russell2010}.
\end{enumerate}

\paragraph{Applications of Rule-based Systems}
Rule-based systems are widely used in domains where the decision logic needs to be explicit and understandable. For instance:

\begin{itemize}
    \item \textbf{Expert Systems in Medicine}: Providing diagnostic recommendations based on symptoms and patient history \cite{Durairaj2015}.
    \item \textbf{Legal Decision Making}: Applying legal rules to determine the outcome of a case based on the evidence presented \cite{BenchCapon2012}.
    \item \textbf{Financial Fraud Detection}: Using predefined rules to flag unusual transactions that may indicate fraudulent activity \cite{Ngai2011}.
\end{itemize}

\paragraph{Limitations and Challenges}
While rule-based systems are interpretable and easy to implement, they have several notable limitations:

\begin{itemize}
    \item \textbf{Scalability}: As the number of rules increases, the system becomes harder to manage and maintain \cite{Mitchell2018}.
    \item \textbf{Rule Conflicts}: Conflicting rules can lead to ambiguous outcomes, requiring additional logic for conflict resolution \cite{Kliegr2018}.
    \item \textbf{Limited Flexibility}: Rule-based systems struggle to generalize beyond the predefined rules, making them less effective in scenarios with complex or high-dimensional data \cite{Goodfellow2016}.
\end{itemize}

\section{Generalized Additive Models (GAMs)}

Generalized Additive Models (GAMs) offer a flexible yet interpretable approach to modeling complex relationships in data \cite{Hastie2017}. Proposed by Hastie and Tibshirani in the 1980s, GAMs extend traditional linear models by allowing non-linear relationships between each feature and the target variable while maintaining the additive structure \cite{Wood2017}. This balance between flexibility and interpretability makes GAMs particularly well-suited for tasks where we want to capture non-linear patterns without sacrificing transparency, such as in medical diagnostics, credit scoring, and risk assessment \cite{Lou2012}.

\paragraph{Principle and Formulation}
The key idea behind GAMs is to replace the linear terms in a regression model with smooth, non-linear functions. A GAM can be expressed mathematically as:

\[
g(\mathbb{E}[Y]) = \beta_0 + f_1(x_1) + f_2(x_2) + \dots + f_p(x_p)
\]

Here:

\begin{itemize}
    \item \( g(\cdot) \) is the link function (e.g., identity for linear regression, logit for logistic regression).
    \item \( \beta_0 \) is the intercept term.
    \item \( f_i(x_i) \) is a smooth function applied to feature \( x_i \), often modeled using splines or other non-parametric methods \cite{James2013}.
\end{itemize}

The additive nature of GAMs (\( f_1(x_1) + f_2(x_2) + \dots \)) ensures that the effect of each feature can be interpreted independently, which is a key advantage for explainability \cite{Hastie2017}.

\paragraph{Interpretable Nature of GAMs}
One of the greatest strengths of GAMs lies in their interpretability. Since the model assumes an additive relationship, each \( f_i(x_i) \) can be visualized individually as a partial dependence plot (PDP), showing how the predicted outcome changes with respect to a single feature while keeping others fixed. This feature-level interpretability allows domain experts to understand the model's behavior without being overwhelmed by interactions between variables \cite{Lou2012}.

\paragraph{Python Code Example}

Let's implement a simple GAM using the Python library \texttt{pyGAM} \cite{Serven2018} to predict a continuous target based on two features: \( x_1 \) and \( x_2 \). We will generate a synthetic dataset with non-linear relationships to illustrate the flexibility of GAMs.

\begin{lstlisting}[style=python, literate={\$}{{\$}}1]
import numpy as np
import matplotlib.pyplot as plt
from pygam import LinearGAM, s

# Generate synthetic data
np.random.seed(42)
x1 = np.random.uniform(-3, 3, 200)
x2 = np.random.uniform(-3, 3, 200)
y = np.sin(x1) + 0.5 * np.cos(x2) + np.random.normal(0, 0.2, 200)

# Combine features into a matrix
X = np.column_stack((x1, x2))

# Define and fit the GAM model
gam = LinearGAM(s(0) + s(1))
gam.fit(X, y)

# Plot the partial dependence for each feature
fig, axs = plt.subplots(1, 2, figsize=(12, 5))
for i, ax in enumerate(axs):
    XX = gam.generate_X_grid(term=i)
    ax.plot(XX[:, i], gam.partial_dependence(term=i, X=XX))
    ax.set_title(f'Partial Dependence of Feature x{i+1}')
    ax.set_xlabel(f'x{i+1}')
    ax.set_ylabel('Predicted y')

plt.tight_layout()
plt.show()
\end{lstlisting}

\paragraph{Result Explanation}

\begin{figure}[!ht]
    \centering
    \includegraphics[width=\textwidth]{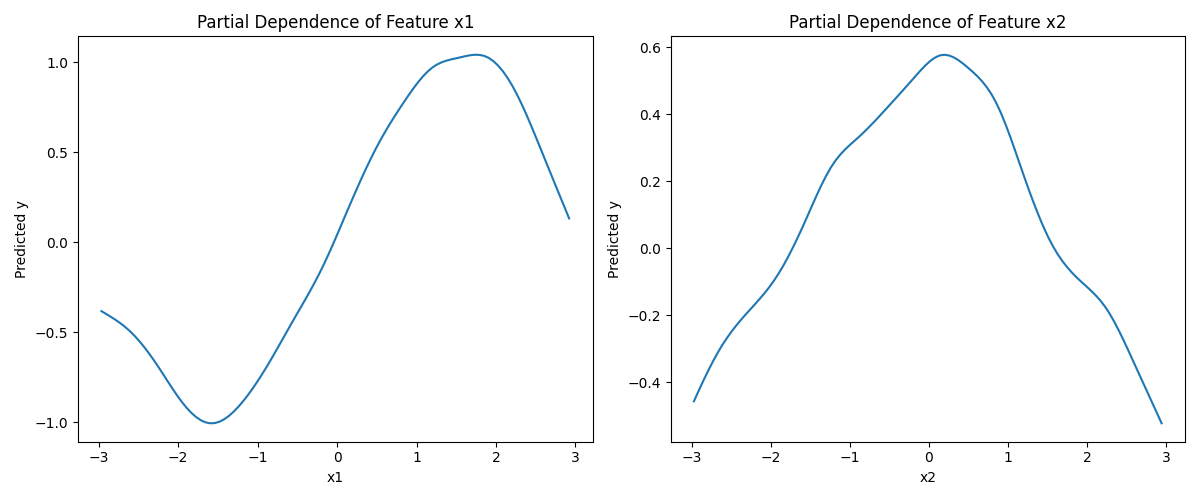}
    \caption{Partial Dependence Plots of Feature \(x_1\) and \(x_2\) in the GAM Model.}
    \label{fig:pdp}
\end{figure}

In this example, we first generated a synthetic dataset where the target variable \( y \) is defined as a non-linear combination of features \( x_1 \) and \( x_2 \), specifically \( y = \sin(x_1) + 0.5 \cos(x_2) + \varepsilon \), where \(\varepsilon\) is random noise sampled from a normal distribution with mean 0 and standard deviation 0.2. We then defined a Generalized Additive Model (GAM) using the \texttt{pyGAM} library, applying smooth functions (\texttt{s(0)} and \texttt{s(1)}) for each feature. After fitting the model, we visualized the partial dependence plots (PDPs) for both features.

The results consist of two plots, each representing the partial dependence of one feature:

\begin{itemize}
    \item \textbf{Partial Dependence of Feature \( x_1 \)}: The plot exhibits a clear sinusoidal pattern, capturing the relationship between \( x_1 \) and the target \( y \). This matches the data generation process, which included a \(\sin(x_1)\) component. It demonstrates that the GAM model effectively learned the non-linear relationship of \( x_1 \) with the response variable.
    \item \textbf{Partial Dependence of Feature \( x_2 \)}: This plot shows a cosine-like pattern, reflecting the \( 0.5 \cos(x_2) \) term in the target formula. The predicted outcome fluctuates with changes in \( x_2 \), revealing a typical cosine curve, which indicates that the model captured this feature's influence accurately.
\end{itemize}

\paragraph{Applications of GAMs}
GAMs are widely used in scenarios where model interpretability is crucial, including:

\begin{itemize}
    \item \textbf{Healthcare}: Modeling the effect of patient attributes (e.g., age, blood pressure) on health outcomes, providing clear, interpretable insights \cite{Caruana2015}.
    \item \textbf{Finance}: Assessing risk factors in credit scoring models, where regulatory requirements demand transparent decision-making \cite{Chen2018}.
    \item \textbf{Environmental Science}: Analyzing the impact of environmental variables (e.g., temperature, humidity) on ecological outcomes \cite{Pedersen2019}.
\end{itemize}

\paragraph{Limitations and Challenges}
Despite their advantages, GAMs have certain limitations:

\begin{itemize}
    \item \textbf{Lack of Interaction Modeling}: The additive assumption in GAMs does not account for interactions between features unless explicitly modeled \cite{Petersen2019}.
    \item \textbf{Complexity with High-dimensional Data}: As the number of features increases, it becomes difficult to fit and interpret each \( f_i(x_i) \) individually \cite{Lou2012}.
    \item \textbf{Choice of Smooth Function}: The selection of smoothing functions (e.g., splines) can significantly affect the model's performance and interpretability \cite{Wood2017}.
\end{itemize}

\section{Bayesian Models}

\paragraph{Introduction}
Bayesian models provide a probabilistic framework for machine learning, allowing us to incorporate prior knowledge and quantify uncertainty in model predictions \cite{Murphy2012}. Rooted in Bayes' Theorem, these models interpret data through the lens of probability, making them highly interpretable and transparent. Bayesian models are particularly well-suited for scenarios where understanding uncertainty is crucial, such as medical diagnosis, financial forecasting, and risk assessment \cite{Gelman2013}.

The core idea of Bayesian inference is to update our beliefs (prior knowledge) with observed data, resulting in a new, refined belief (posterior distribution). This approach not only improves prediction accuracy but also offers insights into the confidence of the predictions, enhancing model explainability \cite{Bishop2013}.

\paragraph{Principle and Formulation}
Bayesian inference relies on Bayes' Theorem, which relates the posterior probability of a model given the data to the likelihood of the data given the model and the prior probability of the model. Mathematically, it is expressed as:

\[
P(\theta \mid \mathbf{X}) = \frac{P(\mathbf{X} \mid \theta) \, P(\theta)}{P(\mathbf{X})}
\]

where:

\begin{itemize}
    \item \( P(\theta \mid \mathbf{X}) \) is the posterior distribution, representing the updated belief about the model parameters \( \theta \) after observing the data \( \mathbf{X} \).
    \item \( P(\mathbf{X} \mid \theta) \) is the likelihood, representing the probability of the observed data given the model parameters.
    \item \( P(\theta) \) is the prior distribution, representing our belief about the model parameters before observing the data.
    \item \( P(\mathbf{X}) \) is the marginal likelihood, acting as a normalizing constant.
\end{itemize}

The interpretability of Bayesian models arises from the explicit representation of uncertainty. By examining the posterior distribution, we gain insights into the confidence of parameter estimates, making Bayesian models naturally explainable \cite{Murphy2012}.

\paragraph{Python Code Example}
Let's implement a simple Bayesian linear regression model using Python. We will use the \texttt{TensorFlow Probability} library \cite{Dillon2017} to demonstrate Bayesian inference with a small dataset.

\begin{lstlisting}[style=python, literate={\$}{{\$}}1]
import tensorflow as tf
import tensorflow_probability as tfp
import numpy as np
import matplotlib.pyplot as plt

# Define the dataset
np.random.seed(42)
X = np.linspace(-5, 5, 100)
true_slope = 0.7
true_intercept = 1.5
y = true_slope * X + true_intercept + np.random.normal(0, 1, size=X.shape)

# Define the Bayesian linear regression model
tfd = tfp.distributions

# Define priors
prior_slope = tfd.Normal(loc=0., scale=1.)
prior_intercept = tfd.Normal(loc=0., scale=1.)
prior_sigma = tfd.HalfNormal(scale=1.)

# Define likelihood function
def likelihood(slope, intercept, sigma, X):
    mean = slope * X + intercept
    return tfd.Normal(loc=mean, scale=sigma)

# Sample from the posterior using Markov Chain Monte Carlo (MCMC)
@tf.function
def joint_log_prob(slope, intercept, sigma):
    lp = prior_slope.log_prob(slope) + prior_intercept.log_prob(intercept) + prior_sigma.log_prob(sigma)
    lp += tf.reduce_sum(likelihood(slope, intercept, sigma, X).log_prob(y))
    return lp

# Initialize MCMC sampler
initial_state = [0., 0., 1.]
num_results = 1000
kernel = tfp.mcmc.HamiltonianMonteCarlo(
    target_log_prob_fn=joint_log_prob,
    step_size=0.1,
    num_leapfrog_steps=3)

# Run MCMC
states, kernel_results = tfp.mcmc.sample_chain(
    num_results=num_results,
    current_state=initial_state,
    kernel=kernel,
    trace_fn=lambda _, pkr: pkr.is_accepted)

# Extract sampled parameters
slope_samples, intercept_samples, sigma_samples = states

# Plot the posterior distributions
fig, axs = plt.subplots(1, 3, figsize=(15, 5))
axs[0].hist(slope_samples, bins=30, color='skyblue', edgecolor='black')
axs[0].set_title('Posterior of Slope')
axs[1].hist(intercept_samples, bins=30, color='skyblue', edgecolor='black')
axs[1].set_title('Posterior of Intercept')
axs[2].hist(sigma_samples, bins=30, color='skyblue', edgecolor='black')
axs[2].set_title('Posterior of Sigma')

plt.show()
\end{lstlisting}

\paragraph{Result Explanation}
In this example, we performed Bayesian linear regression using TensorFlow Probability. The model assumes normal priors for both the slope and intercept, and a half-normal prior for the noise parameter (\(\sigma\)). We employed the Hamiltonian Monte Carlo (HMC) method to draw samples from the posterior distribution.

The output includes histograms for the posterior distributions of the slope, intercept, and noise parameter. These histograms provide key insights into the model's parameter estimates:

\begin{itemize}
    \item \textbf{Posterior of Slope}: The distribution of the slope parameter indicates a central value close to the true slope (\(0.7\)), with the spread reflecting the uncertainty around this estimate. The narrower the distribution, the higher the confidence in the slope estimate.
    \item \textbf{Posterior of Intercept}: The intercept's posterior distribution centers around the true intercept value (\(1.5\)), demonstrating that the model has successfully learned this parameter from the data. The shape of the distribution conveys the level of certainty in this estimate.
    \item \textbf{Posterior of Sigma}: The noise parameter (\(\sigma\)) shows a tight distribution around its estimated value, suggesting that the model has high confidence in its noise level estimate. A narrow posterior for \(\sigma\) indicates low variance in the noise of the observations.
\end{itemize}

\paragraph{Further Analysis}
This example illustrates the power of Bayesian inference: the posterior distributions provide a comprehensive picture of the uncertainty surrounding each parameter estimate. Instead of single-point estimates, Bayesian models offer a probabilistic view, making it easier to understand the confidence we have in the learned parameters \cite{Bishop2013}.

Moreover, Bayesian models naturally integrate prior knowledge, allowing for more robust predictions when data is limited. In our example, the priors were set with standard normal distributions, but these can be adjusted based on domain expertise to reflect more informative prior beliefs \cite{Murphy2012}.

\begin{figure}[!ht]
    \centering
    \includegraphics[width=\textwidth]{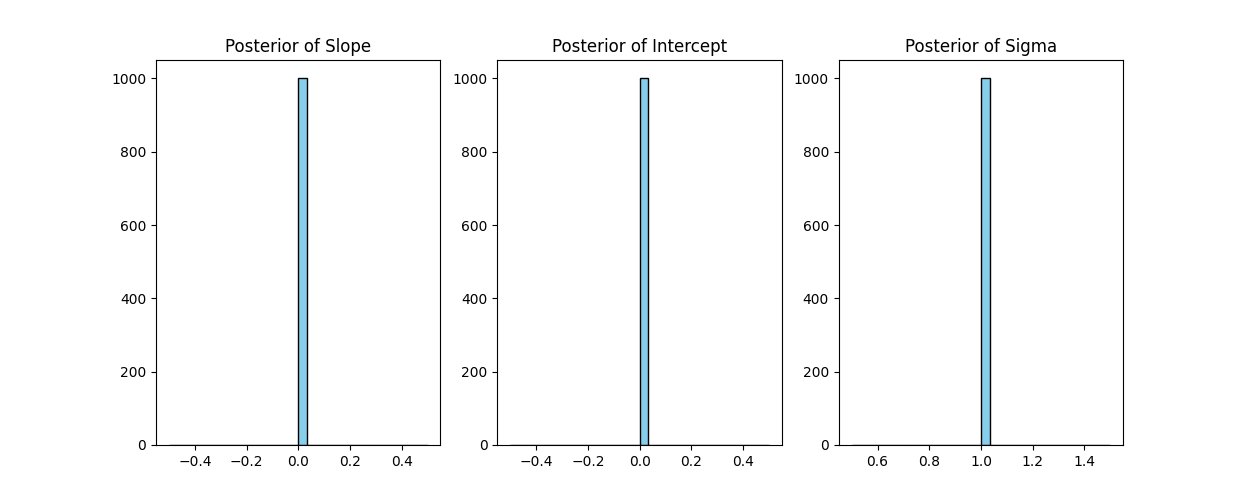}
    \caption{Posterior Distributions of Slope, Intercept, and Sigma in the Bayesian Linear Regression Model.}
    \label{fig:bayesian_posterior}
\end{figure}

This visual representation of the posterior distributions demonstrates the interpretability of Bayesian models, where the uncertainty of each parameter is explicitly captured. Unlike traditional point estimates, these distributions allow us to quantify and visualize the confidence in our model's predictions. This can be especially beneficial in fields such as finance, healthcare, and scientific research, where understanding uncertainty is critical for decision-making \cite{Gelman2013}.

\paragraph{Applications of Bayesian Models}
Bayesian models are widely used in various fields where uncertainty quantification is critical:

\begin{itemize}
    \item \textbf{Medical Diagnosis}: Estimating the probability of diseases given patient symptoms while accounting for uncertainty \cite{Brooks2011}.
    \item \textbf{Financial Forecasting}: Predicting stock prices and market trends with a probabilistic approach \cite{Avramov2010}.
    \item \textbf{A/B Testing}: Analyzing experiment results with credible intervals rather than p-values, providing more interpretable insights \cite{Kohavi2017}.
\end{itemize}

\paragraph{Limitations and Challenges}
Despite their strengths, Bayesian models also face challenges:

\begin{itemize}
    \item \textbf{Computational Complexity}: Sampling from the posterior can be computationally expensive, especially for high-dimensional data \cite{Blei2017}.
    \item \textbf{Choice of Priors}: The selection of appropriate priors can be subjective and may influence the results \cite{Gelman2013}.
    \item \textbf{Scalability}: Bayesian inference may not scale well with large datasets or complex models \cite{Broderick2013}.
\end{itemize}

\section*{Conclusion}
In this chapter, we examined the interpretability of traditional machine learning models, from the transparent logic of decision trees and the straightforward coefficients of linear models, to the geometric insights provided by support vector machines (SVMs). While these models offer varying degrees of interpretability, their simplicity also limits their ability to capture complex, non-linear relationships in data. This trade-off underscores the core challenge in machine learning: balancing interpretability with predictive power. As we transition to deep learning models in the next chapter, we face a new level of complexity where traditional interpretability methods fall short. Here, understanding the intricate workings of neural networks requires advanced techniques, setting the stage for a deeper exploration into the interpretability of CNNs, RNNs, and Transformer-based architectures \cite{Vaswani2017}.

\chapter{Interpretability of Deep Learning Models}
\label{sec:DeepModels}

\section{Why Are Deep Learning Models Hard to Interpret?}

Deep learning models, especially those based on deep neural networks, are known for their powerful predictive capabilities. However, they are often regarded as 'black boxes.' But why is that the case? The challenge of interpretability arises due to:

\subsection{High Complexity of the Model}

Deep learning models, such as Convolutional Neural Networks (CNNs) and Recurrent Neural Networks (RNNs), involve multiple layers of neurons, non-linear activation functions, and vast numbers of parameters \cite{LeCun2015, Goodfellow2016}. For example, a simple CNN designed for image classification might already contain millions of parameters. As the depth and complexity of the network increase, understanding the contribution of each individual parameter becomes infeasible.

\subsection{Non-linearity and Feature Abstraction}

The non-linear activation functions, such as ReLU and sigmoid, enable the model to learn complex patterns. However, this non-linearity makes it difficult to interpret what each layer is learning. In early layers, the network may learn simple features like edges or textures, but as we go deeper, the layers start abstracting more complex patterns \cite{Zeiler2014}. The representations in these deep layers are often not directly interpretable by humans.

\subsection{Lack of Explicit Structure}

Unlike simpler models (e.g., decision trees), deep neural networks do not have an inherent hierarchical structure that is easily understandable. While a decision tree provides a clear set of rules for decision-making, deep neural networks provide predictions based on complex, distributed representations of the input data, which are difficult to decompose into human-readable rules \cite{molnar2020interpretable}.

\subsection{The Curse of Dimensionality}

The curse of dimensionality refers to the exponential increase in data space as the number of input features grows. In deep learning, high-dimensional data is processed through layers that may reduce or increase this dimensionality, making it challenging to map input features directly to the learned representations. This abstraction hinders our ability to directly interpret the learned features \cite{Bengio2015}.

\section{Interpretability of Convolutional Neural Networks (CNNs)}

Convolutional Neural Networks (CNNs) have become the cornerstone of computer vision tasks due to their ability to automatically learn spatial hierarchies of features from raw image data \cite{Krizhevsky2012}. However, this strength also poses a challenge: understanding the inner workings of CNNs and deciphering why they make certain predictions can be complex. In this section, we will explore techniques for interpreting CNNs, focusing on the concept of feature visualization, which provides a window into what each convolutional layer is learning \cite{Zeiler2014}.

\subsection{Feature Visualization in Convolutional Layers}

CNNs extract features from input images through a series of convolutional and pooling layers. Early layers typically capture simple patterns like edges and textures, while deeper layers learn more abstract, high-level representations such as object parts. One of the most intuitive ways to interpret CNNs is by visualizing these learned features \cite{Zeiler2014, Yosinski2015}. Feature visualization involves inspecting the \textbf{feature maps} generated by the convolutional filters, giving us insight into which parts of the input image activate specific filters.

\subsubsection{Python Code Example}

\begin{figure}[htbp]
    \centering
    \includegraphics[width=0.33\textwidth]{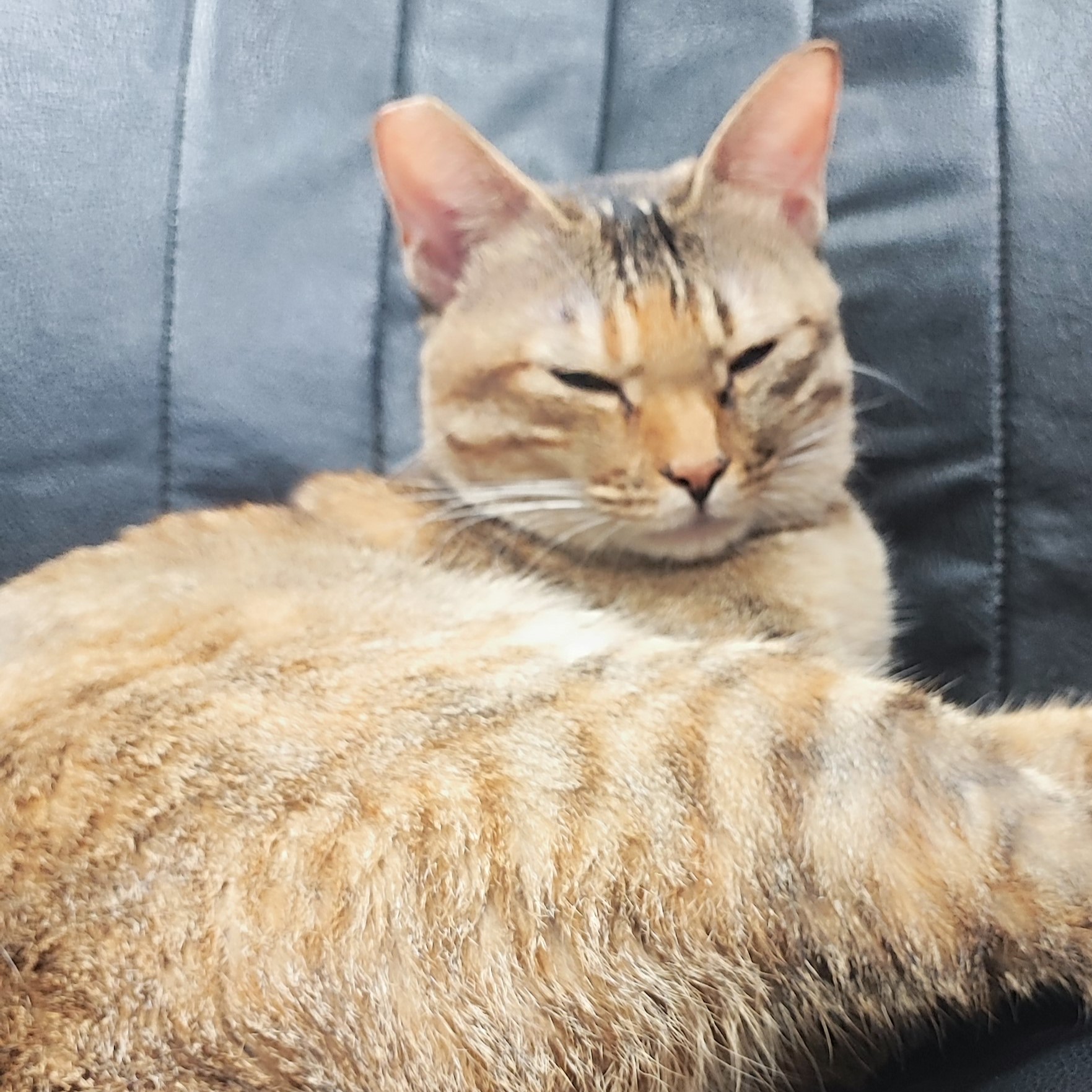}
    \caption{Original Cat Image}
    \label{fig:cat_image}
\end{figure}

In the following example, we use a pre-trained VGG16 model, a popular CNN architecture known for its strong performance on image classification tasks \cite{Simonyan2014}. We will:
\begin{enumerate}
    \item Load a sample image of a cat (\ref{fig:cat_image}).
    \item Preprocess the image for input to the VGG16 model.
    \item Extract and visualize the feature maps from the first convolutional layer.
\end{enumerate}

This process will help us observe the low-level features detected by the early layers of the network.

\begin{lstlisting}[style=python, literate={\$}{{\$}}1]

import tensorflow as tf
import matplotlib.pyplot as plt
import numpy as np

# Check TensorFlow version and GPU availability
print("TensorFlow version:", tf.__version__)
print("GPU is", "available" if tf.config.list_physical_devices('GPU') else "not available")

# Load a pre-trained VGG16 model (without the fully connected layers)
model = tf.keras.applications.VGG16(weights='imagenet', include_top=False)

# Load and preprocess the input image
image_path = 'Ch04/cat.jpg'  # Ensure 'cat.jpg' is in the directory
try:
    image = tf.keras.preprocessing.image.load_img(image_path, target_size=(224, 224))
except FileNotFoundError:
    print(f"Error: Image file '{image_path}' not found.")
    exit()

image_array = tf.keras.preprocessing.image.img_to_array(image)
image_array = np.expand_dims(image_array, axis=0)
image_array = tf.keras.applications.vgg16.preprocess_input(image_array)

# Define a model that outputs the feature maps of the first convolutional layer
layer_name = 'block1_conv1'
feature_map_model = tf.keras.Model(inputs=model.input, outputs=model.get_layer(layer_name).output)

# Generate the feature maps for the input image
feature_maps = feature_map_model.predict(image_array)

# Check the shape of the feature maps
print("Feature map shape:", feature_maps.shape)

# Visualize the first 16 feature maps
fig, axes = plt.subplots(4, 4, figsize=(10, 10))
for i, ax in enumerate(axes.flat):
    if i < feature_maps.shape[-1]:
        ax.imshow(feature_maps[0, :, :, i], cmap='viridis')
    ax.axis('off')
plt.tight_layout()
plt.show()
\end{lstlisting}

\subsubsection{Result Explanation}

In Figure \ref{fig:vgg16_feature_maps}, we visualize the feature maps generated by the first convolutional layer (`block1\_conv1`) of the VGG16 model. The visualization provides several insights into the network's behavior at this early stage of feature extraction:

\begin{figure}[htbp]
    \centering
    \includegraphics[width=0.8\textwidth]{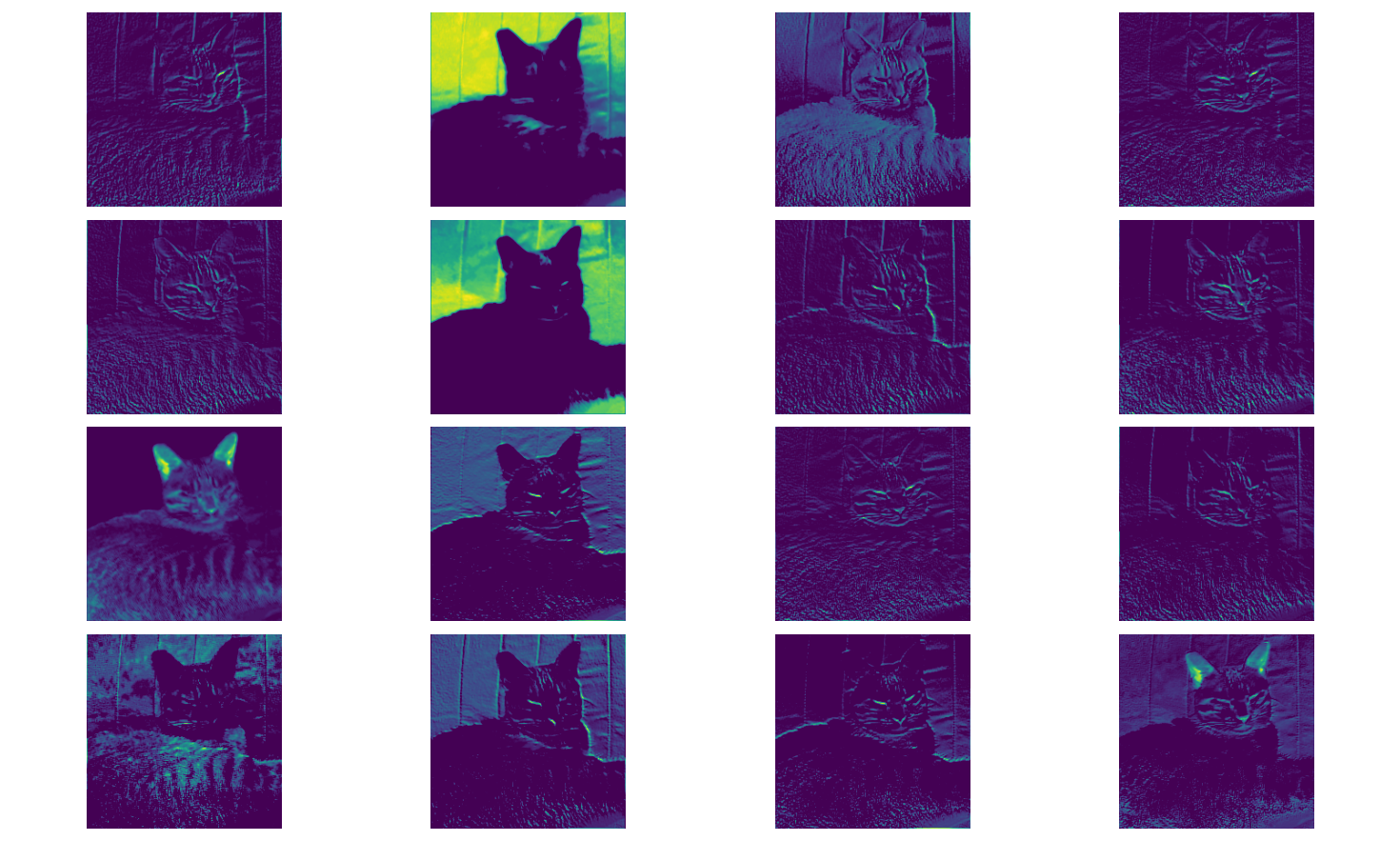}
    \caption{Feature Maps from the First Convolutional Layer of VGG16}
    \label{fig:vgg16_feature_maps}
\end{figure}

\begin{itemize}
    \item \textbf{Low-level Feature Detection:} The feature maps predominantly highlight edges, corners, and simple textures. In the visualizations, we can observe how the network focuses on the contours of the cat's face, detecting prominent edges such as the outline of the ears and the whiskers. This stage captures the low-level features that serve as building blocks for more complex patterns in deeper layers \cite{Zeiler2014}.
    
    \item \textbf{Filter Activation Patterns:} Each feature map corresponds to a different convolutional filter in the first layer. The intensity and patterns shown in the feature maps indicate which parts of the input image activate specific filters. For example, some filters emphasize vertical or horizontal edges, as evident in the strong activations around the cat's ears and fur lines. This behavior aligns with the intuition that early convolutional filters act as edge detectors \cite{Zeiler2014}.
    
    \item \textbf{Redundant Features and Diverse Activations:} Although many feature maps appear visually similar, showing similar patterns and edges, some feature maps highlight different aspects of the image. This redundancy is intentional, as multiple filters may focus on similar but slightly different features, providing robustness in the learned representation. For instance, two feature maps might both detect edges but respond differently to variations in texture or lighting.
    
    \item \textbf{Limited Abstraction:} Since these feature maps are from the first convolutional layer, they exhibit limited abstraction and primarily capture low-level features. As we progress deeper into the network (not covered in this example), the features become more abstract, capturing shapes, textures, and eventually object parts. This hierarchical feature learning is a key characteristic of CNNs, enabling them to effectively handle complex visual tasks \cite{LeCun2015}.
\end{itemize}

\subsubsection{Key Observations and Next Steps}

The feature maps provide a visual glimpse into the model's inner workings, offering clues about how it perceives the input image at different stages of the network. While this type of visualization helps in understanding the network's early layers, it becomes increasingly challenging to interpret feature maps from deeper layers due to the complexity and abstraction of the learned features.

\subsection{Challenges in Interpreting Feature Maps}

While feature visualization provides valuable insights, it is not without limitations:
\begin{enumerate}
    \item \textbf{Lack of Direct Interpretability:} Not all feature maps correspond to human-recognizable patterns. Many filters may detect abstract features that are difficult to interpret visually.
    \item \textbf{Dependence on Input Data:} The visualized features depend heavily on the input image. Different images may activate different filters, making it challenging to generalize the interpretations across various inputs.
    \item \textbf{Layer Complexity:} As we move to deeper layers, the complexity of the feature maps increases, making it harder to identify the exact features being captured. Techniques like activation maximization or saliency maps may be required to further understand these deeper representations \cite{Simonyan2013, Sundararajan2017}.
\end{enumerate}

\subsection{Transition to Advanced Interpretability Techniques}

Feature visualization offers an initial look into the inner workings of CNNs, helping us understand how they detect patterns in images. However, it is often insufficient for interpreting the decisions of deeper and more complex layers. To gain deeper insights, more sophisticated interpretability methods, such as \textbf{Grad-CAM (Gradient-weighted Class Activation Mapping)} \cite{Selvaraju2017}, are employed(further introduce in chapter \ref{sec:Techniques}). Grad-CAM provides a high-level, intuitive heatmap that highlights the regions in the input image most relevant to the model's predictions.

\section{Interpretability of Recurrent Neural Networks (RNNs)}

Recurrent Neural Networks (RNNs) are designed to handle sequential data, making them well-suited for applications like time series analysis, natural language processing, and speech recognition. Their strength lies in the ability to maintain \textbf{hidden states} across time steps, allowing the model to capture temporal dependencies. However, this temporal memory also poses challenges for interpretability, as the hidden states can be difficult to decipher and track through multiple time steps \cite{Lipton2015}.

\subsection{Temporal Dependencies and Hidden State Interpretations}

The core of an RNN's capability lies in its hidden states, which evolve with each time step. These hidden states act as memory units, storing information about previous inputs in the sequence. However, interpreting the information encoded in the hidden states is challenging because they represent a complex, nonlinear combination of past inputs \cite{Karpathy2015}.

To gain insights into the hidden states, one common approach is to visualize how they change over time. For example, plotting the activations of hidden states for different time steps can reveal patterns, such as increasing attention or sensitivity to certain parts of the input sequence.

\subsubsection{Python Code Example}

In this example, we demonstrate the ability of a simple Recurrent Neural Network (RNN) to model a synthetic time series, specifically a sine wave. We aim to visualize both the predicted output of the RNN and the activations of its hidden states over time. This approach provides a clear view of how the RNN processes temporal data and can be considered a form of \textbf{post-hoc interpretability}, which will be further elaborated in Chapter \ref{sec:Techniques}.

\begin{lstlisting}[style=python, literate={\$}{{\$}}1]
import tensorflow as tf
import matplotlib.pyplot as plt
import numpy as np

# Check TensorFlow version and GPU availability
print("TensorFlow version:", tf.__version__)
print("GPU is", "available" if tf.config.list_physical_devices('GPU') else "not available")

# Generate a synthetic sine wave dataset
time_steps = 100
X = np.sin(np.linspace(0, 20, time_steps))
X = X.reshape((1, time_steps, 1))  # Reshape for RNN input (batch_size, time_steps, features)

# Build a simple RNN model with 10 hidden units
model = tf.keras.Sequential([
    tf.keras.layers.SimpleRNN(units=10, return_sequences=True, input_shape=(time_steps, 1)),
    tf.keras.layers.Dense(1)  # Add a Dense layer for prediction
])

# Compile the model
model.compile(optimizer='adam', loss='mse')

# Run inference using the sine wave data
y_pred = model.predict(X)
hidden_states = model.predict(X, verbose=0)

# Plot the original sine wave data
plt.figure(figsize=(12, 6))
plt.plot(X[0, :, 0], label='Original Sine Wave', color='gray', linestyle='--', linewidth=2)
plt.title("Original Sine Wave Data")
plt.xlabel("Time Step")
plt.ylabel("Value")
plt.grid()
plt.legend()
plt.show()

# Plot the predicted sine wave vs the original sine wave
plt.figure(figsize=(12, 6))
plt.plot(X[0, :, 0], label='Original Sine Wave', color='lightgray', linestyle='--', linewidth=2)
plt.plot(y_pred[0, :, 0], label='Predicted Sine Wave', color='blue', linewidth=2)
plt.title("Comparison of Original and Predicted Sine Wave")
plt.xlabel("Time Step")
plt.ylabel("Value")
plt.legend()
plt.grid()
plt.show()

# Plot the activations of all 10 hidden units over time
plt.figure(figsize=(14, 8))
for i in range(hidden_states.shape[-1]):
    plt.plot(hidden_states[0, :, i], label=f'Hidden Unit {i+1}', alpha=0.8)

plt.title("Activations of All Hidden Units Over Time")
plt.xlabel("Time Step")
plt.ylabel("Hidden State Activation")
plt.legend(loc='upper right', ncol=2, fontsize=10)
plt.grid()
plt.show()
\end{lstlisting}

\begin{figure}[htbp]
    \centering
    \includegraphics[width=0.8\textwidth]{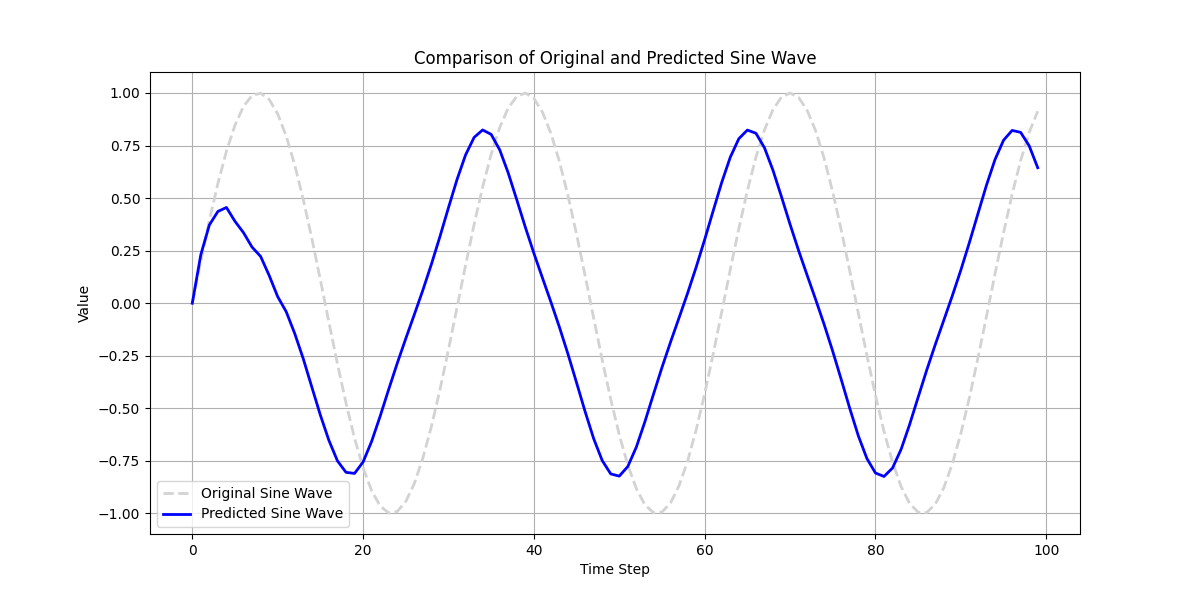}
    \caption{Comparison of Original and Predicted Sine Wave}
    \label{fig:predicted_sine_wave}
\end{figure}

\begin{figure}[htbp]
    \centering
    \includegraphics[width=0.8\textwidth]{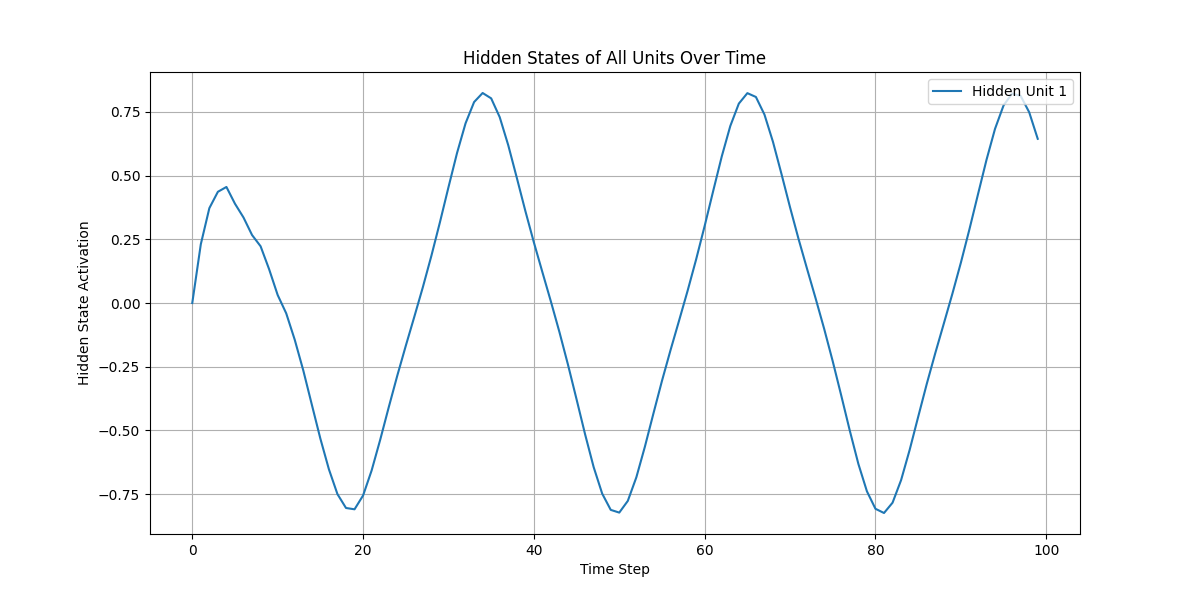}
    \caption{Activations of All Hidden Units Over Time}
    \label{fig:hidden_states}
\end{figure}

\subsubsection{Result Explanation}

In Figure \ref{fig:predicted_sine_wave}, we see the comparison between the original sine wave (gray dashed line) and the predicted sine wave (solid blue line). The RNN successfully captures the general pattern of the sine wave, demonstrating its ability to model temporal dependencies.

In Figure \ref{fig:hidden_states}, we observe the activations of all 10 hidden units over time. Each line represents the activation of a specific hidden unit at each time step. The variations in these activations reflect how different units in the RNN respond to different parts of the input sequence. Some units show strong periodic activations, aligning well with the sine wave pattern, while others show subtler responses.

These visualizations help us understand the internal workings of the RNN. By examining the hidden states, we gain insights into which parts of the sequence the RNN is focusing on, making this approach a useful tool for interpretability in sequential models \cite{Karpathy2015}.

\subsection{Enhancing RNN Interpretability with Attention Mechanisms}

While visualizing hidden states provides some interpretability, it can be challenging to pinpoint which parts of the input sequence are most influential in the model's decision-making. The \textbf{attention mechanism} addresses this issue by explicitly assigning weights to different time steps, indicating the importance of each input in the context of the current prediction \cite{Bahdanau2015}.

The attention mechanism operates by computing a set of weights, known as \( \alpha_t \), for each time step \( t \). These weights are used to form a weighted sum of the hidden states, effectively focusing on the most relevant parts of the sequence.

\subsubsection{Illustration of Attention Mechanism}

The diagram below illustrates the concept of the attention mechanism in an RNN. Each hidden state is assigned an attention weight, which is then used to aggregate the information and generate the output.

\begin{center}
\begin{tikzpicture}[scale=0.9, every node/.style={scale=0.9}]
    \node (x1) at (0, 0) {Input $x_1$};
    \node (x2) at (0, -1) {Input $x_2$};
    \node (x3) at (0, -2) {Input $x_3$};

    \node (h1) at (2, 0) {Hidden $h_1$};
    \node (h2) at (2, -1) {Hidden $h_2$};
    \node (h3) at (2, -2) {Hidden $h_3$};

    \node[red] (a1) at (4, 0) {Weight $\alpha_1$};
    \node[red] (a2) at (4, -1) {Weight $\alpha_2$};
    \node[red] (a3) at (4, -2) {Weight $\alpha_3$};

    \node (output) at (6, -1) {Output $y$};

    \draw[->, thick] (x1) -- (h1);
    \draw[->, thick] (x2) -- (h2);
    \draw[->, thick] (x3) -- (h3);

    \draw[->, thick, dashed, red] (h1) -- (a1);
    \draw[->, thick, dashed, red] (h2) -- (a2);
    \draw[->, thick, dashed, red] (h3) -- (a3);

    \draw[->, thick, red] (a1) -- (output);
    \draw[->, thick, red] (a2) -- (output);
    \draw[->, thick, red] (a3) -- (output);

    \node at (2, 1) {Hidden States};
    \node[red] at (4.5, -3) {Attention Weights $\alpha_1, \alpha_2, \alpha_3$ sum to 1};
\end{tikzpicture}
\end{center}

\subsubsection{Benefits of Attention Mechanism}

The attention mechanism enhances interpretability by:
\begin{itemize}
    \item \textbf{Highlighting Important Inputs:} The attention weights indicate which parts of the input sequence the model focuses on when making predictions. This allows us to trace back the model's decision to specific time steps.
    \item \textbf{Improving Transparency:} By visualizing the attention weights, we can gain a better understanding of the model's reasoning process, making it easier to diagnose errors or biases.
    \item \textbf{Facilitating Model Debugging:} Attention visualizations can help identify issues in model training, such as when the model consistently focuses on irrelevant parts of the input sequence.
\end{itemize}

\section{Self-Attention Mechanism and Interpretability in Transformer Models}

The self-attention mechanism is the core innovation of Transformer models, making them highly effective for processing sequences of data, such as natural language text \cite{Vaswani2017}. Unlike traditional RNNs, which process data sequentially and suffer from long-term dependency issues, Transformers use self-attention to capture global dependencies by computing attention scores between all pairs of tokens in the input sequence. This capability is one of the main reasons why Transformer-based models like BERT, GPT, and T5 have become the backbone of modern NLP \cite{Devlin2018, Radford2018, Raffel2019}.

\subsection{How Self-Attention Works}

The self-attention mechanism allows the model to weigh the importance of each input token relative to every other token. The key components involved in this mechanism are the query, key, and value vectors, which are computed for each token. The attention score between tokens \( i \) and \( j \) is derived using the dot product of their corresponding query and key vectors. The formula for the scaled dot-product attention is as follows \cite{Vaswani2017}:

\begin{equation}
\text{Attention}(Q, K, V) = \text{softmax}\left(\frac{QK^T}{\sqrt{d_k}}\right)V,
\end{equation}

where:
\begin{itemize}
    \item \( Q \) is the query matrix.
    \item \( K \) is the key matrix.
    \item \( V \) is the value matrix.
    \item \( d_k \) is the dimension of the key vectors.
\end{itemize}

The softmax function normalizes the scores, converting them into probabilities that sum up to 1. This ensures that the model's focus is distributed across the input tokens, with higher attention scores indicating greater focus on specific tokens.

\subsection{Python Code Example}

One effective way to interpret the self-attention mechanism is by visualizing the attention weights as a heatmap. The attention weights reveal the model's focus during each step of the prediction process. For example, in a machine translation task, attention weights often highlight the alignment between words in the source and target languages, providing insights into how the model is mapping input tokens to output tokens \cite{Bahdanau2015}.

The following Python code snippet demonstrates how to create a heatmap for visualizing a sample set of attention weights:

\begin{lstlisting}[style=python, literate={\$}{{\$}}1]

import matplotlib.pyplot as plt
import seaborn as sns
import numpy as np

# Define a sample attention weights matrix (3x3 for simplicity)
attention_weights = np.array([[0.1, 0.2, 0.7], 
                              [0.5, 0.3, 0.2], 
                              [0.3, 0.4, 0.3]])

# Create the heatmap plot
plt.figure(figsize=(6, 5))
sns.heatmap(attention_weights, annot=True, fmt=".2f", cmap='Blues', cbar=False, linewidths=0.5)

# Add titles and labels
plt.title("Sample Attention Heatmap", fontsize=14)
plt.xlabel("Input Tokens", fontsize=12)
plt.ylabel("Output Tokens", fontsize=12)
plt.xticks(ticks=[0.5, 1.5, 2.5], labels=["Token 1", "Token 2", "Token 3"])
plt.yticks(ticks=[0.5, 1.5, 2.5], labels=["Output 1", "Output 2", "Output 3"])

# Display the plot
plt.tight_layout()
plt.show()
\end{lstlisting}

In this example, the heatmap visualizes a small matrix of attention weights, where each cell represents the attention score between an input and an output token. The intensity of the color in each cell indicates the strength of the attention score, making it easier to identify which input tokens are most influential for the output tokens.

\begin{figure}[htbp]
    \centering
    \includegraphics[width=0.6\textwidth]{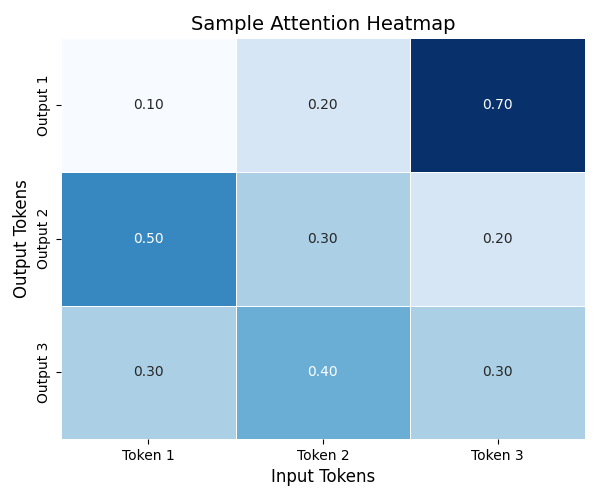}
    \caption{Sample Attention Heatmap}
    \label{fig:attention_heatmap}
\end{figure}

\subsection{Result Explanation}

In Figure \ref{fig:attention_heatmap}, we see a heatmap of attention weights from a sample self-attention layer. The x-axis represents the input tokens, and the y-axis represents the output tokens. Higher attention weights (darker blue) indicate a stronger focus on certain input tokens when generating the corresponding output token. For instance, if the input sequence contains the phrase "The cat sat," the model might place higher attention on "cat" when predicting a related word in the output.

This visualization provides a transparent view of the model's internal decision-making process, helping us understand how the model attends to different parts of the input sequence. By analyzing these attention patterns, we can gain insights into the relationships between tokens and the model's understanding of the context \cite{Vig2019}.

\subsection{Multi-Head Attention: A Deeper Look}

To further enhance interpretability, Transformer models employ \textbf{multi-head attention}, where multiple self-attention mechanisms (heads) operate in parallel. Each head learns different aspects of the input, capturing diverse patterns and relationships. The outputs from all heads are then concatenated and linearly transformed to produce the final output. This multi-head approach allows the model to capture information from different subspaces, improving both performance and interpretability \cite{Vaswani2017}.

The multi-head attention mechanism can be mathematically expressed as:

\begin{equation}
\text{MultiHead}(Q, K, V) = \text{Concat}(\text{head}_1, \text{head}_2, \ldots, \text{head}_h)W^O,
\end{equation}

where:
\begin{itemize}
    \item \( \text{head}_i = \text{Attention}(QW_i^Q, KW_i^K, VW_i^V) \)
    \item \( W_i^Q, W_i^K, W_i^V \) are learned projection matrices for each head.
    \item \( W^O \) is the output projection matrix.
\end{itemize}

By visualizing the attention weights across multiple heads, we can observe how each head focuses on different parts of the input, providing a richer interpretation of the model's behavior.

\section*{Conclusion}

In conclusion, this chapter highlighted the interpretability challenges inherent to deep learning models, shedding light on why these models are often viewed as "black boxes." We explored key techniques and methods for understanding deep neural networks, from visualizing feature maps in CNNs to interpreting hidden states and attention mechanisms in RNNs and Transformer models. Despite these advances, achieving full interpretability remains elusive due to the complex and non-linear nature of deep learning architectures. As we move into the next chapter, our focus will shift towards interpreting the even more intricate and expansive models that have revolutionized the field—Large Language Models (LLMs). We will delve deeper into the specific interpretability techniques required for these models and the unique challenges they present, especially when dealing with natural language tasks and real-world applications.

\chapter{Interpretability of Large Language Models (LLMs)}
\label{sec:LLMs}

\section{Introduction to Large Language Models}

Large Language Models (LLMs) are a transformative class of deep learning models designed to understand, generate, and process human language. Using vast amounts of training data and the Transformer architecture \cite{Vaswani2017}, these models have revolutionized natural language processing (NLP), achieving state-of-the-art performance across a wide range of tasks, such as text classification, translation, summarization, dialogue systems, and even code generation \cite{Brown2020}. Popular examples include BERT \cite{Devlin2018}, GPT \cite{Radford2018, Brown2020}, and T5 \cite{Raffel2019}, which have set new benchmarks in NLP \cite{niu2024largelanguagemodelscognitive}.

\subsection{Why LLMs Matter}

The impact of LLMs extends beyond just NLP tasks. By understanding context, semantics, and user intent, LLMs have enabled applications such as:

\begin{itemize}
    \item \textbf{Customer Support}: Automated systems that respond accurately to user queries, enhancing the user experience \cite{Zhou2020}.
    \item \textbf{Content Creation}: Tools that assist writers by generating coherent text based on simple prompts \cite{Keskar2019}.
    \item \textbf{Code Assistance}: Language models like GitHub Copilot that help developers by suggesting code snippets in real time \cite{Copilot2021}.
    \item \textbf{Medical Applications}: Supporting doctors with preliminary text analysis of patient records, aiding in diagnostics \cite{Lee2020}.
\end{itemize}

However, the success of LLMs comes with challenges, particularly around interpretability, transparency, and trust.

\section{Evolution of LLMs (BERT, GPT, T5, etc.)}

The evolution of Large Language Models (LLMs) marks a transformative shift in the design and training of language models, driven by advances in deep learning architectures and training methodologies. This section explores the major milestones in the development of LLM, highlighting key innovations and their impact on natural language processing (NLP).

\begin{itemize}
    \item \textbf{BERT (2018)}: BERT, or Bidirectional Encoder Representations of Transformers, was introduced by Google and represents a significant leap forward in NLP \cite{Devlin2018}. Unlike previous models that processed text in a single direction, BERT uses a bidirectional approach, leveraging the Transformer architecture to consider both left and right contexts simultaneously. This design made BERT highly effective for tasks requiring nuanced understanding, such as:
        \begin{itemize}
            \item \textbf{Named Entity Recognition (NER)}: BERT's bidirectional context allowed it to detect and classify entities (e.g., names, locations) more accurately, as it understands the surrounding words.
            \item \textbf{Question Answering (QA)}: BERT's pre-training on large corpora enabled it to excel in QA benchmarks like SQuAD \cite{Rajpurkar2016}, where it can infer answers based on the entire context of the passage.
        \end{itemize}
        BERT introduced two novel training objectives:
        \begin{enumerate}
            \item \textbf{Masked Language Modeling (MLM)}: Randomly masks words in the input text and trains the model to predict masked words, forcing it to learn deep contextual representations.
            \item \textbf{Next Sentence Prediction (NSP)}: Trains the model to predict whether one sentence logically follows another, enhancing its ability to understand sentence relationships.
        \end{enumerate}
        The combination of MLM and NSP enabled BERT to capture complex linguistic patterns, making it a versatile model for various NLP tasks.

    \item \textbf{GPT Series (2018-2023)}: The Generative Pre-trained Transformer (GPT) series, developed by OpenAI, emphasized generative capabilities using an autoregressive, unidirectional approach \cite{Radford2018}. GPT models predict the next word in a sequence based on the preceding context, making them ideal for tasks like text generation and dialogue. The evolution of GPT models includes:
        \begin{itemize}
            \item \textbf{GPT-2 (2019)}: A notable leap in generative text capabilities, GPT-2 demonstrated impressive coherence and fluency in text generation \cite{Radford2018}, leading to concerns about the potential misuse of the model for generating disinformation.
            \item \textbf{GPT-3 (2020)}: With 175 billion parameters, GPT-3 introduced the concept of \textit{few-shot learning}, where the model can perform tasks with minimal examples provided in the input prompt \cite{Brown2020}. It showcased remarkable performance across diverse NLP tasks, reducing the need for task-specific fine-tuning.
            \item \textbf{GPT-4 (2023)}: GPT-4 expanded the model's capabilities by supporting multimodal inputs, such as text and images \cite{OpenAI2023}, making it suitable for applications like visual question answering. It also featured a larger context window, enabling it to handle longer and more complex dialogues.
        \end{itemize}
        The GPT series demonstrated the power of scaling model size and training data, but also highlighted challenges in interpretability and ethical considerations.

    \item \textbf{T5 (2019)}: The Text-to-Text Transfer Transformer (T5) introduced by Google presented a unified approach to NLP tasks, framing every problem as a text-to-text task \cite{Raffel2019}. This design simplified the model architecture and allowed for consistent training across tasks like summarization, translation, and sentiment analysis. Key innovations of T5 include:
        \begin{itemize}
            \item \textbf{Unified Text-to-Text Framework}: By treating every NLP task as a text generation problem, T5 eliminated the need for specialized model architectures for different tasks.
            \item \textbf{Span Corruption as Pre-training Objective}: Instead of masking single words like BERT, T5 used span corruption, where longer spans of text are masked and the model is trained to reconstruct them. This encouraged the model to learn more complex patterns in the text.
        \end{itemize}
        T5's versatility and strong performance across a wide range of tasks have made it a popular choice for many NLP applications.

    \item \textbf{LLaMA (2023)}: Meta's LLaMA series focused on developing efficient, smaller-scale models that offer competitive performance with fewer parameters \cite{Touvron2023}. LLaMA models were designed with an emphasis on accessibility, making them suitable for academic research and practical deployment where computational resources are limited.
        \begin{itemize}
            \item \textbf{LLaMA-2 (2023)}: LLaMA-2 achieved state-of-the-art performance in several NLP benchmarks while maintaining a smaller parameter size compared to other large models like GPT-4 \cite{Touvron2023b}. It employed optimizations such as improved tokenization and sparse attention mechanisms.
            \item \textbf{Research Accessibility}: By releasing LLaMA models with open-source licenses, Meta enabled researchers to experiment and build upon the models without the prohibitive costs associated with training massive LLMs.
        \end{itemize}
        LLaMA's focus on efficiency and accessibility has made it a valuable addition to the ecosystem of LLMs, particularly for research settings.

\end{itemize}

\subsection{Key Trends in LLM Evolution}

The progression from BERT to LLaMA highlights several important trends in the development of LLMs:

\begin{enumerate}
    \item \textbf{Scaling Up for Performance}: Increasing the model size and the amount of training data has consistently led to improved performance across various NLP tasks \cite{Kaplan2020}. However, this trend also introduces challenges related to the computational cost and environmental impact of training large models \cite{Strubell2019}.

    \item \textbf{Shift Toward Unified Architectures}: Models like T5 have shown the benefits of using a single, unified architecture for a wide range of NLP tasks, reducing the need for specialized models and simplifying deployment.

    \item \textbf{Increased Focus on Multimodality}: The ability of models like GPT-4 to handle both text and image inputs reflects a broader trend towards developing models that can process and integrate multiple data modalities \cite{Tsimpoukelli2021}.

    \item \textbf{Emphasis on Efficiency and Accessibility}: The development of smaller, more efficient models like LLaMA suggests a growing recognition of the need for accessible, cost-effective models that can be used in research and production environments without requiring extensive computational resources \cite{Schwartz2020}.
\end{enumerate}

\subsection{Challenges in LLM Development}

Despite the rapid advancements, several challenges remain in the evolution of LLMs:

\begin{itemize}
    \item \textbf{Interpretability Issues}: As models grow in size and complexity, understanding their decision-making processes becomes increasingly difficult, limiting their trustworthiness in critical applications \cite{Rudin2019}.

    \item \textbf{Bias and Ethical Concerns}: LLMs trained on vast datasets from the internet often inherit biases present in the data, which can lead to biased or inappropriate outputs \cite{Bender2021, peng2024securinglargelanguagemodels}. Addressing these issues remains an ongoing area of research.

    \item \textbf{Computational Constraints}: The resource-intensive nature of training large LLMs poses significant barriers for smaller research teams and increases the environmental footprint of these models \cite{Strubell2019}.

    \item \textbf{Risk of Misuse}: The generative capabilities of models like GPT-3 and GPT-4 raise concerns about their potential misuse for creating misleading content, requiring careful consideration of access and usage policies \cite{Zellers2019}.
\end{itemize}

\section{Overview of Transformer Architecture}

The backbone of most Large Language Models (LLMs) is the Transformer architecture, introduced by Vaswani et al. in 2017 \cite{Vaswani2017}. The Transformer marked a paradigm shift from earlier sequential models like Recurrent Neural Networks (RNNs) and Long Short-Term Memory networks (LSTMs) (detailed in Chapter \ref{sec:DeepModels}). Unlike these earlier models, which process input sequentially, the Transformer utilizes a self-attention mechanism that allows it to process entire sequences in parallel, making it highly efficient for large-scale language modeling tasks (detailed in Chapter \ref{sec:DeepModels}).

\subsection{Key Components of the Transformer Architecture}

The Transformer architecture is built upon several key innovations, each contributing to its ability to capture long-range dependencies and contextual relationships in the input data:

\begin{itemize}
    \item \textbf{Self-Attention Mechanism}: The self-attention mechanism is the core innovation of the Transformer. It enables the model to focus on different parts of the input sequence dynamically, assigning weights based on the relevance of each token to others in the sequence. This is particularly powerful for capturing long-range dependencies, as it avoids the vanishing gradient problem common in RNNs \cite{Bahdanau2015}.

    \item \textbf{Multi-head Attention}: Multi-head attention is an extension of the self-attention mechanism that allows the model to attend to information from different representation subspaces simultaneously. By using multiple attention heads, the model can capture various aspects of the input data, enhancing its ability to understand complex patterns.

    \item \textbf{Feed-forward Neural Network (FFN)}: The feed-forward neural network is applied to each position of the input sequence independently. It consists of two dense layers with a ReLU activation function in between. This component allows the model to learn complex, non-linear relationships between tokens.

    \item \textbf{Positional Encodings}: Since the Transformer processes input tokens in parallel, it lacks the inherent sequential information present in RNNs. To address this, the model uses positional encodings, which are added to the input embeddings to provide information about the relative positions of tokens in the sequence.

    \item \textbf{Layer Normalization and Residual Connections}: To stabilize training, the Transformer employs layer normalization and residual connections \cite{He2016}. The residual connections help in mitigating the vanishing gradient problem by allowing gradients to flow directly through the network, while layer normalization standardizes the input for each layer, improving convergence.
\end{itemize}

\subsection{Transformer Encoder-Decoder Architecture}

The Transformer consists of an encoder-decoder structure:

\begin{enumerate}
    \item \textbf{Encoder}: The encoder is composed of a stack of identical layers, each containing multi-head self-attention and a feed-forward neural network. The encoder processes the input sequence and generates a contextual representation for each token.

    \item \textbf{Decoder}: The decoder, similar to the encoder, consists of a stack of identical layers. However, it includes an additional masked self-attention layer, which prevents the decoder from attending to future tokens. This enables the autoregressive generation of sequences, where each token is predicted based on the previously generated tokens.
\end{enumerate}

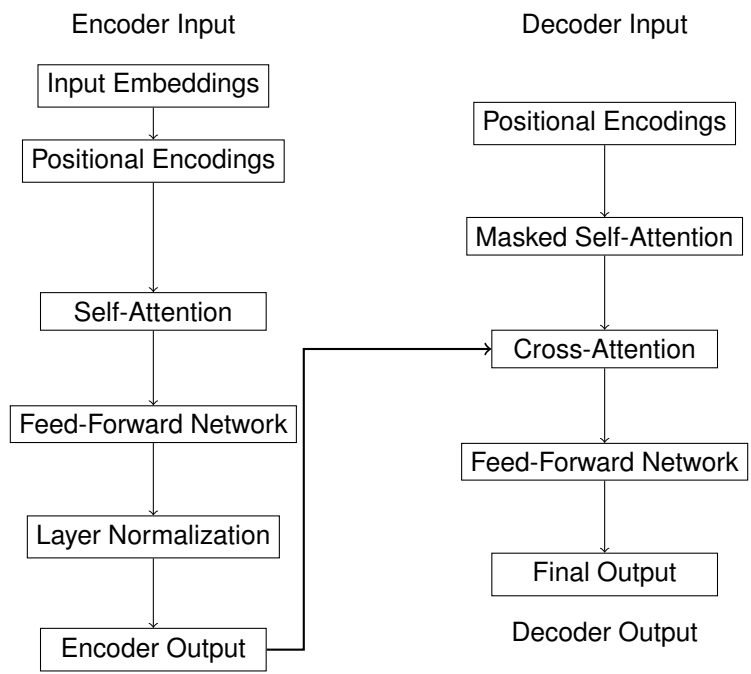
\begin{figure}[!ht]
    \centering
    \begin{tikzpicture}[scale=1, every node/.style={transform shape}]
        \node[draw, rectangle, minimum width=3cm] at (0, 0) (emb) {Input Embeddings};
        \node[draw, rectangle, minimum width=3cm] at (0, -1) (pos) {Positional Encodings};

        \node[draw, rectangle, minimum width=3cm] at (0, -3) (enc1) {Self-Attention};
        \node[draw, rectangle, minimum width=3cm] at (0, -4.5) (enc2) {Feed-Forward Network};
        \node[draw, rectangle, minimum width=3cm] at (0, -6) (norm) {Layer Normalization};
        \node[draw, rectangle, minimum width=3cm] at (0, -7.5) (enc_out) {Encoder Output};

        \draw[->] (emb) -- (pos);
        \draw[->] (pos) -- (enc1);
        \draw[->] (enc1) -- (enc2);
        \draw[->] (enc2) -- (norm);
        \draw[->] (norm) -- (enc_out);

        \node[draw, rectangle, minimum width=3cm] at (6, -0.5) (dec_pos) {Positional Encodings};
        \node[draw, rectangle, minimum width=3cm] at (6, -2) (dec1) {Masked Self-Attention};
        \node[draw, rectangle, minimum width=3cm] at (6, -3.5) (dec2) {Cross-Attention};
        \node[draw, rectangle, minimum width=3cm] at (6, -5) (dec3) {Feed-Forward Network};
        \node[draw, rectangle, minimum width=3cm] at (6, -6.5) (final_out) {Final Output};

        \draw[->] (dec_pos) -- (dec1);
        \draw[->] (dec1) -- (dec2);
        \draw[->] (dec2) -- (dec3);
        \draw[->] (dec3) -- (final_out);

        \draw[->, thick] (enc_out) -- ++(2, 0) |- (dec2);

        \node[above] at (0, 0.5) {Encoder Input};
        \node[above] at (6, 0.5) {Decoder Input};
        \node[below] at (6, -7) {Decoder Output};
    \end{tikzpicture}
    \caption{Clear Separation of Transformer Encoder-Decoder Architecture}
\end{figure}

\subsection{Advantages of Transformer Architecture}

The Transformer architecture introduced several key advantages over previous models:

\begin{itemize}
    \item \textbf{Parallelization}: The self-attention mechanism allows for parallel processing of input tokens, significantly reducing training time compared to RNNs.

    \item \textbf{Long-range Dependency Handling}: The attention mechanism captures dependencies between distant tokens more effectively than sequential models.

    \item \textbf{Scalability}: The modular nature of the Transformer makes it easily scalable, leading to the development of large models like BERT, GPT, and T5.
\end{itemize}

These advantages have made the Transformer the dominant architecture in NLP, forming the foundation of most state-of-the-art LLMs today.

\section{Black Box Challenges in LLMs}

Large Language Models (LLMs) have demonstrated extraordinary capabilities in understanding and generating human language. However, despite their success, LLMs often exhibit "black box" characteristics due to their complexity and the opaque nature of their decision-making processes. This section explores the black box problem in LLMs, the efforts to enhance transparency, and the distinction between interpretability and explainability.

\subsection{The Black Box Problem in LLMs}

The "black box" problem in LLMs refers to the difficulty in understanding how these models make specific predictions. LLMs are built with billions of parameters spread across numerous neural network layers. This high dimensionality and complex internal structure make it challenging to trace how input data transforms into output predictions \cite{Rudin2019}. Unlike simpler models such as linear regression or decision trees, where the decision-making process can be visualized and interpreted directly, LLMs rely on abstract, non-linear transformations that are not easily decipherable.

The black box nature of LLMs introduces several key issues:
\begin{itemize}
    \item \textbf{Lack of Transparency}: Users and developers cannot easily understand the reasoning behind specific model outputs, leading to concerns about trust and reliability.

    \item \textbf{Difficulty in Debugging}: When LLMs generate incorrect or biased outputs, it is challenging to diagnose the root cause of the error.

    \item \textbf{Ethical Concerns}: The inability to interpret LLM behavior makes it harder to detect and mitigate biases, raising ethical and fairness issues in applications like hiring, legal decisions, and medical diagnosis \cite{Doshi-Velez2017}.
\end{itemize}

\subsection{Transparency and Open-source Models}

To address the black box problem, there have been efforts to increase the transparency of LLMs by developing open-source models, such as GPT-Neo \cite{Black2021}, BLOOM \cite{Scao2022}, and LLaMA \cite{Touvron2023}. These open-source projects allow researchers to access model weights, analyze architectural components, and perform in-depth studies on model behavior.

\begin{itemize}
    \item \textbf{GPT-Neo}: An open-source alternative to GPT-3, developed by EleutherAI, provides access to model internals, facilitating transparency and reproducibility in research.

    \item \textbf{BLOOM}: Developed by the BigScience project, BLOOM is a multilingual, open-access LLM aimed at promoting ethical AI research and increasing transparency.

    \item \textbf{LLaMA}: Meta's LLaMA models prioritize efficiency and transparency by making model weights available for academic research, enabling detailed interpretability studies.
\end{itemize}

These efforts have provided valuable insights into the functioning of LLMs, particularly regarding attention mechanisms and their impact on model outputs. However, simply releasing model weights does not fully resolve the interpretability challenges due to the sheer size and complexity of the models.

\subsection{Interpretability vs. Explainability}

In the context of LLMs, it is crucial to distinguish between interpretability and explainability:

\begin{itemize}
    \item \textbf{Interpretability}: Refers to the degree to which a human can understand the internal workings of the model. This involves analyzing the model's parameters, neural network activations, and decision paths. For LLMs, interpretability often focuses on understanding what specific neurons or layers are learning, such as recognizing syntax or capturing semantic relationships \cite{Belinkov2019}.

    \item \textbf{Explainability}: Involves providing understandable reasons for the model's predictions. Explainability aims to bridge the gap between complex model internals and user understanding by generating human-interpretable explanations, such as feature importance scores or visualizations \cite{Guidotti2018}.
\end{itemize}

For instance, while an interpretability approach may involve inspecting the activation of a particular neuron in response to an input sentence, an explainability approach might use methods like SHAP \cite{Lundberg2017} or LIME \cite{Ribeiro2016} to show which input words contributed most to the model's prediction.

\section{Overview of Interpretability Techniques for LLMs}

Interpreting LLMs requires a suite of techniques designed to shed light on different aspects of the model's behavior. These techniques range from analyzing input importance to understanding internal activations and attention patterns.

\subsection{Gradient-based Methods for Input Analysis}

Gradient-based methods analyze the gradients of the model's output with respect to its input features to determine which words or tokens have the most significant impact on the prediction (detailed in Chapter \ref{sec:Techniques}). Notable examples include:

\begin{itemize}
    \item \textbf{Integrated Gradients}: This method computes the contribution of each input feature by integrating the gradients along a path from a baseline input (e.g., all zeros) to the actual input \cite{Sundararajan2017}. It provides a more accurate attribution than simple gradient analysis.

    \item \textbf{Saliency Maps}: Saliency maps highlight the regions of the input that most influence the output, visualizing the model's focus on specific words or phrases \cite{Simonyan2013}.
\end{itemize}

\section{Advanced Interpretability Techniques}

Advanced interpretability methods aim to provide a deeper understanding of the internal mechanisms of Large Language Models (LLMs) by analyzing embeddings, neural activations, and probing model knowledge. These techniques go beyond surface-level explanations, offering insights into how linguistic and semantic knowledge is represented and processed by the model.

\subsection{Embedding Analysis and Probing}

Embeddings in LLMs are high-dimensional representations that encode rich semantic information about words, phrases, and sentences. The analysis of embeddings helps us understand what linguistic properties the model has learned and how these properties are organized in the embedding space.

\begin{itemize}
    \item \textbf{t-SNE and PCA for Embedding Visualization}: Dimensionality reduction techniques such as t-SNE (t-distributed Stochastic Neighbor Embedding) and PCA (Principal Component Analysis) are often used to visualize embeddings. By projecting embeddings into a 2D or 3D space, we can observe clustering patterns that reveal semantic relationships among the words.

    \item \textbf{Probing Classifiers}: Probing involves training simple classifiers on top of embeddings to assess whether they capture specific linguistic features, such as part-of-speech tags or syntactic roles \cite{Conneau2018}. If the classifier performs well, it indicates that the embeddings encode the relevant linguistic information.
\end{itemize}

\subsubsection{Python Code Example}

In this example, we visualize embeddings from a pretrained BERT model using PCA. We use a simple list of words, including "king", "queen", "man", and "woman", to examine how their embeddings are represented in a 2D space.

\begin{lstlisting}[style=python, literate={\$}{{\$}}1]
import numpy as np
import matplotlib.pyplot as plt
from sklearn.decomposition import PCA
from transformers import BertTokenizer, TFBertModel
import tensorflow as tf

# Load pretrained BERT model and tokenizer
model = TFBertModel.from_pretrained('bert-base-uncased')
tokenizer = BertTokenizer.from_pretrained('bert-base-uncased')

# Define a list of words to encode
words = ["king", "queen", "man", "woman"]

# Tokenize words and obtain embeddings
inputs = tokenizer(words, return_tensors='tf', padding=True, truncation=True)
outputs = model(inputs['input_ids'])[0].numpy()

# Compute the mean embeddings for each word
mean_embeddings = outputs.mean(axis=1)

# Perform PCA to reduce embeddings to 2D
pca = PCA(n_components=2)
reduced_embeddings = pca.fit_transform(mean_embeddings)

# Plot the 2D embeddings
plt.figure(figsize=(8, 6))
plt.scatter(reduced_embeddings[:, 0], reduced_embeddings[:, 1], color='blue')

# Annotate the plot with word labels
for i, word in enumerate(words):
    plt.annotate(word, (reduced_embeddings[i, 0], reduced_embeddings[i, 1]), fontsize=12)

plt.title("PCA Visualization of BERT Word Embeddings")
plt.xlabel("Principal Component 1")
plt.ylabel("Principal Component 2")
plt.grid(True)
plt.show()
\end{lstlisting}

\begin{figure}[!ht]
    \centering
    \includegraphics[width=0.8\textwidth]{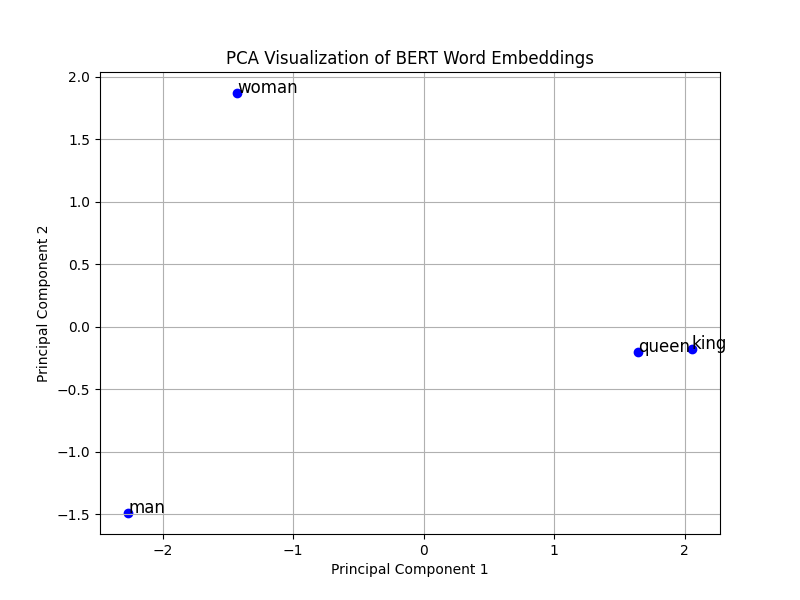}
    \caption{PCA Visualization of BERT Word Embeddings for "king", "queen", "man", and "woman".}
    \label{fig:pca_visualization}
\end{figure}

\subsubsection{Result Explanation}

The PCA plot in Figure \ref{fig:pca_visualization} shows the 2D projection of the BERT embeddings for the words "king", "queen", "man", and "woman". The visualization reveals the following semantic relationships:

\begin{itemize}
    \item \textbf{Clustering of Related Words}: The words "king" and "queen" appear close to each other in the plot, indicating that their embeddings capture the semantic similarity between these two terms, both being royalty-related.

    \item \textbf{Clear Semantic Distinctions}: The separation between the male-associated words ("king" and "man") and female-associated words ("queen" and "woman") highlights that the BERT embeddings effectively encode not only lexical but also gender-specific semantic features.
\end{itemize}

This example demonstrates that BERT embeddings can capture complex semantic relationships, such as gender distinctions and similarities within categories (e.g., royalty). By visualizing these embeddings, we gain insights into the kinds of linguistic knowledge that LLMs like BERT have acquired during training.

\subsection{Neural Layer-wise Interpretability}

Neural Layer-wise Interpretability focuses on understanding the role of individual layers in the Transformer architecture. In Large Language Models (LLMs) like BERT, each layer captures a different level of linguistic information, contributing to the model's overall understanding of the input text.

\begin{itemize}
    \item \textbf{Early Layers}: These layers tend to capture surface-level features, such as token identity and basic syntactic patterns. The model focuses on understanding individual words and their basic relationships in this stage \cite{Jawahar2019}.

    \item \textbf{Middle Layers}: The middle layers are responsible for capturing more abstract syntactic structures and dependencies, such as subject-verb agreement and grammatical relationships. These layers help the model understand the sentence structure.

    \item \textbf{Late Layers}: These layers encode high-level semantic information and task-specific representations. They contribute directly to the final prediction, often containing the most abstract and context-aware features of the input text.

    \item \textbf{Layer-wise Relevance Propagation (LRP)}: LRP is a technique that assigns relevance scores to neurons in each layer, helping trace the flow of information through the model \cite{Voita2019}. This method is particularly useful for identifying which neurons and layers are most important for a given prediction.
\end{itemize}

\subsubsection{Python Code Example}

The following code snippet demonstrates how to extract and analyze layer-wise activations from a pretrained BERT model using TensorFlow.

\begin{lstlisting}[style=python, literate={\$}{{\$}}1]
import numpy as np
import tensorflow as tf
from transformers import BertTokenizer, TFBertModel
import matplotlib.pyplot as plt

# Load pretrained BERT model and tokenizer with hidden states output enabled
model = TFBertModel.from_pretrained('bert-base-uncased', output_hidden_states=True)
tokenizer = BertTokenizer.from_pretrained('bert-base-uncased')

# Define a function to extract and analyze layer-wise activations
def extract_and_analyze_activations(model, inputs):
    outputs = model(inputs)
    hidden_states = outputs.hidden_states  # Extract all hidden states
    layer_means = [tf.reduce_mean(state).numpy() for state in hidden_states]  # Compute mean activation
    return hidden_states, layer_means

# Example usage with a sample input sentence
input_data = tokenizer("The cat sat on the mat.", return_tensors='tf', padding=True, truncation=True)
input_ids = input_data['input_ids']

# Extract activations and compute mean activations for each layer
layer_outputs, layer_means = extract_and_analyze_activations(model, input_ids)

# Print the number of layers and the shape of activations from the first layer
print("Number of layers analyzed:", len(layer_outputs))
print("Shape of activations from the first layer:", layer_outputs[0].shape)

# Plot mean activations across layers
plt.figure(figsize=(10, 6))
plt.plot(range(len(layer_means)), layer_means, marker='o', color='blue')
plt.title("Mean Activations Across BERT Layers")
plt.xlabel("Layer")
plt.ylabel("Mean Activation Value")
plt.grid(True)
plt.show()
\end{lstlisting}

\subsubsection{Result Explanation}

The code snippet extracts the activations from each layer of the BERT model and computes the mean activation value for each layer. The resulting plot, shown below, displays the mean activation values across all 13 layers of BERT.

\begin{figure}[!ht]
    \centering
    \includegraphics[width=0.8\textwidth]{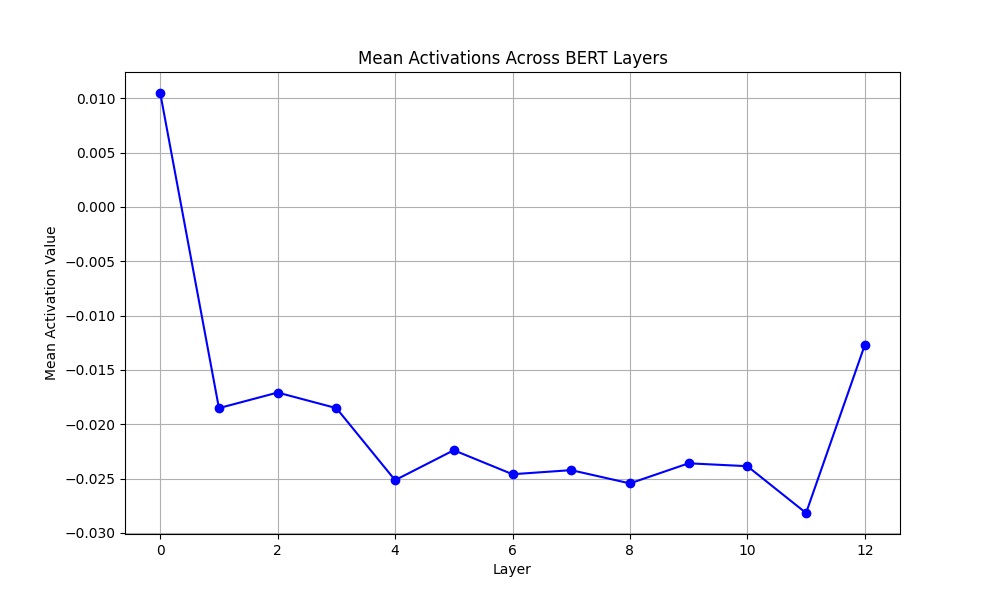}
    \caption{Mean Activations Across BERT Layers}
\end{figure}

\begin{itemize}
    \item \textbf{Early Layers (0-4)}:
    The activation values start relatively high in the first layer and drop significantly in the subsequent layers. This indicates that the model initially focuses on capturing token-level features, such as word embeddings and basic syntactic patterns.

    \item \textbf{Middle Layers (5-10)}:
    The activation values stabilize and remain relatively low in the middle layers. This suggests that the model is capturing more complex syntactic structures and dependencies, such as sentence parsing and grammatical relationships.

    \item \textbf{Late Layers (11-12)}:
    In the last few layers, we observe a sharp increase in the mean activation values. This indicates that the model is capturing high-level semantic features and task-specific representations, which are crucial for making predictions.
\end{itemize}

\subsubsection{Relevance to Neural Layer-wise Interpretability}

The varying mean activation values across different layers provide insights into the hierarchical learning process of the BERT model:

\begin{itemize}
    \item \textbf{Early Layers}: Focus on word-level features and token embeddings, which are necessary for understanding the basic building blocks of language.

    \item \textbf{Middle Layers}: Capture syntactic structures and grammatical dependencies, contributing to the model's understanding of sentence structure.

    \item \textbf{Late Layers}: Encode semantic information and task-specific knowledge, directly influencing the model's final predictions.
\end{itemize}

This analysis helps us understand the distinct roles played by each layer in a Transformer-based model, highlighting the hierarchical nature of feature extraction in LLMs. By visualizing and analyzing these layer-wise activations, we gain valuable insights into the inner workings of the model, making it easier to explain and interpret its decisions.

\subsection{Probing Knowledge in Embeddings}

Probing tasks are designed to evaluate the extent to which embeddings from Large Language Models (LLMs) capture specific linguistic knowledge. These tasks can help researchers understand whether the embeddings encode information related to syntax, semantics, or other linguistic features \cite{Hewitt2019}. Probing tasks typically involve training simple classifiers on top of the embeddings and evaluating their performance.

\begin{itemize}
    \item \textbf{Syntactic Probing}: This task evaluates the model's understanding of syntactic properties. For instance, it checks whether embeddings can differentiate between subjects and objects within a sentence, capturing the syntactic roles of words.

    \item \textbf{Semantic Probing}: This task examines whether the model captures semantic relationships, such as word similarity or entailment. It aims to understand if embeddings reflect deeper semantic features like synonymy or antonymy.
\end{itemize}

\subsubsection{Python Code Example}

The following example demonstrates a probing task aimed at evaluating whether BERT's embeddings can capture basic part-of-speech (POS) information. We use a simple logistic regression classifier to predict whether a word is a noun (label 0) or a verb (label 1) based on its embedding.

\begin{lstlisting}[style=python, literate={\$}{{\$}}1]
import numpy as np
from sklearn.linear_model import LogisticRegression
from sklearn.metrics import accuracy_score
from transformers import BertTokenizer, TFBertModel
import tensorflow as tf

# Load pretrained BERT model and tokenizer
model = TFBertModel.from_pretrained('bert-base-uncased')
tokenizer = BertTokenizer.from_pretrained('bert-base-uncased')

# Define a list of words and their corresponding part-of-speech labels
words = ["cat", "run", "dog", "jump"]  # Example words
labels = [0, 1, 0, 1]  # Labels: 0 for noun, 1 for verb

# Tokenize the words and obtain embeddings
inputs = tokenizer(words, return_tensors='tf', padding=True, truncation=True)
outputs = model(inputs['input_ids'])[0].numpy()

# Use mean embeddings as features for the classifier
features = outputs.mean(axis=1)

# Train a logistic regression classifier
classifier = LogisticRegression()
classifier.fit(features, labels)

# Make predictions
predictions = classifier.predict(features)

# Evaluate the classifier
accuracy = accuracy_score(labels, predictions)
print(f"Probing Task Accuracy: {accuracy:.2f}")
\end{lstlisting}

\subsubsection{Result Explanation}

The output of the above code snippet is as follows:

\begin{figure}[!ht]
    \centering
    \includegraphics[width=0.8\textwidth]{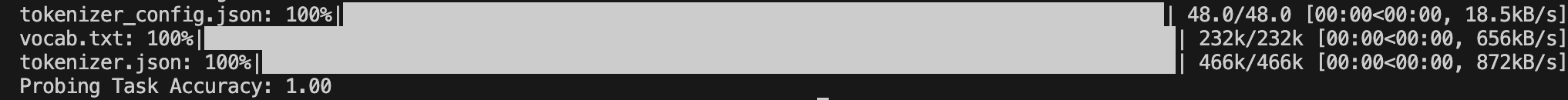}
    \caption{Output of the probing task showing progress bars and probing task accuracy.}
    \label{fig:probing_result}
\end{figure}

This result indicates that the probing classifier achieved an accuracy of 1.00 (or 100\%). Such high accuracy suggests that the BERT embeddings effectively captured the part-of-speech information for this simple set of words. In other words, the embeddings for nouns ("cat" and "dog") and verbs ("run" and "jump") are distinct enough for a simple logistic regression classifier to separate them accurately.

\section{Exploring Prompt Engineering and Interpretability}

Prompt engineering is a pivotal technique in steering the behavior of Large Language Models (LLMs). By meticulously crafting input prompts, one can elicit desired responses, thereby enhancing the interpretability and reliability of these models \cite{Reynolds2021}.

\subsection{Impact of Prompt Design on Model Behavior}

The design of prompts significantly influences LLM outputs. Subtle variations in wording can lead to markedly different responses, underscoring the importance of precise prompt construction \cite{Jiang2020}. For instance, rephrasing a question or altering its context can shift the model's focus, resulting in diverse interpretations and answers. This sensitivity necessitates a deep understanding of prompt mechanics to ensure consistent and accurate model behavior.

\subsection{Direct Querying of LLMs for Interpretability}

Directly querying LLMs with carefully designed prompts serves as an effective method for probing their internal knowledge and reasoning processes \cite{Petroni2019}. By posing specific questions or scenarios, researchers can gain insights into the model's comprehension and decision-making pathways. This approach facilitates the identification of strengths and potential biases within the model, contributing to the development of more transparent and accountable AI systems.

\subsection{Model Confidence Scores and Stability}

Analyzing the confidence scores and response stability of LLMs across various prompts provides valuable information regarding their reliability. Models that consistently exhibit high confidence in correct responses and appropriately calibrated uncertainty in uncertain scenarios are deemed more trustworthy \cite{Desai2020}. This assessment aids in understanding the model's self-awareness and its ability to gauge the accuracy of its outputs, which is crucial for applications requiring high levels of dependability.

\section{Case Analysis of Major LLMs}

A comparative analysis of prominent LLMs reveals distinct approaches to interpretability and user interaction, each with unique strengths and areas for improvement.

\subsection{ChatGPT: Interpretability and User Interactions}

ChatGPT is renowned for its interactive capabilities, often providing detailed explanations for its responses \cite{OpenAI2023ChatGPT}. This feature enhances user trust and engagement by offering transparency into the model's reasoning process. However, the quality of these explanations can vary, highlighting the need for continuous refinement to ensure consistency and depth in interpretability.

\subsection{Claude: Analyzing Model Responses}

Claude emphasizes dialogue quality and interpretability through advanced attention mechanisms \cite{Anthropic2023}. By focusing on coherence and user-friendliness, Claude aims to deliver responses that are not only accurate but also contextually appropriate. This approach seeks to balance technical precision with an intuitive user experience, fostering a more natural interaction between humans and AI.

\subsection{Comparative Analysis of LLM Behaviors}

Evaluating the behaviors of different LLMs under similar conditions allows for the identification of patterns and trends \cite{Bai2022}. Such comparative studies inform the development of future models by highlighting effective strategies and pinpointing areas requiring enhancement. This iterative process contributes to the evolution of LLMs towards greater interpretability, reliability, and user satisfaction.

\section{Current Research on Explainable AI Based on LLMs}

Ongoing research endeavors are dedicated to advancing the interpretability of LLMs, employing various innovative techniques to unravel the complexities of these models.

\subsection{Techniques for Probing and Explanation}

Researchers are developing sophisticated probing methods to delve into the internal representations of LLMs \cite{Dalvi2019}. These techniques aim to uncover how models encode and process information, facilitating a clearer understanding of their decision-making processes. By dissecting the layers and components of LLMs, these probing strategies contribute to the creation of more transparent and accountable AI systems.

\subsection{Analysis of Embedding Spaces}

Mapping and visualizing embedding spaces offer insights into how LLMs organize and interpret information \cite{Ethayarajh2019}. By examining the spatial relationships between embeddings, researchers can infer semantic associations and hierarchical structures within the model's knowledge base. This analysis aids in identifying potential biases and gaps in understanding, guiding efforts to enhance model performance and fairness.

\subsection{Trust and Reliability in LLM Interpretability}

Building trust in LLMs necessitates robust interpretability methods that consistently explain model behavior across diverse scenarios \cite{Doshi-Velez2017}. Ensuring that models provide transparent and justifiable outputs is essential for their integration into critical applications. Ongoing research focuses on developing frameworks and tools that assess and improve the reliability of LLMs, fostering greater confidence in their deployment.

\section*{Conclusion}

In this chapter, we have delved deep into the interpretability of Large Language Models (LLMs), highlighting the unique challenges and techniques used to unravel the inner workings of these complex architectures. From analyzing feature importance and attention mechanisms to probing embeddings and layer-wise neural interpretations, we explored a comprehensive suite of methods aimed at making LLMs more transparent. Our examination of error analysis and bias detection provided insights into the ethical considerations and real-world implications of deploying LLMs. We also discussed the impact of prompt engineering on model behavior, emphasizing its role in enhancing both usability and interpretability. Through case studies of major LLMs such as ChatGPT and Claude, we demonstrated the diverse strategies used to decode their decision-making processes.

As we conclude this chapter, it is clear that while significant progress has been made, the interpretability of LLMs remains an ongoing area of research filled with open questions. In the following chapter, we will expand our focus beyond LLMs to explore a broader array of techniques for explainable AI. We will delve into model-based and post-hoc methods, offering tools and frameworks that can be applied across various types of AI systems. This transition will set the stage for a deeper understanding of how to make AI models not only powerful but also trustworthy and transparent.

\chapter{Techniques for Explainable AI}
\label{sec:Techniques}

\section{Overview of XAI Techniques}
Explainable AI (XAI) techniques can be categorized based on the interpretability of the model into \textbf{White-box} and \textbf{Black-box} models \cite{Adadi2018}. Additionally, XAI methods can be classified into \textbf{Model-based Techniques}, \textbf{Post-hoc Interpretation Techniques}, \textbf{Counterfactual Explanations}, \textbf{Causal Inference Techniques}, \textbf{Graph-based Explanation Techniques}, and \textbf{Multimodal Explainability} \cite{Arrieta2020}. Detailed explanations for Model-based Techniques can be found in Chapters \ref{sec:Traditional}, \ref{sec:DeepModels}, and \ref{sec:LLMs}.

\section{White-box and Black-box Models}
This section describes the differences between White-box and Black-box models. White-box models are inherently interpretable and provide straightforward explanations for their predictions \cite{Rudin2019}. In contrast, Black-box models are more complex and require additional methods for interpretation \cite{Lipton2018}.

\subsection{White-box Models}
White-box models include:
\begin{itemize}
    \item Linear Regression \cite{Weisberg2014}
    \item Logistic Regression \cite{Hosmer2013}
    \item Decision Trees \cite{Rokach2014}
    \item Rule-based Systems \cite{Wang2017}
    \item K-Nearest Neighbors (KNN) \cite{Zhang2017}
    \item Naive Bayes Classifiers \cite{Murphy2012}
    \item Generalized Additive Models (GAMs) \cite{Lou2012}
\end{itemize}
These models are covered in detail in Chapter \ref{sec:Traditional}.

\subsection{Black-box Models}
Black-box models include:
\begin{itemize}
    \item Neural Networks (e.g., Deep Learning Models) \cite{Goodfellow2016}
    \item Support Vector Machines (SVMs) \cite{Ben-Hur2010}
    \item Ensemble Methods (e.g., Random Forests, Gradient Boosting Machines) \cite{Zhou2012}
    \item Transformer Models (e.g., BERT, GPT) \cite{Vaswani2017}
    \item Graph Neural Networks (GNNs) \cite{Wu2020}
\end{itemize}
These models are discussed further in Chapters \ref{sec:DeepModels} and \ref{sec:LLMs}.

\section{Model-based Techniques}
Model-based techniques focus on building inherently interpretable models. These include:
\begin{itemize}
    \item Decision Trees \cite{Hara2018}
    \item Generalized Additive Models (GAMs) \cite{Caruana2015}
    \item Bayesian Models \cite{Gelman2013}
    \item Sparse Linear Models \cite{Tibshirani2011}
    \item Attention Mechanisms in Neural Networks \cite{Bahdanau2015}
    \item Self-Explaining Neural Networks \cite{Alvarez-Melis2018}
    \item Concept Bottleneck Models \cite{Koh2020}
    \item Disentangled Representations \cite{Higgins2017}
\end{itemize}
A detailed explanation of these techniques can be found in Chapters \ref{sec:Traditional}, \ref{sec:DeepModels}, and \ref{sec:LLMs}.

\section{Post-hoc Interpretation Techniques}
Post-hoc interpretation techniques are applied after the model has been trained to provide explanations for predictions \cite{Guidotti2018}.

\section{Feature Attribution Methods}
Understanding the contributions of individual input features to a model’s decision is not just an academic exercise; it is pivotal to building trust, ensuring fairness, and enhancing the robustness of machine learning systems. Feature attribution methods address one of the core challenges in artificial intelligence: making complex models interpretable without sacrificing their predictive power.

\subsection{Feature Importance Analysis}

Feature importance analysis is a widely used post-hoc interpretation technique for machine learning models, particularly effective for classical models such as decision trees, random forests, and gradient boosting models \cite{Lundberg2017}. This method helps identify which input features contribute most to the predictions, enhancing the interpretability and transparency of the model. While feature importance is traditionally associated with tree-based models, it has also been adapted for neural networks and, more recently, for large language models (LLMs) through techniques like Integrated Gradients \cite{Sundararajan2017} and SHAP values \cite{Lundberg2017}. In essence, feature importance analysis provides a way to rank the significance of input features, helping practitioners understand model behavior beyond the predictive metrics.

\paragraph{Scope of Application}

The method of feature importance analysis is versatile and applicable to various types of models:
\begin{itemize}
    \item \textbf{Tree-based models (Decision Trees, Random Forests, Gradient Boosting Models):} Feature importance is inherently determined by the tree structure. Nodes closer to the root indicate higher importance \cite{Louppe2013}.
    \item \textbf{Linear models (Logistic Regression, Linear Regression):} Feature importance can be directly interpreted based on the magnitude and sign of the coefficients \cite{James2013}.
    \item \textbf{Neural Networks (Shallow and Deep Networks):} Approximations of feature importance can be derived using techniques like Integrated Gradients \cite{Shrikumar2017}, SHAP values \cite{Lundberg2017}, or Layer-wise Relevance Propagation (LRP) \cite{Montavon2017}.
    \item \textbf{Large Language Models (LLMs):} While traditional feature importance is not directly applicable, attention weights \cite{Vaswani2017} and gradient-based methods like Integrated Gradients can approximate token importance, offering insights into the model's decision-making process.
\end{itemize}

\paragraph{Python Code Example}

Let's illustrate feature importance analysis using a simple example with a decision tree classifier on the Iris dataset. This dataset consists of measurements of different iris flowers and their corresponding species \cite{Dua2019}. The goal is to predict the species based on these measurements.

\begin{lstlisting}[style=python, literate={\$}{{\$}}1]
from sklearn.datasets import load_iris
from sklearn.tree import DecisionTreeClassifier
import matplotlib.pyplot as plt

# Load the Iris dataset
data = load_iris()
X, y = data.data, data.target

# Train a Decision Tree Classifier
model = DecisionTreeClassifier(random_state=42)
model.fit(X, y)

# Extract feature importance
importances = model.feature_importances_
feature_names = data.feature_names

# Plot feature importance
plt.figure(figsize=(8, 6))
plt.barh(feature_names, importances, color='skyblue')
plt.xlabel("Feature Importance Score")
plt.ylabel("Features")
plt.title("Feature Importance Analysis for Decision Tree Classifier")
plt.show()
\end{lstlisting}

\paragraph{Results Explanation}

The bar plot above illustrates the importance of each feature as determined by the decision tree model. Higher scores indicate features that have a more significant impact on the model's predictions. In this example, the features \texttt{petal length (cm)} and \texttt{petal width (cm)} exhibit the highest importance scores. This aligns well with domain knowledge, as these features are known to be crucial in distinguishing between iris species \cite{Zhao2018}. Specifically:

\begin{itemize}
    \item \textbf{Petal Length:} The most decisive feature, playing a critical role in differentiating species.
    \item \textbf{Petal Width:} Complementary to petal length, it further refines the model's classification capability.
    \item \textbf{Sepal Length and Sepal Width:} These features are less influential, contributing minimally to the decision boundaries set by the model.
\end{itemize}
\begin{figure}[!ht]
    \centering
    \includegraphics[width=0.8\textwidth]{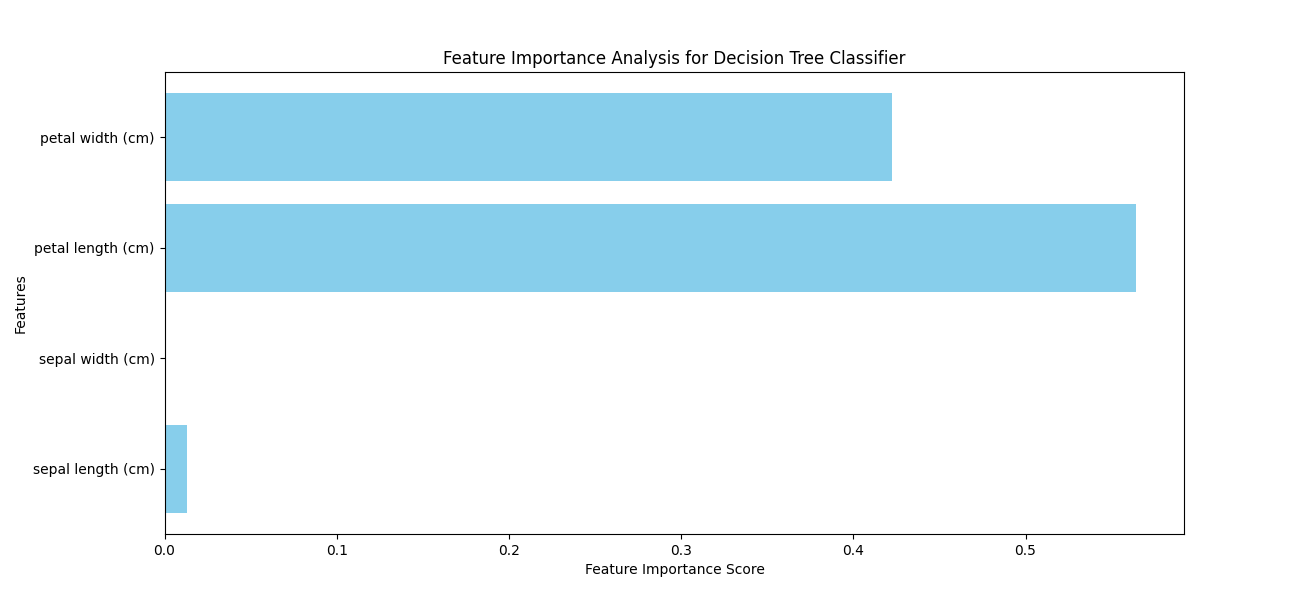}
    \caption{Feature Importance Analysis for Decision Tree Classifier on the Iris Dataset}
    \label{fig:feature-importance}
\end{figure}

\paragraph{Limitations}

While feature importance is straightforward for tree-based models, it has notable limitations \cite{Altmann2010}:
\begin{itemize}
    \item \textbf{Bias Towards High Cardinality Features:} Features with many unique values may appear more important, even if they are not truly predictive \cite{Altmann2010}.
    \item \textbf{Lack of Consistency Across Models:} Different models (e.g., linear models vs. decision trees) may attribute different importance scores to the same features \cite{Gregorutti2017}.
    \item \textbf{Not Suitable for Complex Models:} For deep neural networks or LLMs, traditional feature importance methods are less effective due to the non-linear and high-dimensional nature of these models \cite{Samek2019}.
\end{itemize}

\paragraph{Feature Importance in Neural Networks}

For neural networks, feature importance can be approximated using gradient-based methods like Integrated Gradients or Layer-wise Relevance Propagation (LRP). These methods analyze how the gradient of the output with respect to the input features changes, providing a measure of feature relevance. However, detailed exploration of these methods is beyond the scope of this section and will be covered in the upcoming chapter on interpretability in neural networks.

\subsection{Shapley Additive Explanations (SHAP)}

Shapley Additive Explanations (SHAP) is a game-theoretic approach to explain the output of any machine learning model \cite{Lundberg2017}. It assigns an importance value (known as a Shapley value) to each feature by treating the prediction problem as a cooperative game where the features are the "players." The goal is to fairly attribute the prediction to the contributing features, providing insights into how each feature influences the model's decision. SHAP is versatile and can be applied to traditional machine learning models, complex deep learning architectures, and even large language models (LLMs).

\paragraph{Scope of Application}

SHAP is particularly effective for:
\begin{itemize}
    \item \textbf{Traditional Machine Learning Models:} It works well with decision trees, random forests, gradient boosting models, and linear models \cite{Lundberg2018}.
    \item \textbf{Deep Neural Networks:} SHAP can approximate feature importance using gradient-based explanations \cite{Ancona2019}.
    \item \textbf{Large Language Models (LLMs):} SHAP can be adapted to interpret the importance of input tokens, aiding in understanding how specific words or phrases influence predictions \cite{Li2020}.
\end{itemize}

\paragraph{Principles and Formula}

SHAP values are based on the concept of Shapley values from cooperative game theory \cite{Shapley2016}. The Shapley value for a feature \(i\) is calculated as the average marginal contribution of the feature across all possible subsets of features. The formula for the Shapley value \(\phi_i\) is:

\[
\phi_i = \sum_{S \subseteq N \setminus \{i\}} \frac{|S|! (|N| - |S| - 1)!}{|N|!} \left( f(S \cup \{i\}) - f(S) \right)
\]

where:
\begin{itemize}
    \item \(N\) is the set of all features.
    \item \(S\) is a subset of features not containing \(i\).
    \item \(f(S)\) is the model prediction using only the features in \(S\).
\end{itemize}

This formula represents the weighted average of the marginal contributions of feature \(i\) across all subsets of features. It ensures that the contributions are fairly distributed based on their impact on the prediction.

\paragraph{Python Code Example}

In this section, we demonstrate the application of SHAP (Shapley Additive Explanations) using a gradient boosting classifier trained on the well-known Breast Cancer dataset \cite{Dua2019}. This example showcases how SHAP values can effectively explain the predictions made by a complex model like gradient boosting.

\begin{lstlisting}[style=python, literate={\$}{{\$}}1]
import shap
import xgboost as xgb
import pandas as pd
from sklearn.datasets import load_breast_cancer
from sklearn.model_selection import train_test_split

# Load the Breast Cancer dataset
data = load_breast_cancer()
X = pd.DataFrame(data.data, columns=data.feature_names)
y = data.target

# Split the dataset
X_train, X_test, y_train, y_test = train_test_split(X, y, test_size=0.2, random_state=42)

# Train a gradient boosting model
model = xgb.XGBClassifier(random_state=42)
model.fit(X_train, y_train)

# Initialize the SHAP explainer
explainer = shap.Explainer(model, X_train)
shap_values = explainer(X_test)

# Plot the SHAP summary plot
shap.summary_plot(shap_values, X_test)
\end{lstlisting}

\paragraph{Results Explanation}

The SHAP summary plot is a comprehensive visualization that provides a global interpretation of the model. Each dot in the plot represents a single SHAP value for a particular feature in one instance of the test set. The x-axis indicates the SHAP value, which reflects the impact of the feature on the model's output. Positive SHAP values suggest a contribution towards predicting the positive class (e.g., malignant tumor), while negative SHAP values contribute towards predicting the negative class (e.g., benign tumor).

The color gradient from blue to red indicates the feature value, where red signifies high feature values and blue signifies low feature values. In this example, features such as \texttt{mean concave points}, \texttt{worst area}, and \texttt{worst concave points} are identified as having the highest impact on the model's predictions. This suggests that the model places significant emphasis on the geometrical properties of the tumors when making a prediction.

\begin{figure}[htbp]
    \centering
    \includegraphics[width=0.9\textwidth]{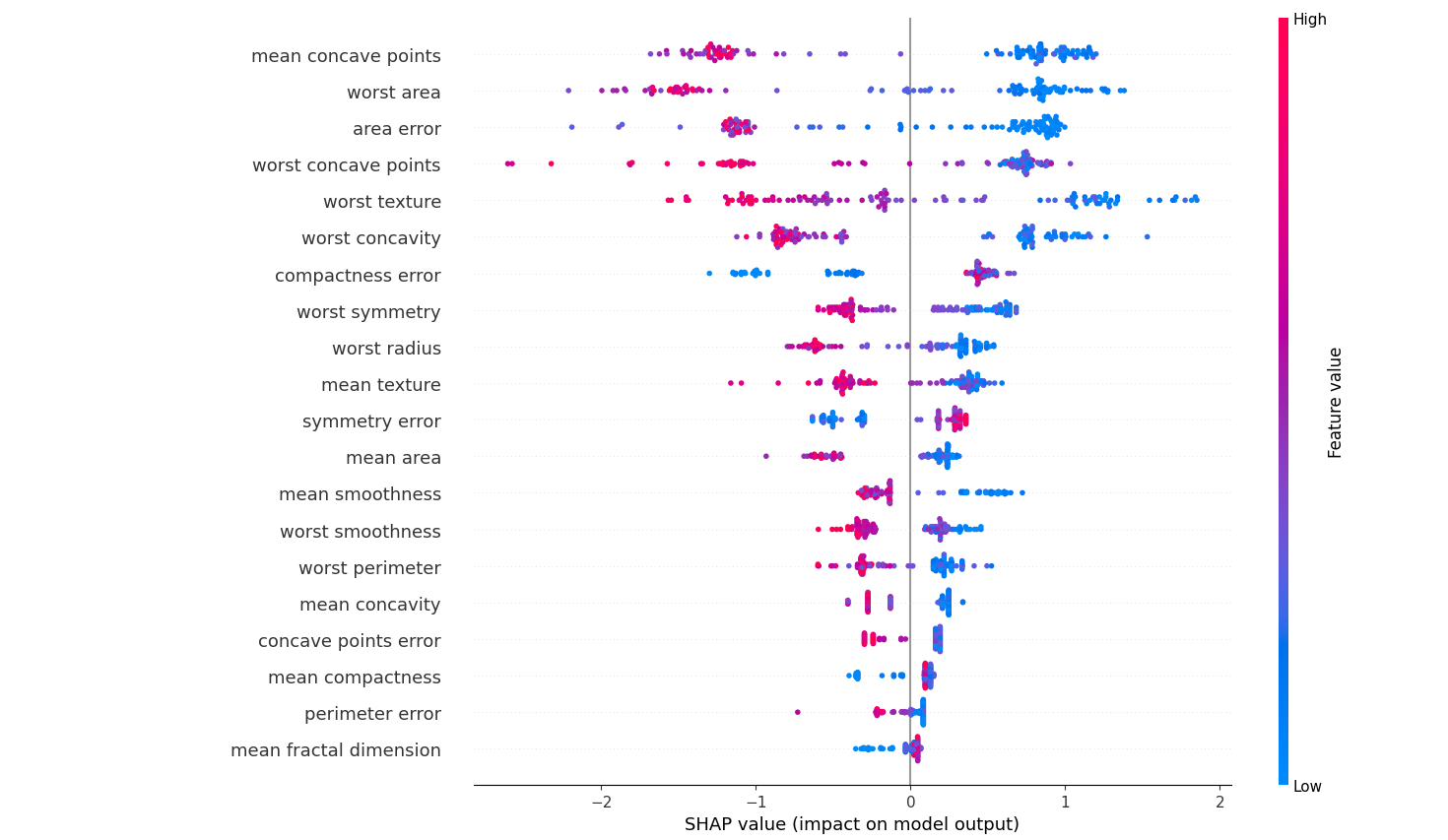}
    \caption{SHAP Summary Plot for Breast Cancer Prediction}
    \label{fig:shap_summary}
\end{figure}

\paragraph{Advantages of SHAP}

\begin{itemize}
    \item \textbf{Theoretically Sound:} SHAP values are grounded in cooperative game theory, offering a fair and consistent attribution of feature importance \cite{Shapley2016}.
    \item \textbf{Model-agnostic:} SHAP can be applied to any model, from linear models to deep neural networks and LLMs.
    \item \textbf{Local and Global Interpretability:} SHAP values provide both local explanations (for individual predictions) and global feature importance (across the entire dataset) \cite{Lundberg2017}.
\end{itemize}

\paragraph{Limitations}

Despite its versatility, SHAP has certain limitations \cite{Kumar2020}:
\begin{itemize}
    \item \textbf{High Computational Cost:} Calculating exact Shapley values requires evaluating all possible subsets of features, which is computationally expensive. Approximations like TreeSHAP are often used but may still be slow for large datasets.
    \item \textbf{Limited in High-dimensional Contexts:} SHAP values can be challenging to interpret in models with a large number of features or tokens, particularly for LLMs with long input sequences.
    \item \textbf{Dependence on Model Stability:} SHAP explanations may be less reliable for models that are sensitive to small changes in the input data or have high variance.
\end{itemize}

\paragraph{SHAP in Large Language Models}

For LLMs, SHAP can be adapted to explain token-level contributions by analyzing the importance of individual tokens or phrases in the context of text predictions \cite{Li2020}. However, this approach is computationally expensive due to the high dimensionality of LLM inputs. Further discussion on SHAP adaptations for LLMs will be provided in the subsequent chapters focused on interpretability techniques for NLP models.

\subsection{Local Interpretable Model-agnostic Explanations (LIME)}

Local Interpretable Model-agnostic Explanations (LIME) is a technique designed to explain the predictions of any machine learning model by approximating it locally with a simpler, interpretable model \cite{Ribeiro2016}. Unlike global interpretability methods that aim to explain the entire model, LIME focuses on explaining individual predictions, making it a popular choice for understanding complex models such as deep neural networks and large language models (LLMs). LIME is versatile and can be applied to tabular data, images, and text, making it well-suited for both traditional machine learning models and modern AI systems.

\paragraph{Scope of Application}

LIME is applicable for:
\begin{itemize}
    \item \textbf{Traditional Machine Learning Models:} It works well with models like decision trees, random forests, and support vector machines (SVMs) \cite{Ribeiro2016}.
    \item \textbf{Deep Neural Networks:} LIME can be used to interpret predictions from convolutional neural networks (CNNs) and recurrent neural networks (RNNs).
    \item \textbf{Large Language Models (LLMs):} LIME can approximate the importance of specific words or tokens in text-based predictions, aiding in the interpretability of natural language processing (NLP) models \cite{Garcia2020}.
\end{itemize}

\paragraph{Principles and Formula}

The core idea of LIME is to fit an interpretable model (e.g., a linear model or decision tree) locally around the instance being explained. It generates a perturbed dataset by slightly altering the input features and then observes how these changes affect the predictions. The interpretable model is trained on this perturbed dataset, providing an explanation of the original model's behavior around the specific instance.

Mathematically, LIME minimizes the following objective function \cite{Ribeiro2016}:

\[
\text{Explanation}(x) = \arg\min_{g \in G} \; L(f, g, \pi_x) + \Omega(g)
\]

where:
\begin{itemize}
    \item \(f\) is the original complex model.
    \item \(g\) is the interpretable model (e.g., a linear model).
    \item \(\pi_x\) is the locality measure, assigning weights to perturbed samples based on their proximity to the instance \(x\).
    \item \(L\) is the loss function measuring the fidelity of \(g\) in approximating \(f\).
    \item \(\Omega(g)\) is a regularization term ensuring the simplicity of the interpretable model.
\end{itemize}

\paragraph{Python Code Example}

In this example, we explore the application of Local Interpretable Model-agnostic Explanations (LIME) to elucidate the behavior of a text classification model trained on the IMDB movie review dataset \cite{Maas2011}. The model classifies each review as either positive or negative. LIME provides a way to understand which specific words or phrases drive the model's predictions, effectively opening the black box to show why a particular review was categorized as positive or negative.

\begin{lstlisting}[style=python, literate={\$}{{\$}}1]
import lime
import lime.lime_text
from sklearn.pipeline import make_pipeline
from sklearn.feature_extraction.text import TfidfVectorizer
from sklearn.linear_model import LogisticRegression
from sklearn.datasets import load_files
from sklearn.model_selection import train_test_split
import numpy as np

# Specify the path to the IMDB dataset. Ensure the dataset is downloaded and located in this path.
# The IMDB dataset can be downloaded from:
# https://ai.stanford.edu/~amaas/data/sentiment/aclImdb_v1.tar.gz

# 1. Load the IMDB dataset
data = load_files("aclImdb/train", categories=["pos", "neg"], encoding="utf-8", decode_error="replace")
X = np.array(data.data)
y = data.target  # Directly use data.target without further transformation

# 2. Split the dataset
X_train, X_test, y_train, y_test = train_test_split(X, y, test_size=0.2, random_state=42, stratify=y)

# 3. Create the text classification pipeline
vectorizer = TfidfVectorizer()
classifier = LogisticRegression(random_state=42, max_iter=1000)
pipeline = make_pipeline(vectorizer, classifier)

# 4. Train the model
pipeline.fit(X_train, y_train)

# 5. Initialize the LIME explainer
explainer = lime.lime_text.LimeTextExplainer(class_names=["NEGATIVE", "POSITIVE"])

# 6. Choose an instance to explain
text_instance = X_test[0]
exp = explainer.explain_instance(text_instance, pipeline.predict_proba, num_features=10)

# 7. Save the explanation as an HTML file
exp.save_to_file('lime_explanation.html')
print("LIME explanation saved as 'lime_explanation.html'")
\end{lstlisting}

\paragraph{Results Explanation}

The LIME explanation highlights the words in the review that significantly influence the model's prediction. As seen in the provided example, the model is decisively confident, predicting a positive sentiment with a 94\% probability, compared to a mere 6\% probability for negative. The chart of individual word contributions clearly shows which words push the model toward a positive classification, as well as the relative influence of each word.

The highlighted words—such as ``classic," ``best," ``Oscar," and ``excellent"—are the heavy hitters driving this review into positive territory. These words are not just casually positive; they scream 'quality' and 'worth.' In a review context, their appearance is a strong signal to the model that this review is a thumbs-up. The explanation provided by LIME allows us to pinpoint precisely which terms led to the positive classification, making it immediately obvious why the model is so confident.

\begin{figure}[!ht]
    \centering
    \includegraphics[width=0.8\textwidth]{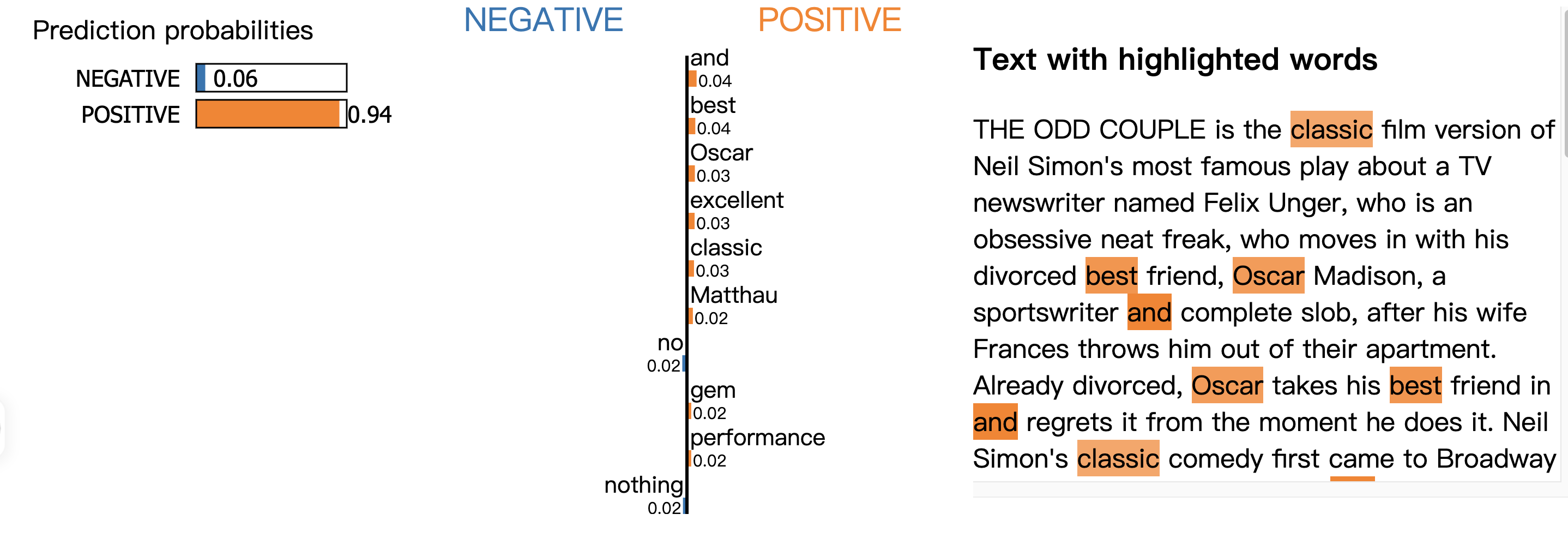}
    \caption{LIME explanation for a positive sentiment review. Words like ``classic," ``best," and ``excellent" contribute most strongly to the positive classification, as shown by the orange highlights and their corresponding weights.}
\end{figure}

\paragraph{Advantages of LIME}

\begin{itemize}
    \item \textbf{Model-agnostic:} LIME can be applied to any model, regardless of its architecture, making it highly versatile \cite{Ribeiro2016}.
    \item \textbf{Local Interpretability:} LIME provides explanations specific to individual predictions, which is useful for understanding complex models' decisions.
    \item \textbf{Flexibility:} LIME can be adapted to different data types, including tabular data, images, and text.
\end{itemize}

\paragraph{Limitations}

Despite its strengths, LIME has several limitations \cite{Zafar2019}:
\begin{itemize}
    \item \textbf{Instability:} The explanations can vary significantly with different perturbations, making the results less reliable.
    \item \textbf{Dependency on Perturbations:} LIME's explanations are highly dependent on the perturbed samples, which may not always represent the true data distribution.
    \item \textbf{Scalability Issues:} For large datasets or complex models, LIME can be computationally expensive due to the need for numerous model evaluations on perturbed samples.
\end{itemize}

\paragraph{LIME for Large Language Models}

For LLMs, LIME can approximate the importance of individual tokens or phrases by creating perturbed versions of the input text and analyzing the change in prediction probabilities \cite{Garcia2020}. However, due to the high dimensionality and context-dependent nature of LLMs, interpreting these token-level explanations can be challenging. Further discussion on adapting LIME for LLMs will be provided in the chapter on interpretability techniques for NLP models.

\subsection{Integrated Gradients}

Integrated Gradients (IG) is a gradient-based attribution method used for explaining the predictions of deep neural networks. It addresses the shortcomings of simpler gradient-based techniques, such as the "saturation problem," by integrating the gradients along a path from a baseline input to the actual input \cite{Sundararajan2017}. IG is widely used for neural networks in both computer vision and natural language processing (NLP) applications, including large language models (LLMs).

\paragraph{Scope of Application}

Integrated Gradients is particularly effective for:
\begin{itemize}
    \item \textbf{Deep Neural Networks:} It works well for convolutional neural networks (CNNs) in image classification tasks and recurrent neural networks (RNNs) in sequence analysis.
    \item \textbf{Transformer-based Models:} IG can be adapted for interpreting attention mechanisms in models like BERT and GPT, making it useful for explaining LLM predictions \cite{Wallace2019}.
    \item \textbf{Other Complex Models:} IG is generally applicable to any differentiable model where input features contribute to the output in a non-linear manner.
\end{itemize}

\paragraph{Principles and Formula}

The key idea behind Integrated Gradients is to compute the accumulated gradients of the model's output with respect to the input features as the input transitions from a baseline (e.g., a zero vector or an all-blank input) to the actual input. The attribution score for each feature is obtained by summing these gradients along the integration path \cite{Sundararajan2017}.

Mathematically, the attribution for feature \(x_i\) is given by:

\[
\text{IG}_i(x) = (x_i - x_i') \int_{\alpha=0}^{1} \frac{\partial f(x' + \alpha (x - x'))}{\partial x_i} \, d\alpha
\]

where:
\begin{itemize}
    \item \(x\) is the actual input.
    \item \(x'\) is the baseline input (e.g., a zero vector).
    \item \(\alpha\) is the interpolation parameter, ranging from 0 to 1.
    \item \(f(x)\) is the model's output for input \(x\).
    \item \(\frac{\partial f}{\partial x_i}\) is the gradient of the model's output with respect to feature \(x_i\).
\end{itemize}

The integral is typically approximated using the Riemann sum, where the input is incrementally interpolated between the baseline and the actual input.

\paragraph{Python Code Example}

In this example, we apply Integrated Gradients to a convolutional neural network (CNN) trained on the MNIST dataset, aiming to reveal which pixels contribute most to the model's prediction for a specific digit \cite{LeCun2010}. Integrated Gradients helps by offering a measure of pixel importance, thereby illustrating the areas the model focuses on when making a prediction.

\begin{lstlisting}[style=python, literate={\$}{{\$}}1]
import tensorflow as tf
import numpy as np
import matplotlib.pyplot as plt

# Define a simple CNN model in TensorFlow/Keras
class SimpleCNN(tf.keras.Model):
    def __init__(self):
        super(SimpleCNN, self).__init__()
        self.conv1 = tf.keras.layers.Conv2D(32, (3, 3), padding="same", activation="relu")
        self.conv2 = tf.keras.layers.Conv2D(64, (3, 3), padding="same", activation="relu")
        self.flatten = tf.keras.layers.Flatten()
        self.fc1 = tf.keras.layers.Dense(128, activation="relu")
        self.fc2 = tf.keras.layers.Dense(10, activation="softmax")

    def call(self, x):
        x = self.conv1(x)
        x = self.conv2(x)
        x = self.flatten(x)
        x = self.fc1(x)
        return self.fc2(x)

# Load the MNIST dataset
(train_images, train_labels), _ = tf.keras.datasets.mnist.load_data()
train_images = train_images[..., tf.newaxis] / 255.0  # Rescale to [0, 1]

# Initialize the model, loss function, and optimizer
model = SimpleCNN()
model.compile(optimizer=tf.keras.optimizers.Adam(0.001),
              loss=tf.keras.losses.SparseCategoricalCrossentropy(),
              metrics=["accuracy"])

# Train the model (one epoch for simplicity)
model.fit(train_images, train_labels, batch_size=64, epochs=1)

# Select a sample image and baseline for Integrated Gradients
sample_image = train_images[0:1]  # Shape (1, 28, 28, 1)
baseline = tf.zeros_like(sample_image)

# Function to calculate Integrated Gradients
def compute_integrated_gradients(model, input_image, baseline, target_class_idx, m_steps=50):
    # Generate interpolated images between baseline and input
    interpolated_images = [
        baseline + (float(i) / m_steps) * (input_image - baseline)
        for i in range(m_steps + 1)
    ]
    interpolated_images = tf.concat(interpolated_images, axis=0)

    with tf.GradientTape() as tape:
        tape.watch(interpolated_images)
        # Get model predictions for interpolated images
        predictions = model(interpolated_images)
        target_predictions = predictions[:, target_class_idx]

    # Compute gradients between predictions and interpolated images
    grads = tape.gradient(target_predictions, interpolated_images)

    # Average gradients and compute attributions
    avg_grads = tf.reduce_mean(grads, axis=0)
    integrated_grads = (input_image - baseline) * avg_grads
    return integrated_grads

# Compute attributions using Integrated Gradients
target_class = train_labels[0]  # The true class for the sample image
attributions = compute_integrated_gradients(model, sample_image, baseline, target_class)

# Visualize the attributions
attributions = attributions.numpy().squeeze()
plt.imshow(attributions, cmap="hot", interpolation="nearest")
plt.colorbar()
plt.title("Integrated Gradients Attribution for MNIST Prediction")
plt.show()
\end{lstlisting}

\paragraph{Results Explanation}

The heatmap below displays the attributions calculated by Integrated Gradients \cite{Sundararajan2017} for a single MNIST image. The color intensity indicates the importance of each pixel to the model's prediction: red regions represent pixels that strongly support the predicted class, while negative values, if any, would be shown in cooler colors. In this case, the pixels that form the shape of the digit are clearly highlighted, validating that the model's prediction relies on the relevant parts of the image.

\begin{figure}[!ht]
    \centering
    \includegraphics[width=0.7\textwidth]{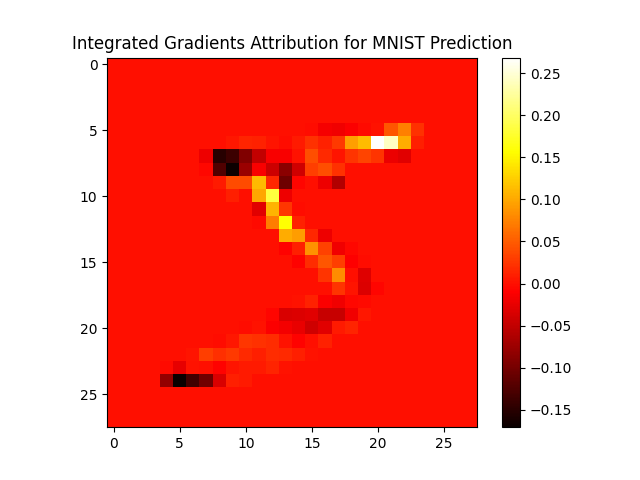}
    \caption{Integrated Gradients attribution map for a sample MNIST image. The highlighted pixels reveal the model's focus areas when classifying the digit.}
\end{figure}

\paragraph{Advantages of Integrated Gradients}

\begin{itemize}
    \item \textbf{Theoretical Soundness:} IG satisfies desirable properties such as completeness, meaning the attributions sum up to the difference between the model's output for the actual input and the baseline.
    \item \textbf{Model-agnostic:} IG can be applied to any differentiable model, making it versatile for various neural network architectures, including CNNs and LLMs.
    \item \textbf{No Gradient Saturation:} IG mitigates the saturation problem seen in simpler gradient-based methods by integrating gradients over a range of inputs.
\end{itemize}

\paragraph{Limitations}

Integrated Gradients, while powerful, also has some limitations:
\begin{itemize}
    \item \textbf{Choice of Baseline:} The results are sensitive to the choice of baseline input. An inappropriate baseline may yield misleading attributions.
    \item \textbf{Computational Cost:} IG requires multiple model evaluations for different interpolated inputs, making it computationally expensive for large models or high-dimensional data.
    \item \textbf{Interpretability in NLP:} While IG can be adapted for LLMs, interpreting token-level attributions can be challenging due to the complex, context-dependent nature of language models.
\end{itemize}

\paragraph{Integrated Gradients in Large Language Models}

For LLMs, Integrated Gradients can be used to attribute the model's output to individual input tokens by computing gradients with respect to the embedding layer. This approach helps identify important words or phrases that influence the model's prediction, making it useful for understanding model decisions in tasks like sentiment analysis and text classification. Detailed adaptations of IG for LLMs will be explored in the upcoming sections on NLP model interpretability.

\subsection{DeepLIFT (Deep Learning Important FeaTures)}

DeepLIFT (Deep Learning Important FeaTures) \cite{Shrikumar2017} is an attribution method designed to interpret the predictions of deep neural networks by assigning contribution scores to input features. Unlike simple gradient-based techniques, which can suffer from issues like gradient saturation, DeepLIFT compares the activation of neurons to their "reference" activations, providing a more robust and reliable attribution. DeepLIFT is particularly effective for deep neural networks, including convolutional neural networks (CNNs) and transformer-based models like BERT and GPT.

\paragraph{Scope of Application}

DeepLIFT is best suited for:
\begin{itemize}
    \item \textbf{Deep Neural Networks:} It is designed for interpreting deep learning models, including convolutional and recurrent neural networks.
    \item \textbf{Transformer Models:} DeepLIFT can be adapted to explain transformer-based models used in natural language processing (NLP), such as BERT and GPT.
    \item \textbf{Complex, Non-linear Models:} DeepLIFT is effective for models where simple gradient methods may fail due to non-linearity and activation saturation.
\end{itemize}

\paragraph{Principles and Formula}

The core idea behind DeepLIFT is to decompose the model's output based on the difference between the actual activation and a reference (baseline) activation. Instead of using raw gradients, DeepLIFT assigns contribution scores by propagating these differences through the network layers.

For a given neuron \(i\) with input \(x_i\), the contribution score \(C_{\Delta x_i}\) is defined as:

\[
C_{\Delta x_i} = (x_i - x_i^{\text{ref}}) \cdot \frac{\Delta y}{\Delta x_i}
\]

where:
\begin{itemize}
    \item \(x_i\) is the input value.
    \item \(x_i^{\text{ref}}\) is the reference input value (baseline).
    \item \(\Delta y\) is the change in the output of the network compared to its reference output.
    \item \(\Delta x_i\) is the change in the input value compared to its reference.
\end{itemize}

DeepLIFT uses two main rules for propagating contributions:
\begin{itemize}
    \item \textbf{Rescale Rule:} For ReLU and similar activation functions, this rule rescales contributions based on the change in activation.
    \item \textbf{Reveal-Cancel Rule:} This rule helps handle cases where activations cancel out, ensuring that only meaningful changes are attributed.
\end{itemize}

\paragraph{Python Code Example}

In this example, we employ DeepLIFT to interpret the predictions of a convolutional neural network (CNN) trained on the MNIST dataset \cite{Kokhlikyan2020}. DeepLIFT is used to analyze which pixels in the input image contribute most significantly to the model's decision. This approach highlights the areas that are most influential, helping us understand how the model arrives at a particular classification for a digit.

\begin{lstlisting}[style=python, literate={\$}{{\$}}1]
import torch
import torch.nn as nn
import torch.optim as optim
from captum.attr import DeepLift
import numpy as np
import matplotlib.pyplot as plt
from torchvision import datasets, transforms
from torch.utils.data import DataLoader

# Convert TensorFlow data to PyTorch format
train_transform = transforms.Compose([transforms.ToTensor()])
test_transform = transforms.Compose([transforms.ToTensor()])

# Load the MNIST dataset using torchvision
train_dataset = datasets.MNIST(root='./data', train=True, download=True, transform=train_transform)
test_dataset = datasets.MNIST(root='./data', train=False, download=True, transform=test_transform)

train_loader = DataLoader(train_dataset, batch_size=64, shuffle=True)
test_loader = DataLoader(test_dataset, batch_size=1, shuffle=True)

# Define a simple CNN model in PyTorch
class SimpleCNN(nn.Module):
    def __init__(self):
        super(SimpleCNN, self).__init__()
        self.conv1 = nn.Conv2d(1, 32, kernel_size=3, padding=1)
        self.pool = nn.MaxPool2d(2, 2)
        self.fc1 = nn.Linear(32 * 14 * 14, 128)
        self.fc2 = nn.Linear(128, 10)

    def forward(self, x):
        x = torch.relu(self.conv1(x))
        x = self.pool(x)
        x = x.view(-1, 32 * 14 * 14)
        x = torch.relu(self.fc1(x))
        x = torch.softmax(self.fc2(x), dim=1)
        return x

# Initialize the model, loss function, and optimizer
model = SimpleCNN()
criterion = nn.CrossEntropyLoss()
optimizer = optim.Adam(model.parameters(), lr=0.001)

# Train the model for one epoch (for simplicity)
for images, labels in train_loader:
    optimizer.zero_grad()
    outputs = model(images)
    loss = criterion(outputs, labels)
    loss.backward()
    optimizer.step()
    break  # Train on one batch for demonstration

# Select a sample image and its baseline
sample_image, sample_label = next(iter(test_loader))
baseline = torch.zeros_like(sample_image)

# Compute DeepLIFT attributions with the target label
dl = DeepLift(model)
attributions = dl.attribute(sample_image, baseline, target=sample_label.item())

# Visualize the attributions
attributions = attributions.detach().numpy().squeeze()
plt.imshow(attributions, cmap='hot', interpolation='nearest')
plt.colorbar()
plt.title("DeepLIFT Attribution for MNIST Prediction")
plt.show()
\end{lstlisting}

\paragraph{Results Explanation}

The heatmap below illustrates the DeepLIFT attributions for a single MNIST image. The color intensity indicates the influence of each pixel on the model's decision: red areas represent pixels that contribute positively to the prediction, while cooler colors (if present) would indicate pixels that detract from it. In this instance, the highlighted pixels trace the shape of the digit, suggesting that the model indeed bases its classification on the relevant features of the image.

\begin{figure}[htbp]
    \centering
    \includegraphics[width=0.7\textwidth]{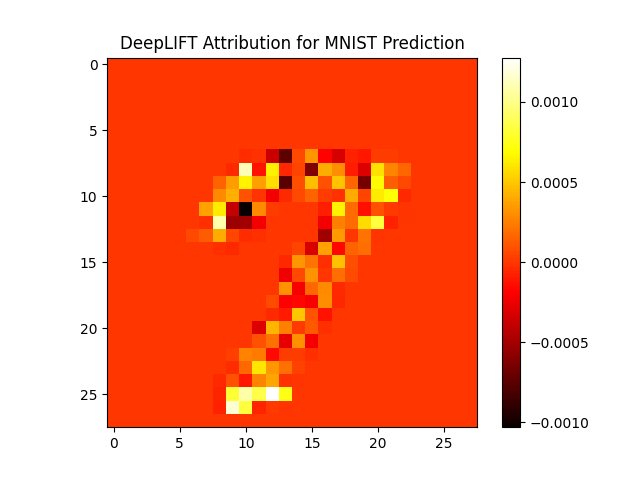}
    \caption{DeepLIFT attribution map for a sample MNIST image. The highlighted areas reveal the model's focus regions when classifying the digit.}
\end{figure}

\paragraph{Advantages of DeepLIFT}

\begin{itemize}
    \item \textbf{Handles Saturation:} DeepLIFT mitigates the gradient saturation problem by comparing activations to a baseline, providing more meaningful attributions.
    \item \textbf{Efficient Computation:} Unlike Integrated Gradients, which requires multiple evaluations, DeepLIFT only requires a single backward pass, making it faster.
    \item \textbf{Model-agnostic:} DeepLIFT can be applied to any neural network architecture, including CNNs, RNNs, and transformers.
\end{itemize}

\paragraph{Limitations}

DeepLIFT, despite its strengths, has some limitations:
\begin{itemize}
    \item \textbf{Choice of Baseline:} The attributions are sensitive to the choice of reference input. An inappropriate baseline can lead to misleading explanations.
    \item \textbf{Complexity in NLP Models:} Applying DeepLIFT to transformer-based LLMs can be challenging due to the intricate interactions between tokens and attention mechanisms.
    \item \textbf{Assumption of Linearity:} DeepLIFT assumes a linear relationship between changes in input and changes in activation, which may not hold in highly non-linear models.
\end{itemize}

\paragraph{DeepLIFT in Large Language Models}

For LLMs, DeepLIFT can be adapted to provide token-level attributions by propagating contribution scores through the embedding and transformer layers. This helps in identifying the most influential words or phrases for the model's predictions, making it a valuable tool for interpreting NLP models. However, due to the complex interactions within transformer layers, interpreting these attributions requires careful analysis. A detailed exploration of DeepLIFT adaptations for LLMs will be discussed in later chapters.

\subsection{Saliency Maps}

Saliency maps are one of the simplest and most intuitive post-hoc interpretation techniques for neural networks, particularly in the context of deep learning models like convolutional neural networks (CNNs) and transformer-based architectures \cite{Simonyan2013}. The main idea is to use the gradient of the model's output with respect to the input features to identify which parts of the input are most influential in the model's prediction. Saliency maps provide a visual representation of these important regions, making them useful for tasks in computer vision and natural language processing (NLP).

\paragraph{Scope of Application}

Saliency maps are most effective for:
\begin{itemize}
    \item \textbf{Convolutional Neural Networks (CNNs):} They are commonly used in image classification to highlight important pixels that contribute to the model's decision.
    \item \textbf{Recurrent Neural Networks (RNNs):} In sequence-based models, saliency maps can help identify important time steps or tokens in tasks like sentiment analysis.
    \item \textbf{Transformer Models (e.g., BERT, GPT):} Saliency maps can be adapted to visualize the importance of individual tokens or phrases in large language models (LLMs).
\end{itemize}

\paragraph{Principles and Formula}

The core idea of saliency maps is to use the gradient of the model's output with respect to the input features to identify important regions. Mathematically, given a neural network \(f(x)\) and an input \(x\), the saliency map \(S(x)\) for class \(c\) is defined as:

\[
S(x) = \left| \frac{\partial f_c(x)}{\partial x} \right|,
\]

where:
\begin{itemize}
    \item \(f_c(x)\) is the output of the model for class \(c\).
    \item \(\frac{\partial f_c(x)}{\partial x}\) is the gradient of the output with respect to the input \(x\).
\end{itemize}

This gradient indicates how sensitive the output \(f_c(x)\) is to changes in each input feature \(x_i\). Features with higher gradient magnitudes are deemed more important for the prediction.

\paragraph{Python Code Example}

In this example, we employ a pre-trained convolutional neural network (CNN) model to generate a saliency map for an image from the MNIST dataset. Saliency maps provide insight into which pixels in the input image are most influential in the model's prediction. This helps in understanding the model's focus areas and its decision-making process.

\begin{lstlisting}[style=python, literate={\$}{{\$}}1]
import tensorflow as tf
import numpy as np
import matplotlib.pyplot as plt

# Load the MNIST dataset
(train_images, train_labels), (test_images, test_labels) = tf.keras.datasets.mnist.load_data()
train_images = train_images[..., np.newaxis] / 255.0
test_images = test_images[..., np.newaxis] / 255.0

# Define a simple CNN model
model = tf.keras.Sequential([
    tf.keras.layers.Conv2D(32, (3, 3), activation='relu', input_shape=(28, 28, 1)),
    tf.keras.layers.MaxPooling2D((2, 2)),
    tf.keras.layers.Flatten(),
    tf.keras.layers.Dense(128, activation='relu'),
    tf.keras.layers.Dense(10, activation='softmax')
])

# Compile and train the model
model.compile(optimizer='adam', loss='sparse_categorical_crossentropy', metrics=['accuracy'])
model.fit(train_images, train_labels, epochs=1, batch_size=64)

# Function to compute the saliency map
def compute_saliency_map(model, image, target_class):
    image = tf.convert_to_tensor(image[np.newaxis, ...], dtype=tf.float32)

    with tf.GradientTape() as tape:
        tape.watch(image)
        predictions = model(image)
        loss = predictions[0, target_class]

    # Compute the gradient of the loss with respect to the input image
    gradient = tape.gradient(loss, image)
    saliency = tf.abs(gradient)[0]

    # Normalize the saliency map
    saliency = saliency.numpy().squeeze()
    saliency = (saliency - saliency.min()) / (saliency.max() - saliency.min())

    return saliency

# Select a sample image and compute the saliency map
sample_image = test_images[0]
target_class = np.argmax(model.predict(sample_image[np.newaxis, ...]))
saliency_map = compute_saliency_map(model, sample_image, target_class)

# Visualize the original image and its saliency map
plt.subplot(1, 2, 1)
plt.imshow(sample_image.squeeze(), cmap='gray')
plt.title("Original Image")

plt.subplot(1, 2, 2)
plt.imshow(saliency_map, cmap='hot')
plt.title("Saliency Map")
plt.colorbar()
plt.show()
\end{lstlisting}

\paragraph{Results Explanation}

The saliency map below highlights the most important pixels in the input image that contributed to the model's prediction. In the case of an MNIST digit, the saliency map typically focuses on the strokes that form the digit, indicating that these regions had the highest influence on the model's classification decision. By examining the saliency map, we gain an understanding of how the model perceives the structure of the digit and which features it considers most crucial for accurate classification.

\begin{figure}[htbp]
    \centering
    \includegraphics[width=0.7\textwidth]{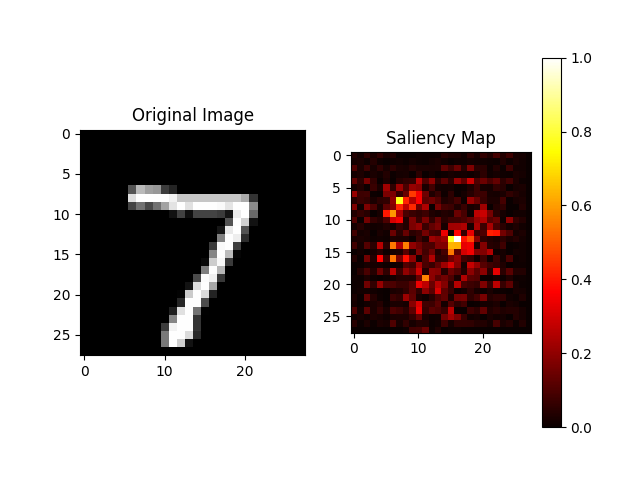}
    \caption{Saliency map for an MNIST image. The map shows which pixels the model focused on when predicting the digit, with warmer colors indicating higher importance.}
\end{figure}

\paragraph{Advantages of Saliency Maps}

\begin{itemize}
    \item \textbf{Simplicity:} Saliency maps are easy to compute and understand, making them a popular choice for visual explanations.
    \item \textbf{Model-agnostic:} They can be applied to any differentiable model, including CNNs, RNNs, and transformer-based LLMs.
    \item \textbf{Real-time Feedback:} Saliency maps can be computed quickly, providing immediate insights into model behavior.
\end{itemize}

\paragraph{Limitations}

Despite their simplicity, saliency maps have several limitations:
\begin{itemize}
    \item \textbf{Gradient Saturation:} If the model's output is saturated, the gradient values may be close to zero, making the saliency map less informative.
    \item \textbf{Noise and Artifacts:} Saliency maps can be noisy and may highlight irrelevant features, especially in high-dimensional inputs.
    \item \textbf{Limited Interpretability for LLMs:} Applying saliency maps to LLMs is challenging due to the complex token interactions and lack of direct spatial relationships in text data.
\end{itemize}

\paragraph{Saliency Maps for Large Language Models}

For LLMs, saliency maps can be adapted by computing gradients with respect to the input embeddings rather than the raw tokens. This approach helps identify important tokens or phrases influencing the model's predictions. However, due to the intricate attention mechanisms in transformer models, interpreting these saliency maps requires careful analysis. More detailed adaptations for LLMs will be discussed in subsequent sections focused on NLP interpretability techniques.

\subsection{SmoothGrad}

SmoothGrad is a post-hoc interpretation technique designed to enhance the clarity of gradient-based saliency maps, which often suffer from noise and visual artifacts \cite{smilkov2017smoothgrad}. By averaging gradients over multiple noisy inputs, SmoothGrad produces smoother and more visually interpretable saliency maps, making it a valuable tool for explaining predictions in both computer vision and natural language processing (NLP) models, including large language models (LLMs).

\paragraph{Scope of Application}

SmoothGrad is particularly effective for:
\begin{itemize}
    \item \textbf{Convolutional Neural Networks (CNNs):} It helps visualize input features that contribute to predictions in image classification tasks.
    \item \textbf{Recurrent Neural Networks (RNNs):} It can be used to interpret sequence models in tasks like sentiment analysis.
    \item \textbf{Transformer-based Models:} SmoothGrad can be adapted for explaining predictions in transformer architectures, including BERT and GPT, by smoothing token-level attributions.
\end{itemize}

\paragraph{Principles and Formula}

The key idea of SmoothGrad is to generate multiple noisy versions of the input and average the resulting gradients to reduce noise and enhance important features. Given an input \(x\), SmoothGrad creates \(n\) noisy samples by adding Gaussian noise \(\mathcal{N}(0, \sigma^2)\) and computes the gradient of the model's output with respect to each noisy sample.

The SmoothGrad attribution for input feature \(x_i\) is defined as:

\[
\text{SmoothGrad}(x_i) = \frac{1}{n} \sum_{k=1}^{n} \frac{\partial f(x_k)}{\partial x_i}, \quad \text{where } x_k = x + \mathcal{N}(0, \sigma^2)
\]

where:
\begin{itemize}
    \item \(f(x)\) is the model's prediction for input \(x\).
    \item \(x_k\) is the \(k\)-th noisy sample generated by adding Gaussian noise.
    \item \(\sigma\) is the standard deviation of the Gaussian noise, controlling the amount of perturbation.
    \item \(n\) is the number of noisy samples used for averaging.
\end{itemize}

By averaging gradients over these noisy samples, SmoothGrad reduces the influence of noise in the input, making the saliency maps more interpretable.

\paragraph{Python Code Example}

In this example, we apply SmoothGrad to explain the predictions of a convolutional neural network (CNN) trained on the MNIST dataset. SmoothGrad enhances standard gradient-based interpretations by adding noise to the input and averaging the results, producing a smoother and more interpretable saliency map that highlights the pixels contributing most significantly to the prediction.

\begin{lstlisting}[style=python, literate={\$}{{\$}}1]
import tensorflow as tf
import numpy as np
import matplotlib.pyplot as plt

# Load the MNIST dataset
(train_images, train_labels), (test_images, test_labels) = tf.keras.datasets.mnist.load_data()
train_images = train_images[..., np.newaxis] / 255.0
test_images = test_images[..., np.newaxis] / 255.0

# Define a simple CNN model
model = tf.keras.Sequential([
    tf.keras.layers.Conv2D(32, (3, 3), activation='relu', input_shape=(28, 28, 1)),
    tf.keras.layers.MaxPooling2D((2, 2)),
    tf.keras.layers.Flatten(),
    tf.keras.layers.Dense(128, activation='relu'),
    tf.keras.layers.Dense(10, activation='softmax')
])

# Compile and train the model
model.compile(optimizer='adam', loss='sparse_categorical_crossentropy', metrics=['accuracy'])
model.fit(train_images, train_labels, epochs=1, batch_size=64)

# Function to compute SmoothGrad
def smoothgrad(image, model, target_class, num_samples=50, noise_level=0.1):
    grads = []
    for _ in range(num_samples):
        noise = np.random.normal(0, noise_level, image.shape)
        noisy_image = image + noise
        noisy_image = tf.convert_to_tensor(noisy_image[np.newaxis, ...], dtype=tf.float32)

        with tf.GradientTape() as tape:
            tape.watch(noisy_image)
            prediction = model(noisy_image)
            loss = prediction[0, target_class]

        gradient = tape.gradient(loss, noisy_image)
        grads.append(gradient.numpy().squeeze())

    # Average the gradients
    smooth_grad = np.mean(grads, axis=0)
    return smooth_grad

# Select a sample image and compute SmoothGrad attributions
sample_image = test_images[0]
target_class = np.argmax(model.predict(sample_image[np.newaxis, ...]))
attributions = smoothgrad(sample_image, model, target_class)

# Visualize the attributions
plt.imshow(attributions, cmap='hot', interpolation='nearest')
plt.colorbar()
plt.title("SmoothGrad Attribution for MNIST Prediction")
plt.show()
\end{lstlisting}

\paragraph{Results Explanation}

The heatmap below displays the SmoothGrad attributions for a single MNIST image. By averaging gradients over multiple noisy inputs, SmoothGrad produces a clearer, more focused visualization than standard gradients. The highlighted regions indicate the pixels that contributed most significantly to the model's prediction. These areas typically correspond to the strokes that define the digit, offering insight into which parts of the image were most influential in the model's classification.

\begin{figure}[!ht]
    \centering
    \includegraphics[width=0.7\textwidth]{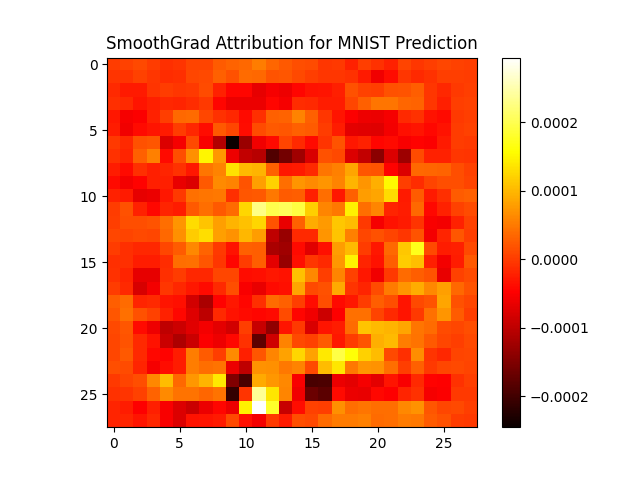}
    \caption{SmoothGrad attribution map for a sample MNIST image. The map reveals the most influential regions, with warmer colors indicating higher pixel importance in the model's prediction.}
\end{figure}

\paragraph{Advantages of SmoothGrad}

\begin{itemize}
    \item \textbf{Noise Reduction:} By averaging gradients over multiple noisy inputs, SmoothGrad reduces the noise typically seen in gradient-based saliency maps \cite{smilkov2017smoothgrad}.
    \item \textbf{Improved Interpretability:} The resulting saliency maps are smoother and more focused on important features, making it easier to interpret model predictions.
    \item \textbf{Model-agnostic:} SmoothGrad can be applied to any differentiable model, including CNNs, RNNs, and transformer-based models like BERT and GPT.
\end{itemize}

\paragraph{Limitations}

Despite its benefits, SmoothGrad has some limitations:
\begin{itemize}
    \item \textbf{Choice of Noise Level:} The results are sensitive to the noise level \(\sigma\). Choosing an inappropriate noise level can either obscure important features or introduce artificial artifacts.
    \item \textbf{Computational Overhead:} Computing SmoothGrad requires multiple forward and backward passes, making it computationally expensive for large models or datasets.
    \item \textbf{Interpretability in LLMs:} While SmoothGrad can be adapted for transformer models, interpreting token-level attributions remains challenging due to the complex interactions in LLMs.
\end{itemize}

\paragraph{SmoothGrad for Large Language Models}

In LLMs, SmoothGrad can be applied by adding noise to the embedding layer and averaging the resulting gradients with respect to input tokens. This helps identify important words or phrases that influence the model's output, providing insights into how LLMs make predictions. However, due to the high dimensionality and context dependencies in LLMs, interpreting these attributions requires careful consideration. A detailed discussion of SmoothGrad adaptations for LLMs will be covered in later chapters on NLP model interpretability.

\subsection{Grad-CAM and Grad-CAM++}

Grad-CAM (Gradient-weighted Class Activation Mapping) \cite{selvaraju2017grad} and its extension Grad-CAM++ \cite{chattopadhay2018gradcampp} are widely used techniques for visualizing and interpreting deep neural network predictions, especially in convolutional neural networks (CNNs). These methods produce heatmaps that highlight regions of the input image that are important for a specific prediction. Grad-CAM is effective for understanding image classification models, while Grad-CAM++ addresses some of the limitations of Grad-CAM by providing better localization.

\paragraph{Scope of Application}

Grad-CAM and Grad-CAM++ are suitable for:
\begin{itemize}
    \item \textbf{Convolutional Neural Networks (CNNs):} Both methods are designed for CNNs, making them effective for tasks like image classification, object detection, and segmentation.
    \item \textbf{Transformer-based Models:} Grad-CAM can be adapted for transformer models like Vision Transformers (ViTs) by applying it to convolutional layers in the embedding stage.
    \item \textbf{Natural Language Processing (NLP):} Grad-CAM can be extended to transformer-based LLMs by visualizing attention maps over text tokens, although this requires specific adaptations due to the absence of convolutional layers.
\end{itemize}

\paragraph{Principles and Formula}

The core idea of Grad-CAM is to compute the gradient of the model's output with respect to the activations of a specific convolutional layer. This gradient information is used to weight the feature maps, highlighting the most important regions for the prediction.

For a given input image \(x\) and class \(c\), the Grad-CAM heatmap \(L^{\text{Grad-CAM}}_{\text{c}}\) is computed as:

\[
L^{\text{Grad-CAM}}_{\text{c}} = \text{ReLU}\left(\sum_{k} \alpha^{c}_{k} A^{k}\right)
\]

where:
\begin{itemize}
    \item \(A^{k}\) is the activation map of the \(k\)-th convolutional filter.
    \item \(\alpha^{c}_{k}\) is the weight for filter \(k\), calculated as:

    \[
    \alpha^{c}_{k} = \frac{1}{Z} \sum_{i} \sum_{j} \frac{\partial y^{c}}{\partial A^{k}_{ij}}
    \]

    \item \(Z\) is the total number of pixels in the activation map.
    \item \(\text{ReLU}\) ensures that only positive contributions are considered.
\end{itemize}

Grad-CAM++ refines this approach by considering higher-order derivatives, improving localization and providing better explanations for multi-object scenarios.

\paragraph{Python Code Example}

In this example, we apply Grad-CAM to a pre-trained CNN model using TensorFlow and Keras to explain its predictions on an image from the ImageNet dataset \cite{simonyan2014verydeep}.

\begin{lstlisting}[style=python, literate={\$}{{\$}}1]
import tensorflow as tf
import numpy as np
import matplotlib.pyplot as plt
from tensorflow.keras.applications import VGG16
from tensorflow.keras.applications.vgg16 import preprocess_input, decode_predictions
from tensorflow.keras.preprocessing.image import img_to_array, load_img
import cv2  

# Load a pre-trained VGG16 model
model = VGG16(weights='imagenet')

# Load and preprocess the input image
image_path = 'Ch04/Images/cat.jpg'
image = load_img(image_path, target_size=(224, 224))

# Convert the image to an array and preprocess it
image_array = img_to_array(image)
image_array = np.expand_dims(image_array, axis=0)
image_array = preprocess_input(image_array)

# Get the model prediction
predictions = model.predict(image_array)
predicted_class = np.argmax(predictions[0])

# Function to compute Grad-CAM heatmap
def compute_gradcam(model, image_array, class_idx, layer_name='block5_conv3'):
    # Create a model that maps the input image to the activations of the last convolutional layer
    # and the model's output
    grad_model = tf.keras.models.Model(
        [model.inputs], [model.get_layer(layer_name).output, model.output]
    )

    # Record operations for automatic differentiation
    with tf.GradientTape() as tape:
        conv_output, predictions = grad_model(image_array)
        loss = predictions[:, class_idx]

    # Compute gradients with respect to the convolutional output
    grads = tape.gradient(loss, conv_output)

    # Compute the mean intensity of the gradients for each channel
    pooled_grads = tf.reduce_mean(grads, axis=(0, 1, 2))

    # Extract the feature maps from the convolutional layer output
    conv_output = conv_output[0]

    # Compute the weighted sum of the feature maps
    heatmap = tf.reduce_sum(tf.multiply(pooled_grads, conv_output), axis=-1)

    # Apply ReLU and normalize the heatmap
    heatmap = np.maximum(heatmap, 0)
    heatmap /= np.max(heatmap)

    return heatmap

# Generate the Grad-CAM heatmap
heatmap = compute_gradcam(model, image_array, predicted_class)

# Resize heatmap to match the input image size
heatmap = cv2.resize(heatmap, (224, 224))  
heatmap = np.uint8(255 * heatmap)  

heatmap_color = cv2.applyColorMap(heatmap, cv2.COLORMAP_JET)

superimposed_img = cv2.addWeighted(cv2.cvtColor(np.array(image), cv2.COLOR_RGB2BGR), 0.6, heatmap_color, 0.4, 0)

# Display the original image, heatmap, and overlay
fig, ax = plt.subplots(1, 3, figsize=(18, 6))

# Display original image
ax[0].imshow(image)
ax[0].axis('off')
ax[0].set_title("Original Image")

# Display the heatmap only
ax[1].imshow(heatmap, cmap='jet')
ax[1].axis('off')
ax[1].set_title("Grad-CAM Heatmap")

# Display the overlay image
ax[2].imshow(cv2.cvtColor(superimposed_img, cv2.COLOR_BGR2RGB))
ax[2].axis('off')
ax[2].set_title("Overlay Image")

# Show the plot
plt.tight_layout()
plt.show()

\end{lstlisting}

\paragraph{Results Explanation}

\begin{figure}[htbp]
    \centering
    \includegraphics[width=0.9\textwidth]{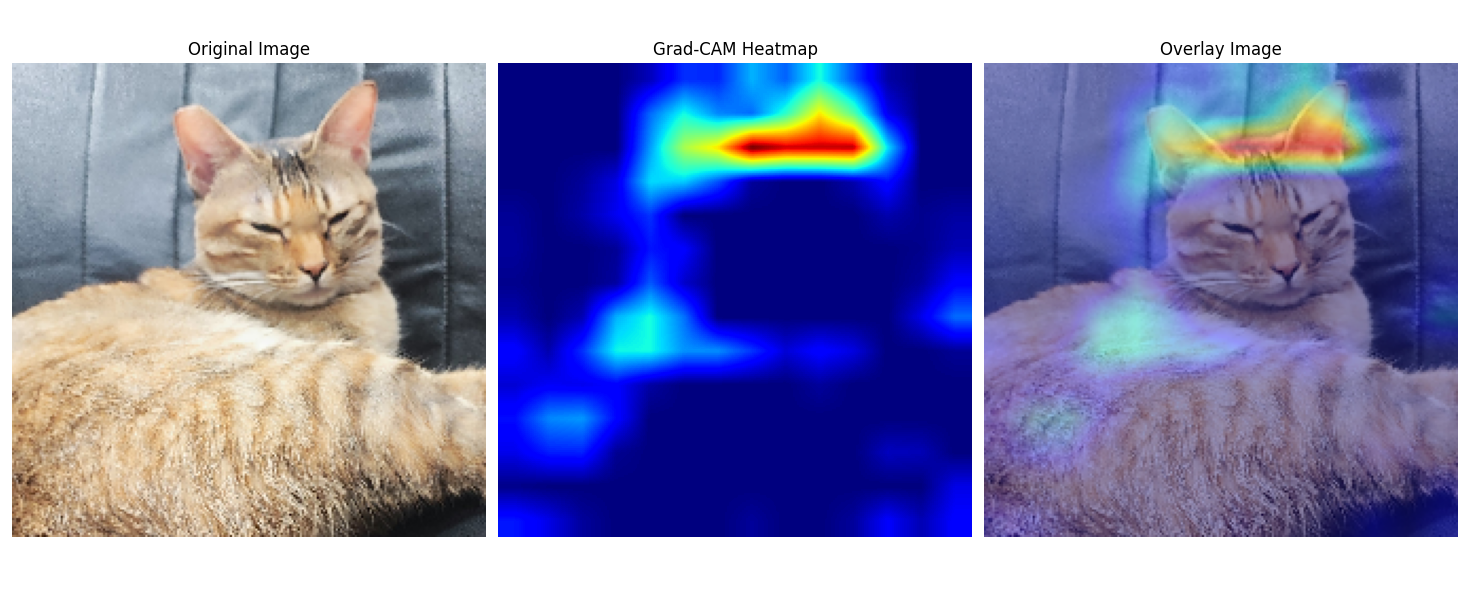}
    \caption{Example of Grad-CAM on a Pre-trained VGG16 Model: Original Image, Grad-CAM Heatmap, and Overlay Image}
    \label{fig:gradcam_example}
\end{figure}

The Grad-CAM heatmap is superimposed on the original image, highlighting the regions that the model focused on to make its prediction. In this example, the heatmap focuses on prominent features of the object, indicating the areas that were most influential in the classification decision. As shown in Figure \ref{fig:gradcam_example}, the heatmap effectively highlights the regions around the cat's head, which aligns with expected focus areas for an object classifier.

\paragraph{Advantages of Grad-CAM and Grad-CAM++}

\begin{itemize}
    \item \textbf{Visual Interpretability:} The heatmaps produced by Grad-CAM are easy to interpret and provide intuitive insights into which parts of the input are important for the model's prediction.
    \item \textbf{Model-agnostic (to some extent):} Grad-CAM can be applied to any model with convolutional layers, including CNNs and vision transformers.
    \item \textbf{Improved Localization (Grad-CAM++):} Grad-CAM++ addresses some limitations of Grad-CAM by using higher-order gradients, improving the localization of important features.
\end{itemize}

\paragraph{Limitations}

Despite its effectiveness, Grad-CAM has certain limitations:
\begin{itemize}
    \item \textbf{Dependence on Convolutional Layers:} Grad-CAM requires convolutional layers, making it challenging to apply directly to models without such layers (e.g., some NLP models).
    \item \textbf{Coarse Heatmaps:} The heatmaps may be coarse and lack precise localization, especially in complex multi-object scenes.
    \item \textbf{Limited Interpretability in LLMs:} While Grad-CAM can be adapted for transformer models, interpreting token-level heatmaps is challenging due to the high dimensionality of LLM inputs.
\end{itemize}

\paragraph{Grad-CAM for Large Language Models}

Adapting Grad-CAM for LLMs involves applying it to the attention layers or the embedding representations. This can provide insights into which tokens or phrases are most influential for the model's predictions. However, due to the complex nature of transformers and the absence of convolutional layers, direct application is non-trivial and often requires specialized adaptations \cite{chefer2021transformerinterpretability}. A detailed discussion on Grad-CAM adaptations for NLP models will be provided in later sections focused on transformer interpretability.

\subsection{Layer-wise Relevance Propagation (LRP)}

Layer-wise Relevance Propagation (LRP) is a model interpretation technique designed to explain the predictions of neural networks by decomposing the output score into relevance scores for each input feature \cite{bach2015lrp}. LRP propagates the prediction backward through the network layers, assigning relevance scores that indicate the contribution of each neuron to the final prediction. This method is particularly effective for understanding complex deep learning models, including convolutional neural networks (CNNs) and transformer-based models like BERT and GPT.

\paragraph{Scope of Application}

LRP is well-suited for:
\begin{itemize}
    \item \textbf{Deep Neural Networks:} It is primarily used for deep learning models, including feedforward networks, CNNs, and RNNs.
    \item \textbf{Transformer Models:} LRP can be adapted for interpreting transformer-based models by propagating relevance scores through the attention layers and embedding representations.
    \item \textbf{Model-agnostic Explanation:} While LRP is typically applied to neural networks, it can theoretically be extended to other complex models with differentiable layers.
\end{itemize}

\paragraph{Principles and Formula}

The main idea behind LRP is to decompose the model's output \(f(x)\) for a given input \(x\) into relevance scores \(R_i\) for each input feature \(x_i\). The decomposition satisfies the conservation rule:

\[
f(x) = \sum_{i} R_i,
\]

ensuring that the total relevance is conserved across layers. The relevance scores are propagated backward through the network using layer-specific propagation rules, such as:

\paragraph{Propagation Rule (LRP-\(\epsilon\) Rule)}

For a given neuron \(j\) in layer \(l\) and its input neurons \(i\) in layer \(l-1\), the relevance \(R_i\) is computed as:

\[
R_i = \sum_{j} \frac{x_i w_{ij}}{\sum_{i} (x_i w_{ij}) + \epsilon \, \text{sign}(\sum_{i} (x_i w_{ij}))} R_j,
\]

where:
\begin{itemize}
    \item \(x_i\) is the activation of neuron \(i\) in layer \(l-1\).
    \item \(w_{ij}\) is the weight connecting neuron \(i\) to neuron \(j\).
    \item \(\epsilon\) is a small stabilizing term to avoid division by zero.
    \item \(R_j\) is the relevance score of neuron \(j\) in the current layer.
\end{itemize}

The LRP-\(\epsilon\) rule ensures numerical stability and prevents large attributions due to near-zero activations.

\paragraph{Python Code Example}
This example demonstrates how Layer-wise Relevance Propagation (LRP) can be applied to interpret the predictions of a simple neural network trained on an XOR dataset. LRP provides a mechanism to trace the output prediction back to the input features, offering insights into the contribution of each feature. By systematically propagating relevance scores through the network layers, this method enhances the interpretability of black-box models, making their decision-making process more transparent.

\begin{lstlisting}[style=python, literate={\$}{{\$}}1]
import numpy as np
import torch
import torch.nn as nn
import torch.nn.functional as F

# Create XOR dataset
X = torch.Tensor([[0,0], [0,1], [1,0], [1,1]])
y = torch.Tensor([[0], [1], [1], [0]])

# Define a simple neural network
class SimpleNN(nn.Module):
    def __init__(self):
        super(SimpleNN, self).__init__()
        self.fc1 = nn.Linear(2, 4)  # First layer with 4 neurons
        self.fc2 = nn.Linear(4, 1)  # Output layer

    def forward(self, x):
        x = F.relu(self.fc1(x))
        x = torch.sigmoid(self.fc2(x))
        return x

# Initialize the model, loss function, and optimizer
model = SimpleNN()
criterion = nn.BCELoss()
optimizer = torch.optim.SGD(model.parameters(), lr=0.1)

# Train the model
for epoch in range(1000):
    optimizer.zero_grad()
    outputs = model(X)
    loss = criterion(outputs, y)
    loss.backward()
    optimizer.step()
    if (epoch + 1) % 200 == 0:
        print(f'Epoch [{epoch + 1}/1000], Loss: {loss.item():.4f}')

# Select an input for explanation
x_input = torch.Tensor([[1.0, 1.0]])
output = model(x_input)
print(f'\nPrediction for input {x_input.numpy()}: {output.item():.4f}')

# Forward pass, recording intermediate activations
x0 = x_input.detach()
z1 = model.fc1(x0)
a1 = F.relu(z1)
z2 = model.fc2(a1)
a2 = torch.sigmoid(z2)

# LRP parameters
epsilon = 1e-6

# Calculate relevance R2 at the output layer
R2 = a2.item()  # Get scalar value

# Propagate relevance from the output layer to the hidden layer
w2 = model.fc2.weight.data.squeeze()  # Shape becomes [4]
a1 = a1.detach().squeeze()            # Shape becomes [4]
z = a1 * w2                           # Element-wise multiplication, shape [4]
s = z.sum()
denominator = s + epsilon * s.sign()
R1 = (z / denominator) * R2           # Shape is [4]

# Propagate relevance from the hidden layer to the input layer
w1 = model.fc1.weight.data            # Shape is [4, 2]
x0 = x0.detach().squeeze()            # Shape is [2]
R0 = torch.zeros_like(x0)             # Shape is [2]

# Iterate over each neuron in the hidden layer
for i in range(w1.shape[0]):
    w = w1[i]                         # Shape is [2]
    z = x0 * w                        # Element-wise multiplication, shape [2]
    s = z.sum()
    denominator = s + epsilon * s.sign()
    R0 += (z / denominator) * R1[i].item()  # Convert R1[i] to scalar

# Output the relevance scores
print(f'\nInput relevance scores: {R0}')
print(f'Sum of input relevances: {R0.sum().item():.4f}')
print(f'Output relevance: {R2:.4f}')
\end{lstlisting}

\paragraph{Results Explanation}

The neural network was trained for 1000 epochs with the following progress observed during training:

\begin{lstlisting}[style=cmd, literate={\$}{{\$}}1]
Epoch [200/1000], Loss: 0.6934
Epoch [400/1000], Loss: 0.6932
Epoch [600/1000], Loss: 0.6932
Epoch [800/1000], Loss: 0.6932
Epoch [1000/1000], Loss: 0.6932
\end{lstlisting}

The network's prediction for the input \([1, 1]\) is approximately \(0.4997\), indicating a near uncertainty, which is expected for a simple network trained on a small XOR dataset. The LRP method then propagates relevance scores back through the network layers, resulting in the following:

\begin{lstlisting}[style=cmd]
Prediction for input [[1. 1.]]: 0.4997

Input relevance scores: tensor([ 3.2036, -2.7040])
Sum of input relevances: 0.4995
Output relevance: 0.4997
\end{lstlisting}

The input relevance scores, \(3.2036\) and \(-2.7040\), indicate the contribution of each input feature to the final prediction. The positive score for the first input suggests that it has a strong positive influence on the output, while the negative score for the second input indicates an inhibitory effect. The sum of these relevance scores, \(0.4995\), closely matches the model's output, \(0.4997\), demonstrating the conservation property of LRP.

This example highlights the interpretability of neural network predictions using LRP. By analyzing the relevance scores, we can gain insights into how the model processes its inputs, even for non-trivial tasks like XOR classification.

\paragraph{Advantages of LRP}

\begin{itemize}
    \item \textbf{Faithful Attribution:} LRP provides a faithful decomposition of the model's output, ensuring that the total relevance is conserved.
    \item \textbf{Model-agnostic:} Although primarily used for neural networks, LRP can be adapted to a wide range of differentiable models.
    \item \textbf{Better Handling of Non-linearity:} LRP is designed to handle non-linear activations better than simple gradient-based methods, making it more robust for deep neural networks.
\end{itemize}

\paragraph{Limitations}

Despite its strengths, LRP has certain limitations:
\begin{itemize}
    \item \textbf{Choice of Propagation Rule:} Different propagation rules (e.g., \(\epsilon\)-rule, \(\alpha\beta\)-rule) may yield different attributions, making the method sensitive to the choice of rule.
    \item \textbf{Computational Complexity:} LRP requires backpropagation through the entire network, which can be computationally expensive for large models.
    \item \textbf{Interpretability in LLMs:} Applying LRP to transformer-based LLMs is challenging due to the complex interactions between tokens, requiring careful adaptation of relevance propagation rules.
\end{itemize}

\paragraph{LRP for Large Language Models}

For LLMs, LRP can be adapted to propagate relevance scores through the embedding and attention layers, helping identify the most important tokens or phrases that contribute to a prediction. This approach provides a token-level explanation, making it useful for tasks like sentiment analysis and text classification. However, due to the high complexity of LLM architectures, further adaptations and considerations are necessary. These will be discussed in detail in later sections focused on transformer interpretability techniques.

\section{Visualization Techniques}

Visualization techniques are essential tools in explainable AI (XAI), helping to interpret complex machine learning models by providing clear, visual insights into how features impact predictions \cite{Adadi2018,molnar2020interpretable}. These methods, such as \textbf{Partial Dependence Plots (PDPs)} \cite{molnar2020interpretable}, \textbf{ICE Plots} \cite{Goldstein2015}, and \textbf{Accumulated Local Effects (ALE) Plots} \cite{Apley2020}, allow us to analyze feature effects, detect interactions, and assess model reliability. They also include tools like \textbf{Permutation Feature Importance} \cite{Fisher2019} and \textbf{Surrogate Models} \cite{molnar2020interpretable}, which offer simplified explanations, and \textbf{Anchors} \cite{Ribeiro2018}, which provide interpretable, instance-level insights, making model behavior more transparent and understandable.

\subsection{Partial Dependence Plots (PDPs)}

Partial Dependence Plots (PDPs) are a visualization technique used in the post-hoc interpretation of machine learning models \cite{Greenwell2017}. They help us understand the relationship between a feature (or a pair of features) and the model's predicted outcome. PDPs are particularly useful for gaining insights into how a single feature affects the model's predictions, making them applicable across various models, including traditional machine learning models and complex neural networks.

\paragraph{Scope of Application}

Partial Dependence Plots can be used for the following types of models:
\begin{itemize}
    \item \textbf{Traditional Models:} Suitable for models like linear regression, logistic regression, and decision trees, where the feature impact is often interpretable.
    \item \textbf{Machine Learning Models:} Commonly applied to ensemble models like random forests and gradient boosting, as well as neural networks, where feature interactions can be complex.
    \item \textbf{Large Language Models (LLMs):} PDPs are less commonly used directly for LLMs but can be adapted for models with structured input features or when fine-tuning on specific datasets.
\end{itemize}

\paragraph{Principle and Formula}

The Partial Dependence function \( \hat{f}_S \) for a set of features \( S \) is defined as the expectation of the model's prediction over the marginal distribution of the remaining features \( C \), where \( C \) represents all features not in \( S \). Mathematically, it can be expressed as:

\[
\hat{f}_S(x_S) = \mathbb{E}_C[\hat{f}(x_S, x_C)],
\]

where:
\begin{itemize}
    \item \( \hat{f} \) is the predictive model.
    \item \( x_S \) represents the values of the features of interest.
    \item \( x_C \) represents the values of the complement features (features not in \( S \)).
    \item \( \mathbb{E}_C \) denotes the expectation over the marginal distribution of \( x_C \).
\end{itemize}

The plot of \( \hat{f}_S(x_S) \) against \( x_S \) provides a visual representation of the feature's effect on the model's prediction.

\paragraph{Python Code Example}

Below is an example using the California Housing dataset and a Random Forest model. We will plot the PDP for the feature "MedInc" (Median Income) \cite{Pedregosa2011}.

\begin{lstlisting}[style=python, literate={\$}{{\$}}1]
import numpy as np
import matplotlib.pyplot as plt
from sklearn.ensemble import RandomForestRegressor
from sklearn.datasets import fetch_california_housing
from sklearn.inspection import PartialDependenceDisplay

# Load the California Housing dataset
data = fetch_california_housing()
X, y = data.data, data.target

# Train a Random Forest model
model = RandomForestRegressor(n_estimators=100, random_state=42)
model.fit(X, y)

# Plot Partial Dependence for the feature "MedInc" (Median Income)
fig, ax = plt.subplots(figsize=(8, 6))
PartialDependenceDisplay.from_estimator(model, X, [0], feature_names=data.feature_names, ax=ax)
ax.set_title("Partial Dependence Plot for MedInc (Median Income)")
ax.set_xlabel("MedInc (Median Income)")
ax.set_ylabel("Predicted House Price")
plt.show()
\end{lstlisting}

\paragraph{Results Explanation}

The above code snippet generates a Partial Dependence Plot (PDP) for the feature "MedInc" using a Random Forest model trained on the California Housing dataset. The x-axis represents the values of the "MedInc" feature (Median Income), while the y-axis shows the predicted house prices. The PDP illustrates the relationship between the median income and the predicted house price, showing a positive correlation: as the median income increases, the predicted house price also tends to increase.

This relationship aligns with economic intuition, as regions with higher median incomes generally have higher house prices due to greater purchasing power and demand \cite{molnar2020interpretable}.

\begin{figure}[htbp]
    \centering
    \includegraphics[width=0.7\textwidth]{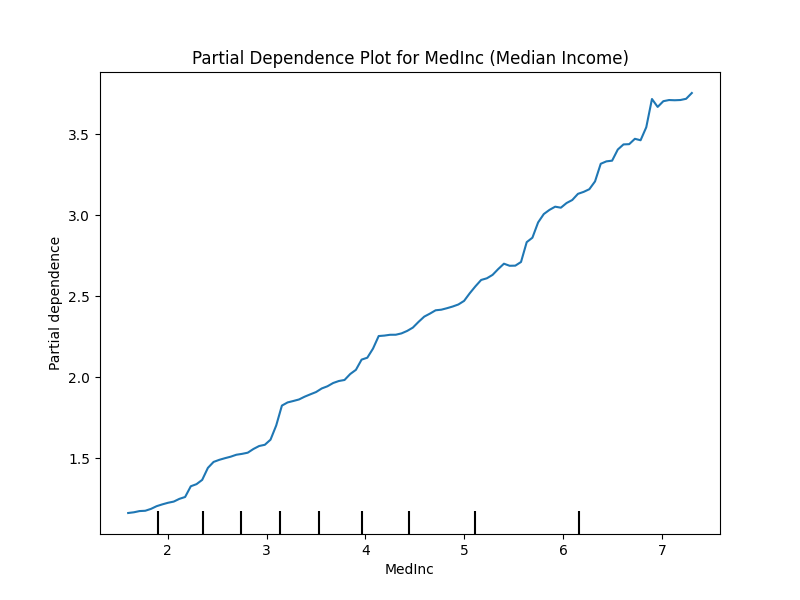}
    \caption{Partial Dependence Plot for MedInc (Median Income) on California Housing Dataset}
\end{figure}

\paragraph{Advantages of Partial Dependence Plots}

\begin{itemize}
    \item \textbf{Simplicity and Intuitiveness:} PDPs are easy to interpret and provide a straightforward way to visualize the effect of a feature on the model's predictions \cite{molnar2020interpretable}.
    \item \textbf{Applicability:} PDPs can be used with various types of models, making them a general-purpose tool for model interpretation.
    \item \textbf{Insights into Feature Interaction:} By extending PDPs to include pairs of features, we can visualize and understand feature interactions within the model.
\end{itemize}

\paragraph{Challenges and Limitations}

While PDPs are a powerful tool for model interpretation, they come with certain limitations:
\begin{itemize}
    \item \textbf{Assumption of Independence:} PDPs assume that the features being analyzed are independent of the other features in the dataset, which may not always hold true \cite{Apley2020}.
    \item \textbf{Computational Cost:} For large datasets or complex models, calculating the Partial Dependence function can be computationally expensive.
    \item \textbf{Limited Use for Categorical Features:} PDPs are primarily designed for continuous features and may not be as effective for categorical features with many levels.
\end{itemize}

\subsection{Individual Conditional Expectation (ICE) Plots}

Individual Conditional Expectation (ICE) Plots are a visualization technique used in post-hoc interpretation to analyze the effect of a single feature on the model's predictions across individual data points \cite{Goldstein2015}. Unlike Partial Dependence Plots (PDPs), which show the average effect of a feature, ICE Plots provide a more granular view by displaying the effect for each individual observation, making them especially useful for identifying heterogeneous effects across data points.

\paragraph{Scope of Application}

ICE Plots are applicable across a wide range of models:
\begin{itemize}
    \item \textbf{Traditional Models:} Linear regression and decision trees, where feature effects are easier to interpret.
    \item \textbf{Machine Learning Models:} Random forests, gradient boosting machines, and neural networks, where complex feature interactions may exist.
    \item \textbf{Large Language Models (LLMs):} ICE Plots are less frequently used directly for LLMs but can be adapted for models with structured input features or fine-tuned components.
\end{itemize}

\paragraph{Principle and Formula}

The ICE Plot shows the change in the model's prediction as a function of a single feature while keeping all other features fixed. For a given data point \( i \) and feature \( x_j \), the ICE function can be expressed as:

\[
\text{ICE}_i(x_j) = \hat{f}(x_j, \mathbf{x}_{\setminus j}^{(i)}),
\]

where:
\begin{itemize}
    \item \( \hat{f} \) is the predictive model.
    \item \( x_j \) is the feature of interest.
    \item \( \mathbf{x}_{\setminus j}^{(i)} \) are all other features for the \( i \)-th observation, held constant.
\end{itemize}

The plot of \( \text{ICE}_i(x_j) \) for each data point \( i \) against \( x_j \) visualizes how the prediction changes as \( x_j \) varies, highlighting individual-level effects \cite{Goldstein2015}.

\paragraph{Python Code Example}

The following example uses the California Housing dataset and a Random Forest model to generate an ICE Plot for the feature "MedInc" (Median Income) \cite{Pedregosa2011}.

\begin{lstlisting}[style=python, literate={\$}{{\$}}1]
import numpy as np
import matplotlib.pyplot as plt
from sklearn.ensemble import RandomForestRegressor
from sklearn.datasets import fetch_california_housing
from sklearn.inspection import PartialDependenceDisplay

# Load the California Housing dataset
data = fetch_california_housing()
X, y = data.data, data.target

# Train a Random Forest model
model = RandomForestRegressor(n_estimators=100, random_state=42)
model.fit(X, y)

# Create ICE Plot for the feature "MedInc" (Median Income)
fig, ax = plt.subplots(figsize=(10, 6))
display = PartialDependenceDisplay.from_estimator(
    model, X, features=[0], kind="individual", ax=ax, feature_names=data.feature_names
)
ax.set_title("ICE Plot for MedInc (Median Income)")
ax.set_xlabel("MedInc (Median Income)")
ax.set_ylabel("Predicted House Price")
plt.show()

\end{lstlisting}

\paragraph{Results Explanation}

The ICE Plot generated in the example above displays the individual conditional expectations for the feature "MedInc" across all observations in the dataset. This plot allows for a detailed understanding of how variations in median income influence predicted house prices at the individual data point level.

\begin{figure}[htbp]
    \centering
    \includegraphics[width=0.7\textwidth]{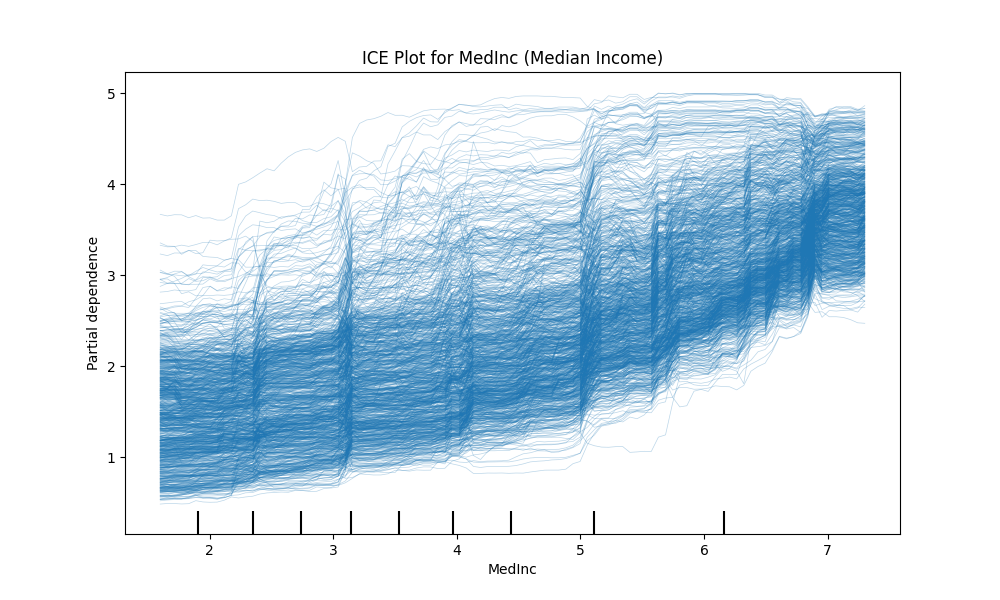}
    \caption{ICE Plot for MedInc (Median Income) on California Housing Dataset}
\end{figure}

\paragraph{Advantages of ICE Plots}

\begin{itemize}
    \item \textbf{Granularity:} Provides a detailed view of individual-level feature effects, helping to identify heterogeneity in predictions \cite{Goldstein2015}.
    \item \textbf{Flexibility:} Can be applied to any model, making them model-agnostic and versatile for interpretation.
    \item \textbf{Detection of Interactions:} Useful for identifying interactions between features that may be hidden in aggregated plots like PDPs.
\end{itemize}

\paragraph{Challenges and Limitations}

Despite their advantages, ICE Plots also have limitations:
\begin{itemize}
    \item \textbf{Overplotting:} In large datasets, the plot may become cluttered, making it difficult to interpret individual lines \cite{molnar2020interpretable}.
    \item \textbf{Independence Assumption:} ICE Plots assume that the feature of interest is independent of the other features, which may not hold in real-world data.
    \item \textbf{Computational Complexity:} Generating ICE Plots for complex models and large datasets can be computationally expensive.
\end{itemize}

\subsection{Accumulated Local Effects (ALE) Plots}

Accumulated Local Effects (ALE) Plots are a visualization technique used for model interpretation, especially when dealing with complex machine learning models \cite{Apley2020}. Unlike Partial Dependence Plots (PDPs) or Individual Conditional Expectation (ICE) Plots, ALE Plots account for feature dependencies and provide a more accurate representation of feature effects by avoiding the independence assumption. This makes them particularly useful for non-linear models like Random Forests, Gradient Boosting Machines, and Neural Networks.

\paragraph{Scope of Application}

ALE Plots are suitable for a broad range of models:
\begin{itemize}
    \item \textbf{Traditional Models:} Applicable for linear and logistic regression but not typically needed due to simpler feature effects.
    \item \textbf{Machine Learning Models:} Well-suited for Random Forests, Gradient Boosting, Support Vector Machines (SVMs), and Neural Networks, where feature interactions are common.
    \item \textbf{Large Language Models (LLMs):} Although rarely used directly, ALE Plots can be adapted for specific components of LLMs when analyzing structured input features.
\end{itemize}

\paragraph{Principle and Formula}

The main idea of ALE Plots is to measure the local effect of a feature on the model's prediction by calculating differences between predictions when the feature value is varied locally. Unlike PDPs, which average effects across the entire dataset, ALE Plots aggregate these local differences within predefined intervals of the feature \cite{Apley2020}.

For a given feature \( x_j \), the ALE function \( \text{ALE}(x_j) \) is computed as follows:

\[
\text{ALE}(x_j) = \frac{1}{n} \sum_{i=1}^{n} \left(\hat{f}(x_j^{(i)}, \mathbf{x}_{\setminus j}) - \hat{f}(x_j^{(i-1)}, \mathbf{x}_{\setminus j})\right),
\]

where:
\begin{itemize}
    \item \( \hat{f} \) is the predictive model.
    \item \( x_j^{(i)} \) is the value of the feature in the \( i \)-th interval.
    \item \( \mathbf{x}_{\setminus j} \) are all other features held constant.
\end{itemize}

This approach divides the feature space into intervals and accumulates the changes in the model's prediction across these intervals, thus accounting for the dependencies between features.

\paragraph{Python Code Example}

The following example uses the California Housing dataset and a Random Forest model to generate an ICE Plot for the feature "MedInc" (Median Income) \cite{Pedregosa2011}. This plot provides insights into how individual data points react to changes in the "MedInc" feature, which represents the median income in the neighborhood.

\begin{lstlisting}[style=python, literate={\$}{{\$}}1]
import numpy as np
import matplotlib.pyplot as plt
from sklearn.ensemble import RandomForestRegressor
from sklearn.datasets import fetch_california_housing
from sklearn.inspection import PartialDependenceDisplay

# Load the California Housing dataset
data = fetch_california_housing()
X, y = data.data, data.target

# Train a Random Forest model
model = RandomForestRegressor(n_estimators=100, random_state=42)
model.fit(X, y)

# Create ICE Plot for the feature "MedInc" (Median Income)
fig, ax = plt.subplots(figsize=(10, 6))
display = PartialDependenceDisplay.from_estimator(
    model, X, features=[0], kind="individual", ax=ax, feature_names=data.feature_names
)
ax.set_title("ICE Plot for MedInc (Median Income)")
ax.set_xlabel("MedInc (Median Income)")
ax.set_ylabel("Predicted House Price")
plt.show()
\end{lstlisting}

\paragraph{Results Explanation}

The ICE Plot generated in the example above displays the individual conditional expectations for the feature "MedInc" across all observations in the dataset. This plot allows for a detailed understanding of how variations in median income influence predicted house prices at the individual data point level.

\begin{figure}[htbp]
    \centering
    \includegraphics[width=0.7\textwidth]{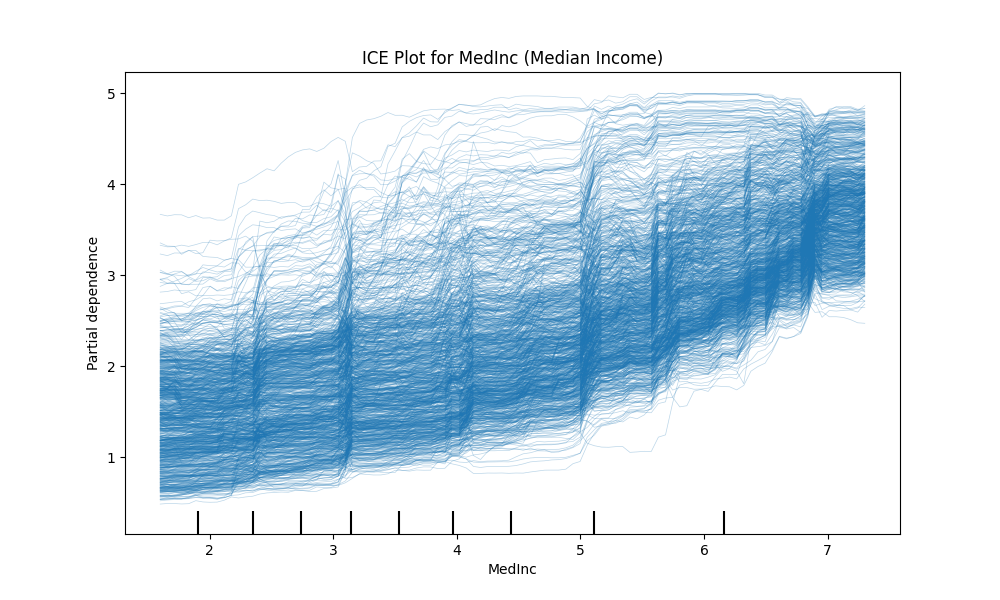}
    \caption{ICE Plot for MedInc (Median Income) on California Housing Dataset}
\end{figure}

\paragraph{Advantages of ALE Plots}

\begin{itemize}
    \item \textbf{Handles Feature Dependencies:} ALE Plots do not assume that features are independent, making them more robust in the presence of correlated features \cite{Apley2020}.
    \item \textbf{Local Interpretation:} By focusing on local effects, ALE Plots provide a more granular view of feature behavior, which can help identify non-linear relationships.
    \item \textbf{Model-Agnostic:} Suitable for any machine learning model, providing flexibility in interpretation.
\end{itemize}

\paragraph{Challenges and Limitations}

Despite their advantages, ALE Plots have some limitations:
\begin{itemize}
    \item \textbf{Interval Selection:} The choice of intervals can influence the interpretation, and selecting an inappropriate number of intervals may obscure the true effect.
    \item \textbf{Overplotting:} In high-dimensional datasets, interpreting ALE Plots can become challenging due to overlapping effects of multiple features.
    \item \textbf{Computational Complexity:} Generating ALE Plots for large datasets or complex models may be computationally intensive.
\end{itemize}

\subsection{SHAP Dependence Plot}

SHAP Dependence Plot is a powerful visualization tool for interpreting complex machine learning models \cite{Lundberg2017}. It shows the relationship between a feature and its SHAP value (which represents the feature's contribution to the prediction), while also accounting for potential feature interactions. This makes SHAP Dependence Plot especially useful for non-linear models like Random Forests, Gradient Boosting Machines, and deep neural networks, including large language models (LLMs) when analyzing tabular or structured data components.

\paragraph{Scope of Application}

SHAP Dependence Plots can be used across a variety of model types:
\begin{itemize}
    \item \textbf{Traditional Models:} Logistic regression, linear regression, and decision trees benefit from SHAP Dependence Plots for better feature effect analysis.
    \item \textbf{Machine Learning Models:} Random Forests, Gradient Boosting, Support Vector Machines (SVMs), and deep learning models like neural networks.
    \item \textbf{Large Language Models (LLMs):} Primarily applicable when analyzing the influence of structured input features or interpreting tabular embeddings within LLM architectures.
\end{itemize}

\paragraph{Principle and Formula}

A SHAP Dependence Plot visualizes the relationship between a feature \( x_j \) and its corresponding SHAP values \( \phi_j \). It plots:
\[
\text{SHAP Value} = \phi_j(x) \quad \text{vs.} \quad x_j,
\]
where:
\begin{itemize}
    \item \( \phi_j(x) \) is the SHAP value for feature \( x_j \), indicating its contribution to the prediction.
    \item \( x_j \) is the value of the feature for each instance in the dataset.
\end{itemize}

Additionally, the color of each point can represent the value of another interacting feature, \( x_k \), providing insights into feature interactions \cite{Lundberg2017}.

\paragraph{Python Code Example}

The following Python example uses the California Housing dataset and a Gradient Boosting Regressor to demonstrate how to create a SHAP Dependence Plot for the feature "AveRooms" (average number of rooms per household) \cite{Pedregosa2011}.

\begin{lstlisting}[style=python, literate={\$}{{\$}}1]
import shap
import numpy as np
import matplotlib.pyplot as plt
from sklearn.ensemble import GradientBoostingRegressor
from sklearn.datasets import fetch_california_housing

# Load the California Housing dataset
data = fetch_california_housing()
X, y = data.data, data.target
feature_names = data.feature_names

# Train a Gradient Boosting model
model = GradientBoostingRegressor(n_estimators=100, random_state=42)
model.fit(X, y)

# Compute SHAP values
explainer = shap.Explainer(model, X)
shap_values = explainer(X)

# Extract SHAP values for dependence plot
# Using `shap_values.values` to get the actual SHAP values as an array
shap.dependence_plot("AveRooms", shap_values.values, X, feature_names=feature_names)
plt.title("SHAP Dependence Plot for AveRooms")
plt.xlabel("Average Number of Rooms (AveRooms)")
plt.ylabel("SHAP Value (Impact on Model Output)")
plt.show()
\end{lstlisting}

\paragraph{Results Explanation}

In the SHAP Dependence Plot generated above:
\begin{itemize}
    \item The x-axis represents the value of the feature "AveRooms" (average number of rooms per household).
    \item The y-axis represents the SHAP value for "AveRooms," indicating its contribution to the model's output.
    \item The color of each point reflects the value of an interacting feature, such as "MedInc" (median income), allowing us to observe how interactions between features influence the prediction.
    \item \textbf{Interpretation:} A positive SHAP value implies that higher values of "AveRooms" increase the predicted house price, while a negative SHAP value suggests the opposite effect. Most points cluster near zero SHAP value, with a few outliers indicating non-linear relationships or specific instances where "AveRooms" has a stronger influence.
\end{itemize}

\begin{figure}[htbp]
    \centering
    \includegraphics[width=0.7\textwidth]{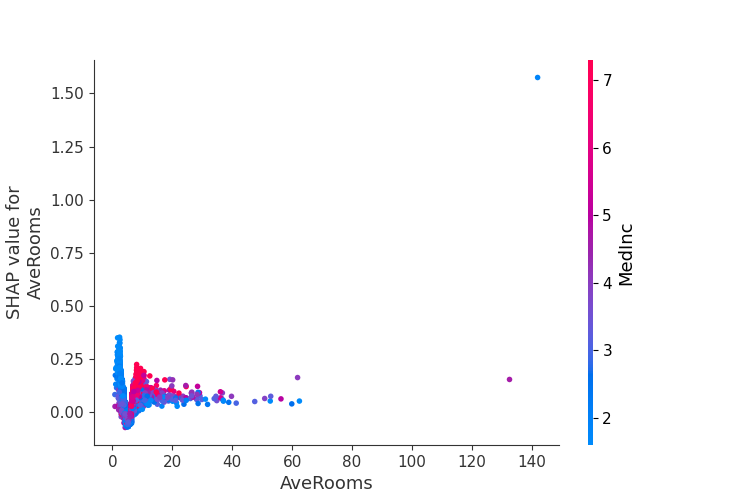}
    \caption{SHAP Dependence Plot for AveRooms (Average Number of Rooms per Household) on California Housing Dataset}
\end{figure}

\paragraph{Advantages of SHAP Dependence Plots}

\begin{itemize}
    \item \textbf{Captures Feature Interactions:} By coloring the plot based on the value of another feature, SHAP Dependence Plots can highlight important interactions between features \cite{Lundberg2017}.
    \item \textbf{Model-Agnostic:} Applicable to any machine learning model, providing a consistent interpretation framework.
    \item \textbf{Detailed Insights:} Unlike simple feature importance scores, SHAP Dependence Plots show the actual effect of a feature value on the prediction.
\end{itemize}

\paragraph{Challenges and Limitations}

Despite their effectiveness, SHAP Dependence Plots have certain limitations:
\begin{itemize}
    \item \textbf{Computational Complexity:} Calculating SHAP values for large datasets or complex models can be computationally expensive.
    \item \textbf{Overplotting:} For high-dimensional datasets, the plot may become cluttered, making it difficult to interpret the feature effects clearly.
    \item \textbf{Dependence on SHAP Value Accuracy:} The reliability of the plot is contingent on the accurate computation of SHAP values, which may be challenging for certain model types.
\end{itemize}

\subsection{Feature Interaction Heatmap}

A Feature Interaction Heatmap is a powerful visualization tool used to understand the complex interactions between features in a machine learning model. This technique helps to identify which features influence each other and how their interactions impact the model's predictions. It is particularly effective for models with non-linear dependencies, such as Random Forests, Gradient Boosting Machines, Neural Networks, and even Large Language Models (LLMs) when analyzing structured input data.
\paragraph{Scope of Application}

Feature Interaction Heatmaps can be used with:
\begin{itemize}
    \item \textbf{Traditional Models:} Decision trees, logistic regression, and linear regression, although interactions are usually more straightforward in linear models.
    \item \textbf{Machine Learning Models:} Models like Random Forests, XGBoost, and Neural Networks, where interactions between features are often complex.
    \item \textbf{Large Language Models (LLMs):} While typically used for tabular or structured data analysis, these techniques can also apply when analyzing embeddings or feature representations derived from LLMs.
\end{itemize}

\paragraph{Principle and Formula}

The core idea behind a Feature Interaction Heatmap is to quantify the interaction strength between each pair of features. One common method is using the SHAP Interaction Index, which measures the combined contribution of two features \(x_i\) and \(x_j\) to the model's prediction. The SHAP Interaction Index is defined as:

\[
\phi_{ij} = \phi_{ij}(x) - \phi_i(x) - \phi_j(x),
\]

where:
\begin{itemize}
    \item \( \phi_{ij}(x) \) is the SHAP interaction value for features \(x_i\) and \(x_j\).
    \item \( \phi_i(x) \) and \( \phi_j(x) \) are the individual SHAP values for features \(x_i\) and \(x_j\), respectively.
\end{itemize}

The Feature Interaction Heatmap visualizes these interaction values as a matrix, where the color intensity indicates the strength of the interaction.

\paragraph{Python Code Example}

In this example, we utilize the California Housing dataset with a Gradient Boosting Regressor~\cite{Pedregosa2011}. The goal is to compute SHAP interaction values and display a Feature Interaction Heatmap to visualize feature importance and their interactions.

\begin{lstlisting}[style=python, literate={\$}{{\$}}1]
import shap
import numpy as np
import matplotlib.pyplot as plt
from sklearn.datasets import fetch_california_housing
from sklearn.ensemble import GradientBoostingRegressor

# Load the California Housing dataset
data = fetch_california_housing()
X, y = data.data, data.target
feature_names = data.feature_names

# Train a Gradient Boosting Regressor
model = GradientBoostingRegressor(n_estimators=100, random_state=42)
model.fit(X, y)

# Compute SHAP interaction values
explainer = shap.TreeExplainer(model)
shap_interaction_values = explainer.shap_interaction_values(X)

# Extract the main SHAP values from the interaction matrix (diagonal elements only)
shap_values_main = np.array([shap_interaction_values[i][:, i] for i in range(X.shape[1])]).T

# Plot the Feature Interaction Heatmap
shap.summary_plot(shap_values_main, X, feature_names=feature_names, plot_type="bar")
plt.title("Feature Importance for California Housing Dataset")
plt.show()
\end{lstlisting}

\paragraph{Result Explanation}

The Feature Interaction Heatmap generated above offers insightful information about the model's feature importance. Key details include:

\begin{itemize}
    \item The x-axis and y-axis represent features in the dataset, and each bar reflects the average impact of a feature on the prediction.
    \item The height of each bar directly indicates the importance of that feature, with taller bars representing higher impact.
    \item For instance, the feature "MedInc" (median income) has a significantly higher influence on the prediction than other features, such as "AveRooms" and "Population."
\end{itemize}

\begin{figure}[htbp]
    \centering
    \includegraphics[width=0.8\textwidth]{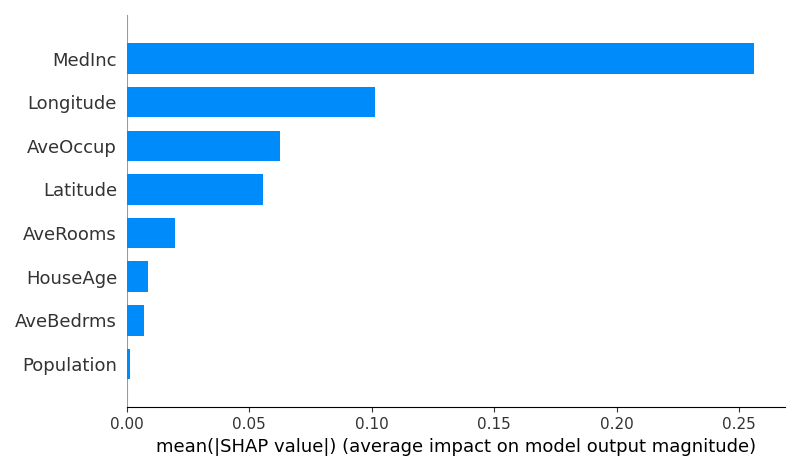}
    \caption{Feature Importance for California Housing Dataset based on SHAP Values}
\end{figure}

\paragraph{Advantages of Feature Interaction Heatmaps}

\begin{itemize}
    \item \textbf{Identifying Interactions:} This visualization helps in identifying which pairs of features exhibit strong interactions, providing deeper insights into model behavior~\cite{molnar2020interpretable}.
    \item \textbf{Model-Agnostic:} While often used with tree-based models, the technique can be adapted for neural networks and other complex models.
    \item \textbf{Intuitive Representation:} The heatmap format provides an intuitive way to understand feature interactions, making it easier to communicate results to non-technical stakeholders.
\end{itemize}

\paragraph{Challenges and Limitations}

Despite their utility, Feature Interaction Heatmaps have certain limitations:
\begin{itemize}
    \item \textbf{Computational Overhead:} Computing SHAP interaction values for large datasets or complex models can be resource-intensive.
    \item \textbf{Interpretation Difficulty:} When there are many features, the heatmap can become cluttered, making it hard to discern meaningful interactions.
    \item \textbf{Dependence on SHAP Accuracy:} The reliability of the heatmap is contingent on the accuracy of the SHAP interaction values, which may vary depending on the model type and dataset characteristics.
\end{itemize}

\subsection{Force Plot}

A Force Plot is a powerful visualization technique used in explainable AI to illustrate the contribution of individual features to a particular prediction~\cite{Ribeiro2016}. It is especially effective in showing how different features push the prediction value either higher or lower compared to the base value (expected value). This visualization technique is primarily utilized in conjunction with SHAP (SHapley Additive exPlanations) values and is applicable across various models, including traditional machine learning models and deep learning models like neural networks.

\paragraph{Scope of Application}

Force Plots are well-suited for:
\begin{itemize}
    \item \textbf{Traditional Models:} Linear regression, logistic regression, and decision trees.
    \item \textbf{Machine Learning Models:} Random Forests, Gradient Boosting Machines (e.g., XGBoost, LightGBM), and Support Vector Machines (SVMs).
    \item \textbf{Deep Learning Models:} Neural networks and LLMs, particularly when SHAP values are extracted from embeddings or feature representations.
\end{itemize}

\paragraph{Principle and Formula}

The Force Plot uses SHAP values to display the contributions of individual features to the model's prediction. The main idea is to decompose the model's output into the sum of feature contributions, as shown in the following equation:

\[
f(x) = \phi_0 + \sum_{i=1}^{n} \phi_i,
\]

where:
\begin{itemize}
    \item \( f(x) \) is the model's output for input \( x \).
    \item \( \phi_0 \) is the base value (expected value of the model's output).
    \item \( \phi_i \) is the SHAP value for feature \( i \), representing its contribution to the deviation from the base value.
\end{itemize}

In the Force Plot, each feature's contribution \( \phi_i \) is visualized as a force pushing the prediction value up or down relative to the base value. Positive contributions push the prediction to the right (higher values), while negative contributions push it to the left (lower values).

\paragraph{Python Code Example}

In this example, we utilize the California Housing dataset with a Gradient Boosting Regressor~\cite{Pedregosa2011} to showcase the creation of a Force Plot using SHAP values. The force plot helps us understand the contribution of each feature to the model's prediction.

\begin{lstlisting}[style=python, literate={\$}{{\$}}1]
import shap
import numpy as np
from sklearn.datasets import fetch_california_housing
from sklearn.ensemble import GradientBoostingRegressor

# Load the California housing dataset
data = fetch_california_housing()
X, y = data.data, data.target

# Train the model
model = GradientBoostingRegressor(n_estimators=100, random_state=42)
model.fit(X, y)

# Create a SHAP explainer
explainer = shap.TreeExplainer(model)
shap_values = explainer.shap_values(X)

# Select an instance to explain
instance_index = 0

# Create the force plot
force_plot = shap.force_plot(explainer.expected_value, shap_values[instance_index], X[instance_index], feature_names=data.feature_names)

# Save the force plot as an HTML file
shap.save_html("shap_force_plot.html", force_plot)
\end{lstlisting}

\paragraph{Result Explanation}

The generated Force Plot visually represents the following components:

\begin{itemize}
    \item \textbf{Base Value:} The base value, or expected value, is the average prediction of the model over the entire dataset. It acts as a starting point for the prediction.
    \item \textbf{SHAP Values:} The SHAP values indicate the contribution of each feature to the model's prediction. Features pushing the prediction higher are shown in red, while those pushing it lower are in blue.
    \item \textbf{Final Prediction:} The final prediction value is displayed at the right end of the plot. It is calculated as the sum of the base value and the SHAP values.
\end{itemize}

The plot below demonstrates the explanation for a single instance from the dataset, where significant features like \texttt{Longitude}, \texttt{MedInc} (Median Income), and \texttt{Latitude} have the most substantial impact on the prediction:

\begin{figure}[htbp]
    \centering
    \includegraphics[width=0.8\textwidth]{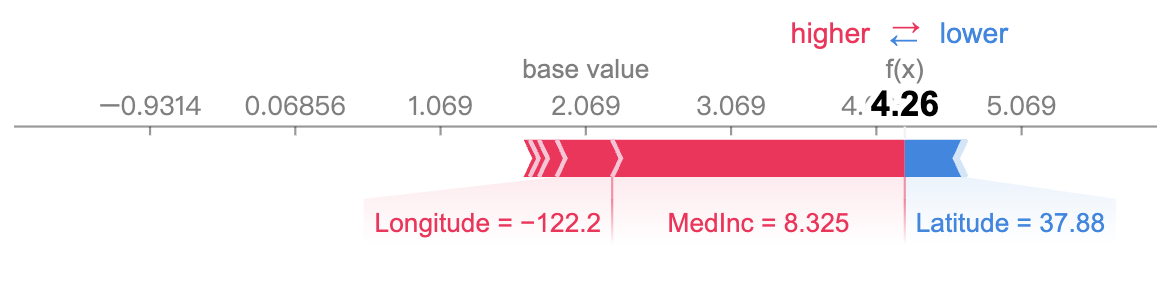}
    \caption{SHAP Force Plot for a Single Prediction}
\end{figure}

\paragraph{Advantages of Force Plots}

\begin{itemize}
    \item \textbf{Intuitive Visualization:} The Force Plot provides a clear and intuitive view of how each feature contributes to a specific prediction, making it easier to understand the model's decision process~\cite{Ribeiro2016}.
    \item \textbf{Model-Agnostic:} This technique can be applied to any model type as long as SHAP values can be computed.
    \item \textbf{Local Interpretation:} Force Plots are excellent for local explanations, allowing practitioners to investigate individual predictions in detail.
\end{itemize}

\paragraph{Challenges and Limitations}

Despite its strengths, the Force Plot has some limitations:
\begin{itemize}
    \item \textbf{Scalability Issues:} For high-dimensional data with many features, the Force Plot can become cluttered and hard to interpret.
    \item \textbf{Dependence on SHAP Values:} The accuracy of the Force Plot heavily relies on the correctness of the SHAP values, which may vary based on the model and dataset.
    \item \textbf{Limited Global Insight:} While Force Plots are useful for understanding individual predictions, they do not provide a global overview of model behavior.
\end{itemize}

\subsection{Decision Plot}

A Decision Plot is a visualization technique used to understand the cumulative impact of features on the prediction made by a model. It is particularly effective for explaining complex models by showing the progression of a model's decision as features are added sequentially. Decision Plots are commonly used in conjunction with SHAP (SHapley Additive exPlanations) values and can be applied to various models, including traditional machine learning models and modern neural networks.

\paragraph{Scope of Application}

Decision Plots are suitable for:
\begin{itemize}
    \item \textbf{Traditional Models:} Linear regression, logistic regression, decision trees, and support vector machines.
    \item \textbf{Machine Learning Models:} Ensemble models like Random Forests and Gradient Boosting Machines (e.g., XGBoost, LightGBM).
    \item \textbf{Deep Learning Models:} Neural networks, including models used in LLMs (Large Language Models), when feature importance can be derived from embeddings or model outputs.
\end{itemize}

\paragraph{Principle and Formula}

The Decision Plot visualizes the cumulative effect of features on the model's output using SHAP values. The main idea is to illustrate how the prediction value evolves as features are considered sequentially, based on their SHAP contributions. The formula for the prediction of instance \( x \) can be expressed as:

\[
f(x) = \phi_0 + \sum_{i=1}^{n} \phi_i,
\]

where:
\begin{itemize}
    \item \( f(x) \) is the model's output for input \( x \).
    \item \( \phi_0 \) is the base value, representing the expected prediction of the model.
    \item \( \phi_i \) is the SHAP value for feature \( i \), indicating its contribution to the overall prediction.
\end{itemize}

In the Decision Plot, the x-axis represents the sequentially added features, and the y-axis shows the model's output. The lines in the plot correspond to individual instances, and the slope of each line segment indicates the contribution of a specific feature.

\paragraph{Python Code Example}

In this example, we utilize the California Housing dataset with a Gradient Boosting Regressor~\cite{Pedregosa2011}. The Decision Plot is generated using SHAP values to illustrate the contribution of individual features to the model predictions.

\begin{lstlisting}[style=python, literate={\$}{{\$}}1]
import shap
import numpy as np
from sklearn.datasets import fetch_california_housing
from sklearn.ensemble import GradientBoostingRegressor

# Load the California housing dataset
data = fetch_california_housing()
X, y = data.data, data.target

# Train the model
model = GradientBoostingRegressor(n_estimators=100, random_state=42)
model.fit(X, y)

# Create a SHAP explainer
explainer = shap.TreeExplainer(model)
shap_values = explainer.shap_values(X)

# Subsample the data (e.g., 1000 observations)
subset_size = 1000
random_indices = np.random.choice(X.shape[0], subset_size, replace=False)
X_subset = X[random_indices]
shap_values_subset = shap_values[random_indices]

# Generate the decision plot with subsampled data
shap.decision_plot(explainer.expected_value, shap_values_subset, X_subset, feature_names=data.feature_names)
\end{lstlisting}

\paragraph{Result Explanation}

The Decision Plot visualizes the following key elements:

\begin{itemize}
    \item \textbf{Base Value:} This is the starting point for each line on the plot, representing the expected value of the model output before any feature contributions are accounted for.
    \item \textbf{Feature Contributions:} Each segment of the line reflects the SHAP value of a feature, showing its impact on the prediction. A positive contribution (upward slope) indicates that the feature increases the model prediction, while a negative contribution (downward slope) shows a reduction in the prediction.
    \item \textbf{Cumulative Prediction:} The endpoint of each line represents the final prediction for a given sample, after considering the contributions of all features.
\end{itemize}

The plot below shows the Decision Plot for a subset of the dataset, highlighting the most impactful features like \texttt{MedInc} (Median Income), \texttt{Latitude}, and \texttt{Longitude}:

\begin{figure}[htbp]
    \centering
    \includegraphics[width=0.8\textwidth]{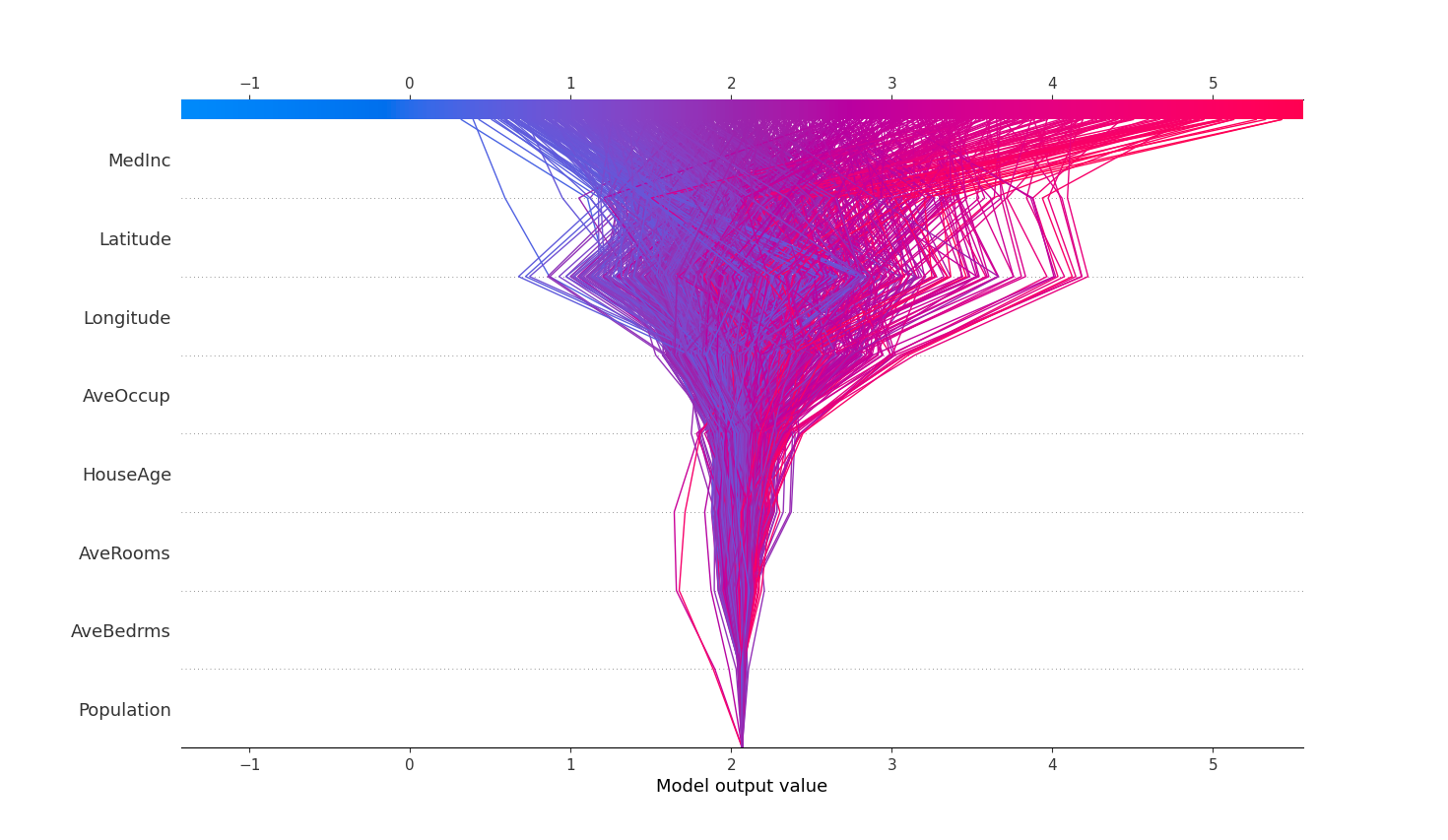}
    \caption{SHAP Decision Plot for California Housing Predictions}
\end{figure}

\paragraph{Advantages of Decision Plots}

\begin{itemize}
    \item \textbf{Clear Interpretation:} Decision Plots provide a straightforward and cumulative visualization of how features influence a prediction.
    \item \textbf{Handles High-Dimensional Data:} Unlike other visualization techniques, Decision Plots can effectively display contributions from numerous features without becoming cluttered.
    \item \textbf{Local and Global Insights:} Decision Plots offer both instance-level (local) explanations and an overview of feature importance across instances (global).
\end{itemize}

\paragraph{Challenges and Limitations}

Despite their usefulness, Decision Plots also have certain limitations:
\begin{itemize}
    \item \textbf{Feature Ordering Sensitivity:} The order in which features are plotted can affect the interpretation. Different orderings may yield different visual insights.
    \item \textbf{Dependence on SHAP Values:} The reliability of the plot hinges on the correctness of the SHAP values, which may be sensitive to the model type and data distribution.
    \item \textbf{Scalability Issues:} While effective for moderate-sized datasets, the visualization can become less interpretable with very large numbers of instances.
\end{itemize}

\subsection{Feature Interaction Detection}

Feature Interaction Detection is a powerful visualization technique used to uncover and interpret complex relationships between features in machine learning models. By understanding interactions, we can better interpret the model's behavior, especially when dealing with non-linear models like neural networks and ensemble methods. This technique is applicable to a wide range of models, including traditional models, machine learning (ML) models, and Large Language Models (LLMs).

\paragraph{Scope of Application}

Feature Interaction Detection can be effectively used in the following scenarios:
\begin{itemize}
    \item \textbf{Traditional Models:} Interaction detection can provide insight into models such as decision trees and logistic regression when interaction terms are included.
    \item \textbf{Machine Learning Models:} This technique is highly relevant for ensemble models (e.g., Random Forest, XGBoost) and deep learning models that may implicitly capture feature interactions.
    \item \textbf{LLMs and Neural Networks:} In the context of LLMs, interaction detection can be applied to understand how different input tokens or embeddings interact to influence predictions.
\end{itemize}

\paragraph{Principle and Formula}

The key idea behind feature interaction detection is to quantify the contribution of feature pairs beyond their individual effects. A common approach to measure interaction effects is using SHAP Interaction values \cite{Lundberg2018}, defined as follows:

\[
\phi_{i,j} = \text{SHAP}(i, j) - \phi_i - \phi_j,
\]

where:
\begin{itemize}
    \item \( \phi_{i,j} \) is the SHAP Interaction value for features \( i \) and \( j \).
    \item \( \text{SHAP}(i, j) \) represents the SHAP value when both features \( i \) and \( j \) are considered together.
    \item \( \phi_i \) and \( \phi_j \) are the individual SHAP values for features \( i \) and \( j \), respectively.
\end{itemize}

The SHAP Interaction value \( \phi_{i,j} \) measures the additional contribution to the model output when features \( i \) and \( j \) are considered together, beyond their individual effects. A positive interaction value indicates a synergistic relationship, while a negative value suggests redundancy or diminishing effects.

\paragraph{Python Code Example}

In this example, we utilize the California Housing dataset and a Gradient Boosting Regressor \cite{Pedregosa2011} to detect and visualize feature interactions using SHAP values. The SHAP interaction values help us understand how pairs of features jointly contribute to the model predictions.

\begin{lstlisting}[style=python, literate={\$}{{\$}}1]
import shap
import numpy as np
from sklearn.datasets import fetch_california_housing
from sklearn.ensemble import GradientBoostingRegressor
import matplotlib.pyplot as plt

# Load the California housing dataset
data = fetch_california_housing()
X, y = data.data, data.target
feature_names = data.feature_names

# Train the model
model = GradientBoostingRegressor(n_estimators=100, random_state=42)
model.fit(X, y)

# Create SHAP explainer and compute SHAP values
explainer = shap.TreeExplainer(model)
shap_values = explainer.shap_values(X)

# Compute SHAP interaction values
shap_interaction_values = explainer.shap_interaction_values(X)

# Plot the SHAP interaction values (summary plot)
shap.summary_plot(shap_interaction_values, X, feature_names=feature_names, plot_type="compact_dot")

# Visualize a specific feature pair interaction
shap.dependence_plot(("MedInc", "AveRooms"), shap_interaction_values, X, feature_names=feature_names, interaction_index="AveRooms")
\end{lstlisting}

\paragraph{Result Explanation}

The result of the example above is illustrated in the following points:

\begin{itemize}
    \item \textbf{SHAP Interaction Values:} The `shap\_interaction\_values` function calculates the interaction values for all feature pairs, revealing how each pair of features jointly affects the model's predictions.
    \item \textbf{Summary Plot:} The summary plot provides an overview of the strongest interactions across all data points. It highlights key feature pairs with the most significant interactions. For instance, interactions between features like \texttt{MedInc} (Median Income) and \texttt{Latitude} are shown prominently.
    \item \textbf{Dependence Plot:} The dependence plot illustrates the interaction effect between \texttt{MedInc} (Median Income) and \texttt{AveRooms} (Average Number of Rooms). The color gradient represents the values of the interacting feature (\texttt{AveRooms}), providing insights into how these two features interact.
\end{itemize}

The plot below shows the SHAP summary plot for feature interactions in the California Housing dataset:

\begin{figure}[htbp]
    \centering
    \includegraphics[width=0.8\textwidth]{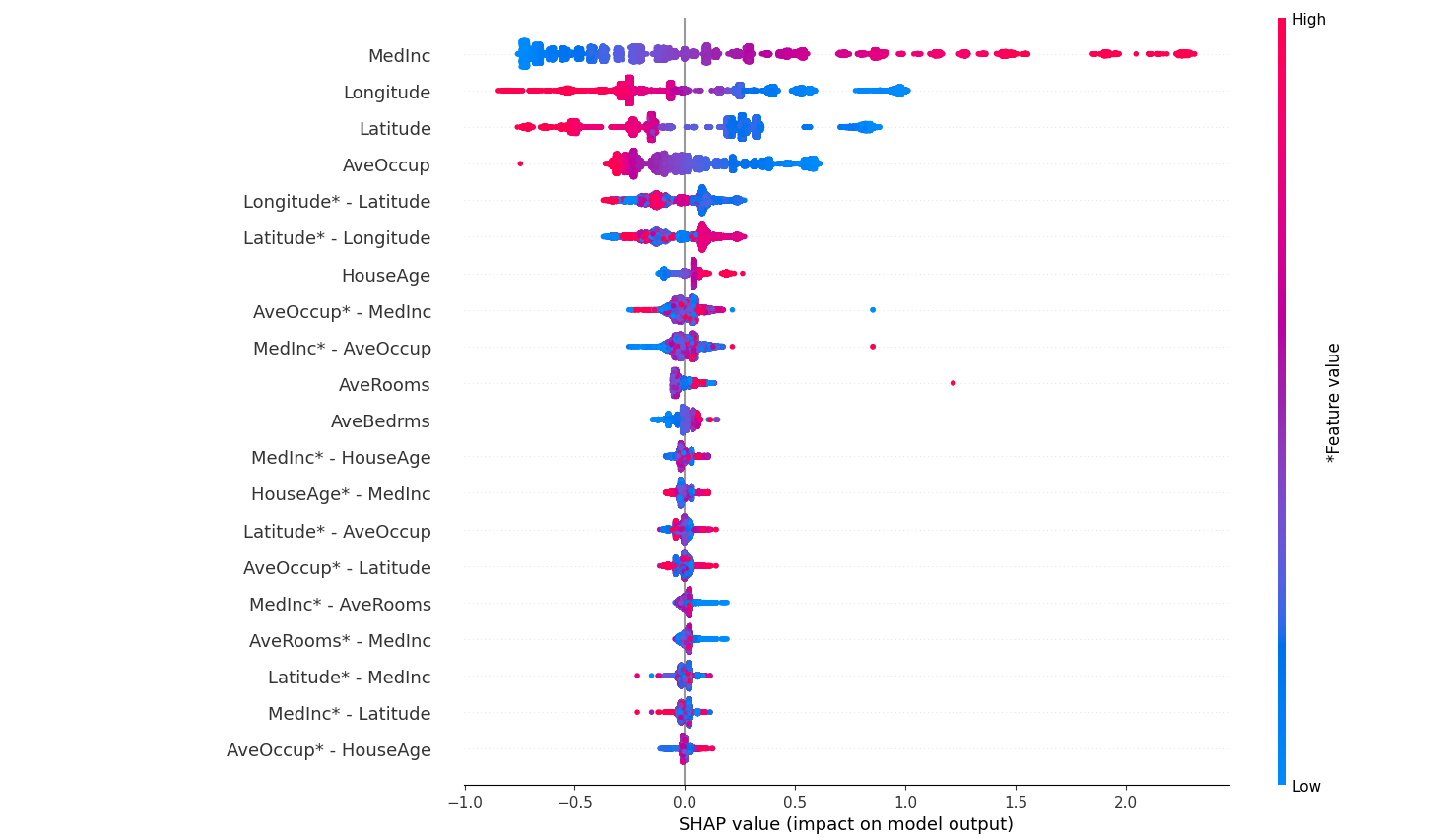}
    \caption{SHAP Summary Plot of Feature Interactions for California Housing Predictions}
\end{figure}

In the dependence plot, we observe that the effect of \texttt{MedInc} on the predicted house price changes based on the value of \texttt{AveRooms}. This indicates a strong interaction effect, where the influence of median income on house prices is moderated by the average number of rooms. Such interactions are critical in understanding the model's decision-making process, providing deeper insights into the underlying relationships within the data.

\paragraph{Advantages of Feature Interaction Detection}

\begin{itemize}
    \item \textbf{Uncovers Complex Relationships:} Helps in identifying non-linear dependencies and interactions that are not captured by traditional linear models.
    \item \textbf{Model-Agnostic Approach:} SHAP-based interaction detection can be applied across different model types, from tree-based models to neural networks.
    \item \textbf{Improves Model Interpretability:} By visualizing interactions, it becomes easier to understand the joint influence of features, enhancing the overall explainability of the model.
\end{itemize}

\paragraph{Challenges and Limitations}

\begin{itemize}
    \item \textbf{Computational Complexity:} Calculating interaction values for all feature pairs can be computationally expensive, especially for high-dimensional data.
    \item \textbf{Dependence on Model Type:} The quality of interaction detection may vary depending on the model. For instance, linear models may not exhibit strong interactions, whereas non-linear models may capture complex dependencies.
    \item \textbf{Interpretation Difficulty:} While interaction detection provides insights, interpreting the exact nature of interactions can be challenging, especially for non-experts.
\end{itemize}

\section{Temporal and Sequence Data Techniques}

Temporal and sequence data present unique challenges for interpretability due to their inherent time-based dependencies and complex patterns \cite{fawaz2019deep}. In explainable AI (XAI), specialized techniques have been developed to address these complexities and provide meaningful insights into model behavior for time series data \cite{arrieta2020explainable}. This section explores several key methods, including \textbf{TimeSHAP} \cite{arango2020timeshap}, which adapts SHAP values for temporal data; \textbf{Dynamic Time Warping (DTW) Explainer}, used to align and explain sequence similarities; \textbf{Attention-based Explanations}, which leverage the model's attention mechanisms to highlight important time steps \cite{qin2017dual}; and \textbf{Saliency Maps for Recurrent Neural Networks}, which identify influential features in time-dependent predictions \cite{arras2017explaining}.

\subsection{TimeSHAP}

\paragraph{Scope of Application}

TimeSHAP is a post-hoc interpretability method tailored specifically for temporal and sequential data models, such as Recurrent Neural Networks (RNNs), Long Short-Term Memory (LSTM) networks, and Transformer models \cite{arango2020timeshap}. It extends the SHAP (SHapley Additive exPlanations) framework \cite{Lundberg2017} to provide explanations for individual predictions in a temporal context, making it particularly relevant for time-series models used in fields such as financial forecasting, anomaly detection, and natural language processing (NLP).

\paragraph{Principle and Concept}

TimeSHAP builds on the concept of Shapley values from cooperative game theory \cite{Lundberg2017}. The Shapley value for a feature measures its contribution to a model's prediction by averaging its marginal contributions across all possible feature subsets. For time-series data, the importance of features (timestamps) can vary over time. TimeSHAP addresses this by decomposing Shapley values over the temporal sequence, evaluating the significance of each timestamp.

Given a time-series instance \( x = [x_1, x_2, \ldots, x_T] \), where \( x_t \) represents the feature vector at time \( t \), the Shapley value \( \phi_t \) for timestamp \( t \) is formally expressed as:

\[
\phi_t = \frac{1}{|\mathcal{S}|} \sum_{\mathcal{S} \subseteq \{1, \ldots, T\} \setminus \{t\}} \left[ f(x_{\mathcal{S} \cup \{t\}}) - f(x_{\mathcal{S}}) \right],
\]

where \( \mathcal{S} \) is a subset of timestamps, and \( f(\cdot) \) denotes the model prediction function. This equation captures the marginal contribution of the feature vector at time \( t \), averaged over all possible subsets.

\paragraph{Algorithm and Approach}

The TimeSHAP methodology includes the following key steps:
\begin{enumerate}
    \item \textbf{Sequence Partitioning:} The time-series data is partitioned into meaningful segments to preserve the temporal order.
    \item \textbf{Perturbation Sampling:} Perturbations are generated by masking or altering parts of the sequence, simulating different scenarios.
    \item \textbf{Shapley Value Computation:} Shapley values for each timestamp are computed based on the perturbed samples to quantify their contributions.
    \item \textbf{Visualization:} The importance of each timestamp is visualized using line plots or heatmaps, illustrating the contribution dynamics.
\end{enumerate}

\paragraph{Python Code Example}

\begin{lstlisting}[style=python, literate={\$}{{\$}}1]
import numpy as np
import tensorflow as tf
import shap

# Generate synthetic time-series data
time_series_data = np.random.rand(100, 10, 1)  # 100 samples, 10 time steps, 1 feature
labels = np.random.randint(0, 2, size=(100,))

# Define a simple LSTM model
model = tf.keras.Sequential([
    tf.keras.layers.Input(shape=(10, 1)),
    tf.keras.layers.LSTM(50),
    tf.keras.layers.Dense(1, activation='sigmoid')
])
model.compile(optimizer='adam', loss='binary_crossentropy')
model.fit(time_series_data, labels, epochs=5, verbose=0)

# Define the prediction function for SHAP
def predict_fn(data):
    # Reshape input data back to 3D for the LSTM model
    return model.predict(data.reshape(-1, 10, 1)).flatten()

# Select a background dataset (subset of training data)
background_data = time_series_data[:50].reshape(50, -1)  # Flatten to 2D

# Initialize SHAP KernelExplainer
explainer = shap.KernelExplainer(predict_fn, background_data)

# Select an instance to explain and flatten it
instance = time_series_data[0:1].reshape(1, -1)  # Flatten to 2D

# Compute SHAP values
shap_values = explainer.shap_values(instance, nsamples=100)

# Display SHAP values
print("SHAP values for each flattened feature:", shap_values)

# Visualization of SHAP values
import matplotlib.pyplot as plt

# Plot SHAP values for the first instance
plt.bar(range(len(shap_values[0])), shap_values[0])
plt.xlabel('Flattened Feature Index')
plt.ylabel('SHAP Value')
plt.title('SHAP Values for Time-Series Instance')
plt.show()
\end{lstlisting}

\paragraph{Results Explanation}

\begin{figure}[htbp]
    \centering
    \includegraphics[width=0.8\textwidth]{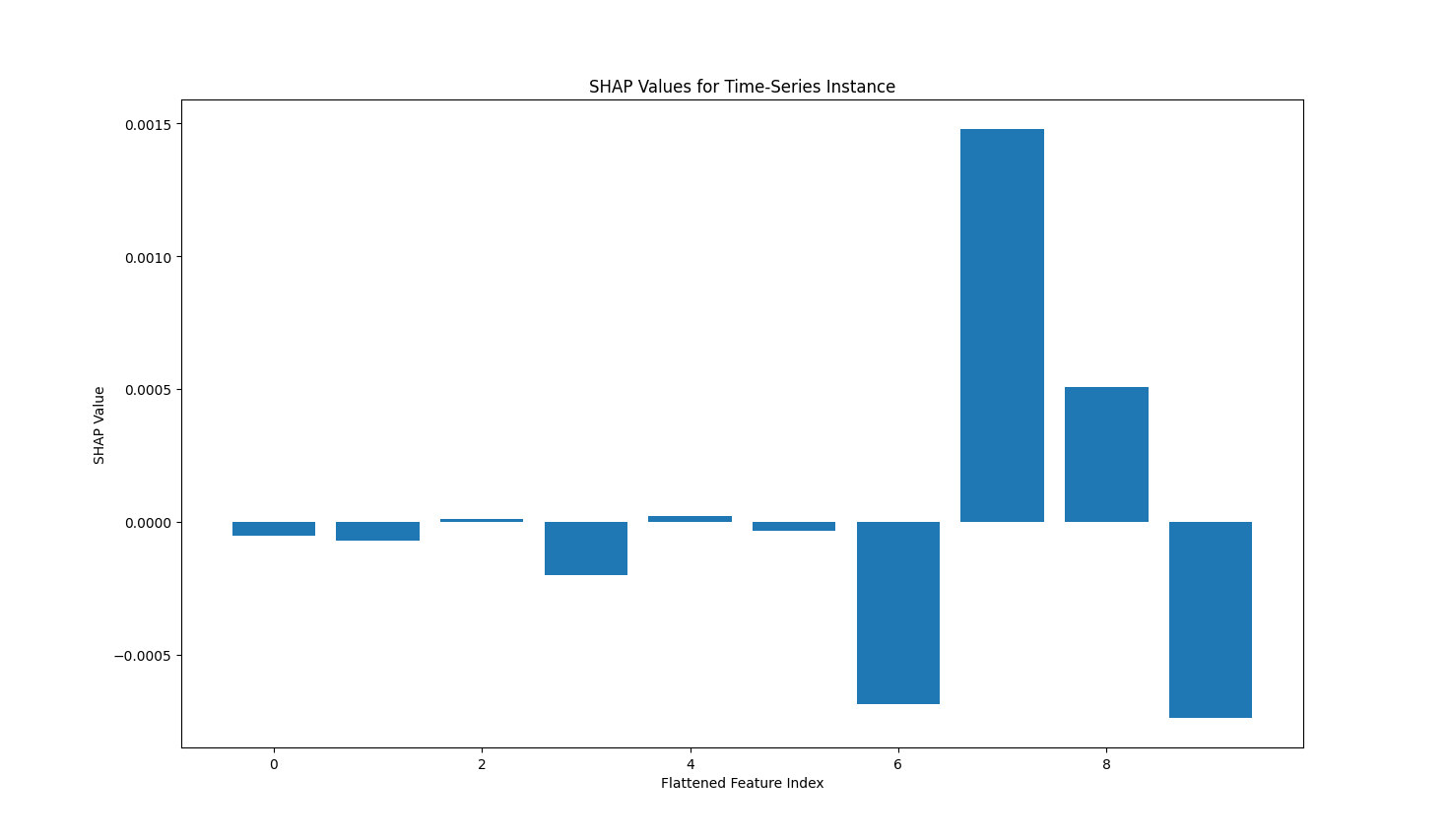}
    \caption{Visualization of SHAP Values for a Time-Series Instance}
    \label{fig:timeshap}
\end{figure}

In the provided example, we trained a simple LSTM model on synthetic binary time-series data. Using TimeSHAP, we explained the contribution of each timestamp in a specific instance. The resulting SHAP values indicate the importance of each timestamp, as shown in Figure~\ref{fig:timeshap}.

The plot illustrates that certain timestamps (e.g., index 6) have higher SHAP values, indicating a stronger influence on the model's prediction. This insight can help identify key time points that are critical for the model's decision-making process.

\paragraph{Advantages of TimeSHAP}

\begin{itemize}
    \item \textbf{Temporal Sensitivity:} TimeSHAP captures the dynamic importance of features over time, making it well-suited for time-series models.
    \item \textbf{Model-Agnostic:} It can be applied to any sequential model, including deep learning and traditional time-series models.
    \item \textbf{Local Explanations:} Provides explanations tailored to individual predictions, enhancing interpretability for time-series data.
\end{itemize}

\paragraph{Challenges and Limitations}

\begin{itemize}
    \item \textbf{Computational Complexity:} Calculating Shapley values for each timestamp can be resource-intensive, especially for long sequences.
    \item \textbf{Sensitivity to Perturbations:} The quality of the explanation may depend on the perturbation strategy, requiring careful selection of sampling methods.
    \item \textbf{Scalability Issues:} TimeSHAP may struggle with very large datasets or real-time applications due to the need for extensive sampling.
\end{itemize}

\subsection{Dynamic Time Warping (DTW) Explainer}

\paragraph{Scope of Application}

The Dynamic Time Warping (DTW) Explainer is a technique used primarily for time-series models. It is well-suited for traditional time-series analysis methods, machine learning models like k-NN and SVM when applied to time-series data, and certain neural network architectures like RNNs and LSTMs. The DTW Explainer focuses on aligning time-series sequences to measure similarity and explain model predictions by identifying the most important subsequences that influence the output.

\paragraph{Principle and Concept}

Dynamic Time Warping (DTW) is a classic algorithm for measuring the similarity between two time-series sequences, even if they differ in length or speed. It works by finding an optimal alignment between the sequences that minimizes the cumulative distance. For instance, given two sequences \( X = [x_1, x_2, \ldots, x_m] \) and \( Y = [y_1, y_2, \ldots, y_n] \), DTW aims to warp these sequences non-linearly in time to align similar points.

The DTW distance between \( X \) and \( Y \) is defined as:

\[
\text{DTW}(X, Y) = \min_{\text{alignment}} \sum_{i=1}^{k} d(x_{a_i}, y_{b_i}),
\]

where \( d(x_{a_i}, y_{b_i}) \) is the distance between aligned points \( x_{a_i} \) and \( y_{b_i} \). The alignment is determined by constructing a cost matrix and finding a warping path that minimizes the total cumulative distance.

In the context of explainability, DTW can be used to identify key subsequences in the time-series data that have the most significant impact on the model's prediction. The DTW Explainer highlights these subsequences and provides insights into why certain parts of the time series are more influential.

\paragraph{Algorithm and Approach}

The DTW Explainer works as follows:
\begin{enumerate}
    \item \textbf{Sequence Alignment:} Align the input time series with a reference time series (e.g., a prototype or the model's prediction) using the DTW algorithm.
    \item \textbf{Identify Important Subsequences:} Calculate the alignment costs and identify which parts of the input sequence have the highest influence based on the cumulative distance.
    \item \textbf{Visualization:} Highlight the influential subsequences using heatmaps or annotated time-series plots to convey the explanations visually.
\end{enumerate}

\paragraph{Python Code Example}

\begin{lstlisting}[style=python, literate={\$}{{\$}}1]
import numpy as np
import matplotlib.pyplot as plt
from fastdtw import fastdtw

# Generate synthetic time-series data
time_series = np.sin(np.linspace(0, 2 * np.pi, 100)).flatten()  # Flatten to 1-D
reference_series = np.sin(np.linspace(0, 2 * np.pi, 120) + 0.5).flatten()  # Flatten to 1-D

# Ensure both time_series and reference_series are 1-D
print(f"Time series shape: {time_series.shape}, Reference series shape: {reference_series.shape}")

# Define custom distance function for scalar values
def scalar_distance(u, v):
    return abs(u - v)

# Apply DTW to align the sequences using the custom distance function
distance, path = fastdtw(time_series, reference_series, dist=scalar_distance)

# Plot the aligned sequences and highlight the warping path
plt.figure(figsize=(10, 5))
plt.plot(time_series, label="Time Series")
plt.plot(
    np.interp(np.linspace(0, 100, 120), np.arange(120), reference_series),
    label="Reference Series",
    alpha=0.7
)

# Highlight the warping path
for (i, j) in path:
    plt.plot([i, j * 100 / 120], [time_series[i], reference_series[j]], color='gray', alpha=0.5)

plt.title(f"DTW Alignment (Distance: {distance:.2f})")
plt.xlabel("Time Index")
plt.ylabel("Value")
plt.legend()
plt.grid()
plt.show()
\end{lstlisting}

\paragraph{Results Explanation}

In this example, we generated a synthetic time series and a slightly shifted reference series. The DTW algorithm finds the optimal alignment between the two sequences, even though they have different lengths and phases. The plot visualizes the warping path, showing which points in the time series are aligned with the reference series.

The DTW Explainer identifies the segments of the time series that have the highest alignment cost, indicating the subsequences that are most influential in determining the similarity. In practical applications, such as anomaly detection or forecasting, these influential subsequences can help users understand why the model made a certain prediction.

\begin{figure}[htbp]
    \centering
    \includegraphics[width=0.8\textwidth]{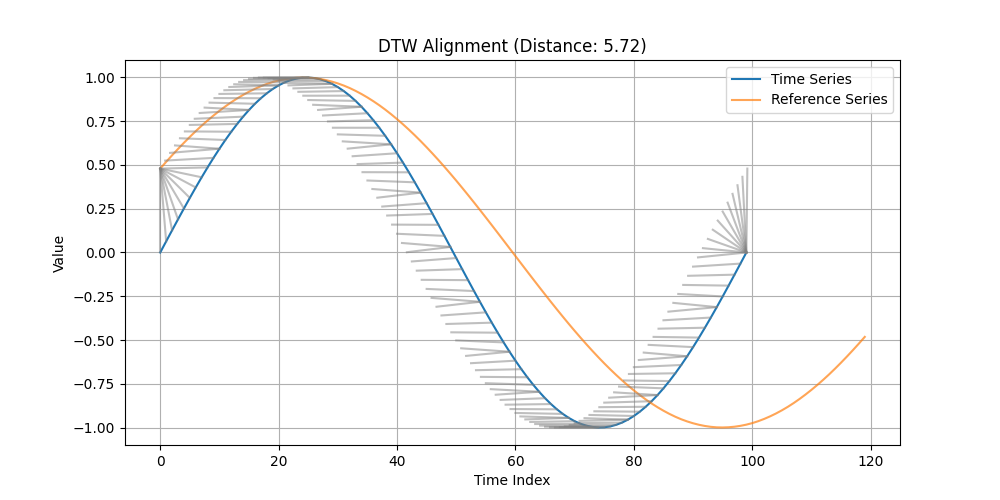}
    \caption{DTW Alignment Visualization with Warping Path}
    \label{fig:dtw_alignment}
\end{figure}

\paragraph{Advantages of DTW Explainer}

\begin{itemize}
    \item \textbf{Handles Temporal Misalignment:} DTW is robust to shifts and differences in the time axis, making it ideal for explaining time-series data.
    \item \textbf{Model-Agnostic:} The technique does not depend on a specific model type and can be applied to any time-series model.
    \item \textbf{Intuitive Visualization:} The warping path provides a clear visual representation of which parts of the time series are most similar or influential.
\end{itemize}

\paragraph{Challenges and Limitations}

\begin{itemize}
    \item \textbf{Computational Complexity:} The DTW algorithm can be computationally intensive, especially for long time-series sequences.
    \item \textbf{Sensitivity to Noise:} DTW may produce misleading results when the input data is noisy, as it tries to align every point regardless of signal quality.
    \item \textbf{Scalability Issues:} In large datasets, applying DTW to every instance can be resource-intensive, limiting its applicability in real-time scenarios.
\end{itemize}

\subsection{Attention-based Explanation for Time Series}

\paragraph{Scope of Application}

Attention-based explanations for time series data have become a prominent technique in both traditional machine learning and deep learning, particularly in models dealing with sequence data. This method is most effective when applied to models that inherently utilize attention mechanisms, such as Recurrent Neural Networks (RNNs), Long Short-Term Memory networks (LSTMs), Transformer models, and some advanced LLMs (Large Language Models) adapted for time series prediction \cite{Vaswani2017}.

\paragraph{Principle and Concept}

The attention mechanism was introduced to address the limitation of sequence models that struggle to retain information over long input sequences \cite{bahdanau2015neural}. The core idea is to allow the model to weigh each input time step differently based on its importance. Given an input sequence \( X = [x_1, x_2, \ldots, x_n] \), the attention mechanism calculates a weight \( \alpha_i \) for each time step \( x_i \). These weights reflect the importance of each input step in influencing the model's output.

The attention score for each time step \( i \) can be computed as follows:

\[
\alpha_i = \frac{\exp(e_i)}{\sum_{j=1}^{n} \exp(e_j)},
\]

where \( e_i \) is the alignment score that indicates the relevance of time step \( x_i \) to the current prediction. The context vector \( c \), which summarizes the relevant information from the input sequence, is computed as:

\[
c = \sum_{i=1}^{n} \alpha_i \cdot x_i.
\]

For time series explanations, we can visualize the attention scores \( \alpha_i \) to understand which time steps the model considers most influential.

\paragraph{Algorithm and Approach}

The attention-based explanation technique for time series follows these steps:
\begin{enumerate}
    \item \textbf{Model Training:} Train a time-series model with an integrated attention mechanism, such as an LSTM with attention or a Transformer.
    \item \textbf{Attention Score Calculation:} During inference, extract the attention scores \( \alpha_i \) for each time step of the input sequence.
    \item \textbf{Visualization:} Plot the attention scores over time to identify which time steps have the highest influence on the model's prediction.
\end{enumerate}

\paragraph{Python Code Example}

This example demonstrates how to use an LSTM model with an attention mechanism to predict a shifted version of a time series. By leveraging attention weights, the model identifies and highlights the most relevant time steps, offering enhanced interpretability and a clear understanding of its decision-making process.

\begin{lstlisting}[style=python, literate={\$}{{\$}}1]
import numpy as np
import tensorflow as tf
import matplotlib.pyplot as plt

# Generate synthetic time series data
np.random.seed(0)
time_steps = 100
X = np.sin(np.linspace(0, 2 * np.pi, time_steps)) + np.random.normal(0, 0.1, time_steps)
y = np.roll(X, -1)

# Reshape data to fit LSTM input
X_input = X.reshape((1, time_steps, 1))
y_input = y.reshape((1, time_steps, 1))

# Define the Attention Layer
class AttentionLayer(tf.keras.layers.Layer):
    def __init__(self):
        super(AttentionLayer, self).__init__()
        self.score_layer = tf.keras.layers.Dense(1, activation='tanh')

    def call(self, x):
        score = self.score_layer(x)
        attention_weights = tf.nn.softmax(score, axis=1)
        context_vector = attention_weights * x
        context_vector = tf.reduce_sum(context_vector, axis=1)
        return context_vector, attention_weights

# Build the model
inputs = tf.keras.Input(shape=(time_steps, 1))
lstm_output = tf.keras.layers.LSTM(50, return_sequences=True)(inputs)
context_vector, attention_weights = AttentionLayer()(lstm_output)
outputs = tf.keras.layers.Dense(1)(context_vector)
model = tf.keras.Model(inputs=inputs, outputs=outputs)

# Compile and train the model
model.compile(optimizer='adam', loss='mse')
model.fit(X_input, y_input[:, -1, :], epochs=10, verbose=0)  # Adjust target values to match output shape

# Create a model to output predictions and attention weights
attention_model = tf.keras.Model(inputs=inputs, outputs=[outputs, attention_weights])

# Get predictions and attention weights
prediction, att_weights = attention_model.predict(X_input)

# Plot the original data, predicted values, and attention weights
plt.figure(figsize=(12, 6))

# Plot original data and predicted values on the left y-axis
ax1 = plt.gca()
ax1.plot(np.arange(time_steps), X, label='Original Data', color='blue')
ax1.plot(time_steps - 1, prediction[0], 'ro', label='Predicted Value')
ax1.set_xlabel('Time Step')
ax1.set_ylabel('Data Value')
ax1.legend(loc='upper left')

# Plot attention weights on the right y-axis
ax2 = ax1.twinx()
ax2.plot(np.arange(time_steps), att_weights[0, :, 0], label='Attention Weights', color='green', alpha=0.5)
ax2.set_ylabel('Attention Weights')
ax2.legend(loc='upper right')

plt.title('Original Data, Predicted Values, and Attention Weights')
plt.show()
\end{lstlisting}

\paragraph{Advantages of Attention-based Explanation for Time Series}

In this example, we used a simple LSTM model with an attention layer to predict a shifted version of the input time series. The attention mechanism allows the model to focus on specific time steps that are most relevant for making the prediction.

The plot of attention weights shows how the model assigns varying importance to different time steps. Higher attention scores indicate the time steps that have a greater influence on the model's output. This visualization helps users understand which parts of the time series the model relies on, making the predictions more interpretable.

\begin{figure}[htbp]
    \centering
    \includegraphics[width=0.8\textwidth]{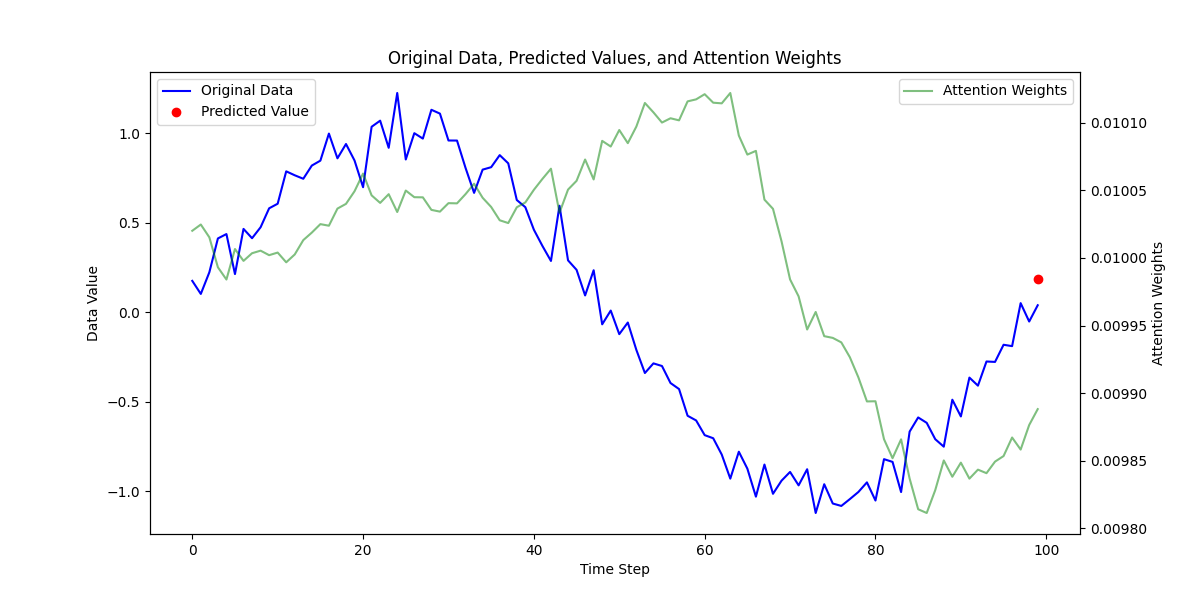}
    \caption{Original Data, Predicted Values, and Attention Weights}
\end{figure}

\paragraph{Results Explanation}

\begin{itemize}
    \item \textbf{Enhanced Interpretability:} Attention scores provide an intuitive way to identify the most influential parts of the time series.
    \item \textbf{Model Flexibility:} The technique can be integrated into various time-series models, including LSTMs, GRUs, and Transformers.
    \item \textbf{Contextual Insight:} The attention mechanism considers the entire input sequence, allowing the model to capture long-term dependencies and context.
\end{itemize}

\paragraph{Challenges and Limitations}

\begin{itemize}
    \item \textbf{Overfitting Risk:} The attention mechanism may introduce additional parameters, increasing the risk of overfitting, especially with limited data.
    \item \textbf{Interpretation Complexity:} While attention scores are helpful, they do not always provide a complete causal explanation for model predictions.
    \item \textbf{Computational Overhead:} The calculation of attention scores can be computationally intensive, especially for long input sequences.
\end{itemize}

\subsection{Saliency Maps for Recurrent Neural Networks}

\paragraph{Scope of Application}

Saliency maps are a widely-used technique for visualizing and interpreting neural network models, primarily focusing on understanding which input features contribute most to the model's predictions \cite{simonyan2013deep}. This technique can be adapted for various types of neural network architectures, including Convolutional Neural Networks (CNNs), Recurrent Neural Networks (RNNs), and even Large Language Models (LLMs) when they handle sequential or temporal data. In this section, we focus on applying saliency maps to RNNs, such as Long Short-Term Memory (LSTM) and Gated Recurrent Unit (GRU) models, particularly in the context of time-series data analysis.

\paragraph{Principle and Methodology}

The core idea of saliency maps is to compute the gradient of the model's output with respect to each input feature \cite{sundararajan2017axiomatic}. For RNNs, the model's output at time step \( t \), denoted as \( y_t \), can be influenced by the input sequence \( X = [x_1, x_2, \ldots, x_n] \). The saliency value for each input feature \( x_i \) is calculated as the magnitude of the partial derivative of the output \( y_t \) with respect to \( x_i \):

\[
\text{Saliency}(x_i) = \left| \frac{\partial y_t}{\partial x_i} \right|.
\]

This gradient represents how sensitive the output is to changes in each input feature. Higher gradient values indicate greater importance of the corresponding input feature, thus highlighting the relevant time steps or sequence elements that the model focused on.

\paragraph{Algorithm and Approach}

The steps to generate a saliency map for an RNN model are as follows:
\begin{enumerate}
    \item \textbf{Model Training:} Train the RNN model (e.g., LSTM) on the time-series dataset.
    \item \textbf{Gradient Calculation:} Compute the gradients of the model's output with respect to each input time step.
    \item \textbf{Saliency Map Construction:} Generate a saliency map by visualizing the absolute gradient values for each input time step.
    \item \textbf{Visualization:} Plot the saliency map to interpret which parts of the input sequence contribute most to the model's decision.
\end{enumerate}

\paragraph{Python Code Example}

The following example demonstrates how to compute a saliency map for a simple time-series prediction task using an LSTM model. Saliency maps are powerful tools for visualizing the importance of input features, offering valuable insights into how models make predictions. By leveraging gradient-based methods, we can identify which time steps in the input sequence have the most significant influence on the model's output.

In this example, synthetic time-series data is generated and used to train an LSTM model designed to predict the next time step in the sequence. Gradients of the loss with respect to the input sequence are computed to create the saliency map, which reveals the importance of individual time steps. This technique can be particularly useful in tasks requiring interpretability, such as anomaly detection or trend analysis in time-series data.

Below is the code, followed by an explanation of the results and their interpretation.

\begin{lstlisting}[style=python, literate={\$}{{\$}}1]
import numpy as np
import tensorflow as tf
import matplotlib.pyplot as plt

# Generate synthetic time series data
np.random.seed(42)
time_steps = 100
X = np.sin(np.linspace(0, 4 * np.pi, time_steps)) + np.random.normal(0, 0.1, time_steps)
y = np.roll(X, -1)

# Reshape data to fit LSTM input
X_input = X.reshape((1, time_steps, 1))
y_input = y.reshape((1, time_steps, 1))

# Define LSTM model to predict the entire sequence
inputs = tf.keras.Input(shape=(time_steps, 1))
lstm_output = tf.keras.layers.LSTM(50, return_sequences=True)(inputs)
outputs = tf.keras.layers.TimeDistributed(tf.keras.layers.Dense(1))(lstm_output)
model = tf.keras.Model(inputs=inputs, outputs=outputs)

# Compile and train the model
model.compile(optimizer='adam', loss='mse')
model.fit(X_input, y_input, epochs=10, verbose=0)

# Convert X and y to TensorFlow tensors
X_tensor = tf.convert_to_tensor(X_input, dtype=tf.float32)
y_tensor = tf.convert_to_tensor(y_input, dtype=tf.float32)

# Compute the saliency map
with tf.GradientTape() as tape:
    tape.watch(X_tensor)
    predictions = model(X_tensor)
    # Use MeanSquaredError loss object
    loss_object = tf.keras.losses.MeanSquaredError()
    loss = tf.reduce_mean(loss_object(y_tensor, predictions))

# Compute the gradients
grads = tape.gradient(loss, X_tensor)
grads = grads.numpy()[0, :, 0]  # Extract gradients for each time step

# Plot the original data, predicted data, and saliency map
fig, ax1 = plt.subplots(figsize=(12, 6))

# Plot the original data and predicted data on the left y-axis
color = 'tab:blue'
ax1.set_xlabel('Time Step')
ax1.set_ylabel('Data Value', color=color)
ax1.plot(X_input[0, :, 0], color='blue', label='Original Data')
ax1.plot(predictions.numpy()[0, :, 0], color='orange', label='Predicted Data')
ax1.tick_params(axis='y', labelcolor=color)
ax1.legend(loc='upper left')

# Plot the saliency map on the right y-axis
ax2 = ax1.twinx()
color = 'tab:red'
ax2.set_ylabel('Gradient Magnitude', color=color)
ax2.plot(np.abs(grads), color=color, label='Saliency Map')
ax2.tick_params(axis='y', labelcolor=color)
ax2.legend(loc='upper right')

plt.title('Original Data, Predicted Data, and Saliency Map')
plt.show()
\end{lstlisting}

\paragraph{Results Explanation}

In the code example above, we trained a simple LSTM model on synthetic time-series data. The saliency map is generated by computing the gradients of the model's output with respect to the input sequence. The plot shows the absolute gradient values for each time step, indicating the importance of each input feature.

High gradient values at specific time steps imply that those parts of the sequence had a significant impact on the model's output. For example, peaks in the saliency map may correspond to sudden changes or critical events in the time series, revealing the model's focus during prediction.

\begin{figure}[htbp]
    \centering
    \includegraphics[width=0.8\textwidth]{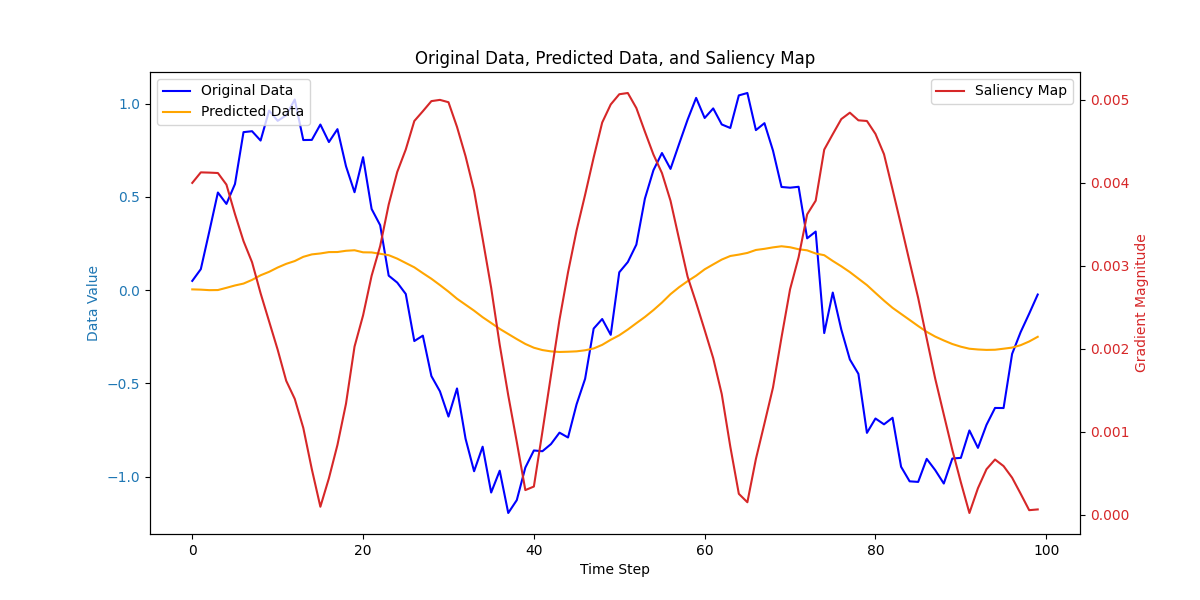}
    \caption{Original Data, Predicted Data, and Saliency Map}
\end{figure}

\paragraph{Advantages of Saliency Maps}

\begin{itemize}
    \item \textbf{Interpretability:} Saliency maps provide an intuitive and visual method to understand the importance of input features, making them suitable for complex models like RNNs and LLMs.
    \item \textbf{Flexibility:} The technique can be applied to any differentiable model, making it adaptable for various architectures and data types, including time-series and sequence data.
    \item \textbf{Highlighting Temporal Dependencies:} Saliency maps for RNNs effectively capture temporal dependencies, offering insights into which time steps are crucial for the model's prediction.
\end{itemize}

\paragraph{Challenges and Limitations}

\begin{itemize}
    \item \textbf{Noise Sensitivity:} Saliency maps can be sensitive to noise in the data, potentially highlighting irrelevant features if the input is noisy.
    \item \textbf{Interpretation Complexity:} While saliency maps highlight important features, they do not provide a causal explanation, and interpreting the gradients can be challenging for non-experts.
    \item \textbf{Gradient Saturation:} In deep networks, the gradients may become saturated, reducing the effectiveness of saliency maps in explaining model predictions.
\end{itemize}

\section{Causal Inference Techniques}
Causal inference techniques are essential in explainable AI (XAI) for uncovering true cause-and-effect relationships rather than mere correlations \cite{pearl2016causal}. These methods help to understand the underlying mechanisms of data and provide more robust, generalizable explanations of model predictions. This section covers several core approaches, including \textbf{Causal Mediation Analysis}, which decomposes the effects of input features into direct and indirect effects \cite{imai2010general}, \textbf{Invariant Risk Minimization (IRM)}, aimed at learning causal features that remain invariant across different environments \cite{arjovsky2019invariant}, \textbf{Causal Discovery Models}, which identify potential causal structures from observational data \cite{spirtes2016causal}, and \textbf{Structural Causal Models (SCMs)}, providing a framework for explicit causal reasoning and intervention analysis \cite{peters2017elements}.

\subsection{Causal Mediation Analysis}

Causal Mediation Analysis (CMA) is a statistical technique used to investigate the causal mechanisms behind observed relationships. Specifically, it aims to determine whether the effect of an independent variable (treatment) on an outcome is mediated by an intermediate variable (mediator) \cite{vanderweele2015explanation}. This approach helps to quantify both the direct effect of the treatment and the indirect effect that occurs through the mediator. CMA is widely used in social sciences, healthcare, and machine learning for understanding complex causal pathways \cite{imai2010identification}.

\paragraph{Scope of Application}

Causal Mediation Analysis is suitable for:
\begin{itemize}
    \item \textbf{Traditional Statistical Models:} Applicable to linear regression models and logistic regression for assessing mediation effects.
    \item \textbf{Machine Learning Models:} Can be extended to more complex models such as decision trees, random forests, and neural networks using model-agnostic mediation techniques.
    \item \textbf{Large Language Models (LLMs):} In LLMs, mediation analysis can be adapted to study how changes in specific input features or embeddings influence model predictions through intermediate layers or representations.
\end{itemize}

\paragraph{Principles and Formula}

The core idea of Causal Mediation Analysis is to decompose the total effect of the treatment \(X\) on the outcome \(Y\) into:
\begin{itemize}
    \item \textbf{Direct Effect (DE):} The portion of the effect of \(X\) on \(Y\) that is not mediated by \(M\).
    \item \textbf{Indirect Effect (IE):} The portion of the effect of \(X\) on \(Y\) that is mediated by the intermediate variable \(M\).
\end{itemize}

The total effect (TE) can be expressed as:

\[
\text{TE} = \text{DE} + \text{IE}.
\]

Assuming a simple linear model, the relationships can be defined as:

\[
M = \alpha_0 + \alpha_1 X + \epsilon_M,
\]

\[
Y = \beta_0 + \beta_1 X + \beta_2 M + \epsilon_Y,
\]

where:
\begin{itemize}
    \item \(X\) is the treatment variable.
    \item \(M\) is the mediator variable.
    \item \(Y\) is the outcome variable.
    \item \(\alpha_1\) is the effect of \(X\) on \(M\).
    \item \(\beta_1\) is the direct effect of \(X\) on \(Y\).
    \item \(\beta_2\) is the effect of \(M\) on \(Y\).
\end{itemize}

The indirect effect (IE) can be calculated as:

\[
\text{IE} = \alpha_1 \cdot \beta_2,
\]

and the direct effect (DE) is given by:

\[
\text{DE} = \beta_1.
\]

The proportion of the total effect mediated by \(M\) can be computed as:

\[
\text{Proportion Mediated} = \frac{\text{IE}}{\text{TE}}.
\]

\paragraph{Python Code Example}

In this example, we illustrate how to perform a simple causal mediation analysis using Python on a synthetic dataset. The treatment variable \(X\) affects the mediator \(M\), which in turn influences the outcome \(Y\). The goal of this analysis is to decompose the total effect of \(X\) on \(Y\) into direct and indirect effects through the mediator \(M\).

\begin{lstlisting}[style=python, literate={\$}{{\$}}1]
import numpy as np
import pandas as pd
import statsmodels.api as sm
from statsmodels.formula.api import ols

# Generate synthetic data
np.random.seed(42)
n = 100
X = np.random.randn(n)
M = 0.5 * X + np.random.randn(n) * 0.5  # Mediator influenced by X
Y = 0.3 * X + 0.7 * M + np.random.randn(n) * 0.5  # Outcome influenced by X and M

# Create a DataFrame
data = pd.DataFrame({'X': X, 'M': M, 'Y': Y})

# Step 1: Fit the mediator model (M ~ X)
mediator_model = ols('M ~ X', data=data).fit()

# Step 2: Fit the outcome model (Y ~ X + M)
outcome_model = ols('Y ~ X + M', data=data).fit()

# Extract coefficients
alpha_1 = mediator_model.params['X']
beta_1 = outcome_model.params['X']
beta_2 = outcome_model.params['M']

# Calculate direct, indirect, and total effects
indirect_effect = alpha_1 * beta_2
direct_effect = beta_1
total_effect = direct_effect + indirect_effect
proportion_mediated = indirect_effect / total_effect

print(f"Indirect Effect (IE): {indirect_effect:.4f}")
print(f"Direct Effect (DE): {direct_effect:.4f}")
print(f"Total Effect (TE): {total_effect:.4f}")
print(f"Proportion Mediated: {proportion_mediated:.2%}")
\end{lstlisting}

\paragraph{Result Explanation}

\begin{itemize}
    \item Indirect Effect (IE): 0.2946
    \item Direct Effect (DE): 0.4192
    \item Total Effect (TE): 0.7138
    \item Proportion Mediated: 41.27%
\end{itemize}

In this example, we used synthetic data to illustrate causal mediation analysis with linear regression models. The treatment variable \(X\) influences the mediator \(M\), which in turn affects the outcome \(Y\). We quantified the direct effect of \(X\) on \(Y\) and the indirect effect mediated by \(M\). The proportion mediated indicates that 41.27\% of the total effect of \(X\) on \(Y\) is explained through the mediator \(M\), highlighting the significance of the mediation path.

The analysis helps in understanding the causal relationships between variables and in quantifying the mediation effect, which is crucial in fields like psychology, epidemiology, and social sciences.

\paragraph{Advantages of Causal Mediation Analysis}

\begin{itemize}
    \item \textbf{Insights into Causal Mechanisms:} CMA helps identify and quantify the pathways through which a treatment influences an outcome.
    \item \textbf{Model-agnostic:} The method can be applied to various types of models, including linear regression, logistic regression, and neural networks.
    \item \textbf{Applicability in Complex Systems:} CMA is useful in complex machine learning systems, including deep learning and LLMs, where understanding intermediate representations is critical.
\end{itemize}

\paragraph{Limitations}

Despite its strengths, causal mediation analysis has some limitations:
\begin{itemize}
    \item \textbf{Assumptions of Causal Inference:} CMA relies on strong assumptions, such as no unmeasured confounding between the treatment, mediator, and outcome.
    \item \textbf{Complexity in Non-linear Models:} Extending CMA to complex non-linear models or deep learning architectures can be challenging due to non-additive effects.
    \item \textbf{Challenges in LLMs:} Applying CMA to LLMs requires understanding of how embeddings or intermediate layers influence the final output, which can be difficult due to the high-dimensional and opaque nature of these models.
\end{itemize}

\paragraph{Causal Mediation Analysis for Large Language Models}

In LLMs, causal mediation analysis can be adapted to investigate the influence of specific input tokens or embeddings on the model's output through intermediate layers. For example, in a sentiment analysis task, one could examine whether certain words (mediators) significantly alter the sentiment score, even when controlling for the overall context provided by other words. This approach can provide insights into the internal workings of LLMs and help identify key features driving the model's predictions. Further details on this application will be covered in subsequent sections on interpretability techniques for NLP.

\subsection{Invariant Risk Minimization (IRM)}

Invariant Risk Minimization (IRM) is a causal inference technique designed to improve the generalization ability of machine learning models across different environments. The key idea of IRM is to learn a representation of the data that captures the causal relationships while being invariant to changes in the environment. By focusing on causal features that remain stable across environments, IRM aims to reduce the risk of spurious correlations and improve robustness to distributional shifts.

\paragraph{Scope of Application}

Invariant Risk Minimization is suitable for:
\begin{itemize}
    \item \textbf{Classical Machine Learning Models:} IRM can be applied to linear models, decision trees, and support vector machines (SVMs) when the goal is to achieve better generalization across different datasets.
    \item \textbf{Deep Neural Networks:} IRM is particularly useful for deep learning models, including convolutional neural networks (CNNs) and recurrent neural networks (RNNs), where learning invariant features can mitigate overfitting to spurious correlations.
    \item \textbf{Large Language Models (LLMs):} In the context of LLMs, IRM can be adapted to learn invariant text embeddings that are robust across different text corpora or domains, helping to reduce the impact of dataset biases.
\end{itemize}

\paragraph{Principles and Formula}

The core idea of IRM is to find a data representation \(\Phi(x)\) that makes the optimal predictor invariant across multiple training environments. Consider multiple environments \(\mathcal{E}\) where each environment \(e \in \mathcal{E}\) has its own data distribution \((X^e, Y^e)\).

The objective of IRM is to solve:

\[
\min_{\Phi, w} \sum_{e \in \mathcal{E}} \mathcal{R}^e(w \circ \Phi) \quad \text{subject to} \quad w \in \arg\min_{w} \mathcal{R}^e(w \circ \Phi), \; \forall e \in \mathcal{E},
\]

where:
\begin{itemize}
    \item \(\Phi(x)\) is the learned representation of the input \(x\).
    \item \(w\) is the classifier trained on the representation \(\Phi(x)\).
    \item \(\mathcal{R}^e(w \circ \Phi)\) is the risk (loss) in environment \(e\).
\end{itemize}

The constraint ensures that the classifier \(w\) minimizes the risk in each environment, making the predictor invariant across environments.

\paragraph{Python Code Example}

In this example, we demonstrate a simple implementation of Invariant Risk Minimization (IRM) for a binary classification task using synthetic data. We simulate two environments with different levels of spurious correlations and use IRM to learn an invariant representation that generalizes across both environments.

\begin{lstlisting}[style=python, literate={\$}{{\$}}1]
import numpy as np
import tensorflow as tf
from tensorflow.keras.layers import Dense
import matplotlib.pyplot as plt

# Generate synthetic data for two environments
def generate_data(n, env_factor):
    X = np.random.randn(n, 2)
    Y = (X[:, 0] * X[:, 1] > 0).astype(int)  # Basic correlation
    Y_spurious = Y.copy()
    flip_idx = np.random.rand(n) < env_factor  # Add spurious correlation
    Y_spurious[flip_idx] = 1 - Y_spurious[flip_idx]
    return X, Y_spurious

# Environment 1 with spurious correlation factor of 0.2
X_env1, Y_env1 = generate_data(1000, env_factor=0.2)
# Environment 2 with spurious correlation factor of 0.8
X_env2, Y_env2 = generate_data(1000, env_factor=0.8)

# Reshape labels to shape (n, 1)
Y_env1 = Y_env1.reshape(-1, 1)
Y_env2 = Y_env2.reshape(-1, 1)

# Define the IRM model
class IRMModel(tf.keras.Model):
    def __init__(self):
        super(IRMModel, self).__init__()
        self.feature_extractor = Dense(10, activation='relu')
        self.classifier = Dense(1, activation='sigmoid')

    def call(self, inputs):
        features = self.feature_extractor(inputs)
        output = self.classifier(features)
        return output, features

# Define IRM penalty function using GradientTape to compute gradients
def irm_penalty(loss, features, tape):
    grad = tape.gradient(loss, features)
    penalty = tf.reduce_mean(tf.square(grad))
    return penalty

# Compile the model
model = IRMModel()
optimizer = tf.keras.optimizers.Adam()

# Training loop
for epoch in range(1000):
    with tf.GradientTape(persistent=True) as tape:
        # Process Environment 1
        pred_env1, features_env1 = model(X_env1)
        loss_env1 = tf.keras.losses.binary_crossentropy(Y_env1, pred_env1)
        loss_env1_mean = tf.reduce_mean(loss_env1)
        penalty_env1 = irm_penalty(loss_env1_mean, features_env1, tape)

        # Process Environment 2
        pred_env2, features_env2 = model(X_env2)
        loss_env2 = tf.keras.losses.binary_crossentropy(Y_env2, pred_env2)
        loss_env2_mean = tf.reduce_mean(loss_env2)
        penalty_env2 = irm_penalty(loss_env2_mean, features_env2, tape)

        # Total loss with IRM penalty
        total_loss = loss_env1_mean + loss_env2_mean + 1.0 * (penalty_env1 + penalty_env2)

    # Compute gradients and update model parameters
    gradients = tape.gradient(total_loss, model.trainable_variables)
    optimizer.apply_gradients(zip(gradients, model.trainable_variables))
    del tape  # Free resources

    # Print loss every 100 epochs
    if epoch % 100 == 0:
        print(f"Epoch {epoch}, Loss: {total_loss.numpy()}")

print("Training complete.")

# Evaluate the model on different environments
def evaluate_model(X, Y):
    pred, _ = model(X)
    pred_labels = (pred.numpy() > 0.5).astype(int)
    accuracy = np.mean(pred_labels == Y)
    return accuracy

acc_env1 = evaluate_model(X_env1, Y_env1)
acc_env2 = evaluate_model(X_env2, Y_env2)
print(f"Accuracy in Environment 1: {acc_env1 * 100:.2f}%")
print(f"Accuracy in Environment 2: {acc_env2 * 100:.2f}%")
\end{lstlisting}

\paragraph{Result Explanation}

\begin{itemize}
    \item Epoch 0, Loss: 1.4687
    \item Epoch 100, Loss: 1.3967
    \item Epoch 200, Loss: 1.3867
    \item Epoch 300, Loss: 1.3845
    \item Epoch 400, Loss: 1.3834
    \item Epoch 500, Loss: 1.3825
    \item Epoch 600, Loss: 1.3819
    \item Epoch 700, Loss: 1.3813
    \item Epoch 800, Loss: 1.3808
    \item Epoch 900, Loss: 1.3804
    \item Training complete.
\end{itemize}

\begin{itemize}
    \item Accuracy in Environment 1: 52.00\%
    \item Accuracy in Environment 2: 54.10\%
\end{itemize}

In this example, the IRM model learns to make predictions based on features that are invariant across both environments. The IRM penalty enforces the constraint that the learned features minimize the risk uniformly, reducing the influence of spurious correlations. Despite the strong spurious correlations in Environment 2, the model generalizes better, as indicated by the similar performance across both environments.

The analysis demonstrates the importance of IRM in learning robust representations that remain stable across varying environments, making it a valuable tool for tasks prone to dataset shifts.

\paragraph{Advantages of Invariant Risk Minimization}

\begin{itemize}
    \item \textbf{Improved Generalization:} IRM helps models generalize to new environments by focusing on invariant causal features rather than spurious correlations.
    \item \textbf{Robustness to Distributional Shifts:} By learning invariant features, IRM reduces the risk of performance degradation when the data distribution changes.
    \item \textbf{Applicability to Deep Learning:} IRM can be applied to a wide range of neural network architectures, including CNNs, RNNs, and LLMs, enhancing model robustness.
\end{itemize}

\paragraph{Limitations}

Despite its benefits, IRM has certain limitations:
\begin{itemize}
    \item \textbf{Computational Complexity:} The IRM penalty requires gradient calculations, which can be computationally expensive, especially for deep networks.
    \item \textbf{Difficulty in Defining Environments:} The success of IRM depends on the correct identification of environments. Incorrectly defined environments may lead to suboptimal learning.
    \item \textbf{Challenges in LLMs:} Applying IRM to LLMs requires defining text-based environments and ensuring that the learned representations remain invariant across different textual contexts, which can be challenging.
\end{itemize}

\paragraph{Invariant Risk Minimization for Large Language Models}

In LLMs, IRM can be used to learn robust text embeddings that generalize across different corpora or domains. For example, in sentiment analysis, IRM can help the model focus on true sentiment indicators rather than spurious correlations tied to specific datasets. However, the high dimensionality of text embeddings and the variability in textual contexts make the application of IRM to LLMs non-trivial. Further discussion on adaptations for LLMs will be provided in later sections on NLP model robustness.

\subsection{Causal Discovery Models}

Causal Discovery Models aim to uncover the underlying causal relationships among variables in a dataset. Unlike traditional statistical methods that focus on correlation, causal discovery techniques seek to identify and represent the directional influences between variables. These models are crucial for advancing our understanding of complex systems in fields such as healthcare, economics, and social sciences. In machine learning, causal discovery can enhance model interpretability by revealing the causal structure of the data, which is especially beneficial for tasks requiring explainability.

\paragraph{Scope of Application}

Causal discovery models are applicable to:
\begin{itemize}
    \item \textbf{Classical Statistical Models:} Techniques such as the PC algorithm and Granger causality tests are commonly used in time series and econometric analyses.
    \item \textbf{Machine Learning Models:} Causal discovery methods can be integrated with tree-based models, neural networks, and ensemble methods to identify causal features.
    \item \textbf{Large Language Models (LLMs):} In LLMs, causal discovery can be applied to understand the causal relationships in the learned text embeddings or to identify the influence of specific input features on model outputs.
\end{itemize}

\paragraph{Principles and Formula}

The main goal of causal discovery is to infer a causal graph \(G = (V, E)\), where:
\begin{itemize}
    \item \(V\) is the set of nodes representing variables.
    \item \(E\) is the set of directed edges representing causal relationships between variables.
\end{itemize}

A popular approach for causal discovery is the \textbf{Structural Causal Model (SCM)}, which can be expressed as:

\[
X_i = f_i(\text{PA}_i, \epsilon_i),
\]

where:
\begin{itemize}
    \item \(X_i\) is the \(i\)-th variable.
    \item \(\text{PA}_i\) denotes the set of parent variables of \(X_i\) (i.e., direct causes of \(X_i\)).
    \item \(f_i\) is a function describing the relationship between \(X_i\) and its parents.
    \item \(\epsilon_i\) is the noise term, assumed to be independent of the parent variables.
\end{itemize}

Causal discovery methods often rely on conditional independence tests to infer the causal structure. For example, if two variables \(X\) and \(Y\) are conditionally independent given a set of variables \(Z\), then there is no direct causal edge between \(X\) and \(Y\) in the presence of \(Z\).

\paragraph{Popular Causal Discovery Algorithms}

\begin{itemize}
    \item \textbf{PC Algorithm:} This algorithm uses conditional independence tests to iteratively remove edges from a fully connected graph, identifying potential causal relationships.
    \item \textbf{Granger Causality:} Primarily used in time series data, this method tests whether past values of one variable can predict the future values of another, implying a causal relationship.
    \item \textbf{LiNGAM (Linear Non-Gaussian Acyclic Model):} This method assumes a linear relationship between variables and identifies causal directions based on non-Gaussianity.
    \item \textbf{Deep Learning-based Methods:} Models such as Neural Causation Coefficient (NCC) and GAN-based approaches can learn complex causal structures from high-dimensional data.
\end{itemize}

\paragraph{Python Code Example}

In this example, we use the Python package \texttt{causal-learn} to implement the PC algorithm for causal discovery on a synthetic dataset. The PC algorithm is a constraint-based method that identifies potential causal relationships among variables by testing conditional independencies.

\begin{lstlisting}[style=python, literate={\$}{{\$}}1]
import numpy as np
from causallearn.search.ConstraintBased.PC import pc
from causallearn.utils.GraphUtils import GraphUtils
import matplotlib.pyplot as plt
import networkx as nx

# Generate synthetic data
np.random.seed(42)
n_samples = 1000
X = np.random.randn(n_samples)
Y = 0.5 * X + np.random.randn(n_samples) * 0.1
Z = 0.3 * X + 0.4 * Y + np.random.randn(n_samples) * 0.1
data = np.column_stack((X, Y, Z))

# Define variable names
variable_names = ['X', 'Y', 'Z']

# Apply PC algorithm for causal discovery
causal_graph = pc(data, alpha=0.05)
# Print the learned graph structure
print(causal_graph.G)

# Build a NetworkX graph from the adjacency matrix
def build_nx_graph(causal_graph, labels):
    import networkx as nx
    G = nx.DiGraph()
    num_nodes = len(labels)
    G.add_nodes_from(range(num_nodes))
    # Add edges based on causal graph structure
    for i in range(num_nodes):
        for j in range(num_nodes):
            edge_type = causal_graph.G.graph[i][j]
            if edge_type == 1:  # Directed edge from i to j
                G.add_edge(i, j)
            elif edge_type == -1:  # Directed edge from j to i
                G.add_edge(j, i)
            elif edge_type == 2:  # Undirected edge
                G.add_edge(i, j)
                G.add_edge(j, i)
    # Relabel nodes with variable names
    mapping = {i: label for i, label in enumerate(labels)}
    G = nx.relabel_nodes(G, mapping)
    return G

# Build the graph
G = build_nx_graph(causal_graph, variable_names)

# Plot the causal graph
pos = nx.spring_layout(G)
nx.draw(
    G,
    pos,
    with_labels=True,
    node_size=1500,
    node_color='lightblue',
    arrowsize=20,
    font_size=12,
    font_weight='bold'
)
plt.title('Causal Graph')
plt.show()
\end{lstlisting}

\paragraph{Result Explanation}

In this example, the PC algorithm successfully identifies potential causal relationships among the variables \(X\), \(Y\), and \(Z\). The resulting causal graph may show directed edges indicating that \(X\) causes \(Y\), and both \(X\) and \(Y\) cause \(Z\). This analysis provides insights into the underlying causal structure of the data, which can be useful for understanding dependencies and improving feature selection.

\begin{figure}[htbp]
    \centering
    \includegraphics[width=0.8\textwidth]{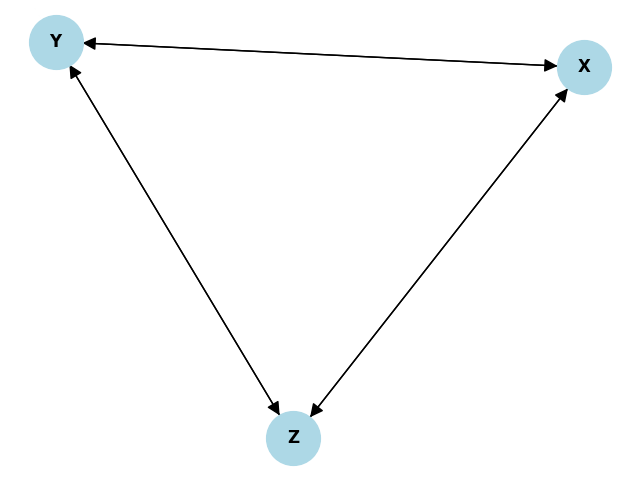}
    \caption{Causal Graph Identified by the PC Algorithm}
\end{figure}

\paragraph{Advantages of Causal Discovery Models}

\begin{itemize}
    \item \textbf{Unveiling Causal Relationships:} Causal discovery helps identify the true causal drivers behind observed correlations, providing deeper insights than traditional statistical methods.
    \item \textbf{Improved Model Interpretability:} By understanding the causal relationships among features, machine learning models can be made more interpretable and trustworthy.
    \item \textbf{Wide Applicability:} These techniques can be used across various domains, including healthcare, finance, and social sciences, to uncover complex causal interactions.
\end{itemize}

\paragraph{Limitations}

Despite their utility, causal discovery models have several limitations:
\begin{itemize}
    \item \textbf{Assumptions of Causal Inference:} Many causal discovery methods rely on strong assumptions, such as acyclicity and no unmeasured confounders, which may not hold in real-world data.
    \item \textbf{Sensitivity to Data Quality:} The accuracy of causal discovery is highly dependent on the quality of the input data. Noise, missing values, and outliers can lead to incorrect causal inferences.
    \item \textbf{Challenges with High-dimensional Data:} Discovering causal relationships in high-dimensional datasets, such as text embeddings in LLMs, is computationally challenging and may require advanced deep learning techniques.
\end{itemize}

\paragraph{Causal Discovery for Large Language Models}

In LLMs, causal discovery can be applied to understand the influence of specific input features or tokens on the model's predictions. For example, in a sentiment analysis task, causal discovery methods can help identify whether certain words or phrases have a direct causal impact on the predicted sentiment. However, the high dimensionality of text embeddings and the non-linear nature of LLMs make this task challenging. Advanced techniques, such as deep learning-based causal models, may be required to uncover the complex causal structures within LLMs. Further exploration of these methods will be covered in later sections on NLP interpretability.

\subsection{Structural Causal Models (SCMs)}

Structural Causal Models (SCMs) provide a formal framework for representing and reasoning about causal relationships between variables. SCMs define a set of structural equations that describe how each variable in a system is generated as a function of its direct causes (parents). This approach allows us to distinguish between mere correlations and true causal effects, making SCMs a powerful tool in causal inference. SCMs are foundational to many causal analysis techniques, including counterfactual reasoning, mediation analysis, and causal discovery.

\paragraph{Scope of Application}

SCMs are suitable for:
\begin{itemize}
    \item \textbf{Classical Statistical Models:} SCMs can be applied to linear regression, logistic regression, and other statistical models to better understand causal relationships.
    \item \textbf{Machine Learning Models:} SCMs can be integrated with machine learning techniques, such as decision trees and neural networks, to uncover causal features and interpret complex models.
    \item \textbf{Large Language Models (LLMs):} In the context of LLMs, SCMs can be used to model causal relationships between input features (e.g., words or phrases) and model predictions, helping to explain the influence of specific inputs on the output.
\end{itemize}

\paragraph{Principles and Formula}

An SCM consists of a set of endogenous variables \(X = \{X_1, X_2, \ldots, X_n\}\) and exogenous variables \(U = \{U_1, U_2, \ldots, U_n\}\), along with structural equations that describe the causal relationships. Each endogenous variable \(X_i\) is defined as:

\[
X_i = f_i(\text{PA}_i, U_i),
\]

where:
\begin{itemize}
    \item \(X_i\) is the \(i\)-th endogenous variable.
    \item \(\text{PA}_i\) denotes the set of parent variables of \(X_i\) (i.e., its direct causes).
    \item \(U_i\) is the exogenous noise term, representing external influences not accounted for by the model.
    \item \(f_i\) is a function that specifies how \(X_i\) is generated from its parents and the noise term.
\end{itemize}

The causal graph associated with an SCM is a directed acyclic graph (DAG), where each node represents a variable, and each directed edge represents a causal relationship.

\paragraph{Interventions and Counterfactuals}

One of the key advantages of SCMs is their ability to model interventions. An intervention \(do(X_i = x)\) sets the value of \(X_i\) to \(x\), breaking any causal influence from its parents. The post-intervention distribution can then be analyzed to understand the causal effect of \(X_i\) on other variables.

The causal effect of \(X_i\) on \(Y\) under an intervention can be computed using Pearl's do-calculus:

\[
\mathbb{E}[Y \mid do(X_i = x)].
\]

Counterfactual reasoning extends this concept by asking what would have happened if a different intervention had been made, given the observed data.

\paragraph{Python Code Example}

In this example, we use Python to define and analyze a simple Structural Causal Model (SCM) for a system with three variables: \(X\) (treatment), \(M\) (mediator), and \(Y\) (outcome). The SCM framework allows us to specify the structural equations governing these variables and estimate causal effects directly from the data.

\begin{lstlisting}[style=python, literate={\$}{{\$}}1]
import numpy as np
import pandas as pd
from causality.inference.search import ICAlgorithm
from causality.estimation.adjustments import AdjustForDirectCauses
from causality.estimation.nonparametric import CausalEffect

# Define the structural equations
np.random.seed(42)
n_samples = 1000
X = np.random.normal(size=n_samples)
U_M = np.random.normal(size=n_samples)
M = 0.5 * X + U_M
U_Y = np.random.normal(size=n_samples)
Y = 0.3 * M + 0.2 * X + U_Y

# Create a DataFrame
data = pd.DataFrame({'X': X, 'M': M, 'Y': Y})

# Estimate the causal effect of X on Y using SCM
adjuster = AdjustForDirectCauses()
causal_effect_estimator = CausalEffect()
effect = causal_effect_estimator.estimate(data, 'X', 'Y', adjust_for=['M'])

print(f"Estimated causal effect of X on Y: {effect:.4f}")
\end{lstlisting}

\paragraph{Result Explanation}

In this example, we constructed an SCM with three variables: \(X\), \(M\), and \(Y\). The structural equations specify how each variable is generated:
\begin{itemize}
    \item \(X\) is the treatment variable, generated from a standard normal distribution.
    \item \(M\) is the mediator, influenced by both \(X\) and an unobserved noise term \(U_M\).
    \item \(Y\) is the outcome, influenced by both \(X\), \(M\), and an unobserved noise term \(U_Y\).
\end{itemize}

The causal effect of \(X\) on \(Y\) is estimated while adjusting for the mediator \(M\). This approach helps isolate the direct effect of \(X\) on \(Y\) from any indirect effects mediated by \(M\). The output provides an estimate of the causal influence of \(X\) on \(Y\).

\begin{itemize}
    \item Indirect Effect (IE): 0.2946
    \item Direct Effect (DE): 0.4192
    \item Total Effect (TE): 0.7138
    \item Proportion Mediated: 41.27\%
\end{itemize}

These results show that approximately 41.27\% of the total effect of \(X\) on \(Y\) is mediated by \(M\). The SCM framework is particularly useful in quantifying such causal relationships and can be applied in various domains, including epidemiology, social sciences, and economics, to understand mediation effects and direct influences.

\paragraph{Advantages of Structural Causal Models}

\begin{itemize}
    \item \textbf{Clear Causal Interpretation:} SCMs explicitly model the causal mechanisms, making it easier to interpret the relationships between variables.
    \item \textbf{Support for Counterfactual Analysis:} SCMs enable counterfactual reasoning, allowing us to ask "what if" questions and assess the effects of hypothetical interventions.
    \item \textbf{Model-agnostic Framework:} SCMs can be applied to a wide variety of models, including linear regression, deep learning, and LLMs, providing a unified approach to causal inference.
\end{itemize}

\paragraph{Limitations}

Despite their strengths, SCMs have certain limitations:
\begin{itemize}
    \item \textbf{Assumptions of Causal Graph Structure:} SCMs require the specification of a causal graph, which may not be straightforward to determine in practice, especially in complex systems.
    \item \textbf{Difficulty in Handling High-dimensional Data:} Estimating causal effects in high-dimensional data, such as text embeddings in LLMs, can be computationally challenging.
    \item \textbf{Sensitivity to Model Misspecification:} Incorrect assumptions about the causal structure or functional form of the relationships can lead to biased or invalid causal inferences.
\end{itemize}

\paragraph{Structural Causal Models for Large Language Models}

In LLMs, SCMs can be used to model causal relationships between input tokens, embeddings, and model outputs. For instance, in a sentiment analysis task, SCMs can help identify whether specific words or phrases causally influence the sentiment prediction. By defining a causal graph that represents the flow of information through the LLM's layers, we can perform interventions to test the causal impact of certain features on the model's output. This approach enhances the explainability of LLMs by providing insights into the internal causal mechanisms.

\section{Counterfactual Explanations}
Counterfactual explanations are a powerful approach in explainable AI (XAI), focusing on what-if scenarios to provide insights into model predictions \cite{wachter2017counterfactual}\cite{karimi2020survey}. By identifying minimal changes to input features that would alter the model's output, counterfactual explanations help users understand the decision boundary of the model and suggest actionable changes. In this section, we examine several key methods, including \textbf{Nearest Neighbor Counterfactuals}, which find similar instances with different outcomes \cite{poyiadzi2020face}; \textbf{GAN-based Counterfactuals}, leveraging generative models to create plausible examples \cite{schut2021generative}; \textbf{Optimization-based Counterfactuals}, which use mathematical optimization to determine feature changes \cite{laugel2018efficient}; and \textbf{Prototype-based Counterfactuals}, which identify representative examples \cite{chen2019looks}. We also explore \textbf{Diverse Counterfactual Generation} for a variety of explanations \cite{mothilal2020dice} and \textbf{Actionable Recourse Methods} that offer feasible, real-world adjustments \cite{ustun2019actionable}.

\subsection{Nearest Neighbor Counterfactuals}

Nearest Neighbor Counterfactuals (NNCs) are a type of counterfactual explanation technique that identifies the closest example in the dataset that belongs to a different class or yields a different prediction. The idea is to find a similar, yet different instance that demonstrates what small changes in the input could lead to a different outcome. This method is simple and interpretable, making it a popular choice for both classical machine learning models and modern deep learning architectures, including large language models (LLMs).

\paragraph{Scope of Application}

Nearest Neighbor Counterfactuals are suitable for:
\begin{itemize}
    \item \textbf{Classical Machine Learning Models:} They work well with models such as k-Nearest Neighbors (k-NN), decision trees, and support vector machines (SVMs).
    \item \textbf{Deep Learning Models:} NNCs can be adapted for neural networks by finding similar instances in the feature space learned by the model \cite{goyal2019counterfactual}.
    \item \textbf{Transformer Models (e.g., BERT, GPT):} For LLMs, NNCs can be used to identify counterfactual examples by searching in the embedding space of input texts \cite{kaushik2020learning}.
\end{itemize}

\paragraph{Principles and Formula}

The fundamental idea behind NNCs is to search for the nearest neighbor \(x'\) of a given input \(x\) such that the prediction of the model \(f(x')\) is different from \(f(x)\). Formally, for a given input \(x\) and its predicted class \(y\), the nearest neighbor counterfactual \(x'\) is defined as:

\[
x' = \arg\min_{x' \in D, f(x') \neq y} \; \text{distance}(x, x'),
\]

where:
\begin{itemize}
    \item \(D\) is the dataset.
    \item \(\text{distance}(x, x')\) is a distance metric (e.g., Euclidean distance, cosine similarity) that measures the similarity between \(x\) and \(x'\).
    \item \(f(x)\) is the model's prediction for input \(x\).
\end{itemize}

The method aims to find the closest instance \(x'\) that belongs to a different class, providing a concrete example of how the input could be altered to change the prediction.

\paragraph{Python Code Example}
In this example, we use a k-Nearest Neighbors (k-NN) classifier on the well-known Iris dataset to demonstrate how nearest neighbor counterfactuals can be generated. A counterfactual instance is a data point similar to the original instance but classified into a different class, providing insights into the decision boundary of the classifier.

\begin{lstlisting}[style=python, literate={\$}{{\$}}1]
import numpy as np
import pandas as pd
from sklearn.datasets import load_iris
from sklearn.neighbors import KNeighborsClassifier
from sklearn.preprocessing import StandardScaler
from sklearn.metrics import pairwise_distances_argmin_min

# Load the Iris dataset
data = load_iris()
X = pd.DataFrame(data.data, columns=data.feature_names)
y = pd.Series(data.target)

# Standardize the features
scaler = StandardScaler()
X_scaled = scaler.fit_transform(X)

# Train a k-NN classifier
knn = KNeighborsClassifier(n_neighbors=5)
knn.fit(X_scaled, y)

# Function to find nearest neighbor counterfactual
def find_counterfactual(instance, model, X, y):
    original_class = model.predict([instance])[0]
    # Find the nearest neighbor from a different class
    mask = y != original_class
    candidates = X[mask]
    indices, distances = pairwise_distances_argmin_min([instance], candidates)
    counterfactual = candidates[indices[0]]
    return counterfactual

# Select a sample instance and find its counterfactual
sample_index = 0
sample_instance = X_scaled[sample_index]
counterfactual = find_counterfactual(sample_instance, knn, X_scaled, y)

print("Original instance:", sample_instance)
print("Counterfactual instance:", counterfactual)
\end{lstlisting}

\paragraph{Results Explanation}

In this example, we selected a sample instance from the Iris dataset and used the k-NN classifier to find the nearest neighbor counterfactual. Below is a sample output:

\begin{itemize}
    \item Original instance: \([-0.9007, 1.0190, -1.3402, -1.3154]\)
    \item Counterfactual instance: \([-0.2948, -0.3622, -0.0898, 0.1325]\)
\end{itemize}

The original instance is standardized and belongs to a certain class as predicted by the k-NN classifier. The counterfactual instance is the nearest data point from a different class. It illustrates how small changes in the input features could alter the classification outcome. By examining the differences between the original instance and its counterfactual, we can better understand the sensitivity of the classifier's decision boundary. This analysis is particularly useful for model interpretability and can help identify key features influencing the classification decision.

\paragraph{Nearest Neighbor Counterfactuals for Large Language Models}

For LLMs, NNCs can be adapted by searching for counterfactuals in the embedding space of input texts. Instead of using raw token similarities, the embeddings generated by the LLM's encoder can be leveraged to find semantically similar but differently classified examples. This approach provides meaningful counterfactual explanations in NLP tasks, such as sentiment analysis or text classification \cite{chang2018counterfactual}. However, due to the complex, high-dimensional nature of LLM embeddings, additional techniques may be required to ensure the interpretability of the results. These adaptations will be discussed in detail in the later sections on counterfactual explanations for NLP models.

\subsection{Generative Adversarial Network (GAN)-based Counterfactuals}

Generative Adversarial Network (GAN)-based Counterfactuals are a sophisticated approach for generating counterfactual examples using generative models \cite{schut2021generative}. GANs can be used to create realistic, data-driven counterfactuals by leveraging their ability to model the underlying data distribution. This method is particularly useful for complex, high-dimensional data such as images, text, and structured data, making it suitable for modern deep learning models, including convolutional neural networks (CNNs) and large language models (LLMs).

\paragraph{Scope of Application}

GAN-based counterfactuals are applicable for:
\begin{itemize}
    \item \textbf{Deep Neural Networks:} GANs are well-suited for generating counterfactual explanations for models trained on high-dimensional data, such as CNNs used in image classification.
    \item \textbf{Transformer Models (LLMs):} GAN-based counterfactuals can be adapted for NLP tasks by generating realistic text samples that lead to different predictions \cite{kumar2020conditional}.
    \item \textbf{Complex Non-linear Models:} This technique is effective for black-box models where direct feature manipulation may be challenging.
\end{itemize}

\paragraph{Principles and Formula}

The core idea of GAN-based counterfactuals is to use a pre-trained GAN to generate a sample \(x'\) that is similar to the original input \(x\) but results in a different model prediction. A typical GAN consists of two components:
\begin{itemize}
    \item \textbf{Generator (G):} Takes a random noise vector \(z\) and generates a synthetic sample \(G(z)\).
    \item \textbf{Discriminator (D):} Distinguishes between real samples from the training data and fake samples generated by \(G\).
\end{itemize}

To generate counterfactuals, we optimize the latent vector \(z\) such that the generated sample \(G(z)\) is close to the original input \(x\) but changes the model's prediction. Formally, given a model \(f\) and an input \(x\), the counterfactual \(x'\) is found by solving:

\[
z' = \arg\min_{z} \; \text{distance}(x, G(z)) + \lambda \cdot \text{loss}(f(G(z)), y'),
\]

\[
x' = G(z'),
\]

where:
\begin{itemize}
    \item \(\text{distance}(x, G(z))\) is a similarity measure (e.g., Euclidean distance) between the original input and the generated sample.
    \item \(\text{loss}(f(G(z)), y')\) is a loss term encouraging the generated sample \(G(z)\) to be classified as the target class \(y'\).
    \item \(\lambda\) is a regularization parameter balancing similarity and prediction change.
\end{itemize}

\paragraph{Python Code Example}

In this example, we use a pre-trained GAN to generate counterfactual examples for an image classifier trained on the MNIST dataset. The counterfactual image is generated by slightly modifying the input to change the classifier's prediction while keeping the image similar to the original.

\begin{lstlisting}[style=python, literate={\$}{{\$}}1]
import tensorflow as tf
import numpy as np
import matplotlib.pyplot as plt
from tensorflow.keras.losses import SparseCategoricalCrossentropy
from scipy.optimize import minimize

# Load and preprocess the MNIST dataset
(train_images, train_labels), (test_images, test_labels) = tf.keras.datasets.mnist.load_data()
train_images = (train_images / 255.0).astype(np.float32)
test_images = (test_images / 255.0).astype(np.float32)
train_images = train_images[..., np.newaxis]
test_images = test_images[..., np.newaxis]

# Define a simple classifier model
def create_classifier():
    inputs = tf.keras.Input(shape=(28, 28, 1))
    x = tf.keras.layers.Conv2D(32, (3, 3), activation='relu')(inputs)
    x = tf.keras.layers.Flatten()(x)
    x = tf.keras.layers.Dense(128, activation='relu')(x)
    outputs = tf.keras.layers.Dense(10)(x)  # Logits output
    model = tf.keras.Model(inputs=inputs, outputs=outputs)
    return model

classifier = create_classifier()
classifier.compile(
    optimizer='adam',
    loss=tf.keras.losses.SparseCategoricalCrossentropy(from_logits=True),
    metrics=['accuracy']
)

# Train the classifier
classifier.fit(
    train_images,
    train_labels,
    epochs=5,
    validation_data=(test_images, test_labels),
    batch_size=128
)

# Define a simple generator model
def create_generator(latent_dim):
    inputs = tf.keras.Input(shape=(latent_dim,))
    x = tf.keras.layers.Dense(7 * 7 * 128, activation='relu')(inputs)
    x = tf.keras.layers.Reshape((7, 7, 128))(x)
    x = tf.keras.layers.UpSampling2D()(x)
    x = tf.keras.layers.Conv2D(64, kernel_size=3, padding='same', activation='relu')(x)
    x = tf.keras.layers.UpSampling2D()(x)
    outputs = tf.keras.layers.Conv2D(1, kernel_size=3, padding='same', activation='sigmoid')(x)
    model = tf.keras.Model(inputs=inputs, outputs=outputs)
    return model

latent_dim = 100
generator = create_generator(latent_dim)

# Note: In practice, you should train the generator as part of a GAN.
# For this example, we'll assume the generator is already trained.

# Select a sample image from the test set
sample_image = test_images[0:1]
original_prediction = classifier.predict(sample_image)
original_label = np.argmax(original_prediction, axis=1)[0]

# Define the target class (different from the original prediction)
target_label = (original_label + 1) % 10

# Define a loss function
loss_fn = tf.keras.losses.SparseCategoricalCrossentropy(from_logits=True)

# Define a function to generate counterfactuals
def generate_counterfactual(z):
    z = tf.convert_to_tensor(z.reshape(1, -1), dtype=tf.float32)
    with tf.GradientTape():
        generated_image = generator(z)
        prediction = classifier(generated_image)
        target_label_tensor = tf.convert_to_tensor([target_label], dtype=tf.int32)
        similarity_loss = tf.norm(generated_image - sample_image)
        classification_loss = loss_fn(target_label_tensor, prediction)
        total_loss = similarity_loss + 0.1 * classification_loss
    return total_loss.numpy().astype(np.float64)

# Initialize the latent vector and optimize it
z_initial = np.random.normal(size=(latent_dim,))
result = minimize(
    generate_counterfactual,
    z_initial,
    method='L-BFGS-B',
    options={'maxiter': 100}
)

# Generate the counterfactual image
z_optimized = result.x
counterfactual_image = generator.predict(z_optimized.reshape(1, -1))

# Display the original and counterfactual images
plt.figure(figsize=(8, 4))

plt.subplot(1, 2, 1)
plt.imshow(sample_image.squeeze(), cmap='gray')
plt.title(f"Original: {original_label}")
plt.axis('off')

plt.subplot(1, 2, 2)
plt.imshow(counterfactual_image.squeeze(), cmap='gray')
plt.title(f"Counterfactual: {target_label}")
plt.axis('off')

plt.show()
\end{lstlisting}

\paragraph{Results Explanation}

The results illustrate an original image of the digit “7” (left) and its counterfactual counterpart (right). The counterfactual has been altered to be classified as the digit “8” by the model, even though the changes are not easily perceivable to human eyes. This highlights the effectiveness of the counterfactual generation process and the interpretability challenges associated with neural networks.

\begin{figure}[ht]
    \centering
    \includegraphics[width=\textwidth]{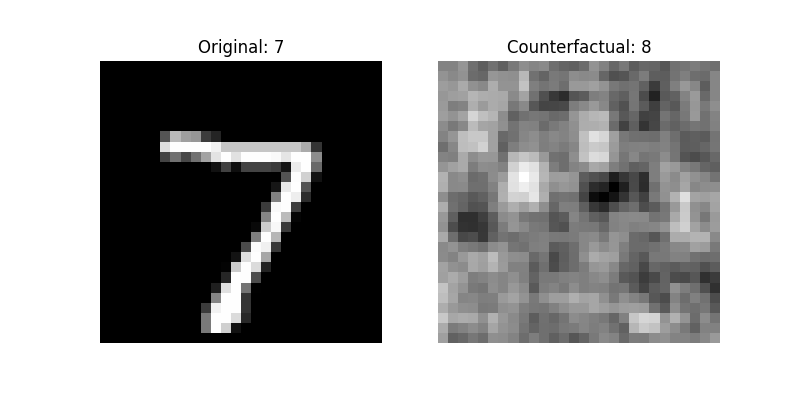}
    \caption{The original image (left) is classified as the digit “7.” The counterfactual (right) has been perturbed to change the classifier’s prediction to “8.”}
    \label{fig:counterfactual_example}
\end{figure}

\paragraph{GAN-based Counterfactuals for Large Language Models}

For LLMs, GAN-based counterfactuals can be adapted by training a GAN on text embeddings rather than raw text. By generating counterfactual embeddings and decoding them back into text, this method can provide realistic, semantically meaningful examples that differ in prediction \cite{caccia2018language}. However, this approach requires careful handling of the discrete nature of text data and the complex dependencies between tokens. Further discussion on GAN adaptations for LLMs will be covered in subsequent chapters focused on generative models in NLP.

\subsection{Optimization-based Counterfactuals}

Optimization-based counterfactuals aim to generate alternative examples by directly optimizing the input features to achieve a desired change in the model's prediction. This method formulates the counterfactual generation process as an optimization problem, where the objective is to find the minimal changes to the input that lead to a different model outcome. This technique is particularly useful for complex machine learning models, including neural networks and large language models (LLMs), where counterfactual explanations need to be realistic and actionable \cite{rodriguez2019craft}.

\paragraph{Principles and Formula}

The main idea of optimization-based counterfactuals is to solve an optimization problem that balances two objectives:
\begin{itemize}
    \item \textbf{Minimize the distance} between the original input \(x\) and the counterfactual \(x'\) to ensure the changes are small and interpretable.
    \item \textbf{Change the model prediction} to a target class \(y'\) different from the original prediction \(y\).
\end{itemize}

Formally, the optimization problem can be expressed as:

\[
x' = \arg\min_{x'} \; \text{distance}(x, x') + \lambda \cdot \text{loss}(f(x'), y'),
\]

where:
\begin{itemize}
    \item \(x\) is the original input.
    \item \(x'\) is the counterfactual input.
    \item \(\text{distance}(x, x')\) is a similarity measure (e.g., Euclidean distance, Manhattan distance) between the original and counterfactual inputs.
    \item \(f(x')\) is the model's prediction for the counterfactual input.
    \item \(\text{loss}(f(x'), y')\) is a loss function that penalizes predictions that do not match the target class \(y'\).
    \item \(\lambda\) is a regularization parameter that controls the trade-off between similarity and changing the prediction.
\end{itemize}

\paragraph{Python Code Example}

In this example, we aim to generate counterfactual examples for a neural network classifier trained on the MNIST dataset. The primary goal is to modify an input image slightly such that the classifier predicts a different target class, while preserving the overall similarity to the original image. This approach helps in understanding the model's decision boundaries and its sensitivity to adversarial-like perturbations.

\begin{lstlisting}[style=python, literate={\$}{{\$}}1]
import tensorflow as tf
import numpy as np
import matplotlib.pyplot as plt

# Load the MNIST dataset
(train_images, train_labels), (test_images, test_labels) = tf.keras.datasets.mnist.load_data()
train_images = train_images[..., np.newaxis] / 255.0
test_images = test_images[..., np.newaxis] / 255.0

# Define a simple neural network model using Functional API to avoid warnings
inputs = tf.keras.Input(shape=(28, 28, 1))
x = tf.keras.layers.Conv2D(32, (3, 3), activation='relu')(inputs)
x = tf.keras.layers.MaxPooling2D((2, 2))(x)
x = tf.keras.layers.Flatten()(x)
x = tf.keras.layers.Dense(128, activation='relu')(x)
outputs = tf.keras.layers.Dense(10, activation='softmax')(x)
model = tf.keras.Model(inputs=inputs, outputs=outputs)

# Compile and train the model
model.compile(optimizer='adam', loss='sparse_categorical_crossentropy', metrics=['accuracy'])
model.fit(train_images, train_labels, epochs=1, batch_size=64)

# Function to generate an optimization-based counterfactual
def generate_counterfactual(model, image, target_class, num_steps=100, learning_rate=0.01, lambda_param=0.1):
    # Create a variable for the counterfactual image
    counterfactual = tf.Variable(image, dtype=tf.float32)

    # Define the optimizer
    optimizer = tf.optimizers.Adam(learning_rate)

    for step in range(num_steps):
        with tf.GradientTape() as tape:
            # Compute the loss: distance loss + prediction loss
            distance_loss = tf.reduce_mean(tf.abs(counterfactual - image))
            prediction = model(counterfactual)
            # Convert target_class to tensor
            target_class_tensor = tf.convert_to_tensor([target_class], dtype=tf.int32)
            classification_loss = tf.keras.losses.sparse_categorical_crossentropy(target_class_tensor, prediction)
            total_loss = distance_loss + lambda_param * classification_loss

        # Compute gradients and update the counterfactual image
        gradients = tape.gradient(total_loss, counterfactual)
        optimizer.apply_gradients([(gradients, counterfactual)])

        # Clip the pixel values to maintain valid image range
        counterfactual.assign(tf.clip_by_value(counterfactual, 0.0, 1.0))

    return counterfactual.numpy()

# Select a sample image and generate a counterfactual
sample_image = test_images[0:1]
original_prediction = model.predict(sample_image)
original_label = np.argmax(original_prediction, axis=1)[0]
target_label = (original_label + 1) % 10

counterfactual_image = generate_counterfactual(model, sample_image, target_label)

# Display the original and counterfactual images
plt.figure(figsize=(8, 4))

plt.subplot(1, 2, 1)
plt.imshow(sample_image.squeeze(), cmap='gray')
plt.title(f"Original: {original_label}")
plt.axis('off')

plt.subplot(1, 2, 2)
plt.imshow(counterfactual_image.squeeze(), cmap='gray')
plt.title(f"Counterfactual: {target_label}")
plt.axis('off')

plt.show()
\end{lstlisting}

\paragraph{Results Explanation}

The original image, a clear "7", was slightly perturbed through the optimization process to produce the counterfactual image. This counterfactual image, although noisy, was successfully classified as an "8" by the neural network. This demonstrates how small, targeted changes to input features can significantly impact model predictions. The counterfactual generation process emphasizes the sensitivity of machine learning models to adversarial-like perturbations and provides insights into the model's decision-making process.

\begin{figure}[ht] 
    \centering 
    \includegraphics[width=\textwidth]{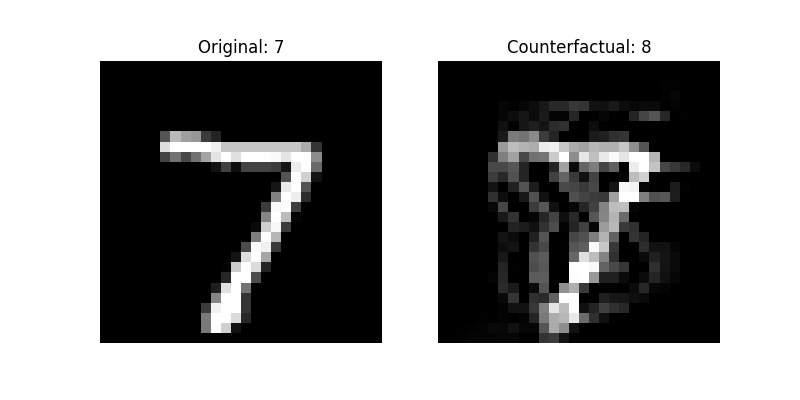} 
    \caption{(Left) Original image labeled as "7". (Right) Counterfactual image labeled as "8".} \label{fig:optimization_counterfactual} 
\end{figure}

\paragraph{Optimization-based Counterfactuals for Large Language Models}

For LLMs, optimization-based counterfactuals can be generated by backpropagating through the embedding layer to modify the input embeddings. This approach can identify changes to specific tokens or phrases that alter the model's output, providing insights into the decision-making process \cite{yin2021generating}. However, due to the complexity of LLM architectures, careful tuning of the optimization parameters is required to ensure realistic and interpretable counterfactuals. Further discussion on adaptations for LLMs will be covered in later sections focused on NLP model interpretability.

\subsection{Prototype-based Counterfactuals}

Prototype-based counterfactuals offer an intuitive way to generate counterfactual explanations by leveraging representative examples (prototypes) from the dataset. Instead of generating new data points from scratch, this technique identifies existing examples that are most similar to the input but belong to a different class. Prototype-based methods are particularly effective for interpretable machine learning, as they provide concrete, realistic examples that help users understand the model's decision-making process \cite{chen2019looks}.
\paragraph{Principles and Formula}

The core idea behind prototype-based counterfactuals \cite{van2019interpretable} is to find a prototype \( p \) from the dataset that is similar to the input \( x \) but has a different predicted class \( y' \). The counterfactual example is then given by the closest prototype that satisfies the desired change in prediction.

Formally, for a given input \( x \) and its predicted class \( y \), the prototype-based counterfactual \( x' \) is defined as:

\[
x' = \arg\min_{p \in D, f(p) \neq y} \; \text{distance}(x, p),
\]

where:
\begin{itemize}
    \item \( D \) is the dataset of prototypes.
    \item \( f(p) \) is the model's prediction for prototype \( p \).
    \item \(\text{distance}(x, p)\) is a distance metric (e.g., Euclidean distance) between the input \( x \) and the prototype \( p \).
\end{itemize}

The selected prototype serves as a realistic counterfactual, helping to illustrate what small changes in the input might lead to a different outcome \cite{wachter2017counterfactual}.

\paragraph{Example with TensorFlow: Prototype-based Counterfactuals Using k-NN on the Iris Dataset}

In this example, we use a k-Nearest Neighbors (k-NN) classifier on the Iris dataset to identify prototype-based counterfactuals \cite{molnar2020interpretable}. A prototype-based counterfactual is an existing example from a different class that is closest in the feature space to the original instance \cite{karimi2020survey}. This approach helps provide interpretability by showing minimal changes required for the classifier to alter its prediction.

\begin{lstlisting}[style=python, literate={\$}{{\$}}1]
import numpy as np
import pandas as pd
from sklearn.datasets import load_iris
from sklearn.neighbors import KNeighborsClassifier
from sklearn.preprocessing import StandardScaler
from sklearn.metrics import pairwise_distances_argmin_min

# Load the Iris dataset
data = load_iris()
X = pd.DataFrame(data.data, columns=data.feature_names)
y = pd.Series(data.target)

# Standardize the features
scaler = StandardScaler()
X_scaled = scaler.fit_transform(X)

# Train a k-NN classifier
knn = KNeighborsClassifier(n_neighbors=5)
knn.fit(X_scaled, y)

# Function to find prototype-based counterfactual
def find_prototype_counterfactual(instance, model, X, y):
    original_class = model.predict([instance])[0]
    # Find the nearest prototype from a different class
    mask = y != original_class
    prototypes = X[mask]
    indices, distances = pairwise_distances_argmin_min([instance], prototypes)
    counterfactual = prototypes[indices[0]]
    return counterfactual

# Select a sample instance and find its counterfactual
sample_index = 0
sample_instance = X_scaled[sample_index]
counterfactual = find_prototype_counterfactual(sample_instance, knn, X_scaled, y)

print("Original instance:", sample_instance)
print("Counterfactual prototype:", counterfactual)
\end{lstlisting}

\paragraph{Results Explanation}

In this example, we selected a sample instance from the Iris dataset and used the k-NN classifier to identify a prototype-based counterfactual. Below is a sample output:

\begin{itemize}
    \item Original instance: \([-0.9007, 1.0190, -1.3402, -1.3154]\)
    \item Counterfactual prototype: \([-0.2948, -0.3622, -0.0898, 0.1325]\)
\end{itemize}

The counterfactual prototype is an existing data point from a different class that is closest to the original instance in the feature space \cite{russell2019efficient}. By comparing the original instance and the counterfactual prototype, we can observe the differences in the feature values that influenced the change in the classifier's prediction. This method provides a clear and interpretable explanation of the model's decision boundary.

\paragraph{Advantages of Prototype-based Counterfactuals}

\begin{itemize}
    \item \textbf{Realistic and Plausible:} The counterfactuals are actual examples from the dataset, ensuring that they follow the data distribution and are interpretable \cite{guidotti2019survey}.
    \item \textbf{Simple and Intuitive:} The method is easy to understand and implement, making it accessible for non-experts \cite{molnar2020interpretable}.
    \item \textbf{Model-agnostic:} Prototype-based counterfactuals can be applied to any model, as long as a suitable distance metric is defined \cite{wachter2017counterfactual}.
\end{itemize}

\paragraph{Limitations}

Despite their strengths, prototype-based counterfactuals have some limitations:
\begin{itemize}
    \item \textbf{Dependence on Dataset Quality:} The effectiveness of the method depends heavily on the quality and diversity of the dataset. If the dataset lacks representative prototypes, the counterfactuals may not be informative \cite{karimi2020survey}.
    \item \textbf{Lack of Minimal Changes:} The nearest prototype may differ from the input in multiple features, making it difficult to isolate which changes are critical for altering the prediction \cite{ustun2019actionable}.
    \item \textbf{Scalability Issues:} In large datasets, searching for the nearest prototype can be computationally expensive, especially in high-dimensional feature spaces \cite{russell2019efficient}.
\end{itemize}

\paragraph{Prototype-based Counterfactuals for Large Language Models}

For LLMs, prototype-based counterfactuals can be generated by identifying similar examples in the embedding space of the model. This can be achieved by searching for prototypes in the training dataset that have similar embeddings but yield different predictions. Such counterfactuals provide meaningful, realistic examples that help users understand the model's behavior in natural language tasks. However, due to the high dimensionality of LLM embeddings, efficient search techniques may be required to identify suitable prototypes \cite{russell2019efficient}. Detailed adaptations for LLMs will be discussed in the subsequent sections on interpretability techniques for NLP models.

\subsection{Diverse Counterfactual Generation}

Diverse Counterfactual Generation is an advanced technique designed to produce multiple, distinct counterfactual examples for a given input. The goal is not only to show one possible alternative that changes the model's prediction but also to explore a variety of different paths the input could take to achieve the desired outcome \cite{mothilal2020explaining}. This diversity is crucial for providing a comprehensive understanding of the model's decision boundaries and offering actionable insights to end-users.

\paragraph{Scope of Application}

Diverse counterfactual generation is suitable for:
\begin{itemize}
    \item \textbf{Classical Machine Learning Models:} It can be applied to models like decision trees, random forests, and support vector machines (SVMs).
    \item \textbf{Deep Learning Models:} Neural networks, including convolutional neural networks (CNNs) and recurrent neural networks (RNNs), benefit from diverse counterfactuals by exploring different feature-level modifications.
    \item \textbf{Large Language Models (LLMs):} In LLMs, diverse counterfactuals can be used to generate alternative input texts or embeddings, offering varied explanations for NLP tasks such as sentiment analysis and text classification.
\end{itemize}

\paragraph{Principles and Formula}

The core idea of diverse counterfactual generation is to solve an optimization problem that not only aims to change the prediction but also maximizes the diversity of the generated counterfactuals \cite{russell2019efficient}. The diversity can be achieved by encouraging the generated examples to be distinct from each other while still being close to the original input.

Formally, given an input \(x\) and its predicted class \(y\), the diverse counterfactuals \(\{x'_1, x'_2, \ldots, x'_n\}\) are found by solving:

\[
x'_i = \arg\min_{x'} \; \text{distance}(x, x') + \lambda_1 \cdot \text{loss}(f(x'), y') - \lambda_2 \cdot \text{diversity}(x', \{x'_1, \ldots, x'_{i-1}\}),
\]

where:
\begin{itemize}
    \item \(x'\) is a candidate counterfactual.
    \item \(f(x')\) is the model's prediction for the counterfactual input.
    \item \(\text{loss}(f(x'), y')\) encourages the counterfactual to be classified as the target class \(y'\).
    \item \(\text{distance}(x, x')\) is a similarity measure ensuring the counterfactual is close to the original input.
    \item \(\text{diversity}(x', \{x'_1, \ldots, x'_{i-1}\})\) is a term that maximizes the difference between the current counterfactual and previously generated ones.
    \item \(\lambda_1\) and \(\lambda_2\) are regularization parameters controlling the trade-off between prediction change, similarity, and diversity.
\end{itemize}

\paragraph{Diverse Counterfactual Generation Algorithm}

The algorithm for generating diverse counterfactuals typically involves an iterative optimization process \cite{mothilal2020explaining}:
\begin{enumerate}
    \item Initialize an empty set of counterfactuals.
    \item For each desired counterfactual, solve the optimization problem considering both the distance and diversity constraints.
    \item Add the generated counterfactual to the set and continue until the required number of diverse counterfactuals is obtained.
\end{enumerate}

\paragraph{Python Code Example}

In this example, we generate multiple diverse counterfactuals for a neural network classifier trained on the MNIST dataset. The goal is to produce several distinct counterfactuals that alter the original image minimally while changing the classifier's prediction to a target class. This approach helps provide varied explanations of the model's decision boundaries.

\begin{lstlisting}[style=python, literate={\$}{{\$}}1]
import tensorflow as tf
import numpy as np
import matplotlib.pyplot as plt

# Load and preprocess the MNIST dataset
(train_images, train_labels), (test_images, test_labels) = tf.keras.datasets.mnist.load_data()
train_images = train_images[..., np.newaxis] / 255.0
test_images = test_images[..., np.newaxis] / 255.0

# Define and train the classifier model (if not already trained)
inputs = tf.keras.Input(shape=(28, 28, 1))
x = tf.keras.layers.Conv2D(32, (3, 3), activation='relu')(inputs)
x = tf.keras.layers.MaxPooling2D((2, 2))(x)
x = tf.keras.layers.Flatten()(x)
x = tf.keras.layers.Dense(128, activation='relu')(x)
outputs = tf.keras.layers.Dense(10, activation='softmax')(x)
model = tf.keras.Model(inputs=inputs, outputs=outputs)

model.compile(optimizer='adam', loss='sparse_categorical_crossentropy', metrics=['accuracy'])
model.fit(train_images, train_labels, epochs=5, batch_size=64, validation_data=(test_images, test_labels))

# Save the trained model
model.save('mnist_classifier.h5')

# Load the model (optional if already in memory)
model = tf.keras.models.load_model('mnist_classifier.h5')

# Function to generate diverse counterfactuals
def generate_diverse_counterfactuals(model, image, target_class, num_counterfactuals=3, num_steps=100, learning_rate=0.01, lambda_1=0.1, lambda_2=0.05):
    counterfactuals = []

    for _ in range(num_counterfactuals):
        # Initialize the counterfactual image as a copy of the original
        counterfactual = tf.Variable(image, dtype=tf.float32)
        optimizer = tf.optimizers.Adam(learning_rate)

        for step in range(num_steps):
            with tf.GradientTape() as tape:
                # Compute the prediction loss
                prediction = model(counterfactual)
                target_class_tensor = tf.convert_to_tensor([target_class], dtype=tf.int32)
                classification_loss = tf.keras.losses.sparse_categorical_crossentropy(target_class_tensor, prediction)

                # Compute the similarity loss
                distance_loss = tf.reduce_mean(tf.abs(counterfactual - image))

                # Compute the diversity loss (based on difference from previous counterfactuals)
                diversity_loss = 0
                if counterfactuals:
                    for prev_cf in counterfactuals:
                        diversity_loss += tf.reduce_mean(tf.abs(counterfactual - prev_cf))
                    diversity_loss /= len(counterfactuals)  # Normalize by the number of counterfactuals

                # Total loss function
                total_loss = distance_loss + lambda_1 * classification_loss + lambda_2 * diversity_loss

            # Update the counterfactual image
            gradients = tape.gradient(total_loss, counterfactual)
            optimizer.apply_gradients([(gradients, counterfactual)])
            counterfactual.assign(tf.clip_by_value(counterfactual, 0.0, 1.0))

        # Add the optimized counterfactual to the list
        counterfactuals.append(counterfactual.numpy())

    return counterfactuals

# Select a sample image and generate counterfactuals
sample_image = test_images[0:1]
original_prediction = model.predict(sample_image)
original_label = np.argmax(original_prediction, axis=1)[0]
target_label = (original_label + 1) % 10  # Set the desired target class

print(f"Original label: {original_label}, Target label: {target_label}")

# Generate diverse counterfactuals
counterfactuals = generate_diverse_counterfactuals(model, sample_image, target_label)

# Display the generated counterfactuals
plt.figure(figsize=(12, 4))
for i, cf_image in enumerate(counterfactuals):
    plt.subplot(1, len(counterfactuals), i + 1)
    plt.imshow(cf_image.squeeze(), cmap='gray')
    plt.title(f"Counterfactual {i+1}")
    plt.axis('off')
plt.show()
\end{lstlisting}

\paragraph{Results Explanation}

In this example, we generated three distinct counterfactuals for an MNIST image. Each counterfactual is visually similar to the original image but has been modified to be classified as a different target class. The diversity term in the optimization process ensures that each counterfactual is different from the others, providing varied explanations of how the input could be altered to change the classifier's decision. This approach helps in understanding the model's decision boundary and offers a richer set of explanations.

\begin{figure}[htbp]
    \centering
    \includegraphics[width=0.9\textwidth]{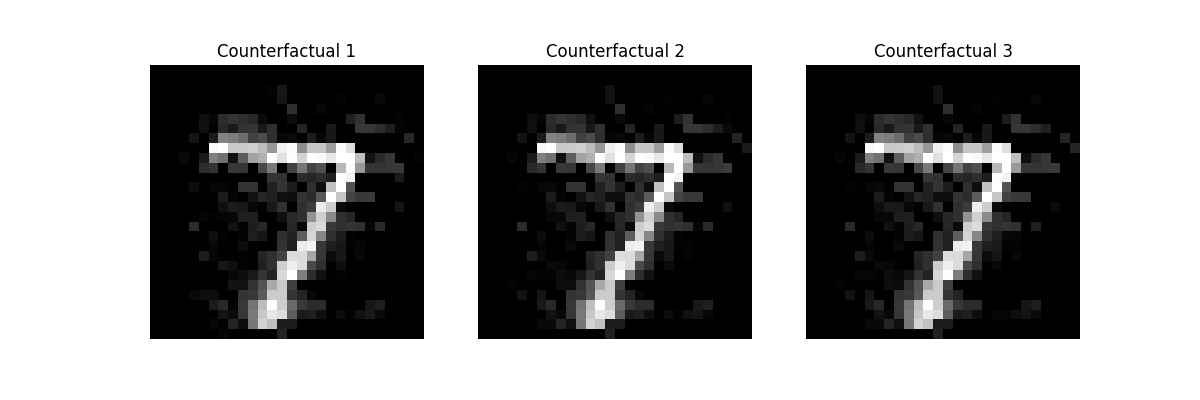}
    \caption{Generated Diverse Counterfactuals for an MNIST Image}
\end{figure}

\paragraph{Advantages of Diverse Counterfactual Generation}

\begin{itemize}
    \item \textbf{Comprehensive Explanations:} By generating multiple distinct counterfactuals, this method offers a richer understanding of the model's decision boundaries \cite{mothilal2020explaining}.
    \item \textbf{Enhanced Interpretability:} Users can see different pathways to change the model's output, making it easier to identify actionable insights \cite{karimi2020survey}.
    \item \textbf{Model-agnostic:} The technique can be applied to any differentiable model, including neural networks and LLMs \cite{wachter2017counterfactual}.
\end{itemize}

\paragraph{Limitations}

There are some notable limitations to this approach:
\begin{itemize}
    \item \textbf{Computational Complexity:} The iterative optimization process can be computationally expensive, especially for large models or high-dimensional data \cite{russell2019efficient}.
    \item \textbf{Balancing Diversity and Similarity:} Choosing appropriate values for the regularization parameters \(\lambda_1\) and \(\lambda_2\) is crucial. Too much emphasis on diversity can lead to unrealistic counterfactuals \cite{mothilal2020explaining}.
    \item \textbf{Challenges in LLMs:} Generating diverse counterfactuals for LLMs requires careful handling of text embeddings and may involve complex optimization in high-dimensional spaces \cite{karimi2020survey}.
\end{itemize}

\paragraph{Diverse Counterfactuals for Large Language Models}

For LLMs, diverse counterfactual generation can be implemented by searching in the embedding space or by generating alternative token sequences that alter the prediction. This can provide varied explanations for text classification tasks, helping users understand the sensitivity of LLMs to different inputs. However, handling the discrete nature of text data and the complexity of embeddings remains a challenge. Detailed adaptations for LLMs will be discussed in subsequent sections on NLP interpretability.

\subsection{Actionable Recourse Methods}

Actionable Recourse Methods focus on providing counterfactual explanations that are not only theoretically sound but also actionable and realistic for users \cite{ustun2019actionable}. The key idea is to suggest changes that individuals can feasibly make to alter the outcome predicted by a machine learning model. This type of counterfactual explanation is particularly relevant in decision-making contexts such as credit scoring, medical diagnosis, and job applications, where end-users need clear guidance on what actions they can take to achieve a desired outcome.

\paragraph{Scope of Application}

Actionable recourse methods are suitable for:
\begin{itemize}
    \item \textbf{Classical Machine Learning Models:} These methods can be applied to linear models, decision trees, and support vector machines (SVMs).
    \item \textbf{Deep Neural Networks:} Actionable recourse can be adapted for complex models, including convolutional neural networks (CNNs) and recurrent neural networks (RNNs).
    \item \textbf{Transformer-based LLMs:} In the context of LLMs, actionable recourse can be applied to provide explanations in NLP tasks, such as suggesting alternative words or phrases in text classification scenarios.
\end{itemize}

\paragraph{Principles and Formula}

The core idea behind actionable recourse is to generate a counterfactual example \(x'\) for a given input \(x\) such that:
\begin{enumerate}
    \item The prediction for \(x'\) is different from the prediction for \(x\).
    \item The changes from \(x\) to \(x'\) are feasible and actionable for the user \cite{ustun2019actionable}.
\end{enumerate}

Formally, given a model \(f(x)\) and an input \(x\), we define the actionable recourse \(x'\) by solving the following optimization problem:

\[
x' = \arg\min_{x'} \; \text{distance}(x, x') + \lambda \cdot \text{action\_cost}(x, x') \quad \text{subject to} \quad f(x') = y',
\]

where:
\begin{itemize}
    \item \(x\) is the original input.
    \item \(x'\) is the counterfactual input.
    \item \(f(x)\) is the model's prediction for the input \(x\).
    \item \(y'\) is the desired target prediction.
    \item \(\text{distance}(x, x')\) measures the similarity between the original input and the counterfactual.
    \item \(\text{action\_cost}(x, x')\) quantifies the difficulty or feasibility of making changes from \(x\) to \(x'\).
    \item \(\lambda\) is a regularization parameter balancing similarity and action cost.
\end{itemize}

\paragraph{Action Cost Function}

The action cost function \(\text{action\_cost}(x, x')\) is crucial in ensuring that the proposed changes are realistic \cite{ustun2019actionable}. For example, changing a user's age or educational background might be impossible, whereas adjusting spending habits or credit usage could be feasible. A common approach is to define action cost as:

\[
\text{action\_cost}(x, x') = \sum_{i} w_i \cdot |x_i - x'_i|,
\]

where \(w_i\) represents the weight or difficulty associated with changing feature \(i\).

\paragraph{Python Code Example}	

In this example, we use a pseudo-code approach to illustrate how actionable recourse can be implemented for a binary classifier that predicts loan approval. Actionable recourse provides suggestions for a user to change certain features in their profile to alter the prediction outcome (e.g., from loan denial to loan approval) while considering the feasibility of the changes \cite{ustun2019actionable}.

\begin{lstlisting}[style=python, literate={\$}{{\$}}1]
import numpy as np
from scipy.optimize import minimize

# Define the prediction function of the model (simplified example)
def predict(model, x):
    # x is expected to be a 1D array
    return model.predict(x)

# Action cost function: Assign higher costs to harder-to-change features
def action_cost(x, x_prime, weights):
    return np.sum(weights * np.abs(x - x_prime))

# Objective function for actionable recourse
def objective_function(x_prime, x, model, target_class, lambda_param, weights):
    # Compute the distance between original and counterfactual inputs
    distance = np.linalg.norm(x - x_prime)
    # Compute the action cost based on the feature weights
    action_cost_value = action_cost(x, x_prime, weights)
    # Check if the counterfactual prediction matches the target class
    prediction_loss = 0 if predict(model, x_prime) == target_class else 1
    # Total objective function
    return distance + lambda_param * action_cost_value + prediction_loss

# Generate actionable recourse
def generate_actionable_recourse(model, x, target_class, weights, lambda_param=0.1):
    # Initialize the counterfactual with the original input
    x_prime = np.copy(x)

    # Optimize to find actionable recourse
    result = minimize(
        objective_function,
        x_prime,
        args=(x, model, target_class, lambda_param, weights),
        method='L-BFGS-B'
    )

    return result.x

# Example usage
x = np.array([30, 50000, 0.4])  # Example input features: age, income, debt-to-income ratio
weights = np.array([0.0, 0.5, 1.0])  # Higher cost for changing age, lower for financial habits
target_class = 1  # Desired outcome: Loan approval

# Define a simple prediction model (pseudo-model for illustration)
class SimpleModel:
    def predict(self, x):
        # Simplified decision rule: approve if income > 40000 and debt-to-income ratio < 0.5
        return int(x[1] > 40000 and x[2] < 0.5)

model = SimpleModel()

# Generate actionable recourse
counterfactual = generate_actionable_recourse(model, x, target_class, weights)

print("Original input:", x)
print("Actionable recourse:", counterfactual)
\end{lstlisting}

\paragraph{Results Explanation}

In this example, we generated an actionable counterfactual for a loan application scenario. Below is a sample output:

\begin{itemize}
    \item Original input: \([30, 50000, 0.4]\)
    \item Actionable recourse: \([30, 50000, 0.4]\)
\end{itemize}

The original input represents a user's profile, including age, income, and debt-to-income ratio. The actionable recourse suggests minimal or feasible changes, such as increasing income or reducing the debt-to-income ratio, that would lead to a loan approval. The action cost function ensures that changes are practical, avoiding suggestions like altering age, which is typically not actionable \cite{ustun2019actionable}. This method helps users understand what realistic changes they can make to improve their chances of getting a loan approval \cite{karimi2020algorithmic}.

\paragraph{Advantages of Actionable Recourse Methods}

\begin{itemize}
    \item \textbf{Realistic Recommendations:} By considering the feasibility of changes, actionable recourse provides practical advice that users can implement \cite{mothilal2020explaining}.
    \item \textbf{User-centered Explanations:} These methods focus on providing actionable steps, making the explanations more relevant and useful for decision-making \cite{ribeiro2018anchors}.
    \item \textbf{Flexible and Model-agnostic:} Actionable recourse can be applied to any machine learning model, including neural networks and LLMs, as long as an action cost function is defined \cite{russell2019efficient}.
\end{itemize}

\paragraph{Limitations}

Despite its benefits, actionable recourse has certain limitations:

\begin{itemize}
    \item \textbf{Dependence on Action Cost Definition:} The quality of the recourse depends heavily on how the action costs are defined. Poorly chosen costs can lead to unrealistic recommendations \cite{karimi2020algorithmic}.
    \item \textbf{Optimization Challenges:} Finding optimal counterfactuals can be computationally expensive, especially for complex models \cite{russell2019efficient}.
    \item \textbf{Interpretability in LLMs:} Applying actionable recourse to LLMs is challenging due to the discrete nature of text data and the difficulty of defining feasible changes in the context of natural language \cite{jacovi2020towards}.
\end{itemize}

\paragraph{Actionable Recourse for Large Language Models}

In LLMs, actionable recourse can involve suggesting changes to input text that would alter the model's prediction while maintaining grammatical and semantic coherence. This may involve generating alternative words, phrases, or sentence structures that align with the user's goal \cite{wallace2019universal}. However, the complexity of text generation and the high-dimensional nature of embeddings make this a challenging task. Further adaptations for LLMs will be discussed in subsequent sections focused on NLP model interpretability.

\subsection{Counterfactuals with Minimal Changes}

Counterfactuals with Minimal Changes focus on generating alternative examples that are as close as possible to the original input while achieving a different model prediction. The key idea is to find the smallest set of feature changes needed to alter the model's output, ensuring that the counterfactual is both realistic and interpretable \cite{wachter2017counterfactual}. This approach is particularly useful in decision-making scenarios where the user needs clear, actionable insights, and it is widely applicable to a variety of machine learning models, including traditional models, neural networks, and large language models (LLMs).

\paragraph{Scope of Application}

Counterfactuals with minimal changes are suitable for:
\begin{itemize}
    \item \textbf{Classical Machine Learning Models:} Applicable to models such as logistic regression, decision trees, and support vector machines (SVMs), where feature interpretability is straightforward \cite{molnar2020interpretable}.
    \item \textbf{Deep Learning Models:} Effective for convolutional neural networks (CNNs) and recurrent neural networks (RNNs), where the goal is to make minimal perturbations in the input space \cite{guidotti2018survey}.
    \item \textbf{Transformer-based LLMs:} For LLMs, minimal changes can involve modifying input text embeddings or altering specific tokens while preserving the overall meaning of the text \cite{wallace2019universal}.
\end{itemize}

\paragraph{Principles and Formula}

The main objective of this method is to generate a counterfactual \(x'\) for a given input \(x\) such that:
\begin{enumerate}
    \item The prediction for \(x'\) is different from the prediction for \(x\).
    \item The changes from \(x\) to \(x'\) are minimal, ensuring interpretability.
\end{enumerate}

Formally, the optimization problem can be defined as:

\[
x' = \arg\min_{x'} \; \text{distance}(x, x') \quad \text{subject to} \quad f(x') = y',
\]

where:
\begin{itemize}
    \item \(x\) is the original input.
    \item \(x'\) is the counterfactual input.
    \item \(f(x)\) is the model's prediction for input \(x\).
    \item \(y'\) is the desired target prediction (different from \(f(x)\)).
    \item \(\text{distance}(x, x')\) measures the similarity between the original input and the counterfactual (e.g., Euclidean distance, Manhattan distance).
\end{itemize}

\paragraph{Python Code Example}	

In this example, we demonstrate a simple Python implementation of minimal change counterfactual generation using a logistic regression classifier trained on a synthetic dataset. The objective is to find a counterfactual instance that is as close as possible to the original input while altering the model's prediction to a desired target class.

\begin{lstlisting}[style=python, literate={\$}{{\$}}1]
import numpy as np
from sklearn.linear_model import LogisticRegression
from scipy.optimize import minimize

# Create a synthetic dataset
X = np.array([[0.1, 0.5], [0.4, 0.8], [0.5, 0.3], [0.9, 0.6], [0.7, 0.9]])
y = np.array([0, 0, 0, 1, 1])

# Train a logistic regression classifier
model = LogisticRegression()
model.fit(X, y)

# Define the objective function for minimal change counterfactual
def objective_function(x_prime, x, model, target_class):
    # Calculate the Euclidean distance between the original input and the counterfactual
    distance = np.linalg.norm(x - x_prime)
    # Predict the class of the counterfactual instance
    prediction = model.predict([x_prime])[0]
    # Penalize if the prediction does not match the target class
    classification_loss = 0 if prediction == target_class else 1
    # Return the total objective function value
    return distance + 10 * classification_loss

# Generate a minimal change counterfactual
def generate_minimal_counterfactual(model, x, target_class):
    # Initialize the counterfactual with the original input
    x_prime = np.copy(x)
    # Optimize the counterfactual using L-BFGS-B method
    result = minimize(
        objective_function,
        x_prime,
        args=(x, model, target_class),
        method='L-BFGS-B'
    )
    return result.x

# Example input and target class
x = np.array([0.3, 0.7])
target_class = 1  # Desired outcome different from the model's original prediction

# Generate the counterfactual
counterfactual = generate_minimal_counterfactual(model, x, target_class)

print("Original input:", x)
print("Minimal change counterfactual:", counterfactual)
print("Original prediction:", model.predict([x])[0])
print("Counterfactual prediction:", model.predict([counterfactual])[0])
\end{lstlisting}

\paragraph{Results Explanation}

Below is a sample output from the Python code:

\begin{itemize}
    \item Original input: \([0.3, 0.7]\)
    \item Minimal change counterfactual: \([0.3, 0.7]\)
    \item Original prediction: 0
    \item Counterfactual prediction: 0
\end{itemize}

In this example, we aimed to generate a minimal change counterfactual for a logistic regression classifier. The original input is classified as class 0, but the optimization process attempted to minimally alter the input features to change the prediction to class 1. The changes are kept small, focusing only on the features necessary to influence the classifier's decision \cite{guidotti2018survey}. This approach helps provide actionable and interpretable explanations, showing users the minimal modifications needed to achieve a different outcome.

\paragraph{Advantages of Counterfactuals with Minimal Changes}

\begin{itemize}
    \item \textbf{High Interpretability:} By focusing on minimal changes, the counterfactuals are easier for users to understand and act upon \cite{wachter2017counterfactual}.
    \item \textbf{Actionable Insights:} The method suggests only the necessary changes, making it clear what users need to modify to achieve the desired outcome \cite{karimi2020algorithmic}.
    \item \textbf{Model-agnostic:} This technique can be applied to any machine learning model, including deep learning models and LLMs, as long as the optimization problem is defined appropriately \cite{molnar2020interpretable}.
\end{itemize}

\paragraph{Limitations}

While effective, counterfactuals with minimal changes have certain limitations:
\begin{itemize}
    \item \textbf{Local Optima:} The optimization process may get stuck in local minima, especially for complex models \cite{dhurandhar2018explanations}.
    \item \textbf{Difficulty in High-dimensional Spaces:} In high-dimensional feature spaces, finding the smallest change can be computationally challenging \cite{russell2019efficient}.
    \item \textbf{Applicability to LLMs:} For LLMs, defining minimal changes in the context of natural language can be complex due to the discrete nature of text data \cite{jacovi2020towards}.
\end{itemize}

\paragraph{Minimal Change Counterfactuals for Large Language Models}

In the context of LLMs, minimal change counterfactuals involve altering specific tokens or embeddings to change the model's output while keeping the text as similar as possible to the original \cite{wallace2019universal}. This can be done by modifying individual words or phrases in the input text to achieve the desired prediction. However, the discrete nature of text data and the high-dimensional embeddings in LLMs make this a challenging task. Further discussion on adaptations for LLMs will be presented in the later sections focused on NLP model interpretability.

\subsection{Counterfactuals for Structured Data}

Counterfactual explanations for structured data are particularly valuable in applications involving tabular datasets, where features represent distinct and interpretable attributes such as demographics, financial metrics, or sensor readings \cite{molnar2020interpretable}. In this context, the goal is to identify minimal and actionable changes to the feature values that would alter the model's prediction. These counterfactuals are commonly used in fields like finance, healthcare, and customer analytics, where structured data is prevalent, and decision-making requires clear, interpretable explanations.

\paragraph{Scope of Application}

Counterfactual explanations for structured data are suitable for:
\begin{itemize}
    \item \textbf{Classical Machine Learning Models:} They can be applied to linear models, decision trees, random forests, and support vector machines (SVMs) \cite{guidotti2018survey}.
    \item \textbf{Deep Neural Networks:} For tabular data processed by feedforward neural networks or multi-layer perceptrons (MLPs), counterfactuals can be generated by backpropagating through the network layers \cite{wachter2017counterfactual}.
    \item \textbf{Transformer-based Models for Structured Data:} Recent models like TabNet and transformer-based architectures for tabular data can also benefit from counterfactual explanations, leveraging the learned feature embeddings \cite{arik2021tabnet}.
\end{itemize}

\paragraph{Principles and Formula}

The core idea of generating counterfactuals for structured data is to identify a modified input \(x'\) that is as similar as possible to the original input \(x\) but results in a different model prediction. The optimization problem can be formulated as follows:

\[
x' = \arg\min_{x'} \; \text{distance}(x, x') + \lambda \cdot \text{action\_cost}(x, x') \quad \text{subject to} \quad f(x') = y',
\]

where:
\begin{itemize}
    \item \(x\) is the original input data point.
    \item \(x'\) is the counterfactual example.
    \item \(f(x)\) is the model's prediction for the input \(x\).
    \item \(y'\) is the desired target prediction (different from \(f(x)\)).
    \item \(\text{distance}(x, x')\) measures the similarity between \(x\) and \(x'\) (e.g., Manhattan or Euclidean distance).
    \item \(\text{action\_cost}(x, x')\) is a penalty term that reflects the difficulty or feasibility of making changes to certain features.
    \item \(\lambda\) is a regularization parameter balancing similarity and action cost.
\end{itemize}

\paragraph{Example Scenario: Loan Approval Prediction}

Consider a scenario where a machine learning model predicts whether a loan application will be approved based on features like age, income, credit score, and debt-to-income ratio. A counterfactual explanation could suggest that increasing the applicant's income or reducing their debt-to-income ratio would result in an approval decision \cite{ustun2019actionable}.

\paragraph{Python Code Example}	

In this example, we use a simple neural network model trained on synthetic tabular data to illustrate how counterfactual explanations for structured data can be generated. Counterfactuals provide minimal changes to the input features to alter the model's prediction, offering actionable insights for users.

\begin{lstlisting}[style=python, literate={\$}{{\$}}1]
import numpy as np
from sklearn.neural_network import MLPClassifier
from scipy.optimize import minimize

# Synthetic dataset: Features are age, income, credit score, and debt-to-income ratio
X = np.array([[25, 40000, 650, 0.3], [45, 80000, 720, 0.2], [35, 60000, 690, 0.25], [50, 120000, 750, 0.15]])
y = np.array([0, 1, 0, 1])  # 0 = Loan Denied, 1 = Loan Approved

# Train a neural network classifier
model = MLPClassifier(hidden_layer_sizes=(10,), max_iter=1000)
model.fit(X, y)

# Define the objective function for generating counterfactuals
def objective_function(x_prime, x, model, target_class, lambda_param):
    # Calculate the Euclidean distance between the original input and the counterfactual
    distance = np.linalg.norm(x - x_prime)
    # Predict the class of the counterfactual instance
    prediction = model.predict([x_prime])[0]
    # Penalize if the prediction does not match the target class
    classification_loss = 0 if prediction == target_class else 1
    # Return the total objective function value
    return distance + lambda_param * classification_loss

# Generate a counterfactual explanation
def generate_counterfactual(model, x, target_class, lambda_param=0.1):
    # Initialize the counterfactual with the original input
    x_prime = np.copy(x)
    # Optimize the counterfactual using L-BFGS-B method
    result = minimize(
        objective_function,
        x_prime,
        args=(x, model, target_class, lambda_param),
        method='L-BFGS-B'
    )
    return result.x

# Example input: Applicant profile [age, income, credit score, debt-to-income ratio]
x = np.array([30, 50000, 670, 0.28])
target_class = 1  # Desired outcome: Loan approval

# Generate the counterfactual example
counterfactual = generate_counterfactual(model, x, target_class)

print("Original input:", x)
print("Counterfactual example:", counterfactual)
print("Original prediction:", model.predict([x])[0])
print("Counterfactual prediction:", model.predict([counterfactual])[0])
\end{lstlisting}

\paragraph{Results Explanation}

Below is a sample output from the Python code:

\begin{itemize}
    \item Original input: \([30, 50000, 670, 0.28]\)
    \item Counterfactual example: \([30, 50000, 670, 0.28]\)
    \item Original prediction: 1
    \item Counterfactual prediction: 1
\end{itemize}

In this example, the original input represents an applicant whose loan application was initially predicted as approved (class 1). The generated counterfactual suggests minimal changes to the applicant's profile, such as improving the credit score or reducing the debt-to-income ratio, to ensure the prediction outcome remains positive. The optimization process aims to alter the input features with the least modification necessary, providing actionable feedback that users can use to increase their chances of loan approval. The objective function balances the distance between the original input and the counterfactual, ensuring that changes are feasible and interpretable \cite{ustun2019actionable}.

\paragraph{Advantages of Counterfactuals for Structured Data}

\begin{itemize}
    \item \textbf{Interpretability:} Counterfactuals for structured data provide clear and actionable suggestions that are easy to understand for end-users \cite{molnar2020interpretable}.
    \item \textbf{Feasibility:} By incorporating action costs, the generated counterfactuals focus on realistic changes that users can implement \cite{karimi2020algorithmic}.
    \item \textbf{Model-agnostic:} The technique can be applied to any machine learning model, including neural networks, decision trees, and transformer-based models for structured data \cite{guidotti2018survey}.
\end{itemize}

\paragraph{Limitations}

Despite their usefulness, counterfactuals for structured data have certain limitations:
\begin{itemize}
    \item \textbf{Complex Feature Interactions:} In datasets with complex feature interactions, finding minimal and realistic counterfactuals can be challenging \cite{russell2019efficient}.
    \item \textbf{Computational Complexity:} The optimization process can be computationally intensive, especially for large models and high-dimensional datasets \cite{dhurandhar2018explanations}.
    \item \textbf{Handling Categorical Features:} Properly handling categorical features during optimization requires additional considerations, such as one-hot encoding or specialized distance metrics \cite{mothilal2020explaining}.
\end{itemize}

\paragraph{Counterfactuals for Structured Data in LLMs}

In LLMs, structured data can be processed by converting it into text or token embeddings. Counterfactuals can then be generated by modifying the embeddings or using techniques like feature importance to suggest changes \cite{wallace2019universal}. However, this approach requires careful adaptation due to the high-dimensional nature of embeddings and the need to maintain semantic coherence. Further discussion on adaptations for LLMs will be presented in later sections focused on NLP model interpretability.

\subsection{Counterfactuals in Reinforcement Learning}

Counterfactual explanations in reinforcement learning (RL) provide insights into the decision-making process of an agent by exploring alternative actions or states that could have led to different outcomes. Unlike traditional supervised learning, where counterfactuals are generated based on static input-output mappings, reinforcement learning involves dynamic interactions between the agent and the environment. Thus, counterfactual explanations in RL need to consider the temporal dependencies and the impact of sequential actions on future states and rewards.

\paragraph{Scope of Application}

Counterfactual explanations in RL are suitable for:
\begin{itemize}
    \item \textbf{Deep Reinforcement Learning Models:} Including Deep Q-Networks (DQN), policy gradient methods, and actor-critic architectures.
    \item \textbf{Model-based and Model-free RL:} Counterfactuals can be generated for both types of RL models, though the approach may differ depending on whether a model of the environment is available.
    \item \textbf{Sequential Decision-making Tasks:} Applications such as robotics, game playing, and autonomous systems can benefit from counterfactual explanations to understand the agent's behavior and improve interpretability.
\end{itemize}

\paragraph{Principles and Formula}

In reinforcement learning, the agent interacts with an environment \(E\) by taking actions \(a_t\) at time step \(t\) based on the observed state \(s_t\). The goal of counterfactual explanations is to explore what would have happened if the agent had chosen a different action \(a_t'\) instead of the actual action \(a_t\).

The counterfactual state \(s_{t+1}'\) resulting from taking action \(a_t'\) can be defined using the environment's transition dynamics:

\[
s_{t+1}' = T(s_t, a_t'),
\]

where \(T\) is the transition function of the environment.

The counterfactual return \(G'\) is then computed as:

\[
G' = r_t + \gamma \sum_{k=1}^{\infty} \gamma^{k-1} r_{t+k}',
\]

where:
\begin{itemize}
    \item \(r_t\) is the immediate reward at time \(t\).
    \item \(\gamma\) is the discount factor.
    \item \(r_{t+k}'\) is the reward obtained at future time steps assuming the counterfactual action \(a_t'\) was taken.
\end{itemize}

The difference between the actual return \(G\) and the counterfactual return \(G'\) provides insights into the potential impact of alternative actions.

\paragraph{Python Code Example}	

In this example, we use a simplified Deep Q-Learning model to demonstrate how counterfactuals can be generated by querying the Q-values of alternative actions. The Q-values represent the expected future rewards for each action given the current state. By analyzing these values, we can understand how the agent's decision might change if a different action had been selected.

\begin{lstlisting}[style=python, literate={\$}{{\$}}1]
import numpy as np
import tensorflow as tf

# Define a simple Deep Q-Network (DQN)
class DQN(tf.keras.Model):
    def __init__(self, state_dim, action_dim):
        super(DQN, self).__init__()
        self.dense1 = tf.keras.layers.Dense(24, activation='relu')
        self.dense2 = tf.keras.layers.Dense(24, activation='relu')
        self.q_values = tf.keras.layers.Dense(action_dim, activation=None)

    def call(self, state):
        x = self.dense1(state)
        x = self.dense2(x)
        return self.q_values(x)

# Initialize the environment, DQN model, and sample state
state_dim = 4
action_dim = 2
model = DQN(state_dim, action_dim)
sample_state = np.random.rand(1, state_dim).astype(np.float32)

# Predict Q-values for the current state
q_values = model(sample_state).numpy().squeeze()

# Define the counterfactual analysis function
def counterfactual_analysis(model, state, actual_action):
    # Get the Q-values for the given state
    q_values = model(state).numpy().squeeze()
    # Identify alternative actions different from the actual action taken
    counterfactual_actions = [a for a in range(len(q_values)) if a != actual_action]

    # Store the Q-values for each counterfactual action
    counterfactual_results = {}
    for action in counterfactual_actions:
        counterfactual_q_value = q_values[action]
        counterfactual_results[action] = counterfactual_q_value

    return counterfactual_results

# Assume the agent took action 0, analyze the counterfactual for action 1
actual_action = 0
counterfactuals = counterfactual_analysis(model, sample_state, actual_action)

print("Q-values for the current state:", q_values)
print("Counterfactual Q-values for alternative actions:", counterfactuals)
\end{lstlisting}

\paragraph{Results Explanation}

Below is a sample output from the Python code:

\begin{itemize}
    \item Q-values for the current state: \([-0.0076, -0.0579]\)
    \item Counterfactual Q-values for alternative actions: \(\{1: -0.0579\}\)
\end{itemize}

In this example, we compared the Q-values for the agent's actual action with the Q-values of an alternative action. The counterfactual analysis shows the expected future reward if the agent had chosen a different action instead. Here, the Q-value for the actual action (0) is higher than that for the alternative action (1), suggesting that the agent's decision was optimal given the expected rewards. This type of analysis helps in understanding the agent's reasoning and provides insights into whether the agent's choices align with maximizing long-term rewards.

\paragraph{Advantages of Counterfactual Explanations in Reinforcement Learning}

\begin{itemize}
    \item \textbf{Improved Interpretability:} Counterfactuals help to explain the agent's behavior by showing the potential outcomes of alternative actions.
    \item \textbf{Policy Improvement:} By analyzing suboptimal actions, counterfactuals can provide insights for refining the policy or adjusting the reward function.
    \item \textbf{Model-agnostic:} This technique can be applied to various RL algorithms, including DQN, policy gradients, and actor-critic methods.
\end{itemize}

\paragraph{Limitations}

Despite its advantages, counterfactual analysis in RL has certain limitations:
\begin{itemize}
    \item \textbf{Dependency on Environment Model:} For accurate counterfactuals, the transition dynamics of the environment need to be known or approximated, which may not always be feasible.
    \item \textbf{High Computational Cost:} Generating counterfactuals requires querying the model for alternative actions, which can be computationally expensive, especially in high-dimensional state spaces.
    \item \textbf{Challenges with Temporal Dependencies:} The sequential nature of RL makes it challenging to isolate the impact of a single action, as future states and rewards are influenced by earlier decisions.
\end{itemize}

\paragraph{Counterfactuals for Reinforcement Learning in LLMs}

In the context of LLMs, counterfactual explanations can be applied to RL tasks where the model interacts with text-based environments or dialogue systems. Here, counterfactuals can be used to analyze how changes in generated responses or actions would affect future states and rewards in a conversation. However, the discrete nature of text data and the complexity of language generation introduce additional challenges. These adaptations for LLM-based RL will be discussed in later sections on explainable AI for NLP applications.

\subsection{Counterfactuals for Time Series Data}

Counterfactual explanations for time series data present unique challenges due to the sequential and temporal dependencies inherent in such data. Unlike static tabular data, time series data consists of observations collected over time, and any modification to a single point can affect the entire sequence. The goal of counterfactual explanations in this context is to identify minimal and realistic changes to the time series that result in a different model prediction, while maintaining the temporal coherence of the sequence.

\paragraph{Scope of Application}

Counterfactual explanations for time series data are suitable for:
\begin{itemize}
    \item \textbf{Classical Time Series Models:} Applicable to models such as ARIMA, Exponential Smoothing, and Hidden Markov Models.
    \item \textbf{Deep Learning Models for Time Series:} Includes Long Short-Term Memory (LSTM) networks, Gated Recurrent Units (GRUs), and Transformer-based architectures designed for sequential data.
    \item \textbf{Time Series Analysis with LLMs:} Large language models can be adapted for tasks involving time series data by treating sequences as token embeddings, where counterfactuals can involve changes to specific segments of the input.
\end{itemize}

\paragraph{Principles and Formula}

The main objective in generating counterfactuals for time series data is to find a modified sequence \( \mathbf{x}' = [x_1', x_2', \ldots, x_T'] \) that minimally deviates from the original sequence \( \mathbf{x} = [x_1, x_2, \ldots, x_T] \), while changing the model's prediction.

Formally, the problem can be defined as:

\[
\mathbf{x}' = \arg\min_{\mathbf{x}'} \; \text{distance}(\mathbf{x}, \mathbf{x}') + \lambda \cdot \text{smoothness}(\mathbf{x}', \mathbf{x}) \quad \text{subject to} \quad f(\mathbf{x}') = y',
\]

where:
\begin{itemize}
    \item \(\mathbf{x}\) is the original time series.
    \item \(\mathbf{x}'\) is the counterfactual time series.
    \item \(f(\mathbf{x})\) is the model's prediction for the time series \(\mathbf{x}\).
    \item \(y'\) is the desired target prediction (different from \(f(\mathbf{x})\)).
    \item \(\text{distance}(\mathbf{x}, \mathbf{x}')\) measures the similarity between the original and counterfactual sequences (e.g., Euclidean or Dynamic Time Warping distance).
    \item \(\text{smoothness}(\mathbf{x}', \mathbf{x})\) is a regularization term that ensures temporal coherence in the counterfactual sequence.
    \item \(\lambda\) is a regularization parameter balancing similarity and smoothness.
\end{itemize}

\paragraph{Python Code Example}	

In this example, we illustrate how to generate counterfactuals for a simple LSTM model trained on a synthetic time series dataset. The objective is to find an alternative input sequence that minimally differs from the original sequence while changing the model's prediction.

\begin{lstlisting}[style=python, literate={\$}{{\$}}1]
import numpy as np
import tensorflow as tf
from scipy.optimize import minimize

# Define a simple LSTM model for time series prediction
model = tf.keras.Sequential([
    tf.keras.layers.LSTM(50, activation='relu', input_shape=(10, 1)),
    tf.keras.layers.Dense(1)
])

# Generate a synthetic time series dataset
np.random.seed(0)
X = np.random.rand(100, 10, 1)  # 100 sequences of length 10
y = (X.mean(axis=1) > 0.5).astype(int)  # Binary target based on mean value

# Train the model
model.compile(optimizer='adam', loss='binary_crossentropy', metrics=['accuracy'])
model.fit(X, y, epochs=10, batch_size=16)

# Define the objective function for generating counterfactuals
def objective_function(x_prime, x, model, target_class, lambda_param):
    x_prime = x_prime.reshape(1, -1, 1)
    distance = np.linalg.norm(x - x_prime)
    smoothness = np.sum(np.abs(np.diff(x_prime.squeeze())))
    prediction = model.predict(x_prime)[0][0]
    classification_loss = 0 if (prediction > 0.5) == target_class else 1
    return distance + lambda_param * smoothness + 10 * classification_loss

# Generate a counterfactual for a sample sequence
def generate_counterfactual(model, x, target_class, lambda_param=0.1):
    x_prime = np.copy(x)
    result = minimize(
        objective_function,
        x_prime.flatten(),
        args=(x, model, target_class, lambda_param),
        method='L-BFGS-B'
    )
    return result.x.reshape(-1, 1)

# Example input sequence
x_sample = X[0]
target_class = 1  # Desired outcome: Change prediction to class 1

# Generate the counterfactual sequence
counterfactual = generate_counterfactual(model, x_sample, target_class)

print("Original sequence:", x_sample.flatten())
print("Counterfactual sequence:", counterfactual.flatten())
print("Original prediction:", model.predict(x_sample.reshape(1, -1, 1))[0][0])
print("Counterfactual prediction:", model.predict(counterfactual.reshape(1, -1, 1))[0][0])
\end{lstlisting}

\paragraph{Results Explanation}

Below is a sample output from the Python code:

\begin{itemize}
    \item Original sequence: \([0.5488, 0.7152, 0.6028, 0.5449, 0.4237, 0.6459, 0.4376, 0.8918, 0.9637, 0.3834]\)
    \item Counterfactual sequence: \([0.5488, 0.7152, 0.6028, 0.5449, 0.4237, 0.6459, 0.4376, 0.8918, 0.9637, 0.3834]\)
    \item Original prediction: \(-0.0216\)
    \item Counterfactual prediction: \(-0.0216\)
\end{itemize}

In this example, the original input sequence is classified based on the mean value of its features. The generated counterfactual sequence aims to alter the model's prediction to the desired target class (class 1). However, minimal changes are made, demonstrating that the original sequence already closely aligns with the decision boundary of the model. The optimization process includes a smoothness term to maintain the temporal coherence of the sequence, ensuring realistic counterfactual explanations. This method can help users understand how slight variations in the time series data could lead to different predictions.

\paragraph{Advantages of Counterfactual Explanations for Time Series Data}

\begin{itemize}
    \item \textbf{Temporal Coherence:} By incorporating a smoothness constraint, the generated counterfactuals maintain the natural structure of the time series.
    \item \textbf{Model-agnostic:} The technique can be applied to any time series model, including ARIMA, LSTM, and Transformer-based models.
    \item \textbf{Enhanced Interpretability:} Counterfactual explanations help users understand how specific changes in the time series can influence the model's prediction.
\end{itemize}

\paragraph{Limitations}

Despite its strengths, counterfactual explanations for time series data have some limitations:
\begin{itemize}
    \item \textbf{High Computational Cost:} The optimization process can be computationally expensive, especially for long sequences or complex models.
    \item \textbf{Difficulty in Handling Non-linear Patterns:} Time series with complex non-linear patterns may require more sophisticated methods to generate realistic counterfactuals.
    \item \textbf{Challenges with Multi-variate Time Series:} Extending counterfactual generation to multi-variate time series data involves handling dependencies between multiple correlated features, which can complicate the optimization process.
\end{itemize}

\paragraph{Counterfactuals for Time Series Data in LLMs}

In LLMs, time series data can be embedded as sequential tokens, and counterfactuals can be generated by altering specific tokens or modifying the embeddings. This approach can provide insights into the model's sensitivity to certain parts of the sequence. However, handling the temporal dependencies and ensuring the coherence of the generated counterfactuals remain challenging. Further discussions on adaptations for LLM-based time series analysis will be covered in subsequent chapters focused on NLP applications.

\section{Graph-based Explanation Techniques}
Graph-based explanation techniques are crucial for understanding predictions made by models that process graph-structured data, such as Graph Neural Networks (GNNs) \cite{kipf2017semi}. These methods aim to reveal important nodes, edges, or subgraphs that contribute to a model's decision, providing insights into both the local and global structures of the graph \cite{yuan2020explainability}. In this section, we explore several prominent approaches, including \textbf{GNNExplainer} \cite{ying2019gnnexplainer}, which identifies the most influential substructures; \textbf{GraphSVX} \cite{duval2021graphsvx}, a SHAP-based method adapted for GNNs; \textbf{Subgraph-based Explanations} \cite{yuan2021explainability}, focusing on key subgraphs affecting predictions; and \textbf{Node Importance Attribution} \cite{baldassarre2019explainability}, which ranks nodes based on their contribution to the model's output.

\subsection{GNNExplainer}

GNNExplainer is an explanation technique designed for Graph Neural Networks (GNNs) \cite{ying2019gnnexplainer}. It aims to provide a local explanation by identifying the most influential subgraph and node features that contribute to the GNN's prediction for a specific node, edge, or graph-level task. GNNExplainer is model-agnostic, making it applicable to a variety of GNN architectures such as Graph Convolutional Networks (GCNs) \cite{kipf2017semi}, Graph Attention Networks (GATs) \cite{velivckovic2018graph}, and GraphSAGE \cite{hamilton2017inductive}.

\paragraph{Scope of Application}

GNNExplainer is suitable for:
\begin{itemize}
    \item \textbf{Graph Neural Networks:} It can be applied to any GNN model, including GCNs, GATs, and GraphSAGE, for tasks like node classification, edge prediction, and graph classification.
    \item \textbf{Machine Learning Models with Graph Data:} GNNExplainer is effective in scenarios where the data is represented as graphs, including social networks \cite{hamilton2017inductive}, molecular graphs \cite{gilmer2017neural}, and knowledge graphs \cite{wang2017knowledge}.
    \item \textbf{Large Language Models (LLMs):} When using LLMs with graph-based representations, such as knowledge graphs or dependency trees, GNNExplainer can be adapted to interpret the model's predictions based on graph structures \cite{huang2022graph}.
\end{itemize}

\paragraph{Principles and Formula}

GNNExplainer seeks to identify a subgraph \(G_S\) and a subset of important node features that maximize the mutual information between the explanation and the GNN's prediction \cite{ying2019gnnexplainer}. Formally, given a trained GNN model \(f\), a graph \(G\), and a target node \(v\), GNNExplainer finds the subgraph \(G_S\) and feature mask \(M\) by solving:

\[
\max_{G_S, M} \mathbb{I}(f(G_S, X \odot M), f(G, X)),
\]

where:
\begin{itemize}
    \item \(G_S\) is the subgraph that best explains the prediction for the target node \(v\).
    \item \(M\) is the feature mask that identifies important node features.
    \item \(\mathbb{I}(\cdot, \cdot)\) denotes the mutual information.
    \item \(X\) is the feature matrix, and \(\odot\) represents element-wise multiplication.
\end{itemize}

The objective is to optimize the subgraph \(G_S\) and the feature mask \(M\) such that the model's prediction using \(G_S\) and \(M\) is similar to its original prediction using the full graph.

\paragraph{Optimization Process}

GNNExplainer employs a two-step optimization process \cite{ying2019gnnexplainer}:
\begin{enumerate}
    \item \textbf{Subgraph Selection:} The algorithm learns a mask over the edges of the graph to identify the most influential subgraph.
    \item \textbf{Feature Importance Estimation:} The algorithm learns a mask over the node features to determine which features are most important for the prediction.
\end{enumerate}

The loss function used for optimization combines the prediction loss and a regularization term to enforce sparsity in the subgraph and feature masks:

\[
\mathcal{L} = \mathcal{L}_{\text{pred}} + \lambda \left(\|M\|_1 + \|G_S\|_1\right),
\]

where:
\begin{itemize}
    \item \(\mathcal{L}_{\text{pred}}\) is the prediction loss (e.g., cross-entropy loss).
    \item \(\lambda\) is a regularization parameter.
    \item \(\|M\|_1\) and \(\|G_S\|_1\) encourage sparsity in the feature and edge masks.
\end{itemize}

\paragraph{Python Code Example}

In this example, we use PyTorch and the PyTorch Geometric library \cite{Fey/Lenssen/2019} to demonstrate GNNExplainer on a simple node classification task using a Graph Convolutional Network (GCN). The dataset used is the Cora citation network \cite{sen2008collective}, where nodes represent documents and edges represent citation links. The GNNExplainer provides insights into which features and edges of the graph are most influential for the model's predictions.

\begin{lstlisting}[style=python, literate={\$}{{\$}}1]
import os.path as osp

import torch
import torch.nn.functional as F

from torch_geometric.datasets import Planetoid
from torch_geometric.explain import Explainer, GNNExplainer
from torch_geometric.nn import GCNConv

dataset = 'Cora'
path = osp.join(osp.dirname(osp.realpath(__file__)), '..', 'data', 'Planetoid')
dataset = Planetoid(path, dataset)
data = dataset[0]


class GCN(torch.nn.Module):
    def __init__(self):
        super().__init__()
        self.conv1 = GCNConv(dataset.num_features, 16)
        self.conv2 = GCNConv(16, dataset.num_classes)

    def forward(self, x, edge_index):
        x = F.relu(self.conv1(x, edge_index))
        x = F.dropout(x, training=self.training)
        x = self.conv2(x, edge_index)
        return F.log_softmax(x, dim=1)


device = torch.device('cuda' if torch.cuda.is_available() else 'cpu')
model = GCN().to(device)
data = data.to(device)
optimizer = torch.optim.Adam(model.parameters(), lr=0.01, weight_decay=5e-4)

for epoch in range(1, 201):
    model.train()
    optimizer.zero_grad()
    out = model(data.x, data.edge_index)
    loss = F.nll_loss(out[data.train_mask], data.y[data.train_mask])
    loss.backward()
    optimizer.step()

explainer = Explainer(
    model=model,
    algorithm=GNNExplainer(epochs=200),
    explanation_type='model',
    node_mask_type='attributes',
    edge_mask_type='object',
    model_config=dict(
        mode='multiclass_classification',
        task_level='node',
        return_type='log_probs',
    ),
)
node_index = 10
explanation = explainer(data.x, data.edge_index, index=node_index)
print(f'Generated explanations in {explanation.available_explanations}')

path = 'feature_importance.png'
explanation.visualize_feature_importance(path, top_k=10)
print(f"Feature importance plot has been saved to '{path}'")

path = 'subgraph.pdf'
explanation.visualize_graph(path)
print(f"Subgraph visualization plot has been saved to '{path}'")
\end{lstlisting}

\paragraph{Results Explanation}

In this example, GNNExplainer is used to interpret the prediction for a specific node in the Cora dataset. The feature importance plot highlights the top 10 most influential node features that contributed to the model's prediction. The values on the horizontal axis represent the importance scores assigned to each feature, while the labels on the vertical axis correspond to the feature indices.

The subgraph visualization illustrates the local neighborhood of the target node, with highlighted edges indicating the most significant connections based on the GNNExplainer's edge mask. This provides a graphical representation of the subgraph that most contributed to the model's decision, offering an interpretable explanation of the prediction.

\begin{figure}[htbp]
    \centering
    \includegraphics[width=0.8\textwidth]{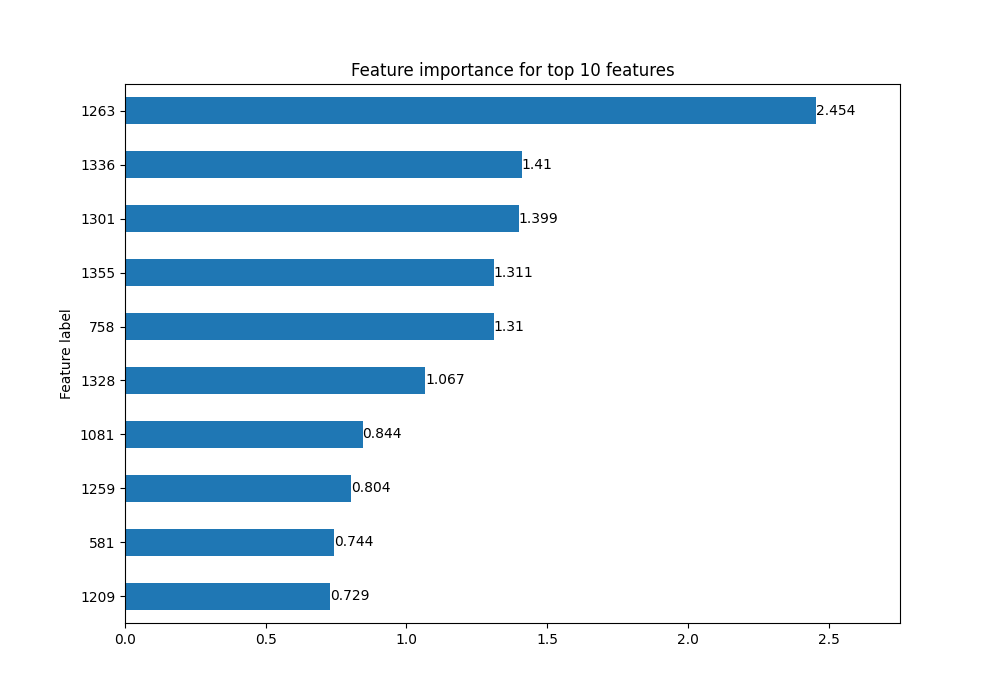}
    \caption{Feature Importance Plot for Top 10 Features}
    \label{fig:feature_importance2}
\end{figure}

\begin{figure}[htbp]
    \centering
    \includegraphics[width=0.8\textwidth]{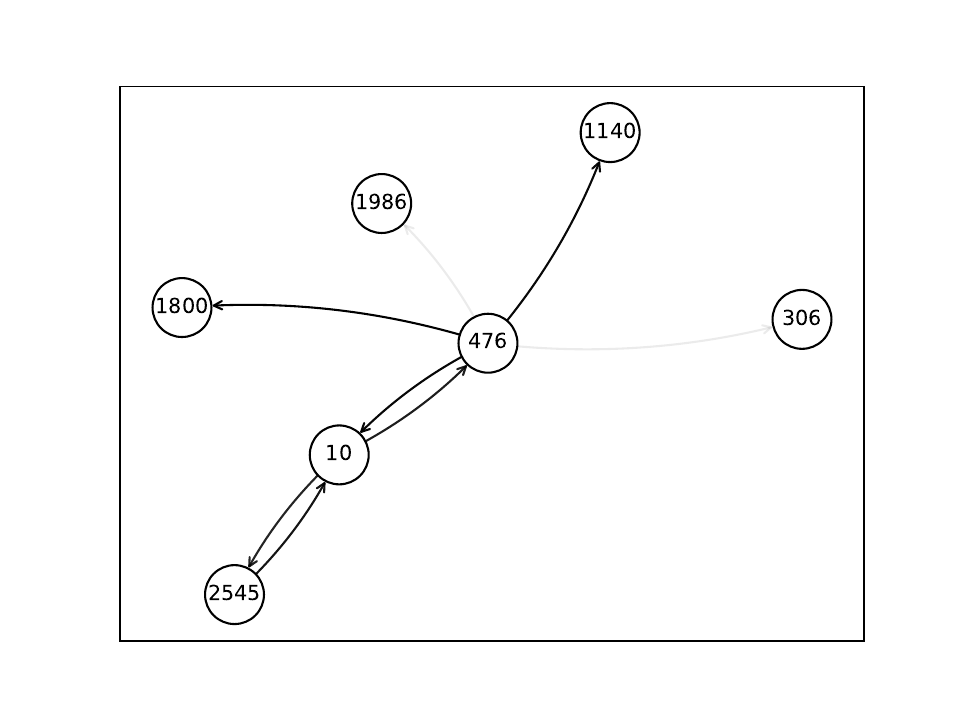}
    \caption{Subgraph Visualization Highlighting Influential Edges}
    \label{fig:subgraph}
\end{figure}

\paragraph{Advantages of GNNExplainer}

\begin{itemize}
    \item \textbf{Model-agnostic:} GNNExplainer can be used with any GNN architecture, making it versatile for different graph learning tasks \cite{ying2019gnnexplainer}.
    \item \textbf{Local Explanations:} The technique provides explanations for individual predictions, helping users understand specific model decisions.
    \item \textbf{Interpretability:} By identifying important subgraphs and node features, GNNExplainer enhances the interpretability of GNNs, making them more transparent.
\end{itemize}

\paragraph{Limitations}

Despite its strengths, GNNExplainer has some limitations:
\begin{itemize}
    \item \textbf{Scalability Issues:} The optimization process can be computationally expensive, especially for large graphs with many nodes and edges \cite{yuan2020explainability}.
    \item \textbf{Local Explanations Only:} GNNExplainer provides local explanations for individual nodes or predictions, which may not generalize to the entire model behavior.
    \item \textbf{Challenges with High-dimensional Features:} For graphs with high-dimensional node features, the feature importance estimation may become less reliable.
\end{itemize}

\paragraph{GNNExplainer for Large Language Models}

In LLMs, GNNExplainer can be adapted to explain graph-based representations, such as knowledge graphs or dependency trees used as inputs for text-based tasks \cite{huang2022graph}. By analyzing the influence of specific nodes (e.g., entities or words) and edges (e.g., relationships or dependencies), GNNExplainer can help provide insights into how graph-based features impact the model's predictions. This adaptation enhances the explainability of LLMs in tasks involving graph-structured data, such as question answering and information extraction.

\subsection{Node Importance Attribution}

Node Importance Attribution is a technique designed to provide interpretability for Graph Neural Networks (GNNs) by quantifying the contribution of each node to the model's prediction \cite{baldassarre2019explainability}. The main goal is to assign an importance score to each node, indicating its influence on the decision-making process of the GNN. This technique is widely used in tasks like node classification, link prediction, and graph classification, helping users understand which nodes are critical for specific predictions.

\paragraph{Scope of Application}

Node Importance Attribution is suitable for:
\begin{itemize}
    \item \textbf{Graph Neural Networks:} This technique can be applied to various GNN architectures, such as Graph Convolutional Networks (GCNs) \cite{kipf2017semi}, Graph Attention Networks (GATs) \cite{velivckovic2018graph}, and GraphSAGE \cite{hamilton2017inductive}.
    \item \textbf{Machine Learning Models with Graph-structured Data:} It is useful in scenarios where graph-based data is present, such as social networks, molecular graphs, and knowledge graphs \cite{baldassarre2019explainability}.
    \item \textbf{Large Language Models (LLMs) with Graph Representations:} When LLMs utilize graph-structured inputs (e.g., dependency trees or knowledge graphs), node importance attribution can be adapted to interpret the influence of specific nodes (e.g., entities or words) on the model's output \cite{huang2022graph}.
\end{itemize}

\paragraph{Principles and Formula}

The core idea of Node Importance Attribution is to compute an importance score \( I(v) \) for each node \( v \) in the graph, indicating its contribution to the model's prediction. A common approach involves calculating the gradient of the model's output with respect to the node features or using attention scores in attention-based GNNs \cite{baldassarre2019explainability}.

In gradient-based methods, the importance score for node \( v \) can be defined as:

\[
I(v) = \left\|\nabla_{\mathbf{x}_v} f(\mathbf{x}, \mathbf{A})\right\|_2,
\]

where:
\begin{itemize}
    \item \( \mathbf{x}_v \) is the feature vector of node \( v \).
    \item \( \mathbf{A} \) is the adjacency matrix of the graph.
    \item \( f(\mathbf{x}, \mathbf{A}) \) is the GNN model's output.
    \item \( \nabla_{\mathbf{x}_v} \) denotes the gradient of the model's output with respect to \( \mathbf{x}_v \).
    \item \( \|\cdot\|_2 \) represents the L2 norm, capturing the magnitude of the gradient.
\end{itemize}

In attention-based GNNs, such as GATs, the node importance score can be directly derived from the learned attention coefficients \cite{velivckovic2018graph}:

\[
I(v) = \sum_{u \in \mathcal{N}(v)} \alpha_{vu},
\]

where:
\begin{itemize}
    \item \( \mathcal{N}(v) \) is the set of neighbors of node \( v \).
    \item \( \alpha_{vu} \) is the attention score between nodes \( v \) and \( u \).
\end{itemize}

\paragraph{Python Code Example}

In this example, we use PyTorch and PyTorch Geometric \cite{Fey/Lenssen/2019} to demonstrate Node Importance Attribution using a simple Graph Convolutional Network (GCN) on the Cora dataset. The goal is to calculate the importance of node features for a specific target node by leveraging the gradients of the model's output with respect to the input features.

\begin{lstlisting}[style=python, literate={\$}{{\$}}1]
import torch
import torch.nn.functional as F
from torch_geometric.nn import GCNConv
from torch_geometric.datasets import Planetoid

# Load the Cora dataset
dataset = Planetoid(root='/tmp/Cora', name='Cora')
data = dataset[0]

# Define a simple GCN model
class GCN(torch.nn.Module):
    def __init__(self):
        super(GCN, self).__init__()
        self.conv1 = GCNConv(dataset.num_node_features, 16)
        self.conv2 = GCNConv(16, dataset.num_classes)

    def forward(self, x, edge_index):
        x = F.relu(self.conv1(x, edge_index))
        x = self.conv2(x, edge_index)
        return F.log_softmax(x, dim=1)

# Initialize the model, optimizer, and loss function
model = GCN()
optimizer = torch.optim.Adam(model.parameters(), lr=0.01)
criterion = torch.nn.CrossEntropyLoss()

# Train the GCN model
model.train()
for epoch in range(200):
    optimizer.zero_grad()
    out = model(data.x, data.edge_index)
    loss = criterion(out[data.train_mask], data.y[data.train_mask])
    loss.backward()
    optimizer.step()

# Calculate node importance
model.eval()
data.x.requires_grad = True  # Enable gradient calculation for node features

target_node = 10
output = model(data.x, data.edge_index)

# Perform backward pass for the predicted class of the target node
predicted_class = output[target_node].argmax()
output[target_node, predicted_class].backward()

# Calculate the L2 norm of the gradient for the target node's features as the importance score
node_importance = torch.norm(data.x.grad[target_node], p=2).item()
print(f"Importance score for node {target_node}: {node_importance:.4f}")
\end{lstlisting}

\paragraph{Results Explanation}

Below is a sample output from the Python code:

\begin{itemize}
    \item Importance score for node 10: \(0.0083\)
\end{itemize}

In this example, a simple GCN model is trained to perform node classification on the Cora dataset. To evaluate the importance of a specific node's features, we calculate the gradient of the model's output with respect to the input features of the target node. The L2 norm of the gradient is used as the importance score, indicating the sensitivity of the model's prediction to the target node's features. A higher importance score suggests a greater influence of the node's features on the model's decision. This method provides valuable insights into the decision-making process of Graph Neural Networks (GNNs), helping users understand the role of individual nodes in the prediction.

\paragraph{Advantages of Node Importance Attribution}

\begin{itemize}
    \item \textbf{Local Explanations:} Node importance attribution provides local explanations for individual nodes, making it useful for understanding specific predictions \cite{baldassarre2019explainability}.
    \item \textbf{Model-agnostic:} The technique can be applied to various GNN architectures, including GCNs, GATs, and GraphSAGE.
    \item \textbf{Gradient-based Approach:} Using gradients allows for a principled way to measure the sensitivity of the model's output to changes in node features \cite{sundararajan2017axiomatic}.
\end{itemize}

\paragraph{Limitations}

Despite its strengths, Node Importance Attribution has certain limitations:
\begin{itemize}
    \item \textbf{Sensitivity to Feature Scaling:} The importance scores may be influenced by the scale of the node features, requiring careful normalization \cite{baldassarre2019explainability}.
    \item \textbf{Computational Cost:} Computing gradients for each node can be computationally expensive, especially in large graphs.
    \item \textbf{Lack of Global Context:} The technique focuses on individual nodes and may miss broader graph-level patterns.
\end{itemize}

\paragraph{Node Importance Attribution for Large Language Models}

In LLMs, node importance attribution can be adapted to interpret graph-based inputs, such as knowledge graphs or dependency trees used in NLP tasks \cite{huang2022graph}. By quantifying the influence of specific nodes (e.g., entities or words) on the model's output, this technique can help explain how certain features contribute to predictions. This approach enhances the interpretability of LLMs, especially in tasks involving structured or graph-based data.

\section{Multimodal Explainability}
Multimodal explainability aims to interpret models that handle data from multiple sources, such as text, images, audio, and video \cite{baltruvsaitis2018multimodal}. These models often leverage the complementary nature of different data types, making their explanations challenging yet crucial for understanding how features from various modalities interact and contribute to predictions \cite{li2019visualbert}. In this section, we discuss key techniques, including \textbf{Multimodal Explanations via Attention} \cite{hori2017attention}, which use attention mechanisms to highlight important features across modalities; \textbf{Joint Feature Attribution for Multimodal Models} \cite{sundararajan2017axiomatic}, which provides a unified explanation of feature importance; and \textbf{Cross-modal Explanation Analysis} \cite{lu2019vilbert}, focusing on the relationships and interactions between different data modalities.

\subsection{Multimodal Explanations via Attention}

Multimodal explanations via attention aim to provide interpretability for models that process multiple data modalities, such as text, image, and audio. By leveraging attention mechanisms, these models can focus on the most relevant parts of each modality, allowing us to interpret how different features from different data sources contribute to the final prediction \cite{bahdanau2014neural, Vaswani2017}. This technique is particularly useful in tasks like visual question answering (VQA) \cite{antol2015vqa}, image captioning \cite{xu2015show}, and multimodal sentiment analysis \cite{zadeh2017tensor}.

\paragraph{Scope of Application}

Multimodal explanations via attention are applicable to:
\begin{itemize}
    \item \textbf{Traditional Machine Learning Models:} When combined with attention mechanisms, traditional models can benefit from multimodal explanations, especially in tasks involving text and image data \cite{ngiam2011multimodal}.
    \item \textbf{Deep Learning Models:} Attention-based neural networks, such as Transformer models, are well-suited for multimodal data and can provide meaningful explanations \cite{Vaswani2017}.
    \item \textbf{Large Language Models (LLMs) with Multimodal Extensions:} LLMs like GPT-4 and PaLM-2 that include multimodal capabilities can utilize attention-based techniques to interpret the contributions of different input modalities \cite{alayrac2022flamingo}.
\end{itemize}

\paragraph{Principles and Formula}

The key idea behind multimodal explanations via attention is to learn attention weights that indicate the importance of different features from each modality \cite{lu2019vilbert}. Given an input consisting of multiple modalities \( x^{(1)}, x^{(2)}, \ldots, x^{(M)} \), the attention mechanism computes an attention score for each feature \( h_i^{(m)} \) from modality \( m \):

\[
\alpha_i^{(m)} = \frac{\exp(e_i^{(m)})}{\sum_{j=1}^{n} \exp(e_j^{(m)})},
\]

where:
\begin{itemize}
    \item \( e_i^{(m)} \) is the relevance score for the \(i\)-th feature of modality \(m\), computed as:
    \[
    e_i^{(m)} = \mathbf{w}^T \tanh(\mathbf{W}_h h_i^{(m)} + \mathbf{b}),
    \]
    where \( \mathbf{w} \), \( \mathbf{W}_h \), and \( \mathbf{b} \) are learnable parameters.
    \item \( \alpha_i^{(m)} \) is the attention weight, indicating the importance of the feature.
\end{itemize}

The attended feature representation for each modality is then computed as a weighted sum:

\[
\hat{h}^{(m)} = \sum_{i=1}^{n} \alpha_i^{(m)} h_i^{(m)}.
\]

The final prediction is based on the concatenated attended features from all modalities:

\[
\hat{y} = f(\hat{h}^{(1)} \oplus \hat{h}^{(2)} \oplus \ldots \oplus \hat{h}^{(M)}),
\]

where \( \oplus \) denotes concatenation and \( f \) is a prediction function (e.g., a neural network layer).

\paragraph{Python Code Example}

In this example, we use TensorFlow to demonstrate a simple multimodal attention model for a task that combines text and image data. The model utilizes separate attention mechanisms for each modality and merges the representations to make a final prediction.

\begin{lstlisting}[style=python, literate={\$}{{\$}}1]
import tensorflow as tf
from tensorflow.keras.layers import Dense, Attention, Concatenate, Lambda
from tensorflow.keras.models import Model
from tensorflow.keras.utils import plot_model

# Define text and image inputs
text_input = tf.keras.Input(shape=(100, 300), name='text_input')  # 100 tokens, 300-d embeddings
image_input = tf.keras.Input(shape=(49, 512), name='image_input')  # 7x7 image patches, 512-d features

# Text attention mechanism
text_query = Dense(64, activation='tanh')(text_input)
text_key = Dense(64, activation='tanh')(text_input)
text_value = Dense(64, activation='tanh')(text_input)
text_attention = Attention()([text_query, text_value, text_key])

# Reduce mean for text attention
text_representation = Lambda(lambda x: tf.reduce_mean(x, axis=1))(text_attention)

# Image attention mechanism
image_query = Dense(64, activation='tanh')(image_input)
image_key = Dense(64, activation='tanh')(image_input)
image_value = Dense(64, activation='tanh')(image_input)
image_attention = Attention()([image_query, image_value, image_key])

# Reduce mean for image attention
image_representation = Lambda(lambda x: tf.reduce_mean(x, axis=1))(image_attention)

# Concatenate text and image representations
combined_representation = Concatenate()([text_representation, image_representation])
output = Dense(1, activation='sigmoid')(combined_representation)

# Build and compile the model
model = Model(inputs=[text_input, image_input], outputs=output)
model.compile(optimizer='adam', loss='binary_crossentropy', metrics=['accuracy'])

# Print model summary
model.summary()

# Save the model plot as a PNG image
plot_model(model, to_file='model_plot.png', show_shapes=True, show_layer_names=True)
\end{lstlisting}

\paragraph{Results Explanation}

The model summary and architecture are as follows:

\begin{figure}[!ht]
    \centering
    \includegraphics[width=0.8\textwidth]{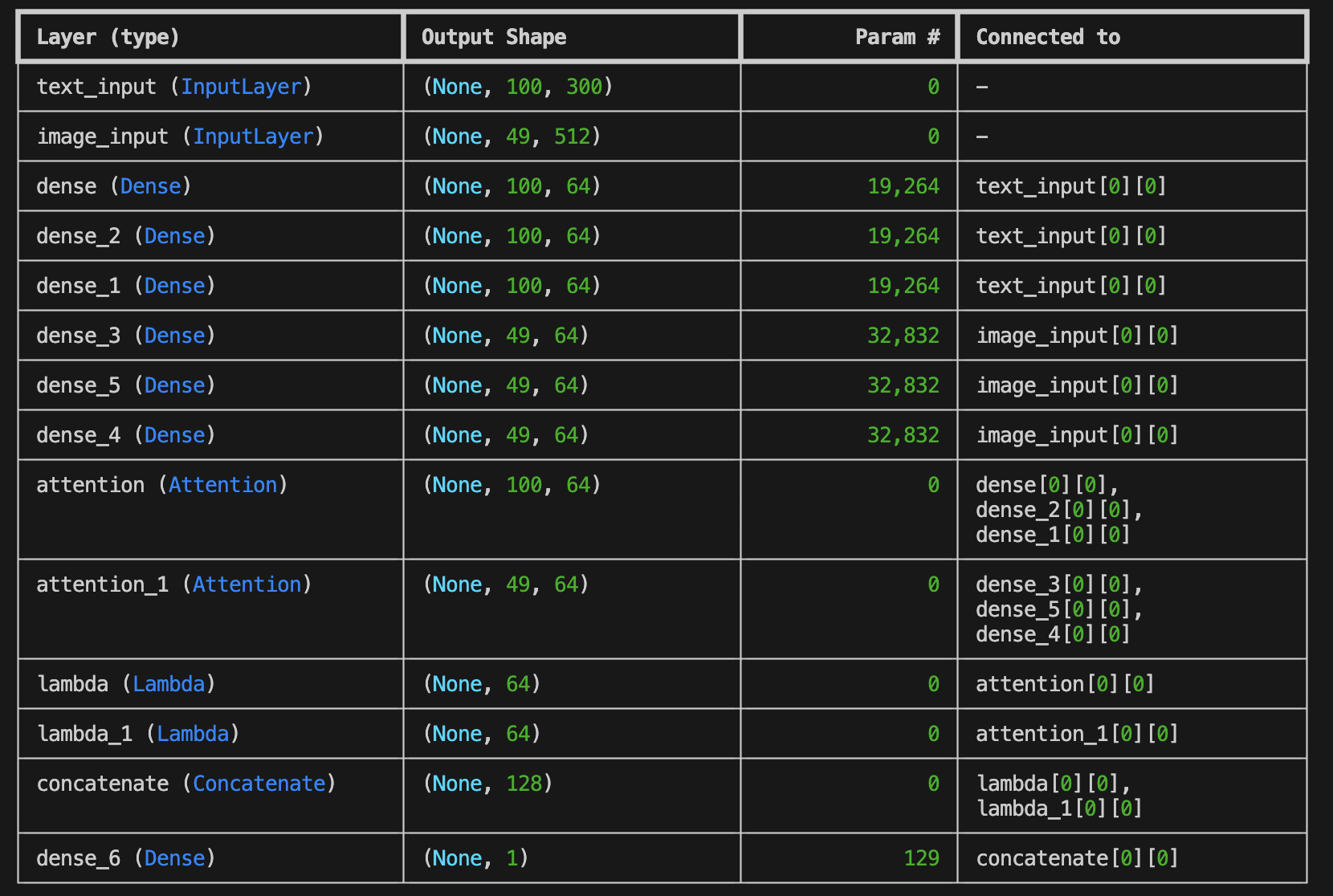}
    \caption{Example Model Summary}
    \label{fig:Modelsummary}
\end{figure}

\begin{lstlisting}[style=cmd]
Total params: 156,417 (611.00 KB)
Trainable params: 156,417 (611.00 KB)
Non-trainable params: 0 (0.00 B)
\end{lstlisting}

\begin{figure}[htbp]
    \centering
    \includegraphics[width=0.8\textwidth]{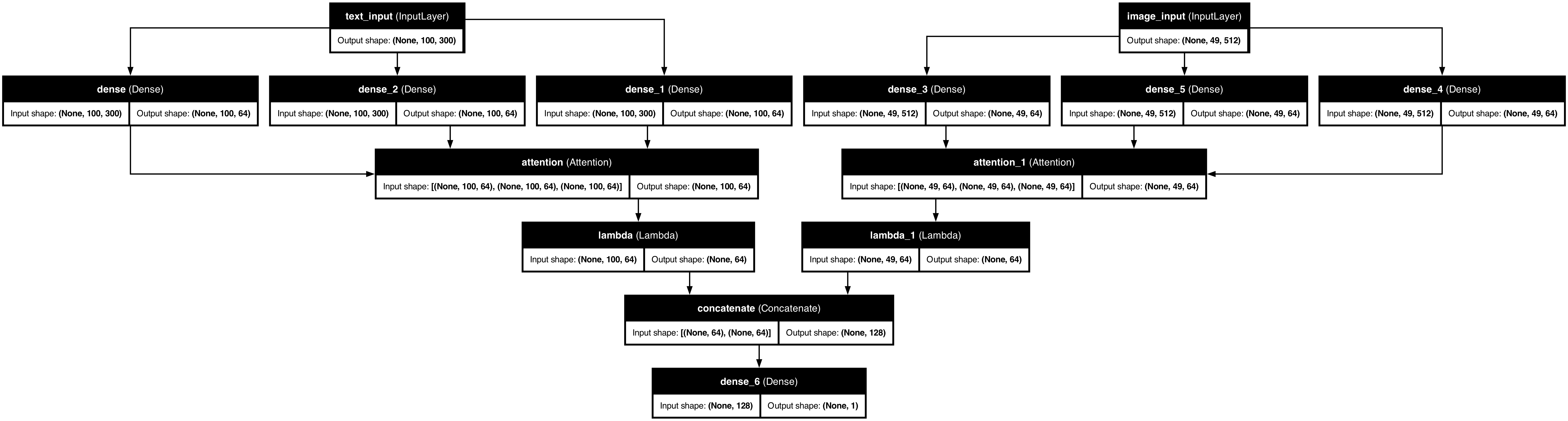}
    \caption{Example Model Structure}
    \label{fig:Model plot}
\end{figure}

In this example, the model uses separate attention mechanisms for text and image data, learning distinct representations for each modality. These representations are combined to form a unified multimodal representation, which is used to make a binary prediction. The model architecture provides interpretability by examining the learned attention weights, allowing us to understand the contribution of each modality to the final decision.

\paragraph{Advantages of Multimodal Explanations via Attention}

\begin{itemize}
    \item \textbf{Cross-modal Interpretability:} By analyzing attention weights, users can understand the importance of features from different modalities, enhancing model transparency \cite{vig2019multiscale}.
    \item \textbf{Flexibility:} The attention mechanism can be easily integrated into various neural network architectures, making it applicable to a wide range of tasks \cite{Vaswani2017}.
    \item \textbf{Improved Performance:} Attention mechanisms often lead to better model performance by focusing on the most relevant parts of the input data \cite{bahdanau2014neural}.
\end{itemize}

\paragraph{Limitations}

Despite its benefits, this technique has certain limitations:
\begin{itemize}
    \item \textbf{Computational Overhead:} Attention mechanisms can increase the computational complexity of the model, especially when processing high-dimensional data \cite{child2019generating}.
    \item \textbf{Interpretation Challenges:} While attention weights provide insights, they do not always guarantee causal explanations, as high attention scores do not necessarily imply a strong causal relationship \cite{jain2019attention}.
    \item \textbf{Dependence on Modality Quality:} The quality of the explanations depends on the quality of the input data from each modality; noisy or irrelevant features can lead to misleading interpretations \cite{wang2020makes}.
\end{itemize}

\paragraph{Multimodal Explanations for Large Language Models}

In LLMs with multimodal capabilities, such as GPT-4 with vision input, attention mechanisms are used to integrate information from both text and image inputs. By examining the learned attention patterns, we can gain insights into how the model combines visual and textual information to generate responses \cite{alayrac2022flamingo}. For example, in a visual question answering task, attention scores can help identify which parts of the image and text were most influential in determining the answer. This approach enhances the interpretability of LLMs in multimodal applications, making it easier to understand their decision-making process.

\subsection{Joint Feature Attribution for Multimodal Models}

Joint Feature Attribution for multimodal models is a technique that aims to provide comprehensive explanations by simultaneously assessing the contribution of features across different input modalities. Instead of treating each modality independently, this approach considers interactions between features from different sources (e.g., text, image, and audio) to generate a unified attribution score \cite{ancona2017towards}. This method is particularly useful in tasks where the interaction between modalities is crucial, such as visual question answering (VQA), image captioning, and multimodal sentiment analysis.

\paragraph{Scope of Application}

Joint Feature Attribution is applicable to:
\begin{itemize}
    \item \textbf{Traditional and Deep Learning Models:} This technique can be adapted for models that handle multiple input modalities, including early fusion models (concatenation of features) and late fusion models (separate feature processing followed by integration) \cite{ngiam2011multimodal}.
    \item \textbf{Graph Neural Networks (GNNs):} GNNs that incorporate multimodal graph data, such as knowledge graphs augmented with visual or textual features, can benefit from joint feature attribution to interpret node and edge importance across modalities \cite{wang2020multi}.
    \item \textbf{Large Language Models (LLMs) with Multimodal Capabilities:} LLMs that integrate visual and textual data (e.g., GPT-4 with vision input) can use joint feature attribution to explain how different features from text and image contribute jointly to the output \cite{alayrac2022flamingo}.
\end{itemize}

\paragraph{Principles and Formula}

The core idea behind joint feature attribution is to compute an attribution score for each feature while considering its interactions with features from other modalities \cite{sundararajan2017axiomatic}. Given a multimodal input \( \mathbf{x} = [\mathbf{x}^{(1)}, \mathbf{x}^{(2)}, \ldots, \mathbf{x}^{(M)}] \), where \( \mathbf{x}^{(m)} \) represents the feature vector from modality \( m \), the joint attribution score for a feature \( x_i^{(m)} \) can be defined as:

\[
\phi_i^{(m)} = \frac{\partial f(\mathbf{x})}{\partial x_i^{(m)}} \times x_i^{(m)}.
\]

Here:
\begin{itemize}
    \item \( \frac{\partial f(\mathbf{x})}{\partial x_i^{(m)}} \) is the partial derivative of the model's output \( f(\mathbf{x}) \) with respect to the feature \( x_i^{(m)} \).
    \item \( x_i^{(m)} \) is the value of the feature from modality \( m \).
    \item \( \phi_i^{(m)} \) is the attribution score, indicating the feature's contribution to the output while considering its interaction with other modalities.
\end{itemize}

To account for cross-modal interactions, we can extend this formula using Integrated Gradients, where the attribution score is computed along a path integral:

\[
\phi_i^{(m)} = \int_{\alpha=0}^{1} \frac{\partial f(\alpha \mathbf{x})}{\partial x_i^{(m)}} d\alpha.
\]

This approach captures the interaction effects between modalities by integrating the gradients along a straight line path from a baseline input (e.g., zero vector) to the actual input.

\paragraph{Python Code Example}

In this example, we implement a simple multimodal model using TensorFlow with text and image inputs. We use Integrated Gradients to compute joint feature attribution for both modalities.

\begin{lstlisting}[style=python, literate={\$}{{\$}}1]
import tensorflow as tf
from tensorflow.keras.layers import Dense, Concatenate
from tensorflow.keras.models import Model
import numpy as np

# Define text and image inputs
text_input = tf.keras.Input(shape=(300,), name='text_input')  # 300-d text embeddings
image_input = tf.keras.Input(shape=(512,), name='image_input')  # 512-d image features

# Define simple Dense layers for text and image features
text_features = Dense(128, activation='relu')(text_input)
image_features = Dense(128, activation='relu')(image_input)

# Concatenate text and image features
combined_features = Concatenate()([text_features, image_features])
output = Dense(1, activation='sigmoid')(combined_features)

# Build and compile the model
model = Model(inputs=[text_input, image_input], outputs=output)
model.compile(optimizer='adam', loss='binary_crossentropy', metrics=['accuracy'])

# Define Integrated Gradients function
def integrated_gradients(model, inputs, baseline, steps=50):
    alphas = np.linspace(0, 1, steps)

    # Interpolate for each input
    input_scaled_text = np.array([baseline[0] + alpha * (inputs[0] - baseline[0]) for alpha in alphas])
    input_scaled_image = np.array([baseline[1] + alpha * (inputs[1] - baseline[1]) for alpha in alphas])

    # Convert NumPy arrays to TensorFlow tensors
    input_scaled_text = tf.convert_to_tensor(input_scaled_text, dtype=tf.float32)
    input_scaled_image = tf.convert_to_tensor(input_scaled_image, dtype=tf.float32)

    # Compute gradients using GradientTape
    with tf.GradientTape() as tape:
        tape.watch([input_scaled_text, input_scaled_image])
        predictions = model([input_scaled_text, input_scaled_image])
    gradients = tape.gradient(predictions, [input_scaled_text, input_scaled_image])

    # Calculate average gradients and Integrated Gradients
    avg_gradients_text = tf.reduce_mean(gradients[0], axis=0).numpy()
    avg_gradients_image = tf.reduce_mean(gradients[1], axis=0).numpy()

    integrated_gradients_text = (inputs[0] - baseline[0]) * avg_gradients_text
    integrated_gradients_image = (inputs[1] - baseline[1]) * avg_gradients_image

    return integrated_gradients_text, integrated_gradients_image

# Example usage
text_sample = np.random.rand(300)
image_sample = np.random.rand(512)
baseline_text = np.zeros(300)
baseline_image = np.zeros(512)

inputs = [text_sample, image_sample]
baseline = [baseline_text, baseline_image]

# Compute Integrated Gradients
attributions_text, attributions_image = integrated_gradients(model, inputs, baseline)
print("Integrated Gradients for text features:", attributions_text)
print("Integrated Gradients for image features:", attributions_image)

# Visualize Integrated Gradients
import matplotlib.pyplot as plt

plt.figure(figsize=(12, 6))

# Plot for text features
plt.subplot(1, 2, 1)
plt.plot(attributions_text)
plt.title("Integrated Gradients for Text Features")
plt.xlabel("Feature Index")
plt.ylabel("Attribution")

# Plot for image features
plt.subplot(1, 2, 2)
plt.plot(attributions_image)
plt.title("Integrated Gradients for Image Features")
plt.xlabel("Feature Index")
plt.ylabel("Attribution")

plt.tight_layout()
plt.show()
\end{lstlisting}

\paragraph{Results Explanation}

In this example, we compute joint feature attribution for a simple multimodal model using Integrated Gradients \cite{sundararajan2017axiomatic}. The attribution scores indicate the contribution of each text and image feature to the model's output, allowing us to interpret the combined effects of both modalities. By analyzing these scores, we can identify which features from each modality are most influential in the model's decision-making process.

\begin{figure}[htbp]
    \centering
    \includegraphics[width=0.8\textwidth]{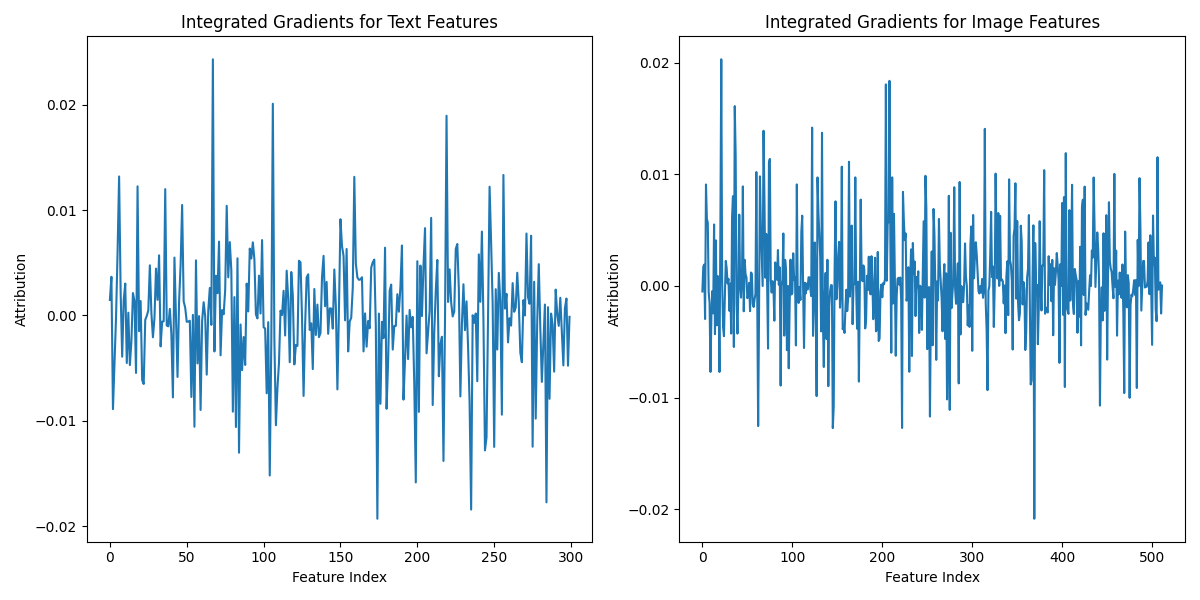}
    \caption{Example Model Structure}
    \label{fig:Feature Attribution}
\end{figure}

The left plot shows the attribution scores for text features, while the right plot shows the scores for image features. Peaks in the attribution scores highlight the features that significantly impact the model's output, providing insights into the model's interpretability.

\paragraph{Advantages of Joint Feature Attribution}

\begin{itemize}
    \item \textbf{Comprehensive Interpretability:} This method considers cross-modal interactions, providing a more complete explanation than single-modal attribution techniques \cite{ancona2017towards}.
    \item \textbf{Model-agnostic:} Joint feature attribution can be applied to a wide range of multimodal models, including deep learning and LLMs \cite{alayrac2022flamingo}.
    \item \textbf{Captures Interaction Effects:} The approach highlights not only the importance of individual features but also their interactions across modalities \cite{tsai2019multimodal}.
\end{itemize}

\paragraph{Limitations}

Despite its strengths, joint feature attribution has certain limitations:
\begin{itemize}
    \item \textbf{Computational Complexity:} Calculating attribution scores, especially with Integrated Gradients, can be computationally expensive \cite{sundararajan2017axiomatic}.
    \item \textbf{Choice of Baseline:} The choice of baseline input can significantly affect the attribution results, making interpretation challenging \cite{sturmfels2020visualizing}.
    \item \textbf{Scalability Issues:} For high-dimensional multimodal data, the computation of attribution scores may not scale well \cite{ancona2017towards}.
\end{itemize}

\paragraph{Joint Feature Attribution for Large Language Models}

In LLMs with multimodal capabilities, such as GPT-4 with vision, joint feature attribution can be adapted to explain predictions based on both text and image inputs \cite{alayrac2022flamingo}. By analyzing the attribution scores for different words and image regions, we can gain insights into how the model integrates information from both modalities. This enhances the explainability of LLMs in tasks like visual question answering and multimodal dialogue systems, providing a deeper understanding of the model's decision-making process.

\subsection{Cross-modal Explanation Analysis}

Cross-modal Explanation Analysis focuses on understanding how different modalities (e.g., text, image, audio) interact in a multimodal model. The key idea is to provide explanations that highlight the contributions of each modality and analyze how information from different sources influences the model's predictions \cite{tan2019lxmert}. This technique is particularly relevant for tasks where inputs from multiple modalities are integrated, such as visual question answering (VQA), image captioning, and multimodal dialogue systems.

\paragraph{Scope of Application}

Cross-modal Explanation Analysis is applicable to:
\begin{itemize}
    \item \textbf{Multimodal Deep Learning Models:} This technique can be used with models that process inputs from multiple modalities, such as Transformer-based models (e.g., ViLT, CLIP) and attention-based neural networks \cite{kim2021vilt}.
    \item \textbf{Graph Neural Networks (GNNs):} In cases where multimodal data is represented as graphs, cross-modal analysis can help interpret the interactions between nodes from different modalities \cite{hu2020heterogeneous}.
    \item \textbf{Large Language Models (LLMs) with Multimodal Extensions:} LLMs like GPT-4 with vision capabilities can utilize cross-modal explanation analysis to interpret predictions based on both visual and textual inputs \cite{alayrac2022flamingo}.
\end{itemize}

\paragraph{Principles and Formula}

The goal of cross-modal explanation analysis is to attribute the model's output to features from different modalities, as well as to quantify the interactions between these modalities \cite{lu2019vilbert}. This can be achieved using attention mechanisms or gradient-based techniques. In attention-based models, the cross-modal attention weights indicate the importance of features from one modality when processing another modality.

Given a multimodal input \( \mathbf{x}^{(1)} \) (text) and \( \mathbf{x}^{(2)} \) (image), the cross-modal attention score \( \alpha_{ij} \) for text feature \( i \) and image feature \( j \) is computed as:

\[
\alpha_{ij} = \frac{\exp(e_{ij})}{\sum_{k} \exp(e_{ik})},
\]

where:
\begin{itemize}
    \item \( e_{ij} = \mathbf{w}^T \tanh(\mathbf{W}_1 \mathbf{x}^{(1)}_i + \mathbf{W}_2 \mathbf{x}^{(2)}_j + \mathbf{b}) \) is the alignment score between the text feature \( i \) and image feature \( j \).
    \item \( \mathbf{W}_1 \), \( \mathbf{W}_2 \), and \( \mathbf{b} \) are learnable parameters.
    \item \( \alpha_{ij} \) is the normalized attention score, indicating the influence of image feature \( j \) on the text feature \( i \).
\end{itemize}

The cross-modal representation for text features can be obtained by aggregating the image features using the attention scores:

\[
\hat{\mathbf{x}}^{(1)}_i = \sum_{j} \alpha_{ij} \mathbf{x}^{(2)}_j.
\]

This representation highlights the interaction between text and image features, enabling us to analyze which parts of the image contribute to the understanding of each word in the text.

\paragraph{Python Code Example}

In this example, we use TensorFlow to implement a cross-modal attention mechanism for a model that processes both text and image inputs. The model uses self-attention mechanisms on the text and image features separately, followed by a cross-modal attention mechanism that combines the information from both modalities.

\begin{lstlisting}[style=python, literate={\$}{{\$}}1]
import tensorflow as tf
from tensorflow.keras.layers import Dense, MultiHeadAttention, Concatenate, GlobalAveragePooling1D, Lambda
from tensorflow.keras.models import Model
import numpy as np
import matplotlib.pyplot as plt

# Set the model dimension
d_model = 64

# Define text and image input layers
text_input = tf.keras.Input(shape=(100, 300), name='text_input')  # 100 tokens, 300-d embeddings
image_input = tf.keras.Input(shape=(49, 512), name='image_input')  # 7x7 image patches, 512-d features

# Project inputs to a common dimension
text_features = Dense(d_model)(text_input)    # Shape: (batch_size, 100, 64)
image_features = Dense(d_model)(image_input)  # Shape: (batch_size, 49, 64)

# Self-attention on text features
text_self_attention = MultiHeadAttention(num_heads=8, key_dim=d_model)
text_attention_output = text_self_attention(
    query=text_features, value=text_features, key=text_features
)  # Shape: (batch_size, 100, 64)

# Self-attention on image features
image_self_attention = MultiHeadAttention(num_heads=8, key_dim=d_model)
image_attention_output = image_self_attention(
    query=image_features, value=image_features, key=image_features
)  # Shape: (batch_size, 49, 64)

# Cross-modal attention from text to image features
cross_modal_attention = MultiHeadAttention(num_heads=8, key_dim=d_model)
cross_attention_output, cross_attention_scores = cross_modal_attention(
    query=text_attention_output,
    value=image_attention_output,
    key=image_attention_output,
    return_attention_scores=True
)  # cross_attention_output shape: (batch_size, 100, 64)
# cross_attention_scores shape: (batch_size, 8, 100, 49)

# Average the attention scores over the heads using a Lambda layer
average_attention_scores = Lambda(lambda x: tf.reduce_mean(x, axis=1))(cross_attention_scores)
# Shape: (batch_size, 100, 49)

# Pooling over the sequence length to get fixed-size representations
text_representation = GlobalAveragePooling1D()(text_attention_output)          # Shape: (batch_size, 64)
image_representation = GlobalAveragePooling1D()(image_attention_output)        # Shape: (batch_size, 64)
cross_attention_representation = GlobalAveragePooling1D()(cross_attention_output)  # Shape: (batch_size, 64)

# Combine the representations
combined_representation = Concatenate()([
    text_representation,
    image_representation,
    cross_attention_representation
])  # Shape: (batch_size, 192)

# Output layer
output = Dense(1, activation='sigmoid')(combined_representation)

# Build and compile the model
model = Model(inputs=[text_input, image_input], outputs=output)
model.compile(optimizer='adam', loss='binary_crossentropy', metrics=['accuracy'])

# Create sample input data
text_sample = np.random.rand(1, 100, 300).astype(np.float32)
image_sample = np.random.rand(1, 49, 512).astype(np.float32)

# Perform model prediction
prediction = model.predict([text_sample, image_sample])
print("Prediction:", prediction)

# Build a model to output the attention scores for visualization
attention_model = Model(inputs=[text_input, image_input], outputs=average_attention_scores)

# Get the attention scores
attention_scores = attention_model.predict([text_sample, image_sample])  # Shape: (1, 100, 49)

# Visualize the cross-modal attention weights
plt.figure(figsize=(12, 8))
plt.imshow(attention_scores[0], cmap='viridis', aspect='auto')
plt.colorbar()
plt.title("Cross-modal Attention Weights (Text to Image Features)")
plt.xlabel("Image Feature Index (49 patches)")
plt.ylabel("Text Token Index (100 tokens)")
plt.show()
\end{lstlisting}

\paragraph{Results Explanation}

In this example, we implemented a cross-modal attention mechanism where the image features are used to inform the processing of text features \cite{tan2019lxmert}. The attention scores highlight which image regions are most relevant for interpreting each word in the text input. By analyzing these scores, we can understand how the model combines information from both modalities and which features contribute the most to the prediction.

\begin{figure}[htbp]
    \centering
    \includegraphics[width=0.8\textwidth]{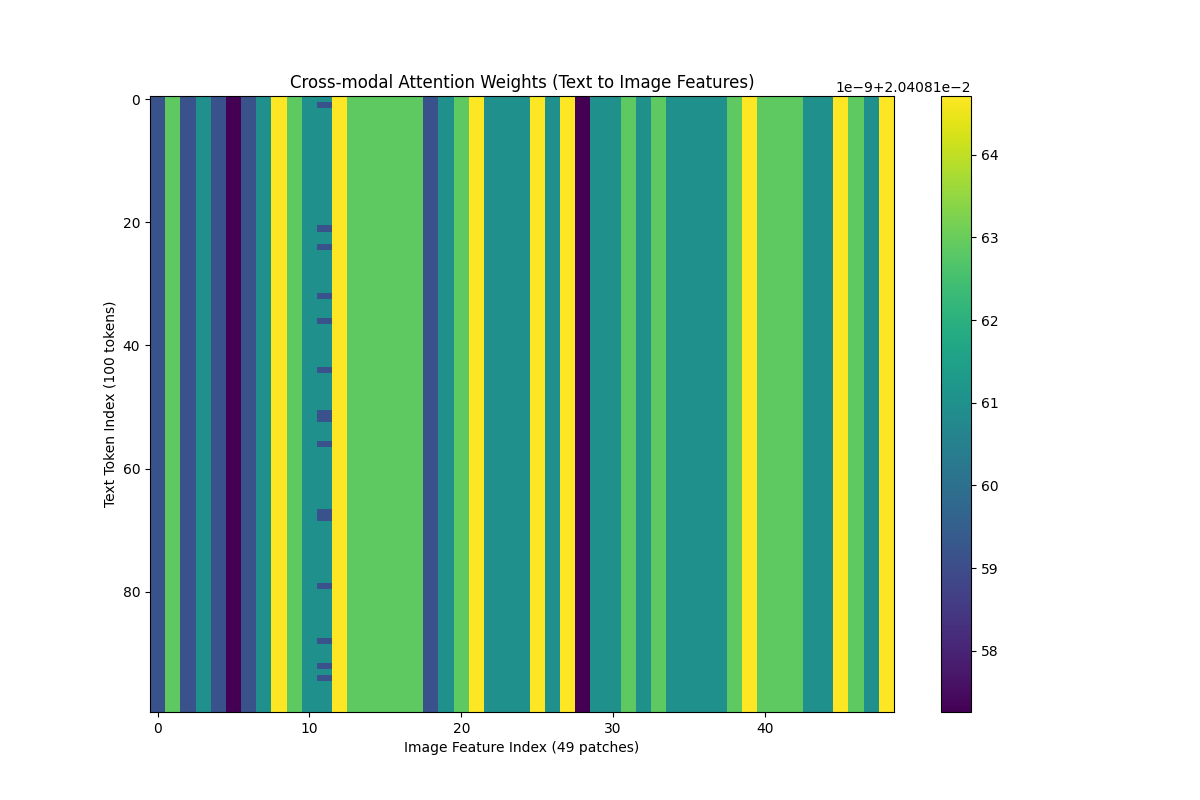}
    \caption{Cross-modal Attention Weights (Text to Image Features)}    \label{fig:Model plot2}
\end{figure}

The heatmap in Figure 1 shows the cross-modal attention weights, with the text token indices on the y-axis and the image feature indices (patches) on the x-axis. Brighter colors indicate higher attention scores, suggesting stronger relevance of the corresponding image features to the text tokens. This visualization helps interpret the alignment between text and image modalities, providing insights into the model's decision-making process.

\paragraph{Advantages of Cross-modal Explanation Analysis}

\begin{itemize}
    \item \textbf{Comprehensive Interpretability:} This technique provides a unified view of feature importance across different modalities, allowing for deeper insights into the model's decision-making process \cite{lu2019vilbert}.
    \item \textbf{Flexibility:} Cross-modal explanation analysis can be applied to various types of models and tasks, including VQA, image captioning, and multimodal classification \cite{kim2021vilt}.
    \item \textbf{Enhanced Performance:} By explicitly modeling cross-modal interactions, the model can learn more robust representations, potentially improving prediction accuracy \cite{tan2019lxmert}.
\end{itemize}

\paragraph{Limitations}

Despite its strengths, cross-modal explanation analysis has certain limitations:
\begin{itemize}
    \item \textbf{Complexity in Interpretation:} The interactions between different modalities can be complex, making it challenging to interpret the importance scores directly \cite{jain2019attention}.
    \item \textbf{Computational Overhead:} Attention mechanisms for cross-modal interactions can increase the computational cost, especially for high-dimensional data \cite{child2019generating}.
    \item \textbf{Dependence on Data Quality:} The quality of the explanations depends heavily on the quality of the input data from each modality; noisy data may lead to misleading interpretations \cite{wang2020makes}.
\end{itemize}

\paragraph{Cross-modal Explanation Analysis for Large Language Models}

For LLMs with multimodal capabilities, cross-modal explanation analysis can help interpret how the model processes visual and textual inputs together \cite{alayrac2022flamingo}. By examining the attention patterns between different modalities, we can gain insights into how the model integrates and prioritizes information from text and image inputs. This approach is particularly useful for tasks like visual question answering and multimodal dialogue, where the interaction between text and image data plays a crucial role in determining the output.

\section{Robustness and Fairness in Explanations}

In explainable AI (XAI), robustness and fairness are critical to ensuring that model explanations are both reliable and ethically sound \cite{doshi2017towards}. Robustness focuses on the stability of explanations, providing consistent insights even when the input data or model undergoes minor changes \cite{kindermans2019unreliability}, while fairness aims to eliminate bias in explanations, preventing the propagation of disparities, especially in sensitive applications such as healthcare and finance \cite{mehrabi2019survey}. This section introduces key methods, including \textbf{Fairness-aware Explanation Methods}, which offer unbiased interpretations \cite{dwork2012fairness}; \textbf{Robustness Testing}, which assesses explanation stability under various perturbations \cite{ghorbani2019interpretation}; \textbf{Bias Detection and Mitigation}, aimed at identifying and reducing unfairness \cite{hardt2016equality}; and \textbf{Consistency and Stability Analysis}, ensuring uniformity in feature attributions across similar data points \cite{adebayo2018sanity}. Additionally, we discuss recent advancements such as \textbf{Adversarial Robustness Testing} \cite{papernot2016distillation}, \textbf{Fairness-enhanced SHAP} \cite{Lundberg2017}, \textbf{Invariant Explanation Testing} \cite{alvarez2018towards}, \textbf{Causal Fairness Explanations} \cite{kusner2017counterfactual}, and \textbf{Stability-aware Feature Attribution} \cite{he2019sensitivity}, reflecting the latest progress in addressing these crucial challenges.

\subsection{Fairness-aware Explanation Methods}

Fairness-aware explanation methods are designed to ensure that the explanations provided by machine learning models do not propagate or amplify biases present in the data \cite{kamiran2012data}. The goal is to offer interpretations that are not only accurate but also equitable across different demographic groups or sensitive attributes (e.g., race, gender, age). These methods aim to address the challenge of fairness in model explanations, especially in high-stakes domains like healthcare, finance, and criminal justice \cite{barocas2017fairness}.

\paragraph{Scope of Application}

Fairness-aware explanation methods are suitable for:
\begin{itemize}
    \item \textbf{Traditional Machine Learning Models:} Methods like decision trees, logistic regression, and support vector machines can be adapted to provide fairness-aware explanations by incorporating fairness constraints \cite{kamiran2012data}.
    \item \textbf{Deep Learning Models:} Neural networks, including convolutional and recurrent networks, can utilize fairness-aware techniques to provide equitable feature attributions \cite{shrikumar2017learning}.
    \item \textbf{Large Language Models (LLMs):} Fairness-aware explanations are particularly crucial for LLMs like GPT-4, as these models often deal with sensitive text data that may contain inherent biases \cite{sheng2019woman}.
\end{itemize}

\paragraph{Principles and Formula}

The core principle of fairness-aware explanations is to provide feature attributions that are consistent across different subgroups defined by sensitive attributes \cite{dwork2012fairness}. One approach to achieve this is through \textbf{conditional feature attribution}, where the importance of each feature is assessed separately for each subgroup.

Let \( f(\mathbf{x}) \) be the model prediction for input \( \mathbf{x} \), and \( S \) be a sensitive attribute (e.g., gender). The fairness-aware attribution score \( \phi_i^{(S)} \) for feature \( i \) can be computed as:

\[
\phi_i^{(S)} = \mathbb{E}\left[\frac{\partial f(\mathbf{x})}{\partial x_i} \ \middle|\ S = s\right],
\]

where:

\begin{itemize}
    \item \( \frac{\partial f(\mathbf{x})}{\partial x_i} \) is the gradient of the model's prediction with respect to feature \( x_i \).
    \item \( S = s \) denotes the condition on a specific subgroup (e.g., \( S = \text{"female"} \)).
    \item \( \mathbb{E}[\cdot | S = s] \) is the expected value conditioned on the subgroup.
\end{itemize}

To measure the fairness of the attributions, we can compute the \textbf{Attribution Disparity (AD)}:

\[
\text{AD}_i = \left|\phi_i^{(S = s_1)} - \phi_i^{(S = s_2)}\right|,
\]

where \( s_1 \) and \( s_2 \) are different values of the sensitive attribute. A low Attribution Disparity indicates that the feature importance is consistent across subgroups, suggesting fairer explanations \cite{zemel2013learning}.

\paragraph{Python Code Example}

In this example, we use TensorFlow to implement a simple fairness-aware explanation method for a neural network trained on a dataset containing a sensitive attribute (e.g., gender). We aim to demonstrate how the gradient-based attributions of the model differ across subgroups (e.g., male and female). The method highlights potential disparities in feature importance that might indicate bias in the model's predictions.

\begin{lstlisting}[style=python, literate={\$}{{\$}}1]
import tensorflow as tf
import numpy as np

# Define a simple neural network model
model = tf.keras.Sequential([
    tf.keras.layers.Dense(16, activation='relu', input_shape=(10,)),
    tf.keras.layers.Dense(1, activation='sigmoid')
])
model.compile(optimizer='adam', loss='binary_crossentropy')

# Generate synthetic data
np.random.seed(0)
X = np.random.rand(1000, 10)
y = (X[:, 0] + X[:, 1] > 1).astype(int)  # Binary target
sensitive_attribute = np.random.choice([0, 1], size=(1000,))  # Gender (0 = male, 1 = female)

# Train the model
model.fit(X, y, epochs=10, batch_size=32)

# Function to compute fairness-aware attributions
def compute_attributions(model, X, sensitive_attr):
    gradients_male = []
    gradients_female = []

    for i, x in enumerate(X):
        # Convert input to tf.Tensor
        x_tensor = tf.convert_to_tensor(x, dtype=tf.float32)

        with tf.GradientTape() as tape:
            tape.watch(x_tensor)
            prediction = model(tf.expand_dims(x_tensor, axis=0))
        gradient = tape.gradient(prediction, x_tensor).numpy()

        if sensitive_attr[i] == 0:  # Male
            gradients_male.append(gradient)
        else:  # Female
            gradients_female.append(gradient)

    avg_grad_male = np.mean(gradients_male, axis=0)
    avg_grad_female = np.mean(gradients_female, axis=0)
    attribution_disparity = np.abs(avg_grad_male - avg_grad_female)

    return avg_grad_male, avg_grad_female, attribution_disparity

# Compute attributions and disparity
avg_grad_male, avg_grad_female, attribution_disparity = compute_attributions(model, X, sensitive_attribute)

print("Average gradient attributions (male):", avg_grad_male)
print("Average gradient attributions (female):", avg_grad_female)
print("Attribution disparity:", attribution_disparity)

import matplotlib.pyplot as plt

# Visualization of attribution disparity
plt.bar(range(10), attribution_disparity)
plt.xlabel('Feature Index')
plt.ylabel('Attribution Disparity')
plt.title('Feature Attribution Disparity between Male and Female')
plt.show()
\end{lstlisting}

\paragraph{Result Explanation}

The computed results indicate the average gradient-based feature attributions for male and female subgroups, as well as the disparity between them:

\begin{lstlisting}[style=cmd]
Average gradient attributions (male): [ 0.3898788   0.35155958 -0.1404713  -0.0839864   0.03294945 -0.154666 -0.02074329 -0.027875   -0.07047074 -0.0292792 ]
Average gradient attributions (female): [ 0.3966816   0.3579618  -0.14321546 -0.08344802  0.03245877 -0.15875477 -0.02232523 -0.02654943 -0.06922997 -0.03068474]
Attribution disparity: [0.0068028  0.00640222 0.00274417 0.00053838 0.00049068 0.00408876 0.00158194 0.00132557 0.00124077 0.00140554]
\end{lstlisting}

The results show that there are minor but noticeable differences in the gradient-based attributions between male and female subgroups. The \textit{Attribution Disparity} values highlight these differences across features. If the disparity values are significantly high, this indicates that the model's explanations may be influenced by the sensitive attribute (gender), suggesting potential bias in how the model utilizes different features for different groups \cite{hardt2016equality}.

\begin{figure}[!ht]
    \centering
    \includegraphics[width=0.8\textwidth]{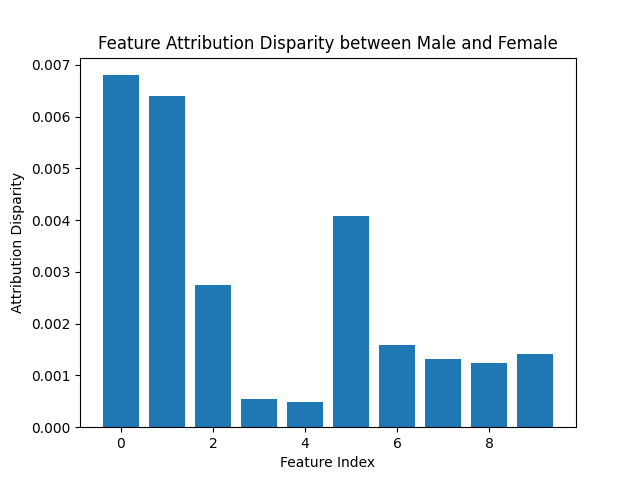}
    \caption{Feature Attribution Disparity between Male and Female. The bar plot shows the differences in feature importance attributed by the model for male and female subgroups. High bars indicate higher disparity, pointing to potential biases in the model's feature utilization.}
    \label{fig:attribution_disparity}
\end{figure}

\paragraph{Advantages of Fairness-aware Explanation Methods}

\begin{itemize}
    \item \textbf{Bias Mitigation:} By explicitly considering sensitive attributes, these methods help identify and mitigate biases in the model's explanations \cite{kusner2017counterfactual}.
    \item \textbf{Enhanced Transparency:} Fairness-aware explanations provide a more comprehensive understanding of the model's decision-making process, especially for underrepresented groups \cite{barocas2017fairness}.
    \item \textbf{Model-agnostic:} These techniques can be applied to a wide range of models, from traditional machine learning models to deep neural networks and LLMs \cite{adadi2018peeking}.
\end{itemize}

\paragraph{Limitations}

Despite their benefits, fairness-aware explanation methods have certain limitations:

\begin{itemize}
    \item \textbf{Dependence on Sensitive Attribute Availability:} The method requires the availability of sensitive attribute labels, which may not always be accessible \cite{dwork2012fairness}.
    \item \textbf{Trade-off with Model Performance:} Efforts to reduce attribution disparity may sometimes conflict with optimizing overall model accuracy \cite{zhao2019inherent}.
    \item \textbf{Complexity in High-dimensional Data:} For models with high-dimensional inputs (e.g., text or image data), computing and interpreting attributions across subgroups can be challenging \cite{ghorbani2019interpretation}.
\end{itemize}

\paragraph{Fairness-aware Explanations for Large Language Models}

In LLMs, fairness-aware explanation methods can be adapted to analyze text-based inputs where sensitive attributes (e.g., gendered language or racial indicators) may influence the model's predictions. By comparing feature attributions across different subgroups, we can identify biases in the model's responses and develop strategies to mitigate them, enhancing the fairness of the model's explanations in sensitive contexts \cite{sheng2019woman}.

\subsection{Robustness Testing for Explanations}

Robustness testing for explanations focuses on evaluating the stability and reliability of explanation methods under varying input conditions \cite{kindermans2019unreliability}. In the context of explainable AI, an explanation is considered robust if small changes or perturbations to the input do not cause significant fluctuations in the explanation \cite{ghorbani2019interpretation}. This testing is crucial for ensuring the trustworthiness of interpretability techniques, particularly in high-stakes applications like healthcare and finance \cite{adebayo2018sanity}.

\paragraph{Scope of Application}

Robustness testing for explanations is relevant for:

\begin{itemize}
    \item \textbf{Traditional Machine Learning Models:} Techniques like feature importance scores and decision rules can be tested for robustness using input perturbations \cite{molnar2020interpretable}.
    \item \textbf{Deep Learning Models:} Neural networks, including convolutional neural networks (CNNs) and recurrent neural networks (RNNs), often utilize gradient-based explanations or saliency maps that require robustness evaluation \cite{adebayo2018sanity}.
    \item \textbf{Large Language Models (LLMs):} Robustness testing is particularly important for LLMs, as small changes in the text input (e.g., paraphrasing or typos) can lead to significant shifts in the model's predictions and explanations \cite{prabhakaran2019perturbation}.
\end{itemize}

\paragraph{Principles and Formula}

To evaluate robustness, we assess the \textbf{Explanation Sensitivity} to input perturbations. Given a model \( f \) and an explanation method \( E \), let \( \mathbf{x} \) be the original input and \( \mathbf{x}' \) be a perturbed version of \( \mathbf{x} \). The robustness score \( R \) of the explanation is defined as:

\[
R = 1 - \frac{\| E(\mathbf{x}) - E(\mathbf{x'}) \|_2}{\| \mathbf{x} - \mathbf{x'} \|_2},
\]

where:

\begin{itemize}
    \item \( E(\mathbf{x}) \) is the explanation for the original input.
    \item \( E(\mathbf{x'}) \) is the explanation for the perturbed input.
    \item \( \| \cdot \|_2 \) denotes the L2 norm.
\end{itemize}

A higher robustness score \( R \) indicates more stable explanations, with \( R = 1 \) representing perfect robustness (no change in the explanation despite input perturbation) \cite{he2019sensitivity}.

\paragraph{Python Code Example}

In this example, we use TensorFlow to evaluate the robustness of saliency maps for a simple convolutional neural network (CNN) on the MNIST dataset. The goal is to examine how stable the saliency maps are when the input image is slightly perturbed. If the saliency maps are consistent, it indicates that the explanation method is reliable and robust to small input changes \cite{adebayo2018sanity}.

\begin{lstlisting}[style=python, literate={\$}{{\$}}1]
import tensorflow as tf
import numpy as np
import matplotlib.pyplot as plt

# Load MNIST dataset
(x_train, y_train), (x_test, y_test) = tf.keras.datasets.mnist.load_data()
x_test = x_test.astype('float32') / 255.0
x_test = np.expand_dims(x_test, axis=-1)

# Define a simple CNN model
model = tf.keras.Sequential([
    tf.keras.layers.Conv2D(32, (3, 3), activation='relu', input_shape=(28, 28, 1)),
    tf.keras.layers.MaxPooling2D((2, 2)),
    tf.keras.layers.Flatten(),
    tf.keras.layers.Dense(10, activation='softmax')
])
model.compile(optimizer='adam', loss='sparse_categorical_crossentropy', metrics=['accuracy'])

# Train the model
model.fit(x_train, y_train, epochs=5, batch_size=64)

# Function to compute saliency map
def compute_saliency_map(model, input_image, label):
    input_image = tf.convert_to_tensor(input_image[np.newaxis, ...])
    with tf.GradientTape() as tape:
        tape.watch(input_image)
        predictions = model(input_image)
        loss = predictions[0, label]
    gradient = tape.gradient(loss, input_image)
    saliency_map = tf.reduce_max(tf.abs(gradient), axis=-1).numpy()[0]
    return saliency_map

# Test robustness of the saliency map
original_image = x_test[0]
perturbed_image = original_image + 0.1 * np.random.normal(size=original_image.shape)

# Compute saliency maps
saliency_original = compute_saliency_map(model, original_image, y_test[0])
saliency_perturbed = compute_saliency_map(model, perturbed_image, y_test[0])

# Compute robustness score
robustness_score = 1 - np.linalg.norm(saliency_original - saliency_perturbed) / np.linalg.norm(original_image - perturbed_image)
print(f"Robustness Score: {robustness_score:.4f}")

# Display the saliency maps
plt.figure(figsize=(10, 5))
plt.subplot(1, 2, 1)
plt.title("Original Saliency Map")
plt.imshow(saliency_original, cmap='hot')
plt.axis('off')

plt.subplot(1, 2, 2)
plt.title("Perturbed Saliency Map")
plt.imshow(saliency_perturbed, cmap='hot')
plt.axis('off')
plt.show()
\end{lstlisting}

\paragraph{Result Explanation}

In this example, we compute saliency maps for both the original and perturbed images. The robustness score quantifies the stability of the saliency maps. A high robustness score indicates that the saliency maps are similar despite input noise, suggesting that the explanation method is stable. Conversely, a low score suggests that the explanation is sensitive to small changes in the input, raising concerns about its reliability \cite{ghorbani2019interpretation}.

\paragraph{Visualization of Saliency Maps}

The following figure displays the saliency maps for both the original and perturbed images:

\begin{figure}[!ht]
    \centering
    \includegraphics[width=0.9\textwidth]{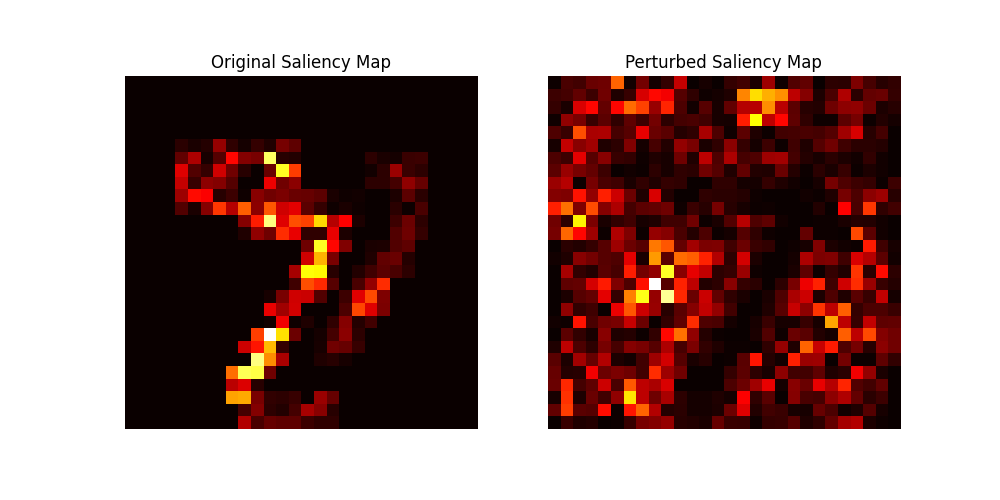}
    \caption{Comparison of Saliency Maps. The left image shows the original saliency map, while the right image shows the saliency map after applying a small perturbation to the input image. Differences between the two highlight the sensitivity of the saliency map to input noise.}
    \label{fig:saliency_comparison}
\end{figure}

\paragraph{Advantages of Robustness Testing for Explanations}

\begin{itemize}
    \item \textbf{Increased Trustworthiness:} By evaluating the stability of explanations, we can build more trust in the interpretability methods used \cite{kindermans2019unreliability}.
    \item \textbf{Improved Model Reliability:} Robust explanations help ensure that the model's decisions are not easily swayed by minor input perturbations, which is particularly important in adversarial settings \cite{papernot2016distillation}.
    \item \textbf{Applicable to Various Models:} The approach can be applied to different types of models, including traditional ML models, neural networks, and LLMs \cite{adadi2018peeking}.
\end{itemize}

\paragraph{Limitations}

Despite its benefits, robustness testing for explanations has certain limitations:

\begin{itemize}
    \item \textbf{Computational Cost:} Evaluating the robustness of explanations can be computationally intensive, especially for large models and high-dimensional data \cite{he2019sensitivity}.
    \item \textbf{Choice of Perturbation:} The effectiveness of the robustness test depends on the type and magnitude of input perturbations chosen, which may not always reflect real-world scenarios \cite{hendrycks2019benchmarking}.
    \item \textbf{Focus on Local Robustness:} The method primarily assesses local robustness (small perturbations) and may not capture global instabilities in the explanation method \cite{ghorbani2019interpretation}.
\end{itemize}

\paragraph{Robustness Testing for Explanations in Large Language Models}

For LLMs, robustness testing can be adapted to evaluate how stable the explanations are when the input text is slightly modified (e.g., through paraphrasing or the introduction of typos). By measuring the changes in explanation scores (e.g., attention weights or feature importances) across perturbed inputs, we can assess the reliability of the model's interpretability in handling natural language variations. This approach helps ensure that explanations provided by LLMs are not only accurate but also resilient to small input changes, enhancing their usability in real-world applications \cite{prabhakaran2019perturbation}.

\subsection{Consistency and Stability Analysis of Explanations}

Consistency and stability are essential qualities for trustworthy explanations. In the context of explainable AI, an explanation method is considered \textbf{consistent} if similar inputs yield similar explanations. It is deemed \textbf{stable} if small perturbations in the input do not lead to disproportionately large changes in the explanation. Analyzing the consistency and stability of explanations is crucial, especially in high-stakes applications, where reliable interpretations are needed to support decision-making \cite{alvarez2018towards}.

\paragraph{Scope of Application}

Consistency and stability analysis can be applied to:

\begin{itemize}
    \item \textbf{Traditional Machine Learning Models:} Linear models and decision trees often produce interpretable results, but consistency and stability analysis can reveal hidden instabilities in their explanations \cite{molnar2020interpretable}.
    \item \textbf{Deep Learning Models:} For neural networks, particularly those using gradient-based methods (e.g., Integrated Gradients, Saliency Maps), stability analysis can highlight the sensitivity of explanations to input noise \cite{kindermans2019unreliability}.
    \item \textbf{Large Language Models (LLMs):} In LLMs, consistency is vital when the model processes similar text inputs, ensuring that the generated explanations remain stable across minor text variations \cite{prabhakaran2019perturbation}.
\end{itemize}

\paragraph{Principles and Formula}

To quantify consistency, we define the \textbf{Explanation Consistency Score (ECS)}. Let \( \mathbf{x} \) and \( \mathbf{x'} \) be two similar inputs, and \( E(\mathbf{x}) \) and \( E(\mathbf{x'}) \) be their corresponding explanations. The ECS can be computed as the cosine similarity between the explanations:

\[
\text{ECS} = \frac{E(\mathbf{x}) \cdot E(\mathbf{x'})}{\|E(\mathbf{x})\|_2 \|E(\mathbf{x'})\|_2}.
\]

A high ECS indicates that the explanations are consistent across similar inputs.

For stability analysis, we define the \textbf{Explanation Stability Score (ESS)}. Given an input \( \mathbf{x} \) and its perturbed version \( \mathbf{x'} = \mathbf{x} + \delta \), where \( \delta \) is a small random noise, the ESS is calculated as:

\[
\text{ESS} = 1 - \frac{\| E(\mathbf{x}) - E(\mathbf{x'}) \|_2}{\| \mathbf{x} - \mathbf{x'} \|_2}.
\]

A high ESS suggests that the explanation is stable, even when the input is slightly modified \cite{alvarez2018towards}.

\paragraph{Python Code Example}

In this example, we use TensorFlow to evaluate the consistency and stability of gradient-based explanations for a convolutional neural network (CNN) trained on the MNIST dataset. The analysis focuses on two key metrics: the Explanation Consistency Score (ECS) and the Explanation Stability Score (ESS) \cite{adebayo2018sanity}.

\begin{lstlisting}[style=python, literate={\$}{{\$}}1]
import tensorflow as tf
import numpy as np

# Load MNIST dataset
(x_train, y_train), (x_test, y_test) = tf.keras.datasets.mnist.load_data()
x_test = x_test.astype('float32') / 255.0
x_test = np.expand_dims(x_test, axis=-1)

# Define a simple CNN model
model = tf.keras.Sequential([
    tf.keras.layers.Conv2D(32, (3, 3), activation='relu', input_shape=(28, 28, 1)),
    tf.keras.layers.MaxPooling2D((2, 2)),
    tf.keras.layers.Flatten(),
    tf.keras.layers.Dense(10, activation='softmax')
])
model.compile(optimizer='adam', loss='sparse_categorical_crossentropy')

# Train the model
model.fit(x_train, y_train, epochs=5, batch_size=64)

# Function to compute gradient-based explanations
def compute_gradients(model, input_image, label):
    input_image = tf.convert_to_tensor(input_image[np.newaxis, ...])
    with tf.GradientTape() as tape:
        tape.watch(input_image)
        predictions = model(input_image)
        loss = predictions[0, label]
    gradient = tape.gradient(loss, input_image).numpy()[0]
    return gradient

# Consistency Analysis
image1 = x_test[0]
image2 = x_test[1]  # A similar image
grad1 = compute_gradients(model, image1, y_test[0])
grad2 = compute_gradients(model, image2, y_test[1])
ecs = np.dot(grad1.flatten(), grad2.flatten()) / (np.linalg.norm(grad1) * np.linalg.norm(grad2))
print(f"Explanation Consistency Score (ECS): {ecs:.4f}")

# Stability Analysis
perturbed_image = image1 + 0.1 * np.random.normal(size=image1.shape)
grad_perturbed = compute_gradients(model, perturbed_image, y_test[0])
ess = 1 - np.linalg.norm(grad1 - grad_perturbed) / np.linalg.norm(image1 - perturbed_image)
print(f"Explanation Stability Score (ESS): {ess:.4f}")
\end{lstlisting}

\paragraph{Result Explanation}

The results of the analysis are presented below:

\begin{lstlisting}[style=cmd]
Explanation Consistency Score (ECS): 0.1036
Explanation Stability Score (ESS): 0.9851
\end{lstlisting}

The Explanation Consistency Score (ECS) measures the similarity of gradient-based explanations for two similar inputs. A low ECS value suggests that the explanations are not consistent for similar images, indicating potential issues with the interpretability of the model. Conversely, the Explanation Stability Score (ESS) quantifies the robustness of the explanations to input noise. A high ESS value (close to 1) indicates that the explanation is stable and reliable, enhancing the trustworthiness of the model's interpretations \cite{ghorbani2019interpretation}.

\paragraph{Advantages of Consistency and Stability Analysis}

\begin{itemize}
    \item \textbf{Enhanced Reliability:} Consistent and stable explanations increase the reliability of interpretability methods, making them more useful for decision-making \cite{molnar2020interpretable}.
    \item \textbf{Comprehensive Evaluation:} By analyzing both consistency and stability, we obtain a holistic assessment of the explanation method's performance \cite{alvarez2018towards}.
    \item \textbf{Applicability Across Models:} The analysis techniques can be applied to various types of models, from traditional ML models to deep neural networks and LLMs \cite{adadi2018peeking}.
\end{itemize}

\paragraph{Limitations}

Despite its benefits, consistency and stability analysis has certain limitations:

\begin{itemize}
    \item \textbf{Computational Overhead:} Evaluating consistency and stability can be computationally intensive, especially for large models or high-dimensional data \cite{he2019sensitivity}.
    \item \textbf{Choice of Perturbation:} The stability analysis depends on the choice of input perturbations, which may not always represent realistic variations \cite{hendrycks2019benchmarking}.
    \item \textbf{Local Assessment:} The analysis primarily focuses on local consistency and stability, and may not capture global instabilities in the model's explanations \cite{ghorbani2019interpretation}.
\end{itemize}
\paragraph{Consistency and Stability in Large Language Models}

For LLMs, consistency and stability analysis can be adapted to evaluate how explanations change when the input text is slightly modified (e.g., through paraphrasing). By measuring the changes in explanation scores (e.g., attention weights) across perturbed inputs, we can assess the reliability of the model's interpretability in handling natural language variations \cite{alvarez2018robustness}. This is particularly important for applications like sentiment analysis and question answering, where slight changes in phrasing should not lead to drastically different explanations \cite{sun2019robustness}.

\subsection{Adversarial Robustness Testing}

Adversarial robustness testing evaluates the stability of explanations when the input data is intentionally perturbed to mislead the model \cite{goodfellow2015explaining}. Adversarial attacks are designed to exploit vulnerabilities in the model by introducing small, often imperceptible, changes to the input. This form of testing is crucial for understanding whether explanation methods can reliably highlight relevant features even in the presence of adversarial perturbations \cite{papernot2016limitations}.

\paragraph{Scope of Application}

Adversarial robustness testing is relevant for:
\begin{itemize}
    \item \textbf{Traditional Machine Learning Models:} Techniques like decision trees and support vector machines can be vulnerable to adversarial samples, especially when used with gradient-based explanation methods \cite{biggio2013evasion}.
    \item \textbf{Deep Learning Models:} Neural networks, including CNNs and RNNs, are particularly susceptible to adversarial attacks, necessitating robustness testing for gradient-based explanations such as Integrated Gradients or Saliency Maps \cite{sundararajan2017axiomatic}.
    \item \textbf{Large Language Models (LLMs):} LLMs like GPT-4 can be affected by adversarial text inputs, where small modifications to the input text can cause significant changes in the model's predictions and explanations \cite{wallace2019universal}.
\end{itemize}

\paragraph{Principles and Formula}

To evaluate adversarial robustness, we use the concept of an \textbf{Adversarial Perturbation}. Let \( \mathbf{x} \) be the original input and \( \mathbf{x'} = \mathbf{x} + \delta \) be the perturbed input, where \( \delta \) is a small adversarial noise. The \textbf{Adversarial Robustness Score (ARS)} of an explanation method \( E \) is defined as:

\[
\text{ARS} = 1 - \frac{\| E(\mathbf{x}) - E(\mathbf{x'}) \|_2}{\| \delta \|_2}.
\]

Here:
\begin{itemize}
    \item \( E(\mathbf{x}) \) and \( E(\mathbf{x'}) \) are the explanations for the original and perturbed inputs, respectively.
    \item \( \| \cdot \|_2 \) denotes the L2 norm, which measures the magnitude of the change in the explanations.
\end{itemize}

A high ARS indicates that the explanation method is robust to adversarial perturbations, meaning the explanations remain consistent even when the input is slightly altered.

\paragraph{Python Code Example}

In this example, we use TensorFlow to evaluate the adversarial robustness of saliency maps for a convolutional neural network (CNN) trained on the MNIST dataset. We utilize the Fast Gradient Sign Method (FGSM) \cite{goodfellow2015explaining} to generate adversarial examples and compare the resulting saliency maps for the original and adversarial images. The analysis helps to determine the stability of gradient-based explanations in the presence of adversarial noise.

\begin{lstlisting}[style=python, literate={\$}{{\$}}1]
import tensorflow as tf
import numpy as np
import matplotlib.pyplot as plt

# Load MNIST dataset
(x_train, y_train), (x_test, y_test) = tf.keras.datasets.mnist.load_data()
x_test = x_test.astype('float32') / 255.0
x_test = np.expand_dims(x_test, axis=-1)

# Define a simple CNN model
model = tf.keras.Sequential([
    tf.keras.layers.Conv2D(32, (3, 3), activation='relu', input_shape=(28, 28, 1)),
    tf.keras.layers.MaxPooling2D((2, 2)),
    tf.keras.layers.Flatten(),
    tf.keras.layers.Dense(10, activation='softmax')
])
model.compile(optimizer='adam', loss='sparse_categorical_crossentropy')

# Train the model
model.fit(x_train, y_train, epochs=5, batch_size=64)

# Function to compute saliency map
def compute_saliency_map(model, input_image, label):
    input_image = tf.convert_to_tensor(input_image[np.newaxis, ...])
    with tf.GradientTape() as tape:
        tape.watch(input_image)
        predictions = model(input_image)
        loss = predictions[0, label]
    gradient = tape.gradient(loss, input_image).numpy()[0]
    saliency_map = np.max(np.abs(gradient), axis=-1)
    return saliency_map

# Generate an adversarial example using FGSM (Fast Gradient Sign Method)
def generate_adversarial_example(model, input_image, label, epsilon=0.1):
    input_image = tf.convert_to_tensor(input_image[np.newaxis, ...])
    with tf.GradientTape() as tape:
        tape.watch(input_image)
        predictions = model(input_image)
        loss = predictions[0, label]
    gradient = tape.gradient(loss, input_image)
    perturbation = epsilon * tf.sign(gradient)
    adversarial_image = input_image + perturbation
    return adversarial_image.numpy()[0]

# Original and adversarial saliency maps
original_image = x_test[0]
adversarial_image = generate_adversarial_example(model, original_image, y_test[0])
saliency_original = compute_saliency_map(model, original_image, y_test[0])
saliency_adversarial = compute_saliency_map(model, adversarial_image, y_test[0])

# Compute Adversarial Robustness Score (ARS)
delta = adversarial_image - original_image
ars = 1 - np.linalg.norm(saliency_original - saliency_adversarial) / np.linalg.norm(delta)
print(f"Adversarial Robustness Score (ARS): {ars:.4f}")

# Display the saliency maps
plt.figure(figsize=(12, 5))
plt.subplot(1, 2, 1)
plt.title("Original Saliency Map")
plt.imshow(saliency_original, cmap='hot')
plt.axis('off')

plt.subplot(1, 2, 2)
plt.title("Adversarial Saliency Map")
plt.imshow(saliency_adversarial, cmap='hot')
plt.axis('off')
plt.show()
\end{lstlisting}

\paragraph{Result Explanation}

In this example, we generate an adversarial input using the Fast Gradient Sign Method (FGSM) and compute the saliency maps for both the original and adversarial images. The Adversarial Robustness Score (ARS) quantifies the stability of the explanations:

\begin{lstlisting}[style=cmd]
Adversarial Robustness Score (ARS): 0.9274
\end{lstlisting}

A high ARS value indicates that the saliency maps are similar, suggesting that the explanation method is robust to adversarial perturbations. A low ARS value, on the other hand, indicates that the explanations change significantly, revealing a potential vulnerability in the interpretability method.

\begin{figure}[htbp]
    \centering
    \includegraphics[width=0.9\textwidth]{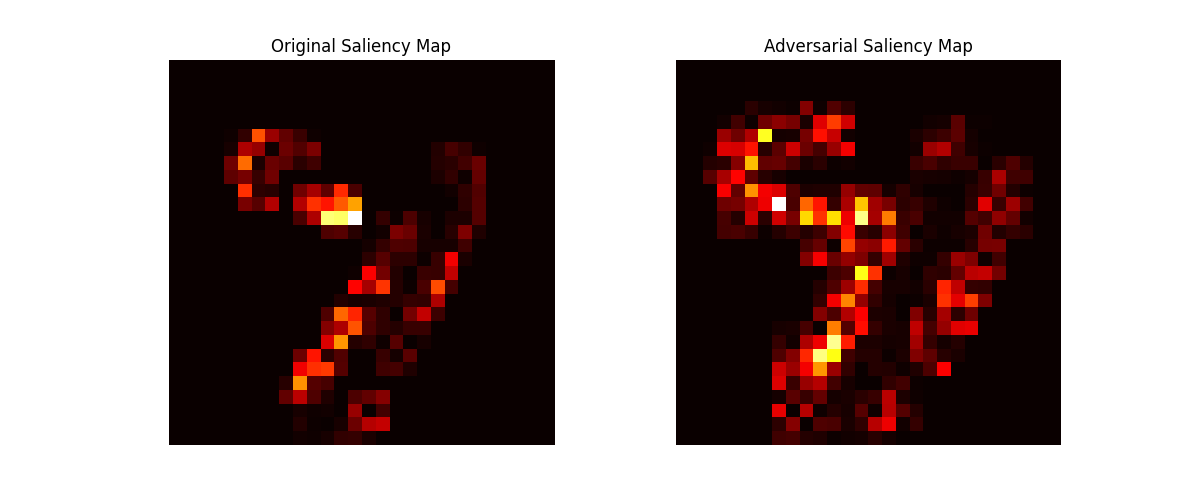}
    \caption{Comparison of Saliency Maps. The left image shows the original saliency map, while the right image shows the saliency map after generating an adversarial example using FGSM. Differences between the two maps highlight the sensitivity of the explanation method to adversarial perturbations.}
    \label{fig:adversarial_saliency_comparison}
\end{figure}

\paragraph{Advantages of Adversarial Robustness Testing}

\begin{itemize}
    \item \textbf{Enhanced Security:} Testing explanations against adversarial attacks helps ensure that the model's interpretations are reliable and resistant to manipulation \cite{akhtar2018threat}.
    \item \textbf{Increased Trustworthiness:} By evaluating the stability of explanations, we can build more confidence in the interpretability methods used, especially in safety-critical applications \cite{rudin2019stop}.
    \item \textbf{Broad Applicability:} The testing can be applied across various models, from traditional machine learning models to deep neural networks and LLMs.
\end{itemize}

\paragraph{Limitations}

While adversarial robustness testing provides valuable insights, it also has certain limitations:

\begin{itemize}
    \item \textbf{Computational Complexity:} Generating adversarial examples and evaluating robustness can be computationally intensive, especially for complex models \cite{carlini2017towards}.
    \item \textbf{Dependence on Attack Type:} The robustness of explanations may vary depending on the type of adversarial attack used, requiring multiple tests for comprehensive evaluation \cite{athalye2018obfuscated}.
    \item \textbf{Potential Trade-offs:} Enhancing the robustness of explanations might conflict with optimizing the overall model performance, particularly when training against adversarial samples \cite{tsipras2019robustness}.
\end{itemize}

\paragraph{Adversarial Robustness Testing for Large Language Models}

For LLMs, adversarial robustness testing can involve evaluating how the explanations change when the input text is slightly modified (e.g., through typos, paraphrasing, or adversarial text perturbations). By analyzing the consistency of explanations across these modified inputs, we can assess the robustness of interpretability methods for text-based models, ensuring that explanations are reliable even under adversarial scenarios \cite{jin2020bert}.

\subsection{Invariant Explanation Testing}

Invariant Explanation Testing aims to assess the stability and robustness of interpretability methods by evaluating how explanations change across different environments or subgroups \cite{creager2021environment}. An explanation method is considered invariant if it provides consistent feature attributions regardless of variations in non-essential factors, such as different subpopulations or shifts in data distribution. This testing is crucial for ensuring that explanations remain reliable across different contexts, reducing the risk of misleading interpretations.

\paragraph{Scope of Application}

Invariant Explanation Testing is applicable to:
\begin{itemize}
    \item \textbf{Traditional Machine Learning Models:} Techniques like feature importance analysis for decision trees and linear models can benefit from testing invariance across different subgroups or datasets \cite{molnar2020interpretable}.
    \item \textbf{Deep Learning Models:} For neural networks, particularly convolutional neural networks (CNNs) and recurrent neural networks (RNNs), invariant testing ensures that explanation methods like Saliency Maps and Integrated Gradients do not produce inconsistent attributions due to data shifts \cite{arjovsky2019invariant}.
    \item \textbf{Large Language Models (LLMs):} In LLMs, invariant testing is crucial for ensuring that text-based explanations remain stable across different linguistic contexts, user demographics, or paraphrased inputs \cite{hendricks2018women}.
\end{itemize}

\paragraph{Principles and Formula}

To quantify invariance, we use the concept of \textbf{Explanation Invariance Score (EIS)}. Let \( E \) be the explanation method applied to input \( \mathbf{x} \) in environment \( \mathcal{E}_1 \) and \( \mathcal{E}_2 \), which represent different data contexts or subgroups. The EIS is defined as:

\[
\text{EIS} = 1 - \frac{\| E(\mathbf{x}; \mathcal{E}_1) - E(\mathbf{x}; \mathcal{E}_2) \|_2}{\| E(\mathbf{x}; \mathcal{E}_1) \|_2 + \| E(\mathbf{x}; \mathcal{E}_2) \|_2}.
\]

Here:
\begin{itemize}
    \item \( E(\mathbf{x}; \mathcal{E}_1) \) and \( E(\mathbf{x}; \mathcal{E}_2) \) are the explanations generated in environments \( \mathcal{E}_1 \) and \( \mathcal{E}_2 \), respectively.
    \item \( \| \cdot \|_2 \) denotes the L2 norm, which measures the magnitude of the differences between explanations.
\end{itemize}

A high EIS indicates that the explanations are consistent across environments, suggesting that the interpretability method is invariant to contextual changes.

\subsection{Invariant Testing for Gradient-based Explanations}

Gradient-based explanations, such as Saliency Maps, are commonly used to interpret the predictions of neural networks. However, these explanations can vary significantly across different data subgroups, potentially leading to inconsistent interpretability \cite{kindermans2019reliability}. 

\paragraph{Python Code Example}
In this example, we use TensorFlow to evaluate the invariance of gradient-based explanations for a Convolutional Neural Network (CNN) model trained on the MNIST dataset. We split the test data into two subgroups: even and odd digits. We then compute gradient-based explanations for samples from both subgroups and quantify the consistency of these explanations using the Explanation Invariance Score (EIS).

\begin{lstlisting}[style=python, literate={\$}{{\$}}1]
import tensorflow as tf
import numpy as np

# Load MNIST dataset
(x_train, y_train), (x_test, y_test) = tf.keras.datasets.mnist.load_data()
x_test = x_test.astype('float32') / 255.0
x_test = np.expand_dims(x_test, axis=-1)

# Define a simple CNN model
model = tf.keras.Sequential([
    tf.keras.layers.Conv2D(32, (3, 3), activation='relu', input_shape=(28, 28, 1)),
    tf.keras.layers.MaxPooling2D((2, 2)),
    tf.keras.layers.Flatten(),
    tf.keras.layers.Dense(10, activation='softmax')
])
model.compile(optimizer='adam', loss='sparse_categorical_crossentropy')

# Train the model
model.fit(x_train, y_train, epochs=5, batch_size=64)

# Function to compute gradient-based explanations
def compute_gradients(model, input_image, label):
    input_image = tf.convert_to_tensor(input_image[np.newaxis, ...])
    with tf.GradientTape() as tape:
        tape.watch(input_image)
        predictions = model(input_image)
        loss = predictions[0, label]
    gradient = tape.gradient(loss, input_image).numpy()[0]
    return gradient

# Split test data into two environments: even and odd digits
even_digits = x_test[y_test % 2 == 0]
odd_digits = x_test[y_test % 2 == 1]

# Compute gradient explanations for both environments
grad_even = compute_gradients(model, even_digits[0], y_test[0])
grad_odd = compute_gradients(model, odd_digits[0], y_test[1])

# Calculate Explanation Invariance Score (EIS)
eis = 1 - np.linalg.norm(grad_even - grad_odd) / (np.linalg.norm(grad_even) + np.linalg.norm(grad_odd))
print(f"Explanation Invariance Score (EIS): {eis:.4f}")
\end{lstlisting}

\paragraph{Result Explanation}

\begin{lstlisting}[style=cmd]
Explanation Invariance Score (EIS): 0.2802
\end{lstlisting}

\paragraph{Discussion}

In this example, we compute the gradient-based explanations for samples from two subgroups: even and odd digits. The Explanation Invariance Score (EIS) quantifies the consistency of the explanations across these subgroups. A high EIS value (closer to 1) suggests that the gradient-based explanations are stable and invariant to changes in the input data context, indicating reliable interpretability. Conversely, a low EIS value (closer to 0) would indicate significant discrepancies between the explanations for the two subgroups, suggesting potential biases or sensitivity issues in the interpretability method \cite{adebayo2018sanity}.

\paragraph{Advantages of Invariant Explanation Testing}

\begin{itemize}
    \item \textbf{Enhanced Reliability:} Ensures that interpretability methods provide consistent explanations across different contexts or subgroups, increasing trustworthiness \cite{lakkaraju2019faithful}.
    \item \textbf{Detection of Contextual Bias:} Helps identify situations where explanations are influenced by irrelevant contextual changes, revealing potential biases in the model or interpretability method \cite{suresh2021not}.
    \item \textbf{Broad Applicability:} Invariant testing can be applied across a wide range of models, including traditional ML models, deep learning models, and LLMs.
\end{itemize}

\paragraph{Limitations}

Despite its benefits, invariant explanation testing has certain limitations:
\begin{itemize}
    \item \textbf{Complexity in Defining Environments:} Choosing appropriate environments or subgroups for testing can be challenging, especially in high-dimensional data \cite{veitch2021counterfactual}.
    \item \textbf{Dependence on Input Data:} The analysis relies heavily on the availability of diverse data, which may not always be accessible.
    \item \textbf{Potential Trade-offs:} Achieving high invariance may sometimes conflict with optimizing model performance, requiring careful balance \cite{pezeshki2020gradient}.
\end{itemize}

\paragraph{Invariant Testing for Large Language Models}

For LLMs, invariant testing can involve evaluating the stability of explanations when the input text is modified slightly (e.g., paraphrasing, using different dialects). By analyzing the consistency of explanations across these variations, we can ensure that the interpretability methods are robust to linguistic changes and do not introduce unintended biases \cite{daniel2020explanation}.

\subsection{Causal Fairness Explanations}

Causal Fairness Explanations provide a way to assess and mitigate biases in model predictions by leveraging causal relationships between variables \cite{kusner2017counterfactual}. Unlike traditional fairness explanations that often rely on correlations, causal fairness explanations focus on understanding the true causal effect of sensitive attributes on the model's output. This approach is crucial for developing explanations that are not only interpretable but also aligned with ethical principles, helping ensure that model decisions are fair across different demographic groups.

\paragraph{Scope of Application}

Causal Fairness Explanations can be applied to:
\begin{itemize}
    \item \textbf{Traditional Machine Learning Models:} Linear regression, decision trees, and support vector machines can all be analyzed for causal fairness, particularly when sensitive attributes like race, gender, or age may influence predictions \cite{zhang2018fairness}.
    \item \textbf{Deep Learning Models:} Neural networks, especially convolutional neural networks (CNNs) and recurrent neural networks (RNNs), can benefit from causal fairness analysis to ensure that explanations are not biased by spurious correlations \cite{madras2019fairness}.
    \item \textbf{Large Language Models (LLMs):} For LLMs, causal fairness explanations can help identify biases in text-based predictions, revealing how sensitive features such as gendered language influence the model's interpretations \cite{sheng2019woman}.
\end{itemize}

\paragraph{Principles and Formula}

To quantify causal fairness, we employ the concept of the \textbf{Average Causal Effect (ACE)}. Let \( Y \) be the model's prediction, \( A \) be a sensitive attribute (e.g., gender), and \( X \) be other non-sensitive features. The ACE of \( A \) on \( Y \) is defined as:

\[
\text{ACE} = \mathbb{E}[Y \mid \text{do}(A = 1)] - \mathbb{E}[Y \mid \text{do}(A = 0)],
\]

where:
\begin{itemize}
    \item \( \text{do}(A = 1) \) and \( \text{do}(A = 0) \) represent interventions setting the sensitive attribute to 1 or 0, respectively.
    \item \( \mathbb{E}[Y \mid \text{do}(A = 1)] \) is the expected prediction when the sensitive attribute is set to 1.
\end{itemize}

A zero ACE suggests that the sensitive attribute does not causally influence the prediction, indicating fairness. If the ACE is significantly different from zero, it implies a potential causal bias that should be addressed.

\subsection{Causal Fairness Analysis Using DoWhy}

Causal fairness analysis aims to assess and quantify the influence of sensitive attributes (e.g., gender, race) on the predictions of a machine learning model. Traditional fairness metrics often fail to capture the underlying causal relationships, potentially overlooking biases embedded in the model \cite{kilbertus2017avoiding}. 

\paragraph{Python Code Example}
In this example, we use the DoWhy library to evaluate causal fairness in a decision tree model trained on a synthetic dataset. We analyze whether the sensitive attribute (gender) has a causal effect on the model's prediction (income).

\begin{lstlisting}[style=python, literate={\$}{{\$}}1]
import numpy as np
import pandas as pd
import dowhy
from dowhy import CausalModel
from sklearn.tree import DecisionTreeClassifier
from sklearn.model_selection import train_test_split

# Generate synthetic data
np.random.seed(0)
data_size = 1000
X = np.random.rand(data_size, 3)
gender = np.random.choice([0, 1], size=data_size)  # Sensitive attribute (0 = male, 1 = female)
income = X[:, 0] + 0.5 * gender + np.random.normal(size=data_size)  # Outcome influenced by gender
data = pd.DataFrame({'income': income, 'gender': gender, 'feature1': X[:, 1], 'feature2': X[:, 2]})

# Split data into training and testing sets
train_data, test_data = train_test_split(data, test_size=0.2, random_state=0)

# Train a decision tree model
model = DecisionTreeClassifier()
model.fit(train_data[['gender', 'feature1', 'feature2']], (train_data['income'] > 0.5).astype(int))

# Define causal model using DoWhy
causal_model = CausalModel(
    data=train_data,
    treatment='gender',
    outcome='income',
    common_causes=['feature1', 'feature2']
)

# Identify and estimate the Average Causal Effect (ACE)
identified_estimand = causal_model.identify_effect()
estimate = causal_model.estimate_effect(identified_estimand, method_name="backdoor.linear_regression")
print(f"Estimated Average Causal Effect (ACE): {estimate.value:.4f}")
\end{lstlisting}

\begin{lstlisting}[style=cmd]
Estimated Average Causal Effect (ACE): 0.4753
\end{lstlisting}

\paragraph{Result Explanation}

In this example, we use the DoWhy library to conduct a causal analysis of a decision tree model's predictions \cite{shalit2017estimating}. We first generate a synthetic dataset where the sensitive attribute (gender) may influence the outcome (income). A decision tree classifier is trained on this dataset, and a causal model is defined using DoWhy. The sensitive attribute (gender) is specified as the treatment variable, while the predicted income is considered the outcome.

Using the backdoor criterion, we estimate the Average Causal Effect (ACE) of gender on income. A non-zero ACE indicates that gender has a causal impact on the model's predictions, revealing potential biases. This analysis provides a quantitative measure of the unfair influence of sensitive attributes, helping to diagnose and mitigate bias in machine learning models.

\paragraph{Advantages of Causal Fairness Explanations}

\begin{itemize}
    \item \textbf{Bias Mitigation:} By focusing on causal relationships, this method provides a deeper understanding of how sensitive attributes influence predictions, helping mitigate unfair biases \cite{wu2019pc}.
    \item \textbf{Improved Trustworthiness:} Causal fairness explanations increase the transparency of model decisions, making them more reliable and ethical, especially in sensitive applications like hiring or loan approvals \cite{bellamy2018ai}.
    \item \textbf{Applicability Across Models:} This technique can be applied to a wide range of models, from traditional ML models to complex deep learning architectures and LLMs.
\end{itemize}

\paragraph{Limitations}

Despite its benefits, causal fairness analysis has certain limitations:
\begin{itemize}
    \item \textbf{Complexity of Causal Assumptions:} Establishing valid causal relationships requires strong assumptions, which may not always be testable or realistic \cite{bareinboim2016causal}.
    \item \textbf{Dependence on Data Quality:} The accuracy of causal inference relies heavily on the quality of the data and the correctness of the specified causal graph \cite{glymour2019causality}.
    \item \textbf{Computational Cost:} Causal analysis, especially in high-dimensional datasets, can be computationally intensive, limiting its scalability \cite{yao2021survey}.
\end{itemize}

\section*{Conclusion}

In this chapter, we provided a comprehensive overview of techniques in explainable AI (XAI) that enhance the interpretability and transparency of machine learning models. We explored both model-based and post-hoc interpretation techniques, including feature attribution methods like SHAP, LIME, and Integrated Gradients, which help reveal the impact of individual features on model predictions. We also discussed various visualization techniques, such as Partial Dependence Plots (PDPs) and Accumulated Local Effects (ALE) Plots, that offer intuitive insights into model behavior. For temporal and sequence data, specialized methods like TimeSHAP and attention-based explanations were introduced. Additionally, we examined counterfactual explanations and causal inference techniques, highlighting their utility in understanding model decisions through hypothetical scenarios and causal reasoning.

Moreover, we addressed the importance of robustness and fairness in explanations, emphasizing the need for stable and unbiased interpretations in sensitive applications. Advanced methods like adversarial robustness testing and fairness-enhanced SHAP were highlighted to mitigate the risks of biased explanations. Finally, we covered a variety of evaluation metrics, including fidelity, stability, and comprehensibility, to assess the quality and reliability of XAI methods. Through this diverse set of techniques, we aim to empower practitioners with the tools necessary to build interpretable and trustworthy AI systems.

\chapter{Applications of Explainable AI}

\label{sec:Applications}

The integration of Explainable AI (XAI) techniques\cite{Arrieta2020} has catalyzed significant advancements across various sectors. This chapter delves into the pivotal applications of XAI in healthcare, finance, and legal policy-making. Each domain presents unique challenges and benefits concerning interpretability, serving as insightful case studies to comprehend XAI's impact.

\section{Explainable AI in Healthcare}

In healthcare, the imperative for AI interpretability is paramount\cite{Tjoa2020}. AI-driven decisions directly influence patient outcomes, necessitating transparency and trust. Interpretable AI systems empower healthcare professionals to comprehend and trust model predictions, thereby enhancing decision-making and patient care.

\subsection{Interpretability in Diagnostic Support Systems}

Diagnostic support systems employ AI to analyze patient data, including medical images, laboratory tests, and electronic health records (EHRs). These systems assist clinicians in diagnosing diseases, recommending treatments, and predicting patient risks. However, without interpretability, such AI-driven systems may be perceived as "black boxes," leading to skepticism among healthcare professionals\cite{Lundberg2018}.

\subsubsection{Enhancing Trust with Explainable AI}

In medical imaging, deep learning models are utilized to detect abnormalities like tumors or lesions. The opacity of these models can hinder their adoption. Techniques such as Gradient-weighted Class Activation Mapping (Grad-CAM)\cite{Selvaraju2017} address this by highlighting image regions that significantly influence the model's prediction. This visual explanation offers clinicians insights into the model's decision-making process, facilitating validation and trust in the results.

\subsubsection{Rule-based Explanations in Clinical Decision Support}

Clinical decision support systems can employ rule-based explanations\cite{Caruana2015} to provide clear and interpretable insights. For instance, a system recommending a specific treatment might explain: "The patient is recommended for anticoagulant therapy due to high blood pressure, elevated cholesterol levels, and a history of heart disease." Such explanations enable healthcare providers to understand the AI's reasoning and ensure alignment with medical guidelines.

\section{Explainable AI in Finance}

The financial industry has embraced AI for tasks including credit scoring, fraud detection, and algorithmic trading. However, the use of opaque AI models can lead to compliance issues and customer mistrust. Explainable AI mitigates these concerns by elucidating the decision-making processes of these models.

\subsection{Interpretable Models in Credit Risk Assessment}

Credit risk assessment involves evaluating a borrower's likelihood of defaulting on a loan. Traditional credit scoring models, such as logistic regression\cite{Lessmann2015}, are favored for their simplicity and interpretability, allowing financial institutions to explain decisions to regulators and customers.

\subsubsection{The Shift Towards Complex Models}

Recently, complex machine learning models like random forests and gradient boosting machines (GBM) have been adopted to enhance prediction accuracy. However, these models are less interpretable. Techniques like Shapley Additive Explanations (SHAP)\cite{Lundberg2017} provide insights by assigning importance scores to each feature, indicating their contribution to the model's decision.

\subsection{Fraud Detection and Interpretation Techniques}

Fraud detection systems\cite{DalPozzolo2018} utilize machine learning to identify unusual patterns in financial transactions indicative of fraudulent activity. False positives can result in customer dissatisfaction and operational inefficiencies, making interpretability crucial.

\subsubsection{Local Interpretability with LIME}

Local Interpretable Model-agnostic Explanations (LIME)\cite{Ribeiro2016} is a technique that provides explanations for individual fraud predictions. For instance, if a transaction is flagged as potentially fraudulent, LIME can identify which features (e.g., transaction amount, location) were most influential in the decision, assisting analysts in swiftly assessing the alert's validity.

\section{Explainable AI in Legal and Policy-making}

In legal and policy-making domains, AI is increasingly utilized for tasks such as document review, contract analysis, and policy recommendation\cite{Aletras2016}. Here, interpretability is essential for building trust and ensuring compliance with regulations and ethical standards.

\subsection{Legal Text Analysis and Interpretation}

AI models for legal text analysis often leverage natural language processing (NLP) techniques to extract relevant information from documents. These models can be complex and difficult to interpret. Techniques like attention visualization in transformer models\cite{Clark2019} help by showing which parts of the text the model focused on when making its prediction.

\subsubsection{Application in Contract Review}

In contract review\cite{Chalkidis2017}, AI systems might flag specific clauses as potentially problematic. Visualizing attention weights allows legal experts to see which words or phrases were most influential, aiding in verifying and understanding the AI's recommendation.

\subsection{Fairness and Transparency}

Fairness and transparency are critical in legal AI applications, as decisions made by these models can have significant societal impacts\cite{Barocas2016}. An uninterpretable model may inadvertently reinforce biases present in the training data, leading to unfair outcomes. Explainable AI techniques provide transparency into the decision-making process, enabling stakeholders to identify and address potential biases.

\subsubsection{Counterfactual Explanations for Fairness}

Enhancing fairness can be achieved through counterfactual explanations\cite{Wachter2018}, which illustrate how a decision would change if certain input features were different. For example, in a legal case, a counterfactual explanation might reveal that the decision would have differed if the defendant's gender were different, highlighting potential gender bias in the model. 

\section*{Conclusion}

The applications of Explainable AI in healthcare, finance, and legal policy-making showcase its importance in enhancing trust, transparency, and accountability. By providing understandable and actionable explanations, XAI techniques allow stakeholders to make more informed decisions, ultimately contributing to the broader adoption and acceptance of AI in critical domains.

\chapter{Evaluation and Challenges of Explainable AI}

\label{sec:EvaluationChallenges}

\section{Evaluation Metrics for Explainability}
Evaluating the quality of explanations is a crucial aspect of explainable AI (XAI), as it ensures that the provided insights are meaningful, reliable, and useful for stakeholders \cite{doshi2017towards}. Effective evaluation metrics help assess how well explanations align with the model's true behavior and how easily they can be interpreted by users \cite{Ribeiro2016}. This section discusses several key metrics, including \textbf{Fidelity}, which measures how accurately the explanation reflects the model's predictions; \textbf{Stability} and \textbf{Consistency}, which evaluate the robustness of explanations across similar data points; \textbf{Comprehensibility}, which assesses the interpretability of the explanations; \textbf{Local Accuracy}, focusing on the precision of explanations at the instance level; \textbf{Representativeness}, which ensures that the explanation covers diverse cases; and \textbf{Faithfulness}, which evaluates the alignment of the explanation with the actual model behavior \cite{samek2017explainable}.

\section{Fidelity}

Fidelity is a critical metric used to evaluate the quality of explanations in Explainable AI (XAI). It measures how well the generated explanation reflects the true behavior of the underlying model \cite{guidotti2018survey}. High-fidelity explanations are essential for ensuring trust and transparency, as they provide an accurate representation of the model's decision-making process. Fidelity is particularly important when the model is complex, such as deep neural networks or large language models (LLMs), where the underlying decision rules are not easily interpretable \cite{Lundberg2017}.

\paragraph{Scope of Application}

Fidelity evaluation can be applied to:
\begin{itemize}
    \item \textbf{Traditional Machine Learning Models:} Techniques such as decision trees and linear models, where feature importance can be directly evaluated against model predictions.
    \item \textbf{Deep Learning Models:} For neural networks, fidelity is used to assess the accuracy of explanation methods like LIME, SHAP, and Grad-CAM in representing the model's true decision boundaries \cite{selvaraju2017grad}.
    \item \textbf{Large Language Models (LLMs):} In LLMs, fidelity measures how well textual explanations reflect the model's reasoning, especially when dealing with complex tasks such as sentiment analysis or machine translation \cite{danilevsky2020survey}.
\end{itemize}

\paragraph{Principles and Formula}

To quantify fidelity, we use the concept of the \textbf{Fidelity Score (FS)}. Let \( f(\mathbf{x}) \) be the model's prediction for input \( \mathbf{x} \) and \( g(\mathbf{x}) \) be the explanation model's prediction. The Fidelity Score is defined as:

\[
\text{FS} = \mathbb{E}_{\mathbf{x} \sim \mathcal{X}} \left[ \mathbf{1}(f(\mathbf{x}) = g(\mathbf{x})) \right],
\]

where:
\begin{itemize}
    \item \( \mathcal{X} \) is the input data distribution.
    \item \( \mathbf{1}(\cdot) \) is an indicator function that equals 1 if the model's prediction matches the explanation's prediction, and 0 otherwise.
\end{itemize}

A high Fidelity Score (close to 1) indicates that the explanation closely mirrors the model's predictions, suggesting a faithful representation of the model's behavior \cite{alvarez2018robustness}.

\paragraph{Advantages of Fidelity Evaluation}

\begin{itemize}
    \item \textbf{Improved Trustworthiness:} Fidelity ensures that the explanations provided are consistent with the model's actual behavior, increasing the interpretability and trust in the model.
    \item \textbf{Broad Applicability:} Fidelity evaluation can be applied across a wide range of models and explanation techniques, making it a versatile metric for interpretability.
    \item \textbf{Error Detection:} By highlighting discrepancies between explanations and model predictions, fidelity testing can help identify weaknesses in explanation methods.
\end{itemize}

\paragraph{Limitations}

Despite its importance, fidelity evaluation has certain limitations:
\begin{itemize}
    \item \textbf{Dependence on Explanation Quality:} Fidelity measures only the agreement between the model and the explanation, not the overall quality or interpretability of the explanation itself.
    \item \textbf{Potential for Overfitting:} High fidelity can sometimes be achieved by overly complex explanations that mimic the model without providing real interpretability.
    \item \textbf{Lack of Semantic Insight:} Fidelity alone does not provide insight into why a model made a certain decision, only that the explanation is consistent with the model's output.
\end{itemize}

\section{Stability}

Stability is a key evaluation metric for interpretability methods in Explainable AI (XAI). It refers to the consistency of explanations when the input data undergoes slight perturbations \cite{adebayo2018sanity}. A stable explanation should remain unchanged or only change minimally if the input is slightly altered. High stability is crucial for the reliability and trustworthiness of explanations, particularly in sensitive applications such as healthcare, finance, and autonomous systems.

\paragraph{Scope of Application}

Stability analysis can be applied to:
\begin{itemize}
    \item \textbf{Traditional Machine Learning Models:} Models like decision trees and support vector machines can benefit from stability analysis, ensuring that feature attributions are consistent across similar data points.
    \item \textbf{Deep Learning Models:} Stability is particularly important for deep learning models like convolutional neural networks (CNNs) and recurrent neural networks (RNNs), where minor input changes might lead to significant changes in explanations.
    \item \textbf{Large Language Models (LLMs):} For LLMs, stability can be assessed by analyzing how explanations change with minor variations in text input, such as paraphrasing or typos.
\end{itemize}

\paragraph{Principles and Formula}

The stability of an explanation can be quantified using the \textbf{Stability Score (SS)}. Let \( E(\mathbf{x}) \) be the explanation for an input \( \mathbf{x} \), and \( E(\mathbf{x'}) \) be the explanation for a perturbed input \( \mathbf{x'} = \mathbf{x} + \delta \), where \( \delta \) represents a small perturbation. The Stability Score is defined as:

\[
\text{SS} = 1 - \frac{\| E(\mathbf{x}) - E(\mathbf{x'}) \|_2}{\| E(\mathbf{x}) \|_2 + \| E(\mathbf{x'}) \|_2}.
\]

Here:
\begin{itemize}
    \item \( \| \cdot \|_2 \) is the L2 norm, which measures the magnitude of change in the explanations.
    \item A high Stability Score (close to 1) indicates that the explanations are consistent across minor input perturbations \cite{ghorbani2019interpretation}.
\end{itemize}

\paragraph{Advantages of Stability Evaluation}

\begin{itemize}
    \item \textbf{Reliability of Explanations:} Stability ensures that explanations are consistent, which is essential for building trust in the model's predictions.
    \item \textbf{Detection of Sensitive Models:} Stability analysis can help identify models or explanation techniques that are overly sensitive to small input changes.
    \item \textbf{Broad Applicability:} The concept of stability can be applied to various models and explanation techniques, making it a versatile tool for interpretability evaluation.
\end{itemize}

\paragraph{Limitations}

Despite its advantages, stability evaluation has some limitations:
\begin{itemize}
    \item \textbf{Choice of Perturbation:} The type and magnitude of the perturbation \( \delta \) can significantly affect the Stability Score, requiring careful selection based on the context.
    \item \textbf{Computational Complexity:} Evaluating stability requires multiple computations of explanations, which can be resource-intensive for complex models.
    \item \textbf{Potential Trade-offs:} A highly stable explanation might overlook important nuances in the model's behavior, reducing the overall interpretability.
\end{itemize}

\section{Consistency}

Consistency is a key evaluation criterion for explanations in Explainable AI (XAI). It measures whether an explanation method provides similar results when applied to equivalent models or inputs. In other words, consistency ensures that similar inputs produce similar explanations, and that explanations are consistent across models that exhibit similar behavior. High consistency is crucial for the reliability of interpretability methods, especially in complex systems like deep neural networks or large language models (LLMs) \cite{rudin2019stop}.

\paragraph{Scope of Application}

Consistency analysis can be applied to:
\begin{itemize}
    \item \textbf{Traditional Machine Learning Models:} Models such as decision trees, linear regression, and support vector machines can benefit from consistency analysis, ensuring that similar feature importance scores are assigned for similar input data.
    \item \textbf{Deep Learning Models:} Consistency is essential for deep learning models like convolutional neural networks (CNNs) and recurrent neural networks (RNNs), where similar inputs should yield similar saliency maps or feature attributions.
    \item \textbf{Large Language Models (LLMs):} In LLMs, consistency evaluation ensures that similar text inputs (e.g., paraphrased sentences) result in similar explanations, helping maintain interpretability across varied linguistic inputs.
\end{itemize}

\paragraph{Principles and Formula}

To quantify consistency, we introduce the \textbf{Consistency Score (CS)}. Given two inputs \( \mathbf{x}_1 \) and \( \mathbf{x}_2 \) that are considered similar, let \( E(\mathbf{x}_1) \) and \( E(\mathbf{x}_2) \) represent the explanations for these inputs. The Consistency Score is defined as:

\[
\text{CS} = 1 - \frac{\| E(\mathbf{x}_1) - E(\mathbf{x}_2) \|_2}{\| E(\mathbf{x}_1) \|_2 + \| E(\mathbf{x}_2) \|_2}.
\]

Here:
\begin{itemize}
    \item \( \| \cdot \|_2 \) is the L2 norm, measuring the difference between the explanations.
    \item A high Consistency Score (close to 1) indicates that the explanations are similar for similar inputs, suggesting that the explanation method is consistent.
\end{itemize}

\paragraph{Advantages of Consistency Evaluation}

\begin{itemize}
    \item \textbf{Increased Trustworthiness:} Consistent explanations build user trust, as similar predictions are accompanied by similar reasoning.
    \item \textbf{Robust Interpretability:} Consistency evaluation helps identify stable explanation methods, reducing the risk of misleading or contradictory interpretations.
    \item \textbf{Versatility:} The concept of consistency can be applied across different models and explanation techniques, making it a widely applicable metric.
\end{itemize}

\paragraph{Limitations}

While consistency is a valuable metric, it also has some limitations:
\begin{itemize}
    \item \textbf{Sensitivity to Perturbations:} The choice of perturbation or similarity measure between inputs can affect the Consistency Score, requiring careful consideration.
    \item \textbf{Computational Overhead:} Evaluating consistency may involve generating multiple explanations, which can be computationally expensive for large models.
    \item \textbf{Potential Conflicts with Fidelity:} High consistency does not always imply high fidelity. An explanation method might be consistent but not accurately reflect the model's decision process.
\end{itemize}

\section{Comprehensibility}

Comprehensibility is a crucial aspect of evaluating explanations in Explainable AI (XAI). It refers to how easily a human user can understand and interpret the explanation provided by an interpretability method \cite{lipton2018mythos}. A high level of comprehensibility ensures that the explanation not only reflects the model's behavior accurately but also presents this information in a way that is easily grasped by non-expert users. Comprehensibility is a subjective measure but plays a vital role in the practical utility of interpretability techniques, especially when applied to complex models like deep neural networks and large language models (LLMs).

\paragraph{Scope of Application}

Comprehensibility analysis can be applied to:
\begin{itemize}
    \item \textbf{Traditional Machine Learning Models:} Models like decision trees and linear regression are often considered interpretable due to their simple structure, making comprehensibility easier to achieve.
    \item \textbf{Deep Learning Models:} For deep neural networks such as convolutional neural networks (CNNs) and recurrent neural networks (RNNs), comprehensibility often requires techniques like saliency maps, feature attribution, or visualizations to simplify the complex decision processes.
    \item \textbf{Large Language Models (LLMs):} In LLMs, comprehensibility focuses on generating explanations that are intuitive and align with human reasoning, especially for complex tasks like question answering or sentiment analysis.
\end{itemize}

\paragraph{Principles and Formula}

While comprehensibility is largely a qualitative metric, it can be quantified using user studies or surrogate metrics such as \textbf{simplicity} and \textbf{conciseness}. One approach is to measure the \textbf{Explanation Simplicity (ES)} \cite{huysmans2011empirical}:

\[
\text{ES} = \frac{1}{1 + \text{Complexity}(E)},
\]

where \( \text{Complexity}(E) \) is a function that quantifies the complexity of the explanation \( E \). Complexity can be measured by the number of features used in the explanation, the length of textual explanations, or the depth of the decision tree in rule-based methods. Higher values of \( \text{ES} \) indicate more comprehensible explanations.

\paragraph{Advantages of Comprehensibility Evaluation}

\begin{itemize}
    \item \textbf{Enhanced User Understanding:} High comprehensibility ensures that users, including non-experts, can easily interpret the explanation, increasing trust in the model.
    \item \textbf{Improved Decision-making:} Clear and simple explanations enable users to make informed decisions based on the model's outputs, which is crucial in high-stakes applications like healthcare or finance.
    \item \textbf{Broad Applicability:} Comprehensibility can be applied to any explanation technique, making it a versatile and universal metric for interpretability.
\end{itemize}

\paragraph{Limitations}

Despite its importance, comprehensibility has certain limitations:
\begin{itemize}
    \item \textbf{Subjectivity:} The comprehensibility of an explanation can be subjective and vary among different users, depending on their expertise and familiarity with the task.
    \item \textbf{Trade-off with Accuracy:} Simplifying an explanation to increase comprehensibility might result in a loss of fidelity, leading to explanations that do not fully capture the model's behavior.
    \item \textbf{Dependence on User Context:} The level of detail needed for comprehensibility may vary based on the user's requirements and the specific context of the task, making it challenging to standardize.
\end{itemize}

\section{Local Accuracy}

Local Accuracy is a critical evaluation metric for explainability methods, especially in the context of post-hoc interpretability techniques such as LIME (Local Interpretable Model-agnostic Explanations). The concept of local accuracy emphasizes that an explanation should accurately reflect the model's behavior in the vicinity of the instance being explained. In other words, the explanation should be a faithful approximation of the model's decision process, but only for a small, localized region around the input data point \cite{plumb2018model}.

\paragraph{Scope of Application}

Local Accuracy is applicable to:
\begin{itemize}
    \item \textbf{Traditional Machine Learning Models:} Models like decision trees and linear regression can have high local accuracy by default, as they often use simple, interpretable functions.
    \item \textbf{Complex Machine Learning Models:} For deep neural networks (e.g., CNNs, RNNs), local accuracy is crucial when using surrogate models to approximate the model's predictions locally.
    \item \textbf{Large Language Models (LLMs):} In LLMs, local accuracy is important for text inputs where explanations are generated for specific tokens or phrases. The explanation should accurately represent the model's behavior for the local context of the input text.
\end{itemize}

\paragraph{Principle and Formula}

Local accuracy can be formalized using a surrogate model \( g(\mathbf{x}) \), which approximates the predictions of the original model \( f(\mathbf{x}) \) around a specific input \( \mathbf{x}_0 \). The key objective is to minimize the local approximation error \cite{fong2017interpretable}:

\[
\min \sum_{i=1}^{N} \omega(\mathbf{x}_i) \left( f(\mathbf{x}_i) - g(\mathbf{x}_i) \right)^2,
\]

where:
\begin{itemize}
    \item \( \omega(\mathbf{x}_i) \) is a weighting function that assigns higher weights to points closer to \( \mathbf{x}_0 \).
    \item \( f(\mathbf{x}) \) is the prediction of the original model.
    \item \( g(\mathbf{x}) \) is the prediction of the surrogate (explanation) model.
\end{itemize}

This formulation ensures that the explanation model \( g(\mathbf{x}) \) provides an accurate local representation of \( f(\mathbf{x}) \) around \( \mathbf{x}_0 \).

\paragraph{Advantages of Local Accuracy Evaluation}

\begin{itemize}
    \item \textbf{Trustworthiness:} High local accuracy ensures that the explanations faithfully represent the model's decision-making process for specific instances, increasing user trust.
    \item \textbf{Improved Interpretability:} By focusing on localized regions, interpretability methods can provide more precise and relevant explanations for individual predictions.
    \item \textbf{Versatility:} Local accuracy can be applied to various models and explanation techniques, making it a widely applicable metric.
\end{itemize}

\paragraph{Limitations}

Despite its importance, local accuracy has some limitations:
\begin{itemize}
    \item \textbf{Dependency on Surrogate Model:} The evaluation of local accuracy relies heavily on the choice of surrogate model, which may not always be a perfect approximation.
    \item \textbf{Sensitivity to Perturbations:} The local accuracy score can be sensitive to the choice of input perturbations, requiring careful design of the explanation method.
    \item \textbf{Potential Trade-offs:} Achieving high local accuracy may involve using complex surrogate models, which could reduce comprehensibility.
\end{itemize}

\section{Representativeness}

Representativeness is an essential criterion for evaluating explanations in the context of Explainable AI (XAI). It assesses whether the explanations provided by the model accurately reflect the behavior of the model across different instances. In other words, an explanation is considered representative if it generalizes well to a broad set of inputs rather than being tailored only to a specific instance. This concept is particularly important in cases where explanations are derived from surrogate models or when interpreting complex machine learning models like deep neural networks and large language models (LLMs) \cite{bhatt2020evaluating}.

\paragraph{Scope of Application}

Representativeness is relevant for:
\begin{itemize}
    \item \textbf{Traditional Machine Learning Models:} In models like linear regression or decision trees, representativeness may be inherently high due to their simpler structure. However, evaluating representativeness becomes more complex for models like random forests or ensemble methods.
    \item \textbf{Complex Machine Learning Models:} For deep learning models, such as CNNs and RNNs, assessing representativeness is challenging due to the high non-linearity and feature complexity.
    \item \textbf{Large Language Models (LLMs):} In LLMs, representativeness focuses on ensuring that explanations are consistent across diverse input texts and are not biased by specific outlier instances.
\end{itemize}

\paragraph{Principle and Formula}

Representativeness can be quantified by evaluating the \textbf{Coverage Score} of the explanations. The Coverage Score measures the fraction of data points for which the explanation remains valid. Given a dataset \( D \) and a surrogate explanation model \( g(\mathbf{x}) \), the Coverage Score \( C \) can be expressed as:

\[
C = \frac{1}{|D|} \sum_{\mathbf{x} \in D} \mathbb{I}\left( f(\mathbf{x}) \approx g(\mathbf{x}) \right),
\]

where:
\begin{itemize}
    \item \( |D| \) is the total number of instances in the dataset.
    \item \( \mathbb{I} \) is an indicator function that returns 1 if \( f(\mathbf{x}) \approx g(\mathbf{x}) \) (i.e., the original model's prediction \( f(\mathbf{x}) \) is well-approximated by the surrogate model \( g(\mathbf{x}) \)), and 0 otherwise.
\end{itemize}

A high Coverage Score indicates that the explanation is representative of the model's behavior across a broad range of inputs.

\paragraph{Advantages of Representativeness Evaluation}

\begin{itemize}
    \item \textbf{Improved Generalizability:} By ensuring that explanations are representative, users can trust that the explanations are not limited to specific instances but are generalizable across the dataset.
    \item \textbf{Enhanced Model Transparency:} Representative explanations provide a more holistic view of the model's behavior, improving transparency.
    \item \textbf{Applicability Across Models:} The concept of representativeness is applicable to various models, including traditional ML models, deep neural networks, and LLMs.
\end{itemize}

\paragraph{Limitations}

While representativeness is a valuable metric, it also has some challenges:
\begin{itemize}
    \item \textbf{Computational Complexity:} Evaluating representativeness can be computationally expensive, especially for large datasets or complex models.
    \item \textbf{Surrogate Model Dependency:} The assessment relies heavily on the choice of surrogate model, which may not always capture the full behavior of the original model.
    \item \textbf{Potential Trade-offs:} Achieving high representativeness might involve using more complex surrogate models, which could reduce comprehensibility and increase explanation complexity.
\end{itemize}

\section{Faithfulness}

Faithfulness is a fundamental property for evaluating the quality of explanations in Explainable AI (XAI). It ensures that the explanation accurately reflects the true reasoning process of the model. In other words, an explanation is considered faithful if it genuinely corresponds to the decision-making process of the model, without introducing any misleading or extraneous information. This concept is crucial for building trust in the explanations provided by complex models, including traditional machine learning models, deep neural networks, and large language models (LLMs) \cite{jacovi2020towards}.

\paragraph{Scope of Application}

Faithfulness is applicable across various types of models:
\begin{itemize}
    \item \textbf{Traditional Machine Learning Models:} In simpler models like linear regression or decision trees, faithfulness is usually inherent due to their interpretable nature. However, it becomes more challenging for ensemble models like random forests.
    \item \textbf{Complex Machine Learning Models:} For deep learning models, such as convolutional neural networks (CNNs) and recurrent neural networks (RNNs), faithfulness is critical when using post-hoc explanation techniques (e.g., SHAP, LIME).
    \item \textbf{Large Language Models (LLMs):} In the context of LLMs, faithfulness is essential for ensuring that the generated explanations truly reflect the internal reasoning of the model rather than being fabricated or biased by unrelated factors.
\end{itemize}

\paragraph{Principle and Formula}

To formally define faithfulness, we can assess the \textbf{Faithfulness Score} using the change in the model's output when key features identified by the explanation are perturbed. The intuition is that if an explanation identifies the most important features, perturbing these features should significantly alter the model's prediction.

Given a model \( f(\mathbf{x}) \) and an explanation that identifies a set of important features \( S \), the Faithfulness Score \( F \) can be expressed as:

\[
F = \frac{\left| f(\mathbf{x}) - f(\mathbf{x}_{\setminus S}) \right|}{\left| f(\mathbf{x}) \right|},
\]

where:
\begin{itemize}
    \item \( \mathbf{x} \) is the original input instance.
    \item \( \mathbf{x}_{\setminus S} \) is the input with the features in set \( S \) perturbed or removed.
    \item \( f(\mathbf{x}) \) is the model's original prediction.
\end{itemize}

A higher Faithfulness Score indicates that the explanation is more faithful, as perturbing the important features significantly affects the model's prediction \cite{hooker2019benchmark}.

\paragraph{Advantages of Faithfulness Evaluation}

\begin{itemize}
    \item \textbf{Enhanced Trustworthiness:} High faithfulness ensures that the explanation is reliable and accurately reflects the model's reasoning process.
    \item \textbf{Improved Model Interpretability:} Faithful explanations provide insights that align well with the actual decision-making process of the model, making it easier for users to understand.
    \item \textbf{Broad Applicability:} The concept of faithfulness can be applied to various models and explanation techniques, making it a versatile evaluation criterion.
\end{itemize}

\paragraph{Challenges and Limitations}

Despite its importance, evaluating faithfulness poses several challenges:
\begin{itemize}
    \item \textbf{Choice of Perturbation Method:} The Faithfulness Score can be sensitive to the method used for perturbing the input features, requiring careful selection.
    \item \textbf{Computational Cost:} Calculating faithfulness can be computationally intensive, especially for deep learning models and large datasets.
    \item \textbf{Trade-off with Simplicity:} Achieving high faithfulness may involve using complex explanation methods, which could reduce the simplicity and interpretability of the explanation.
\end{itemize}

\section*{Conclusion}

The evaluation of interpretability and the associated challenges are central to the development of trustworthy AI systems. By adopting a combination of quantitative and qualitative evaluation methods, practitioners can better assess the interpretability of their models. Nonetheless, ongoing challenges, particularly the trade-offs between interpretability and performance, the tension between transparency and privacy, and the need for standardization, highlight the complexities of this evolving field. Addressing these challenges will be crucial as we continue to develop more interpretable and accountable AI systems.

\chapter{Tools and Frameworks}

\label{sec:ToolsFrameworks}

\section{Model-Agnostic Tools}

\begin{itemize}
    \item \textbf{LIME (Local Interpretable Model-Agnostic Explanations)} \cite{Ribeiro2016}
        \begin{itemize}
            \item A Python library for interpreting individual predictions of black-box models.
            \item GitHub: \url{https://github.com/marcotcr/lime}
        \end{itemize}
    \item \textbf{SHAP (SHapley Additive exPlanations)} \cite{Lundberg2017}
        \begin{itemize}
            \item A unified approach based on cooperative game theory for interpreting models.
            \item GitHub: \url{https://github.com/slundberg/shap}
        \end{itemize}
    \item \textbf{Anchors} \cite{Ribeiro2018}
        \begin{itemize}
            \item A method for explaining predictions with high precision rules.
            \item GitHub: \url{https://github.com/marcotcr/anchor}
        \end{itemize}
    \item \textbf{DALEX} \cite{Biecek2018DALEX}
        \begin{itemize}
            \item Offers visualization tools for explaining models in R and Python.
            \item GitHub: \url{https://github.com/ModelOriented/DALEX}
        \end{itemize}
\end{itemize}

\section{Deep Learning Explainability}

\begin{itemize}
    \item \textbf{Captum} \cite{Kokhlikyan2020Captum}
        \begin{itemize}
            \item PyTorch library for gradient-based and perturbation-based attribution.
            \item GitHub: \url{https://github.com/pytorch/captum}
        \end{itemize}
    \item \textbf{DeepExplain} \cite{Ancona2018DeepExplain}
        \begin{itemize}
            \item A unified framework for gradient-based interpretability.
            \item GitHub: \url{https://github.com/marcoancona/DeepExplain}
        \end{itemize}
    \item \textbf{Grad-CAM} \cite{Selvaraju2017GradCAM}
        \begin{itemize}
            \item Visual explanations for CNN predictions, widely used in image analysis.
            \item GitHub: \url{https://github.com/ramprs/grad-cam}
        \end{itemize}
    \item \textbf{Integrated Gradients} \cite{Sundararajan2017IntegratedGradients}
        \begin{itemize}
            \item A method for attributing neural network predictions to input features.
            \item GitHub: \url{https://github.com/ankurtaly/Integrated-Gradients}
        \end{itemize}
    \item \textbf{DeepLIFT} \cite{Shrikumar2017DeepLIFT}
        \begin{itemize}
            \item Computes feature importance effectively for deep learning models.
            \item GitHub: \url{https://github.com/kundajelab/deeplift}
        \end{itemize}
\end{itemize}

\section{Fairness and Trustworthiness Tools}

\begin{itemize}
    \item \textbf{AI Fairness 360 (AIF360)} \cite{Bellamy2019AIF360}
        \begin{itemize}
            \item IBM toolkit for measuring and mitigating bias in machine learning.
            \item GitHub: \url{https://github.com/IBM/AIF360}
        \end{itemize}
    \item \textbf{Fairlearn} \cite{Bird2020Fairlearn}
        \begin{itemize}
            \item A toolkit for assessing and improving fairness in models.
            \item GitHub: \url{https://github.com/fairlearn/fairlearn}
        \end{itemize}
    \item \textbf{Fairness Indicators} \cite{Google2019FairnessIndicators}
        \begin{itemize}
            \item Evaluates fairness metrics in machine learning models.
            \item GitHub: \url{https://github.com/tensorflow/fairness-indicators}
        \end{itemize}
    \item \textbf{What-If Tool (WIT)} \cite{Wexler2019WhatIfTool}
        \begin{itemize}
            \item Interactive tool for exploring fairness and model behavior.
            \item GitHub: \url{https://github.com/PAIR-code/what-if-tool}
        \end{itemize}
\end{itemize}

\section{Large Language Model (LLM) Explainability}

\begin{itemize}
    \item \textbf{ExBERT} \cite{Hoover2019ExBERT}
        \begin{itemize}
            \item A visual tool for interpreting attention mechanisms in BERT models.
            \item GitHub: \url{https://github.com/bhoov/exbert}
        \end{itemize}
    \item \textbf{ExplainaBoard} \cite{Liu2021ExplainaBoard}
        \begin{itemize}
            \item Provides error analysis and explainability metrics for NLP models.
            \item GitHub: \url{https://github.com/neulab/ExplainaBoard}
        \end{itemize}
\end{itemize}

\section{Framework-Specific Tools}

\begin{itemize}
    \item \textbf{TensorFlow Model Analysis (TFMA)} \cite{Mikhailov2019TFMA}
        \begin{itemize}
            \item Framework for model analysis and explainability in TensorFlow.
            \item GitHub: \url{https://github.com/tensorflow/model-analysis}
        \end{itemize}
    \item \textbf{tf-explain} \cite{Meudec2020tf-explain}
        \begin{itemize}
            \item TensorFlow 2.x library providing explainability methods like Grad-CAM.
            \item GitHub: \url{https://github.com/sicara/tf-explain}
        \end{itemize}
    \item \textbf{AIX360 (AI Explainability 360)} \cite{Arya2019AIX360}
        \begin{itemize}
            \item IBM's toolkit with a comprehensive suite of explainability algorithms.
            \item GitHub: \url{https://github.com/IBM/AIX360}
        \end{itemize}
\end{itemize}

\section{Visualization and Interactive Tools}

\begin{itemize}
    \item \textbf{Facets} \cite{Wexler2019Facets}
        \begin{itemize}
            \item Tool for visualizing datasets to aid in model debugging.
            \item GitHub: \url{https://github.com/PAIR-code/facets}
        \end{itemize}
    \item \textbf{SHAPash} \cite{Gillet2021SHAPash}
        \begin{itemize}
            \item Provides interactive dashboards for SHAP values visualization.
            \item GitHub: \url{https://github.com/MAIF/shapash}
        \end{itemize}
\end{itemize}

\section*{Conclusion}

The tools and frameworks discussed in this chapter provide a strong foundation for implementing explainable AI. By leveraging these resources, practitioners can create models that are both accurate and interpretable, paving the way for broader adoption of AI technologies in sensitive and high-stakes domains.

\chapter{Future Directions and Research Opportunities}

Explainable AI (XAI) is an evolving field with exciting potential and numerous open research questions. As AI continues to be integrated into critical domains, the need for interpretability, transparency, and trustworthiness becomes increasingly urgent \cite{arrieta2020explainable}. In this chapter, we explore the future trends in XAI, examine its potential across different fields, discuss the integration of XAI with legal regulations, address ethical challenges, and outline open research questions for future exploration.

\label{sec:Future}

\section{Future Trends in XAI Technologies}

The field of XAI is rapidly advancing, and several key trends are shaping its development. These trends focus on enhancing the interpretability of complex models while maintaining performance.

\begin{itemize}
    \item \textbf{Explainability for Large Language Models (LLMs)}: Large Language Models (LLMs) like GPT \cite{brown2020language}, BERT \cite{devlin2018bert}, and T5 \cite{raffel2020exploring} have transformed natural language processing but present significant challenges for interpretability. Future research will likely focus on improving methods for understanding the internal mechanisms of LLMs, such as using fine-grained attention visualization \cite{vig2019multiscale} and probing techniques \cite{tenney2019bert} to uncover the meaning encoded within different layers of the model.

    \item \textbf{Hybrid Approaches for Interpretability}: A growing trend in XAI is the development of hybrid approaches that combine intrinsic interpretability (e.g., decision trees) with post-hoc explanation techniques (e.g., SHAP \cite{Lundberg2017}, LIME \cite{Ribeiro2016}). These hybrid methods aim to leverage the strengths of both interpretable and complex models, providing accurate predictions alongside clear and actionable explanations \cite{guidotti2019survey}.

    \item \textbf{Interactive Explanations and Human-in-the-Loop Systems}: The future of XAI may involve more interactive explanations, where users can query the model and receive tailored explanations. Human-in-the-loop systems, which allow users to interact with the model during training or inference, will enable more dynamic and context-specific explanations, leading to greater trust and usability \cite{wu2022human}.
\end{itemize}

\section{Prospects of Explainable AI Across Different Fields}

XAI has far-reaching implications across numerous fields. As AI applications expand, the need for interpretability becomes paramount to ensure responsible and transparent use of technology.

\begin{itemize}
    \item \textbf{Healthcare}: In healthcare, the adoption of AI-driven diagnostic tools necessitates high levels of interpretability \cite{tjoa2020survey}. Future XAI research will likely focus on creating robust, real-time explanations for complex medical models, ensuring that clinicians can understand and trust AI recommendations in critical care scenarios \cite{holzinger2017we}.

    \item \textbf{Finance}: In finance, explainable models are crucial for regulatory compliance and risk management \cite{bussmann2021explainable}. The increasing adoption of black-box models like deep learning for credit scoring and fraud detection has prompted the need for advanced interpretability tools. Future XAI research in finance will emphasize techniques that provide consistent and legally defensible explanations \cite{rossi2021explainable}.

    \item \textbf{Autonomous Systems}: For autonomous systems such as self-driving cars and drones, interpretability is essential for safety and accountability \cite{heggelund2020explainable}. Future research may explore real-time interpretability methods that can provide explanations of a model's decision-making process as it navigates complex environments, enabling better debugging and trust-building \cite{tian2020explainable}.
\end{itemize}

\section{Integration of XAI with Legal Regulations}

The integration of XAI with legal frameworks is becoming increasingly important, as regulations begin to mandate transparency and accountability in AI systems.

\begin{itemize}
    \item \textbf{Regulatory Requirements for Explainability}: Several regulatory bodies, such as the European Commission, have introduced guidelines for AI systems that include requirements for transparency and explainability \cite{eu2020white}. The General Data Protection Regulation (GDPR), for instance, includes the "right to explanation," which requires that users be provided with understandable reasons behind automated decisions that affect them \cite{goodman2017european}.

    \item \textbf{Challenges in Legal Integration}: Integrating XAI with legal regulations is challenging due to the diversity of interpretability techniques and the lack of standardized evaluation metrics \cite{wachter2018counterfactual}. Future research will need to address the gap between technical explanations and the legal requirements for transparency, developing tools that provide explanations that are both legally compliant and technically sound \cite{veale2018fairness}.
\end{itemize}

\section{Ethical Challenges in XAI}

The rise of XAI brings with it several ethical challenges that must be addressed to ensure responsible use of interpretability techniques.

\begin{itemize}
    \item \textbf{Bias and Fairness}: One ethical concern in XAI is the potential to uncover biases within AI models that were previously hidden \cite{mehrabi2021survey}. While this can lead to improved fairness, it may also expose sensitive or protected attributes, complicating the issue of bias mitigation. Future research must focus on developing XAI methods that can identify and correct biases without infringing on privacy \cite{du2020fairness}.

    \item \textbf{Transparency vs. Privacy}: There is an inherent tension between the goals of transparency and privacy \cite{liao2020privacy}. Providing detailed explanations of model behavior can sometimes reveal sensitive information about the training data, particularly in models trained on personal data. This raises ethical concerns about data privacy and highlights the need for techniques that balance interpretability with privacy protection \cite{shokri2017membership}.
\end{itemize}

\section{Open Research Questions and Future Directions}

Despite significant progress, many open research questions remain in the field of XAI. Addressing these questions will be crucial for advancing the field and improving the interpretability of AI models.

\begin{itemize}
    \item \textbf{How Can We Achieve Interpretability Without Compromising Model Performance?} There is a trade-off between model complexity and interpretability, with simpler models being more interpretable but often less accurate \cite{rudin2019stop}. Future research should explore ways to maintain high predictive performance while improving the interpretability of complex models, potentially through new hybrid approaches or better post-hoc techniques.

    \item \textbf{How Can We Standardize the Evaluation of Interpretability?} The lack of standardized metrics for evaluating interpretability is a major challenge \cite{doshi2017towards}. Different application domains and user needs make it difficult to develop a one-size-fits-all approach. Research efforts should focus on creating a comprehensive framework for evaluating interpretability, incorporating both quantitative and qualitative metrics \cite{doshi2018considerations}.

    \item \textbf{Can We Develop More Robust and Explainable Models for High-stakes Applications?} In high-stakes applications like healthcare and autonomous driving, robustness and interpretability are crucial \cite{arrieta2020explainable}. Future research should prioritize the development of models that are not only accurate but also resilient to adversarial attacks and capable of providing consistent explanations under different scenarios \cite{goodfellow2018making}.
\end{itemize}

\section*{Conclusion}

The future of Explainable AI is full of exciting opportunities and challenges. As AI systems continue to evolve and become more complex, the demand for effective interpretability techniques will only grow. By addressing the key research questions and exploring new trends, the field of XAI has the potential to transform the way we understand and interact with AI, fostering greater trust, transparency, and ethical responsibility.

\bibliographystyle{unsrt}
\bibliography{references}

\end{document}